\documentclass[preprint,3p,times]{elsarticle}

\biboptions{numbers,sort&compress}

\usepackage{hyperref}

\usepackage{booktabs}
\usepackage{arydshln} 
\usepackage{graphicx}%
\usepackage{multirow,makecell}%
\usepackage{amsmath,amssymb,amsfonts}%
\usepackage{mathtools}
\usepackage{siunitx} 
\usepackage{mathrsfs}%
\usepackage{algorithm}%
\usepackage{threeparttable}

\usepackage{algpseudocode}%
\usepackage{subcaption}  

\usepackage{caption}
\usepackage{wrapfig}

\makeatletter
\newenvironment{breakablealgorithm}
  {
   \begin{center}
     \refstepcounter{algorithm}
     \hrule height.8pt depth0pt \kern2pt
     \renewcommand{\caption}[2][\relax]{
       {\raggedright\textbf{\ALG@name~\thealgorithm} ##2\par}%
       \ifx\relax##1\relax 
         \addcontentsline{loa}{algorithm}{\protect\numberline{\thealgorithm}##2}%
       \else 
         \addcontentsline{loa}{algorithm}{\protect\numberline{\thealgorithm}##1}%
       \fi
       \kern2pt\hrule\kern2pt
     }
  }{
     \kern2pt\hrule\relax
   \end{center}
  }
\makeatother

\usepackage{pifont}
\newcommand{\cmark}{\ding{51}}%
\newcommand{\xmark}{\ding{55}}%

\usepackage{amssymb}
\usepackage{amsmath}

\usepackage{bm}
\usepackage{xcolor}

\usepackage[normalem]{ulem}
\useunder{\uline}{\ul}{}
\usepackage[nameinlink,capitalize]{cleveref} 
\journal{journal}

\usepackage{xfrac}

\newcommand\x{\bm{x}}

\newcommand\h{\bm{h}}

\begin{document}

\begin{frontmatter}



\title{Rethinking Recurrent Neural Networks for Time Series Forecasting: A Reinforced Recurrent Encoder with Prediction-Oriented Proximal Policy Optimization } 
\author[label1]{Xin Lai}
\author[label1]{Shiming Deng}
\author[label4]{Lu Yu}
\author[label2]{Yumin Lai}
\author[label1]{Shenghao Qiao}
\author[label3]{Xinze Zhang\corref{cor1}}

\cortext[cor1]{Corresponding author: Xinze Zhang. E-mail address: laixin@hust.edu.cn (Xin Lai), 
smdeng@hust.edu.cn (Shiming Deng),
lu.yu@enpc.fr (Lu Yu),
laiyumin2001@126.com (Yumin Lai),
shenghaoqiao@hust.edu.cn (Shenghao Qiao),
xinze@hust.edu.cn (Xinze Zhang).
}

\affiliation[label1]{organization={School of Management, Huazhong University of Science and Technology},
            city={Wuhan},
            postcode={430074},
            state={Hubei},
            country={China}}
            
\affiliation[label4]{organization={Center for Applied Mathematics,  École des Ponts ParisTech},
            city={Paris},
            postcode={77420},
            state={Île-de-France},
            country={France}}

\affiliation[label2]{organization={School of Mechanical Engineering, Dalian Jiaotong University},
            city={Dalian},
            postcode={116000},
            state={Liaoning},
            country={China}}

\affiliation[label3]{organization={School of Computer Science and Technology, Huazhong University of Science and Technology},
            city={Wuhan},
            postcode={430074},
            state={Hubei},
            country={China}}

\begin{abstract}
Time series forecasting plays a crucial role in contemporary engineering information systems for supporting decision-making across various industries, where Recurrent Neural Networks (RNNs) have been widely adopted due to their capability in modeling sequential data. Conventional RNN-based predictors adopt an encoder-only strategy with sliding historical windows as inputs to forecast future values. However, this approach treats all time steps and hidden states equally without considering their distinct contributions to forecasting, leading to suboptimal performance. To address this limitation, we propose a novel Reinforced Recurrent Encoder with Prediction-oriented Proximal Policy Optimization, RRE-PPO4Pred, which significantly improves time series modeling capacity and forecasting accuracy of the RNN models. The core innovations of this method are: (1) A novel Reinforced Recurrent Encoder (RRE) framework that enhances RNNs by formulating their internal adaptation as a Markov Decision Process, creating a unified decision environment capable of learning input feature selection, hidden skip connection, and output target selection; (2) An improved Prediction-oriented Proximal Policy Optimization algorithm, termed PPO4Pred, which is equipped with a Transformer-based agent for temporal reasoning and develops a dynamic transition sampling strategy to enhance sampling efficiency; (3) A co-evolutionary optimization paradigm to facilitate the learning of the RNN predictor and the policy agent, providing adaptive and interactive time series modeling. Comprehensive evaluations on five real-world datasets indicate that our method consistently outperforms existing baselines, and attains accuracy better than state-of-the-art Transformer models, thus providing an advanced time series predictor in engineering informatics.

\end{abstract}


\end{frontmatter}



\section{Introduction\label{sec:sec1}}

Time series forecasting plays a crucial role in analyzing complex temporal dependencies within modern industrial applications by extracting information from historical data to predict future values. 
It has been widely applied across various engineering domains, such as traffic management \citep{ermagun2018spatiotemporal}, electricity demand forecasting \citep{gebremeskel2021long}, and climate modeling \citep{barrera2022rainfall}.
Existing models in time series prediction can be broadly categorized into statistical and Deep Learning (DL)-based methods. 
Traditional statistical approaches, such as ARIMA \citep{schaffer2021interrupted}, often struggle to model complex nonlinear patterns and to capture intricate temporal dependencies \citep{box2015time}.
To address these limitations, DL-based methods, particularly Recurrent Neural Networks (RNNs) \citep{kong2025unlocking} and Transformers \citep{nie2022time,liu2023itransformer}, have gained widespread adoption.

Compared with Transformer-based architectures, RNNs generally demand significantly fewer computational and memory resources due to their non-quadratic complexity \citep{zhuang2023survey}, making them well suited for resource-constrained settings such as edge computing \citep{wang2018deep} and mobile applications \citep{li2020federated}.
Furthermore, RNNs naturally align with the sequential nature of time series data and exhibit clear temporal flow, thereby providing interpretable insights into information propagation through sequences \citep{kong2025unlocking}.
Therefore, advancing RNN capabilities remains crucial for time series forecasting.

RNNs inherently capture temporal dependencies by sequentially processing inputs through recurrent connections and maintaining a hidden state that encodes historical information. 
The most widely used RNN variants include the Long Short-Term Memory (LSTM) networks \citep{hochreiter1997long} and Gated Recurrent Units (GRUs) \citep{cho2014learning}. 
These gated architectures effectively regulate information flow and substantially alleviate the gradient vanishing and exploding issues that affect traditional RNNs \citep{sherstinsky2020fundamentals}. 
Additionally, several specialized RNN variants have been developed, such as Clockwork RNN \citep{koutnik2014clockwork}, phased LSTM \citep{neil2016phased}, IndRNN \citep{Indrnn}, and xLSTM \citep{beck2024xlstm}. 
These architectural innovations substantially improve the capacity to capture complex temporal dependencies, yielding stronger performance in time series forecasting. Despite these advances, effectively adapting RNN-based models to the unique characteristics of time series prediction remains challenging.

\begin{figure}[t]
    \centering
    \begin{subfigure}[b]{0.51\textwidth}
        \centering
        \includegraphics[width=\textwidth]{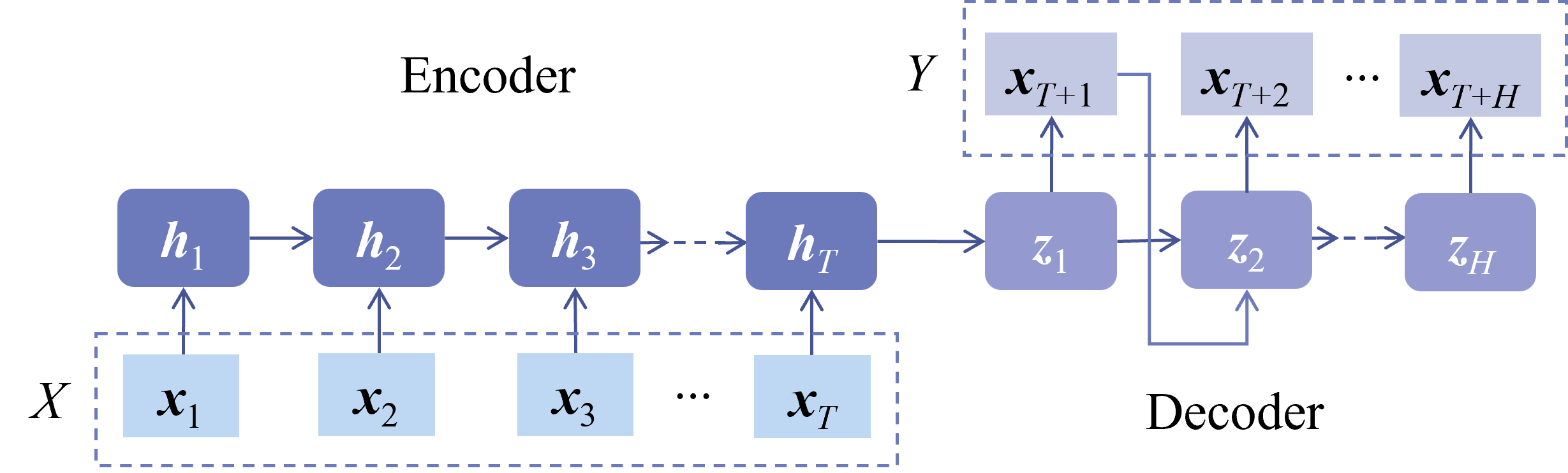} 
        \caption{Seq2Seq RNN architecture}
        \label{fig:seq2seq}
    \end{subfigure}
    \hfill 
    \begin{subfigure}[b]{0.46\textwidth}
        \centering
        \includegraphics[width=\textwidth]{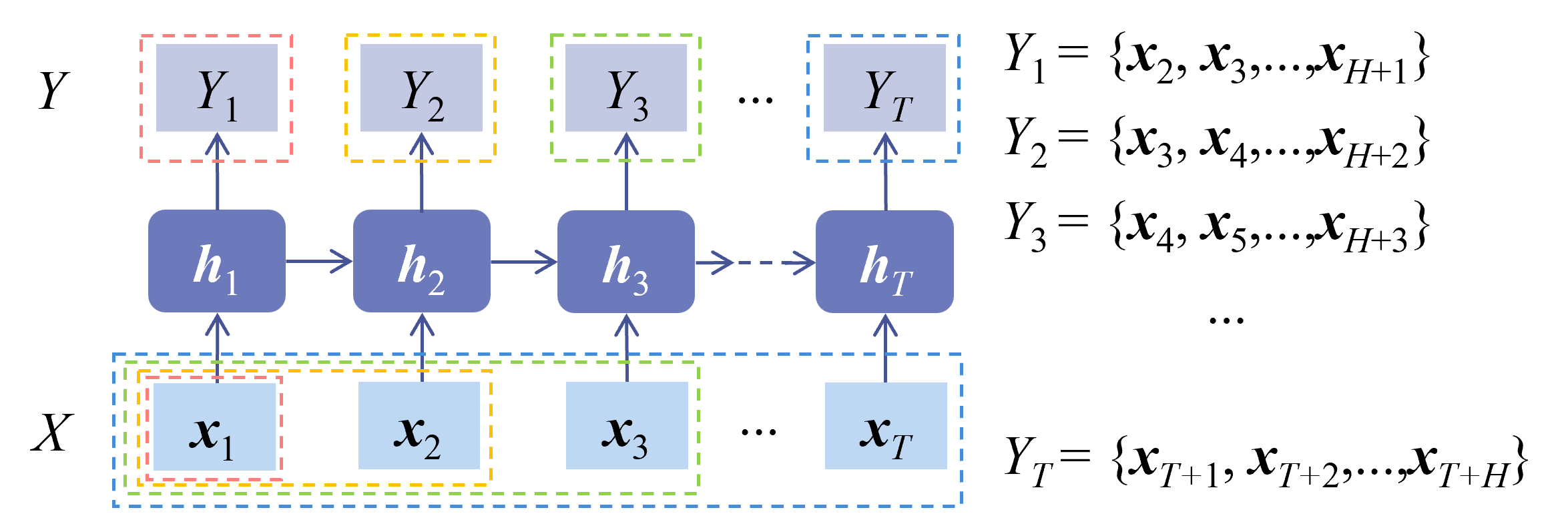} 
        \caption{Encoder-only RNN architecture}
        \label{fig:rnnencoder}
    \end{subfigure}
    \vspace{-2mm}
    \caption{Two RNN architectures for time series forecasting tasks. Solid arrows represent hidden state transitions, and dashed boxes indicate input-output pairs. (a) The Seq2Seq model forms a single training pair (blue dashed box). (b) The Encoder-only model generates multiple training pairs (colored dashed boxes) via an autoregressive structure.}
    \label{fig:architecture}
\end{figure}

Traditional RNN-based time series predictors, largely adapted from Natural Language Processing (NLP) tasks such as neural machine translation \citep{sutskever2014sequence} and abstractive generation \citep{see2017get}, exhibit inherent limitations when directly applied to time series data. 
These models typically adopt a Sequence-to-Sequence (Seq2Seq) structure, as illustrated in \cref{fig:seq2seq}. 
This architecture employs an encoder-decoder framework \citep{cho2014learning}, in which the encoder compresses the full input sequence into a global representation and the decoder iteratively generates future values conditioned on previously predicted outputs. 
While the Seq2Seq paradigm proves effective in NLP tasks, it poses significant challenges in time series forecasting. One key issue is error accumulation, where prediction errors at each step compound during decoding, resulting in escalating inaccuracies for multi-step-ahead forecasting \citep{hewamalage2021recurrent}. 
Moreover, Seq2Seq models commonly depend on ground-truth values as decoder inputs during training to facilitate learning. 
This practice induces exposure bias, where the model must replace ground-truth inputs with its own predictions during inference, creating a training-testing mismatch \citep{ranzato2015sequence}.

Beyond the Seq2Seq style, RNN-based models can adopt an encoder-only strategy for time series forecasting, as illustrated in \cref{fig:rnnencoder}. 
This autoregressive strategy employs sliding historical windows as inputs to predict future horizons.
By generating multiple input-target pairs from a single time series, it leverages diverse historical segments as distinct temporal contexts, effectively augmenting the training data and promoting richer temporal representation learning \citep{hewamalage2021recurrent}. During inference, the encoder-only model directly derives forecasts from the final hidden state, thereby avoiding both error accumulation and exposure bias.

\begin{figure}[t]
    \centering
    \begin{subfigure}[b]{0.49\textwidth}
        \centering
        \includegraphics[width=\textwidth]{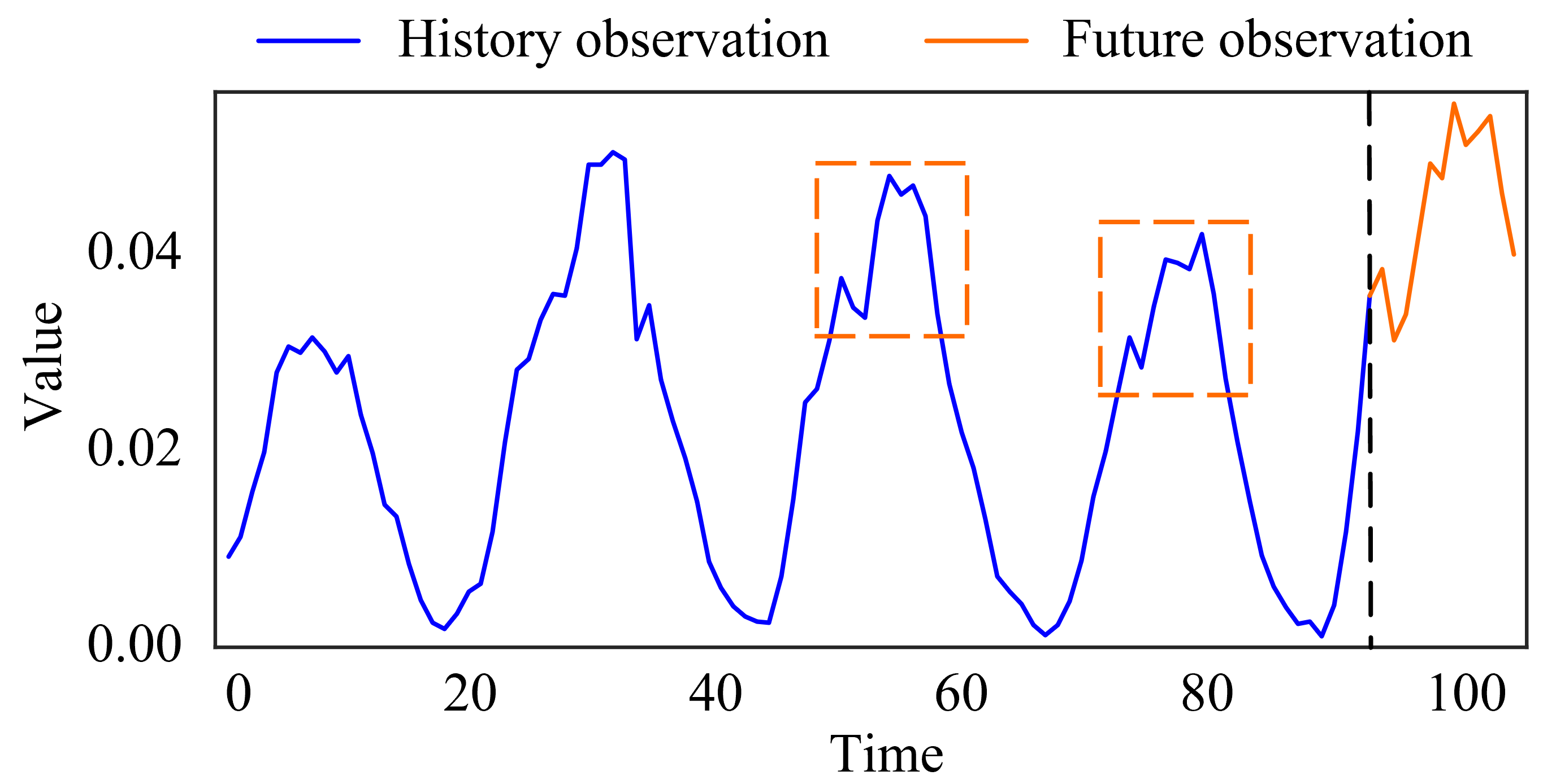} 
        \caption{Forecasting example 1}
        \label{fig:sample1}
    \end{subfigure}
    \hfill 
    \begin{subfigure}[b]{0.49\textwidth}
        \centering
        \includegraphics[width=\textwidth]{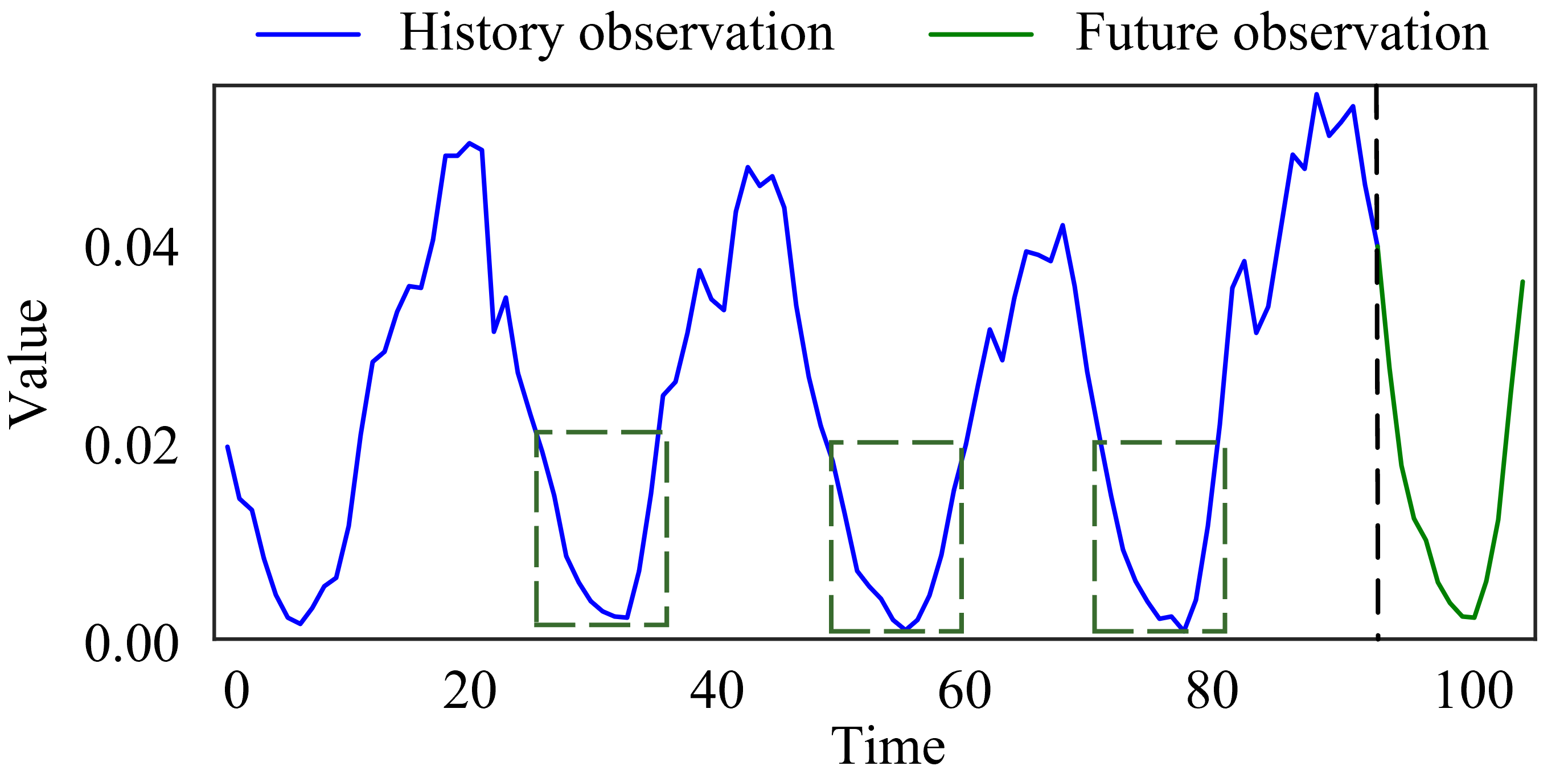} 
        \caption{Forecasting example 2}
        \label{fig:sample2}
    \end{subfigure}
    \vspace{-2mm}
    \caption{Illustration of time series forecasting examples in real-world traffic data \citep{nie2022time}.  Historical observations (96 time steps, blue line) are used to generate future forecasts (12-step horizon, orange and green line), where example 2 is obtained by shifting example 1. The dashed boxes in the historical region highlight segments with patterns similar to the future forecasts. \label{fig:mapping}}
    
\end{figure}

Despite these advances, encoder-only RNN predictors remain limited in capturing the dynamic and diverse nature of temporal dependencies.  As shown in \cref{fig:sample1}, when future observations exhibit peak patterns (orange line), effective forecasting requires adaptively extracting salient characteristics from pattern similar segments in the historical inputs (highlighted by two orange dashed boxes). A shifted example from the same dataset ( \cref{fig:sample2}) further illustrates that different prediction segments rely on distinct historical patterns (highlighted by three green dashed boxes). These contrasting instances highlight the need for adaptive retrieval of relevant historical inputs and dynamic alignment between historical contexts and future targets. However, conventional encoder only RNNs, with their fixed recurrent transitions, lack flexible adjustment mechanisms, which constrains their ability to respond to evolving temporal structures \citep{oreshkin2019n}. 
Therefore, developing adaptive learning mechanisms that enable dynamic alignment to accommodate time-varying patterns is crucial for advancing time series forecasting, which constitutes the focus of this study.

To enhance model adaptability, recent research has introduced input selection mechanisms aimed at reducing the negative impact of irrelevant features on predictive performance \citep{pabuccu2024feature}. 
Heuristic methods typically formulate input selection as a combinatorial optimization problem, employing metaheuristic algorithms such as Genetic Algorithms (GA) \citep{ileberi2022machine}, and Particle Swarm Optimization (PSO) \citep{tijjani2024enhanced} to search for optimal input subsets. 
While effective in exploring large search spaces, heuristic methods typically yield deterministic selection strategies that cannot adapt to the dynamic relevance of inputs across temporal contexts. 
Reinforcement Learning (RL)-based methods treat input selection as a sequential decision-making process, where an agent learns to adaptively select inputs based on historical states and immediate rewards \citep{yoon2018invase,ren2023mafsids}. RL can capture dynamic changes in temporal data and adjust input selection strategies based on environmental feedback, making it well-suited for time series forecasting.

Beyond selecting relevant inputs, a complementary line of work focuses on enhancing hidden-state generation within RNN architectures. Researchers have explored hidden skip connection strategies, such as recurrent skip connections \citep{zhang2016architectural}, dilated skip patterns \citep{chang2017dilated}, and fixed shortcut mappings \citep{campos2017skip}. 
These models introduce direct links between non-adjacent time steps to enlarge the effective receptive field and access information over longer time horizons \citep{tsungnanlearning}. 
However, the skip lengths are predetermined, limiting the ability to dynamically capture temporal dependencies or align hidden states with inputs as temporal patterns evolve. 
To enable adaptive hidden feature generation, recent work has incorporated RL methods  \citep{yu2017learning,weerakody2023policy} through algorithms such as Policy Gradient (PG) \citep{sutton2000policy}, Deep Q-Network (DQN) \citep{mnih2015human}, and Proximal Policy Optimization (PPO) \citep{schulman2017proximal}. These RL-based approaches adaptively regulate skip patterns according to input characteristics, thereby enhancing their capacity to capture complex temporal relationships.

In parallel with these developments, the supervised learning-based mapping mechanism has emerged as another promising direction. This mechanism focuses on transforming hidden states into output predictions by minimizing forecasting errors to train the output weights. Research in this area includes regularization approaches \citep{krueger2016zoneout} that impose sparsity to refine output weights, as well as heuristic methods \citep{wang2019optimizing,sima2025enhancing} that employ backtracking search to optimize the output layer. 
Although these methods enhance the stability and expressiveness of the output target mapping, progress in this line of work remains modest compared with advances in input selection and hidden-state optimization.

Integrating input selection, hidden-state skipping, and output mapping within a single framework would enable end-to-end learning and substantially improve adaptability to time-varying patterns.
However, achieving such integration introduces significant technical challenges. Even for input selection alone, Even for input selection alone, deciding which elements to retain or discard over $T$ time steps yields a combinatorial space of $\mathcal{O}(2^T)$ possibilities. When combined with hidden-state and output-mapping strategies, the overall search space expands exponentially, creating a highly complex and dynamic optimization problem. This inherent complexity has constrained progress toward unified and adaptive temporal modeling frameworks.

To address these challenges, we propose a Reinforced Recurrent Encoder (RRE) framework that formulates encoder-only RNN enhancement as a Markov decision process.
RRE provides a unified environment for dynamically determining three key decisions: input feature selection, hidden skip connection, and output target selection, which enhances the adaptability and predictive capability of diverse encoder-only RNN-based time series forecasting models.
To construct an effective agent within RRE, we develop PPO4Pred, an improved PPO method specifically designed for prediction tasks. 
PPO4Pred introduces two key innovations: a Transformer-based agent that leverages self-attention mechanisms to effectively model the temporal dependencies intrinsic to time series forecasting; a Dynamic Transition Sampling (DTS) strategy that adaptively prioritizes informative transitions to enhance sampling efficiency for training the agent.
Through iterative joint optimization of the RNN environment and the agent, our approach facilitates co-evolution between the encoder-only RNN predictors and the agent, progressively improving both the representational capacity and adaptability of encoder-only RNNs for time series forecasting. 
Together, the RRE framework and PPO4Pred algorithm form the final RRE-PPO4Pred method.
The main contributions of this work are summarized as follows:
\begin{itemize}
\item A novel RRE framework is proposed by integrating input feature selection, hidden skip connection, and output target selection to enhance the adaptability and predictive capability of encoder-only RNN-based time series forecasting models.
\item An innovative PPO4Pred algorithm is developed for RRE framework, featuring two improvements: a Transformer-based agent that leverages self-attention to effectively model the temporal dependencies, and a DTS strategy that substantially improves sample efficiency of the agent.
\item A co-evolutionary optimization paradigm is devised to facilitate the joint evolution of the agent in PPO4Pred and RNN environment in the RRE framework, providing adaptive and interactive time series modeling.
\item Comprehensive experiments across five real-world datasets  demonstrate that our method consistently surpasses state-of-the-art baselines across eight RNN architectures and three prediction horizons. Furthermore, our method achieves comparable performance to advanced Transformer-based models.
\end{itemize}

The rest of this paper is organized as follows. Section \ref{sec:related} introduces the problem formulation and the literature related to this work.
In Section \ref{method}, the proposed method is illustrated in detail. 
Section \ref{setup} introduces the data resources and experimental setup.
Section \ref{experiment} analyzes the experimental results.
Finally, the conclusion is drawn in Section \ref{conclusion}.

\section{Related Work\label{sec:related}}

\subsection{Problem formulation}\label{sec:RNNfore}
In a general time series forecasting task, the objective is to predict a sequence of future  observations based on a given sequence of historical data. 
Formally, let the input and target be represented as $(X, Y) \in (\mathbb{R}^{T \times D_{in}}, \mathbb{R}^{H \times D_{out}})$. 
$X = \{ \bm{x}_1, \ldots, \bm{x}_T \} \in \mathbb{R}^{T \times D_{in}}$ denotes the input sequence comprising $T$ historical observations, where each $\bm{x}_t \in \mathbb{R}^{D_{in}}$ is a $D_{in}$-dimensional vector, and $D_{in}$ denotes the number of input variables at each input time step $t \in \{1, \ldots, T\}$.
Similarly, $Y = \{ \bm{y}_{1}, \ldots, \bm{y}_{H} \} \in \mathbb{R}^{H \times D_{out}}$ represents the corresponding future observations over the forecasting horizon $H$, where each $\bm{y}_{h} \in \mathbb{R}^{D{out}}$, and $D_{out}$ denotes the number of variables to be predicted at each output time step $h \in \{1, \ldots, H\}$. 

Notably, $D_{out}$ can be equal to $D_{in}$ (i.e., multivariate forecasting where $\bm{y}_h = \bm{x}_{T+h}$ as shown in \cref{fig:seq2seq}) or be a subset (e.g., $D_{out} = 1$ and $D_{in} > 1$ for univariate forecasting from multivariate inputs). 
The goal of time series forecasting is to learn a mapping function $\mathcal{F}_{\theta}: \mathbb{R}^{T \times D_{in}} \rightarrow \mathbb{R}^{H \times D_{out}}$, 
parameterized by $\theta$, that minimizes the error between the ground truth target $Y$ and the predicted value $\hat{Y}= \mathcal{F}_{\theta}(X)$.

In contrast to the traditional learning strategy of constructing one input target pair per example,
encoder-only RNNs leverage the auto-regressive nature of time series forecasting to construct $T$ input-target pairs $\{(X_t, Y_t)\}_{t=1}^T$.
For conciseness, we use multivariate forecasting ($D_{out} = D_{in}$ as shown in \cref{fig:rnnencoder}) to illustrate this process.
Specifically, let $\bm{x}_{1:t} =\{\bm{x}_1, \ldots, \bm{x}_t\}$ denote the historical data from the beginning $\bm{x}_1$ to $\bm{x}_t$.
In the training phase of encoder-only RNNs, the hidden state $\bm{h}_t$ and the corresponding output $\hat{Y}_t$ are generated at each time step $t$, which can be formulated as:
\begin{subequations}
    \begin{align}
        \h_t  &= \text{RNNcell}(\x_t, \h_{t-1}),  \label{eq:tht}\\ 
        \hat{Y}_t &= \text{Output}(\h_t), \label{eq:ht}
    \end{align}
\end{subequations}
where $\text{RNNcell}(\cdot)$ denotes the transformation of input-to-hidden and hidden-to-hidden processes\footnote{For simplification, here we ignore the implementation differences of RNN cells in the inner hidden-to-hidden transformation (e.g., LSTM cells and GRU cells). Different RNN cells can be implemented by contextually changing \cref{eq:tht} easily.}.
$\text{Output}(\cdot)$ denotes the output layer which is typically a fully connected layer that projects the hidden state $\bm{h}_t$ to the prediction target $\bm{x}_{t+1:t+H}$.
Let $X_t$ and $Y_t$ be $\bm{x}_{1:t}$ and $\bm{x}_{t+1:t+H}$, respectively.
The above procedures (i.e., \cref{eq:tht} and \cref{eq:ht}) can be briefly formulated as $\hat{Y}_t = \mathcal{F}_{\theta}(X_t)$.

The model $\mathcal{F}_{\theta}(\cdot)$ is trained by minimizing the overall training loss between the targets and predicted values across all time steps:
\begin{equation}
\min \mathcal{L}(\theta) = \min_{\theta} \frac{1}{T} \sum_{t=1}^{T} \ell(\hat{Y}_t, Y_{t}),
\label{eq:lossrnn}
\end{equation}
where $\ell(\hat{Y}_t, Y_{t})$ is typically the Mean Squared Error (MSE) function that measures the prediction error at each forecast step for all output values.
During the inference phase, the last output $\hat{Y}_T$ is generated and used as the final prediction value for evaluating the model performance of the trained model.

This encoder-only strategy effectively augments the number of training pairs and enriches the temporal supervision available to the model.
As introduced in the previous section, many approaches endeavor to improve the learning process of encoder-only RNNs, which can be roughly categorized into three components: input feature selection of $\x_t$, hidden state generation of $\h_t$, and output target selection for $Y_t$. 
The last two components can be collectively viewed as the hidden feature learning of $\h_t \rightarrow Y_t$.

\subsection{Dynamic input feature selection}
In time series forecasting, input selection is critical for determining which variables or temporal segments contribute most to predictive accuracy. 
Unlike static domains, the salience of input variables in time series often varies over time and exhibits complex temporal dependencies. 
This dynamic nature makes input selection particularly challenging, necessitating adaptive approaches.
Within the context of RNN-based forecasting, existing methods can be broadly categorized into heuristic and RL-based approaches.

Heuristic methods formulate input selection as a combinatorial optimization problem aimed at identifying the optimal feature subset. 
Swarm intelligence algorithms, such as Particle Swarm Optimization (PSO) and its Boolean variants, have been integrated with deep learning models to improve generalization performance~\citep{tijjani2024enhanced}.  
Genetic Algorithm (GA) is employed by using encoding, crossover, and mutation to efficiently explore high-dimensional feature spaces~\citep{ileberi2022machine}, while Whale Optimization Algorithm (WOA) is utilized by deriving sparse yet robust feature subsets~\citep{nadimi2022enhanced}. 
However, these methods produce deterministic feature subsets prior to model training, limiting their adaptability to time-varying feature relevance. 
Moreover, their search complexity increases exponentially with feature dimensionality, and their capacity to capture cross-temporal dependencies in non-stationary time series remains limited \citep{xue2015survey}.

RL-based methods conceptualize feature selection as a sequential decision-making process, where agents learn policies to activate or deactivate features based on historical states and feedback. 
\citet{yoon2018invase} employ an Actor-Critic framework to enable sample-specific feature selection. 
\citet{fan2020autofs} propose an interactive reinforced feature selection, where agents learn from both self-exploration and expert guidance. \citet{liu2021automated} extend this paradigm through multi-agent reinforcement learning, in which agents corresponding to individual feature dimensions collaboratively optimize decisions. 
\citet{ren2023mafsids} further enhance the prior works by introducing cooperative agent communication mechanisms. 
However, RL-based methods are inherently sensitive to reward design, state representation, and policy stability, often requiring large sample size to achieve convergence.

Therefore, dynamic input selection mechanisms that explicitly model temporal dependencies and adapt to time-varying feature importance are essential for time series forecasting on improving both predictive accuracy and model interpretability.

\subsection{Hidden feature learning mechanism}
Beyond input feature selection, enhancing encoder-only RNN-based  forecasting models requires addressing two critical components. 
The first is how hidden states are generated across time steps to capture temporal dependencies, and the second is how hidden states are mapped to prediction targets. 
Recent research has explored both fixed and adaptive strategies for these two mechanisms.

A common manner of improving the hidden state generation is skip connection, which establishes direct hidden state connections between non-adjacent time steps, extending the temporal receptive field and mitigating gradient vanishing.
Early fixed skip strategies established predetermined connectivity patterns. 
\citet{zhang2016architectural} show that increasing recurrent skip coefficients significantly enhances forecasting performance on long-term dependency tasks. 
\citet{irie2016lstm} incorporate highway connections with gating mechanisms into LSTMs to enable direct information flow across layers.
Dilated RNNs extend the temporal receptive field through fixed-stride dilated connections that skip intermediate time steps~\citep{chang2017dilated}. 
Skip RNN employs a learned skipping mechanism to determine when to bypass hidden state updates~\citep{campos2017skip}.
DA-SKIP~\citep{huang2021novel} introduces gated recurrent network with fixed skip state connection to solve the long-distance dependence problem in time series forecasting. 
However, these predetermined skip lengths lack adaptability to time-varying patterns in non-stationary time series.

Dynamic hidden-state connection mechanisms have been proposed that adaptively adjust the connectivity between hidden states based on the content of the input sequence.
\citet{huang2019leap} present Leap-LSTM, which uses an MultiLayer Perceptron (MLP) to compute skipping probabilities. 
To address non-differentiability, various approaches have employed differentiable relaxations or reinforcement learning.
Structural-Jump-LSTM \citep{hansen2019neural} utilizes Gumbel-Softmax estimators to dynamically learn hidden state jumping mechanism of LSTMs. 
RL-LSTM \citep{gui2019long} leverages RL to select historical hidden states.
Similarly, LSTMJump \citep{yu2017learning} and PGLSTM \citep{weerakody2023policy} introduce dynamic skip connections using PG algorithm to capture both periodic and aperiodic patterns. 
However, most of these methods are developed for discrete NLP tasks and are not directly applicable to continuous time series forecasting.

Compared to the extensive studies on hidden state skip connections, output target selection has received limited attention from researchers.
In RNN-based forecasting models, output target selection determines the supervised relationships between hidden representations and prediction targets, thereby playing a crucial role in aligning learned temporal features with forecasting objectives.
Traditional RNNs employ fixed linear transformations from final hidden states, which may lead to suboptimal modeling when the representations contain redundant information. 
To enhance adaptive hidden-output mapping, \citet{krueger2016zoneout} introduce Zoneout, which stochastically preserves hidden activations. 
\citet{wang2019optimizing} apply backtracking search to optimize output weights, while \citet{sima2025enhancing} explore output mappings by combining the PSO algorithm with Bayes optimization.
Nevertheless, existing work treats output mapping in isolation, neglecting its synergy with input selection and hidden-state connectivity.  
Therefore, this study aims to bridge this gap by coordinating adaptation across all three components, which is essential for capturing dynamic patterns in time series forecasting.

\section{Methodology}\label{method}
\begin{figure}[t]
    \centering
    \includegraphics[width=0.95\linewidth]{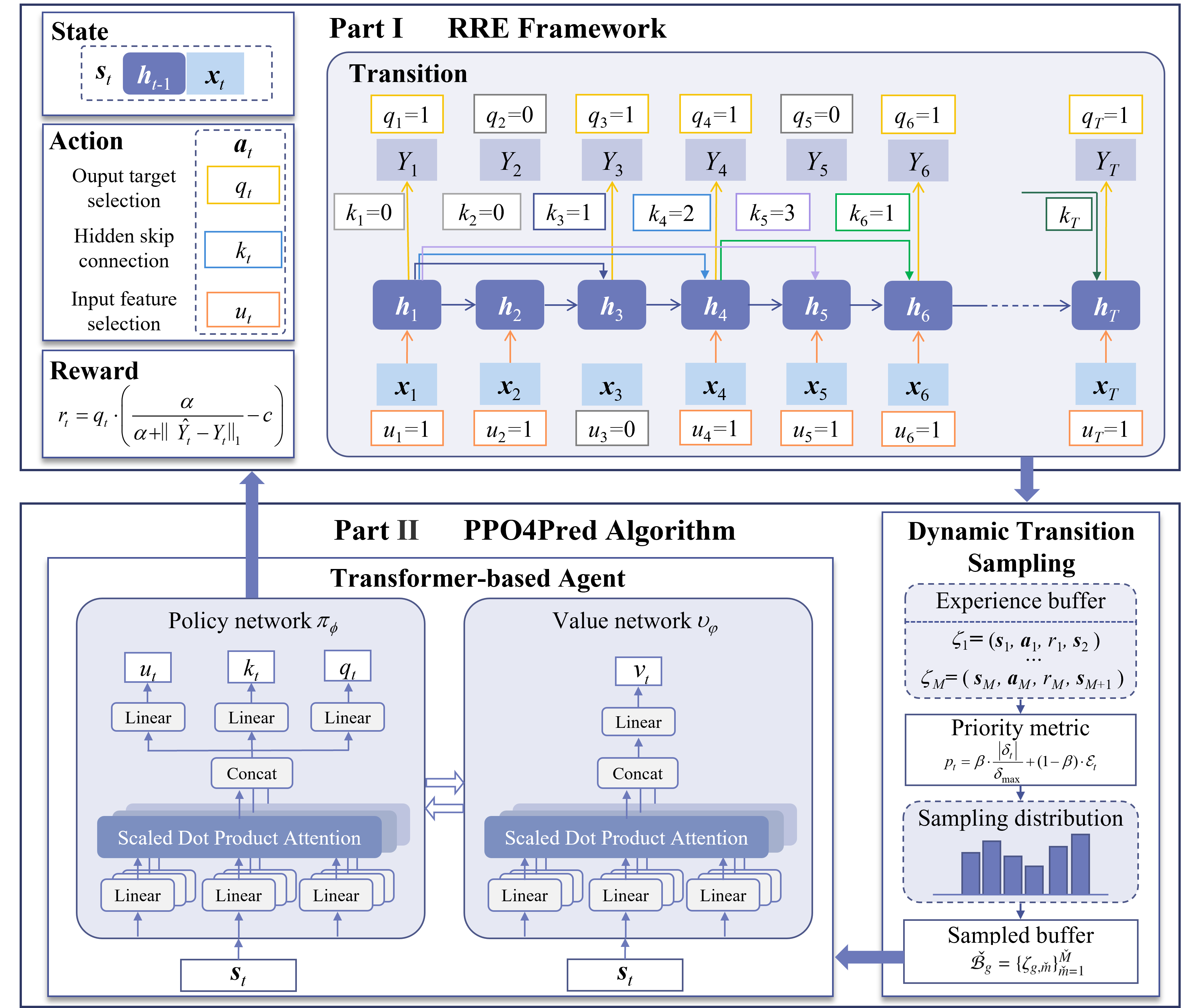}
    \caption{Architecture of the RRE-PPO4Pred method. Part I: RRE framework with sequential decision-making process; Part II: PPO4Pred algorithm with Transformer-based agent and dynamic transition sampling. }
    \label{fig:framework}
\end{figure}

Given the inherent sequential and dynamic nature of encoder-only RNNs for time series forecasting, the optimization problems of input feature selection, hidden skip connection, and output target selection all share the characteristics of sequential decision-making problems.
Therefore, inspired by the remarkable performance of reinforcement learning techniques in such problems, we begin by constructing a reinforced encoder framework, namely RRE, as illustrated in \cref{fig:framework}, to provide a universal  environment for enhancing encoder-only RNNs for forecasting tasks.
We then develop a novel PPO4Pred algorithm with a Transformer-based agent and dynamic transition sampling to facilitate efficient agent training.
Finally, we outline the implementation procedure of the proposed RRE-PPO4Pred method, which combines RRE framework and PPO4Pred algorithm in a co-evolutionary optimization paradigm.

\subsection{Reinforced encoder framework \label{sec:reenc}}
In our RRE framework, 
we aim to design an agent that interacts with the RNN environment, where the agent uses a policy to choose an appropriate action at each time step.
We formulate this dynamic optimization problem as a Markov Decision Process (MDP), which comprises the four-tuple $\mathcal{M}=\{\mathcal{S}, \mathcal{A}, \mathcal{P}, \mathcal{R}\}$, i.e., a state space $\mathcal{S}$, an action space $\mathcal{A}$, a transition probability $\mathcal{P}$, and a reward function $\mathcal{R}$. 

\subsubsection{State and action \label{sec:stateandaction}}

\textbf{State space} $\mathcal{S}$ encapsulates the RNN information available to the agent for decision-making at each time step.
We construct the environmental state $\bm{s}_t \in \mathcal{S}$ by integrating the temporal hidden state with the current input observation.
Specifically, at time step $t$, the state $\bm{s}_t$ is formed by concatenating the previous hidden state $\bm{h}_{t-1}$ (which encodes the historical context) and the current input vector $\bm{x}_t$ (which provides real-time feature information): 
$\bm{s}_t = \text{concat}(\bm{h}_{t-1}, \bm{x}_t)$,
where $\text{concat}(\cdot)$ denotes the concatenation operation.

For the initial time step $t=1$, the hidden state $\bm{h}_0$ is initialized as a zero vector $\bm{0}$ or configured according to the applied the specific RNN cell.
In the environmental state $\bm{s}_{t}$, the first term $\bm{h}_{t-1}$ provides the agent with access to the network's internal memory, enabling it to assess the informativeness of previously processed temporal patterns. 
The second term $\bm{x}_t$ allows the agent to evaluate the characteristics of the current input. 
This state design thus enables the agent to make context-aware architectural decisions that adapt to both historical dynamics and immediate observations, directly supporting the actions introduced below.

\textbf{Action space} $\mathcal{A}$ defines the boundaries within which the agent can act to assist the predictors.
At each time step $t \in \{1, 2, \ldots, T\}$ within an input time series $\x_{1:T}$, the agent observes the current environmental state $\bm{s}_t \in \mathcal{S}$. 
Based on this state, the agent selects a ternary action $\bm{a}_t = \{u_t, k_t, q_t\} \in \mathcal{A}$ according to its current policy, where $u_t$ determines input feature selection, $k_t$ determines hidden skip connection, and $q_t$ determines output target selection. 

\textbf{Input feature selection} is widely used to address noisy or irrelevant features in high-dimensional time series.
To enable selective and dynamic input feature selection, we introduce an input feature selection action $u_t \in \{0, 1\}$ that determines whether the current input $\bm{x}_t$ is fed into the encoder at time step $t$. 
The binary decision mechanism operates as follows. When $u_t = 1$, the input $\bm{x}_t$ is propagated through the network and contributes to the hidden-state update according to standard RNN dynamics. Conversely, when $u_t = 0$, the input is replaced by a zero vector, removing its influence on the hidden-state computation. In this case, the network relies solely on its internal memory, which is provided by the previous hidden state, to preserve temporal context.
Formally, the input-aware hidden state update is given by:
\begin{equation}
\bm{h}_t = \text{RNNcell}(u_t \odot \bm{x}_t, \bm{h}_{t-1}),
\label{eq:gatea}
\end{equation}
where $u_t \odot \bm{x}_t$ denotes element-wise multiplication. This formulation preserves the standard RNN structure while introducing learnable input selectivity.
Moreover, it provides interpretability by identifying informative time steps, enabling post-hoc analysis of temporal importance.

\textbf{Hidden skip connection} creates direct paths from distant past hidden states to the current hidden state, thereby updating and strengthening the memory of longer input patterns for the RNN-based forecasting models.
Unlike fixed-interval skip connections in traditional RNNs, dynamic hidden skip connections enable context-aware adjustment of the temporal receptive field.
However, this mechanism has been seldom investigated for RNN-based forecasting tasks.

In the proposed RRE framework,  a dynamic hidden skip connection action $k_t \in \{0, 1, 2, \ldots, K\}$ is designed. 
At time step $t$, the agent performs a non-skip operation when $k_t = 0$.
When $k_t \in \{1, 2, \ldots, K\}$, the agent selects the corresponding historical hidden state $\bm{h}_{t-k_t -1 }$ from a dynamic candidate set $\mathcal{H}_t = \{\bm{h}_{t-K-1}, \ldots, \bm{h}_{t-3}, \bm{h}_{t-2}\}$ to connect it to the current hidden state $\h_t$, where $\mathcal{H}_t$ contains the $K$ most recent hidden states at time step $t$, and $K$ is the predefined skip window size.
This selection enables flexible and  direct information flow from $\bm{h}_{t-k_t - 1}$ to the current time step.
The action incorporating both input feature selection and hidden skip connection can be formulated by modifying \cref{eq:tht} as follows:
\begin{equation} \label{eq:gateh}
    \bm{h}_t = \begin{cases}
        \text{RNNcell}(u_t \odot \bm{x}_t, \bm{h}_{t-1}), & \text{if}\quad k_t = 0, \\
        \text{RNNcell}(u_t \odot \bm{x}_t, \bm{h}_{t-1}, \bm{h}_{t-k_t-1}), & \text{else},
    \end{cases}
\end{equation}
where the RNN cell $\text{RNNcell}(\cdot)$ is modified to accept three inputs: the gated current input $u_t \odot \bm{x}_t$, the previous hidden state $\bm{h}_{t-1}$, and the selected skip-connected state $\bm{h}_{t-k_t-1}$. 
\begin{figure}[t]
    \centering
    \includegraphics[width=0.5\linewidth]{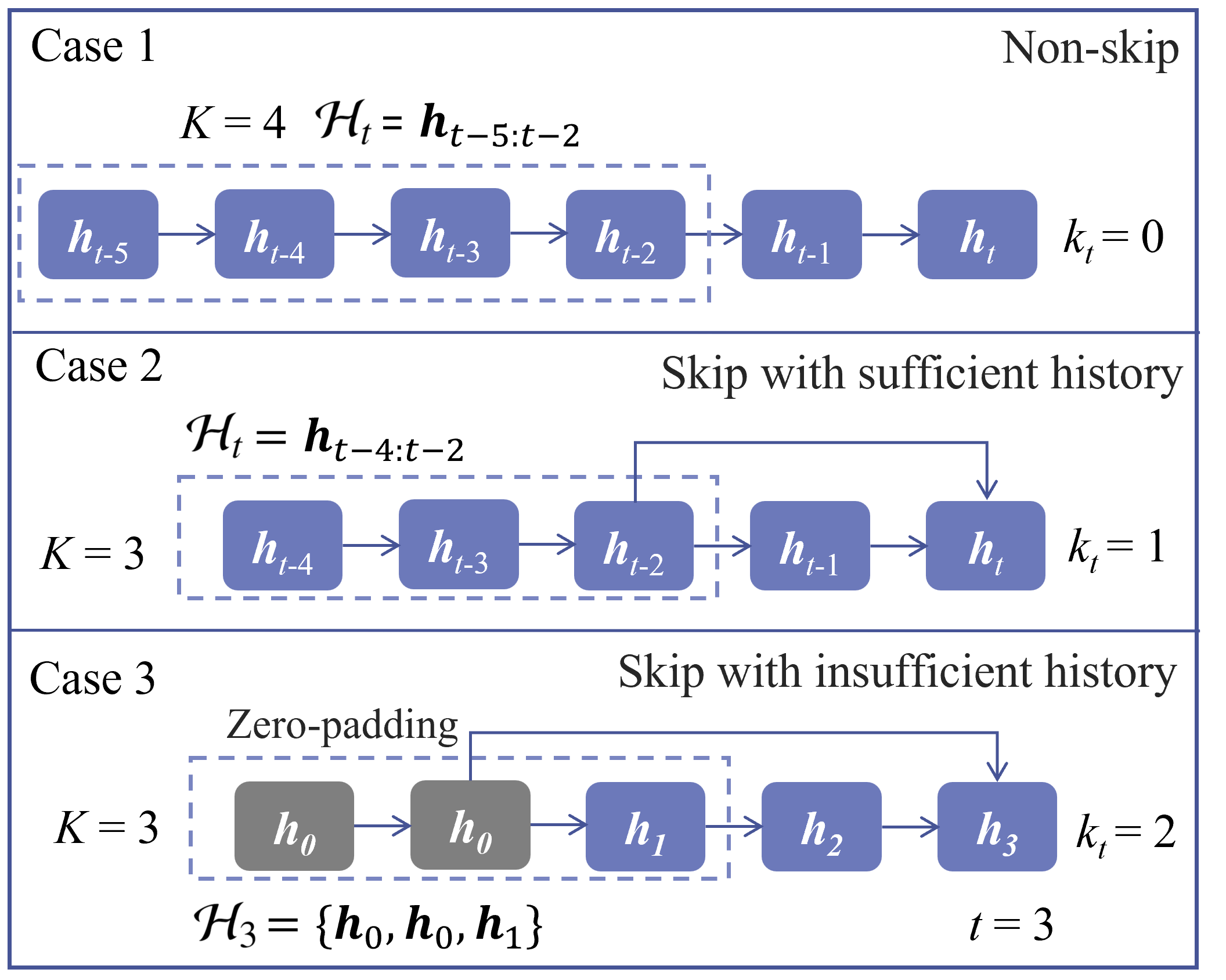}
    \caption{Illustration of dynamic hidden skip connection with three operational cases. 
    Case 1 (Non-skip, $k_t=0$): Standard RNN computation without skip connections. 
    Case 2 (Skip with sufficient history, $k_t \neq 0$, $t-K>1$): Dynamic skip connection selects from full candidate set $\mathcal{H}_t$ of historical hidden states. 
    Case 3 (Skip with insufficient history, $k_t \neq 0$, $t-K \leq 1$): Early time steps require zero-padding (gray blocks) in candidate set due to insufficient historical states. }
    \label{fig:skip}
\end{figure}

Specifically,  when $k_t = 0$, the non-skip operation is performed by replacing $\bm{h}_{t-k_t-1}$ with zero vector $\bm{0}$ of the same dimension  to maintain implementation simplicity.
Let $\h_{t-k-1}$ denote the $k$-th candidate hidden state in $\mathcal{H}_t$.
When $t -K \leq 1$ as shown in case 3 of \cref{fig:skip}, the candidate set $\mathcal{H}_t$ cannot be fully populated with $K$ historical hidden states, since some required hidden states $\{\h_{t-K-1},\ldots, \h_{t-k-2}, \h_{t-k-1}\}$ do not exist at early time steps.
In this case, we pad $\mathcal{H}_t$ with zero hidden states $\bm{0}$ as $\mathcal{H}_t = \{\bm{0},  \ldots, \bm{0}, \bm{h}_{1}, \ldots, \bm{h}_{t-2}\}$, where the number of zero-padded hidden states is $K - t + 2$.
This zero-padding strategy ensures that the action space for the hidden skip connection $k_t$ maintains a consistent candidate size across all time steps.

\textbf{Output target selection} is an important operation in training encoder-only RNNs for time series forecasting.
In the learning process of encoder-only RNN-based forecasting models, the network generates an output $\hat{Y}_t \in \mathbb{R}^{H\times D_{out}}$ at each time step $t$, and evaluates it against the corresponding target $Y_t$, resulting in a total of $T$ output-target supervision pairs per input time series. 
However, in real-world time series forecasting tasks, supervision pairs at intermediate time steps may contribute differently to the overall forecasting objective across different examples, as illustrated in \cref{fig:mapping}. 

To address this challenge, we devise an output target selection action $q_t$ that determines whether an output $\hat{Y}_t$ contributes to the loss at time step $t$. 
This binary scheduling mechanism operates as follows. When $q_t = 1$, the prediction $\hat{Y}_t$ is generated from the current hidden state $\bm{h}_t$ via an output function $\text{Output}(\cdot)$, and the corresponding loss term is included in the training objective. 
Conversely, when $q_t = 0$, the loss term is excluded and its gradient is not backpropagated for this time step, while the network continues to update its hidden state.
Formally, the training objective is formulated by modifying \cref{eq:lossrnn} as:
\begin{equation}
\min \mathcal{L}(\theta) = \min_{\theta} \frac{1}{\|\bm{q}\|_1} \sum_{t=1}^{T} q_t \ell(\hat{Y}_t, Y_{t}), \quad q_t \in \{0, 1\},
\label{eq:gatey}
\end{equation}
where $\bm{q} = [q_1, \ldots,  q_T]$ is the vector of output selection actions across all time steps, and $\|\cdot\|_1$ denotes the $L_1$ norm.

Hence, the action space $\mathcal{A}$ defines the set of architectural control decisions $\bm{a}_t = (u_t, k_t, q_t)$ available to the agent, where each component corresponds to a key optimization objective in encoder-only RNNs for time series forecasting.
As illustrated in \cref{fig:framework}, by using the RNN information as the environment state and three learnable agent actions, our RRE serves as a general framework applicable to various RNN cells.

\subsubsection{Transition and reward}

\textbf{Transition function} $\mathcal{P}(\bm{s}_{t+1} \mid \bm{s}_t, \bm{a}_t)$ defines the probability distribution over next states given the current state and action. 
In this study, the state transition function follows the RNN dynamics and the input time series.
Given the current environment state $\bm{s}_t$ and action $\bm{a}_t = (u_t, k_t, q_t)$, the next state $\bm{s}_{t+1} = \text{concat}(\bm{h}_t, \bm{x}_{t+1})$ is constructed by concatenating the next input $\bm{x}_{t+1}$ and the updated hidden state $\bm{h}_t$, where $\bm{h}_t$ is computed according to the action-conditioned RNN update rule described in \cref{eq:gateh}.

\textbf{Reward function} $\mathcal{R}(\bm{s}_t, \bm{a}_t)$ quantifies the immediate performance of the agent's architectural decisions. 
The reward $r_t$ at time step $t$ depends on the current state $\bm{s}_t$ and action $\bm{a}_t$, reflecting the prediction accuracy resulting from their interaction.
In our RRE framework, the reward aims to provide feedback on time series prediction quality at each time step. 
Specifically, we design a prediction-specific reward ${r}_t$ as:
\begin{equation}
    {r}_t = q_t \cdot \left(\frac{\alpha}{\alpha + \|\hat{Y}_t - Y_t\|_1} - c\right),
    \label{eq:reward}
\end{equation}
where $\alpha > 0$ is the reward sensitivity coefficient.
$\|\cdot\|_1$ denotes the $L_1$ norm, which measures the Mean Absolute Error (MAE) between the prediction outputs $\hat{Y}_t$ and ground-truth target values $Y_t$.
The threshold $c \in (0, 1)$ serves as a critical decision boundary that determines the minimum prediction quality required for  an output to yield a positive reward. 
Mathematically, producing outputs is advantageous (i.e., $r_t > 0$) if and only if:
\begin{equation}
\frac{\alpha}{\alpha + \|\hat{Y}_t - Y_t\|_1} > c \quad \Leftrightarrow \quad \|\hat{Y}_t - Y_t\|_1 < \alpha\left(\frac{1}{c} - 1\right).
\label{eq:reward_condition}
\end{equation}

When the agent decides to select the supervision ($q_t = 1$) and the prediction error $\|\hat{Y}_t - Y_t\|_1$ is small, the quality term ${\alpha}/({\alpha + \|\hat{Y}_t - Y_t\|_1})$ approaches 1. If the prediction quality exceeds the threshold (i.e., ${\alpha}/({\alpha + \|\hat{Y}_t - Y_t\|_1}) > c$), the agent receives a positive reward ${r}_t > 0$. 
This incentivizes the agent to generate outputs when confident about prediction accuracy. 
When $q_t = 1$ but the prediction error is large, the quality term approaches 0, resulting in $r_t \approx -c < 0$. 
This penalty discourages the agent from generating unreliable predictions under high uncertainty, thereby promoting selective output behavior.
When the agent chooses to suppress this supervision ($q_t = 0$), the multiplicative factor ensures ${r}_t = 0$ regardless of the underlying prediction quality. 

Through our prediction-specific reward, the proposed RRE framework provides natural guidance for selecting informative output-target pairs.
The agent learns to produce outputs only when their expected quality surpasses the threshold defined by \cref{eq:reward_condition}, thereby maximizing cumulative rewards by capturing the benefits of accurate predictions while avoiding the costs of erroneous ones.

\subsection{PPO4Pred algorithm \label{sec:ppo}}

To maximize the expected return of the agent in our RRE framework, we propose PPO4Pred, an improved Proximal Policy Optimization (PPO) algorithm tailored for enhancing encoder-only RNN-based time series prediction models.
Compared to the conventional PPO algorithm, PPO4Pred introduces two key improvements: a Transformer-based agent for better environment state learning, and dynamic transition sampling for efficient exploration.

\subsubsection{Transformer-based agent}
To learn the optimal policy $\pi^*$ that maximizes the expected cumulative return, PPO employs two neural networks: a policy network $\pi_\phi$ parameterized by $\phi$ to model the action distribution, and a value network $\upsilon_\varphi$ parameterized by $\varphi$ to estimate state values for variance reduction. 
In PPO4Pred, we introduce a Transformer-based architecture for both networks to effectively capture the temporal dependencies inherent in time series prediction tasks, distinguishing our approach from conventional PPO implementations that typically rely on simple MLP architectures.

\textbf{Policy Network $\pi_\phi(\bm{s}_t)$} models the conditional probability distribution over the ternary action space $\mathcal{A}$ given the current state $\bm{s}_t$. 
Unlike conventional PPO implementations that typically employ simple MLP architectures, our design introduces a Transformer-based architecture.
Specifically, we employ a lightweight Transformer encoder as a backbone to process the current state $\bm{s}_t$, generating a contextual embedding:
\begin{equation}
    \bm{e}_t^\pi = \text{Transformer}_{\phi_e}(\bm{s}_t) \in \mathbb{R}^{D_e},
    \label{eq:trans}
\end{equation}
where $\text{Transformer}_{\phi_e}(\cdot)$ denotes the Transformer encoder parameterized by $\phi_e$, and $D_e$ is the embedding dimension of $\bm{e}_t^\pi$.
Building upon the representation $\bm{e}_t^\pi$, three parallel task-specific MLP heads independently compute logits for each action component.
The logits from each head are then transformed into categorical probability distributions via the softmax function to determine the corresponding action components $u_t$, $k_t$, and $q_t$ as follows:
\begin{equation}
\begin{aligned} \label{eq:a_mlp}
u_t   &= \operatorname{Softmax}(\operatorname{MLP}_{\phi_u}(\bm{e}_t^\pi)), \\
k_t &= \operatorname{Softmax}(\operatorname{MLP}_{\phi_k}(\bm{e}_t^\pi)), \\
q_t &= \operatorname{Softmax}(\operatorname{MLP}_{\phi_{q}}(\bm{e}_t^\pi)),
\end{aligned}
\end{equation}
where $\phi_u$, $\phi_k$, and $\phi_q$ denote the head-specific parameters. The full policy parameter set is $\phi = \{\phi_e, \phi_u, \phi_k, \phi_q\}$. 
The complete action generation process from \cref{eq:trans}--\cref{eq:a_mlp} is briefly formulated as $\bm{a}_t = \pi_{\phi}(\bm{s}_t)$.

\textbf{Value Network $\upsilon_\varphi(\bm{s}_t)$} estimates the expected cumulative return ${v}_t$ from state $\bm{s}_t$ under the current policy. 
Mirroring the policy network, our value network adopts a Transformer-based architecture that consists of a Transformer encoder backbone followed by an MLP regression head. The value estimation $\upsilon_\varphi(\bm{s}_t)$ proceeds as follows:
\begin{equation}
\label{eq:ecv}
\begin{aligned}
\bm{e}_t^{\upsilon} &= \text{Transformer}_{\varphi_e}(\bm{s}_t) \in \mathbb{R}^{D_e}, \\
{v}_t &= \operatorname{MLP}_{\varphi_v}(\bm{e}_t^{\upsilon}) \in \mathbb{R},
\end{aligned}
\end{equation}
where $\text{Transformer}_{\varphi_e}(\cdot)$ denotes the Transformer encoder parameterized by $\varphi_e$, and $\operatorname{MLP}_{\varphi_v}(\cdot)$ represents the value regression head parameterized by $\varphi_v$. 
The complete value network $\upsilon_\varphi(\bm{s}_t)$ is thus parameterized by $\varphi = \{\varphi_e, \varphi_v\}$, and the computation in \cref{eq:ecv} is compactly denoted as ${v}_t = \upsilon_\varphi(\bm{s}_t)$.

\subsubsection{Dynamic transition sampling}
In standard PPO-based algorithms,  experiences are stored in an experience buffer and uniformly sampled to update the policy and value networks. 
To enhance sampling effectiveness, PPO4Pred introduces a novel Dynamic Transition Sampling (DTS) mechanism that adaptively prioritizes informative transitions. 

Unlike conventional uniform sampling, DTS enhances sampling efficiency through a priority metric and an adaptive temperature scheduling strategy. 
The priority metric quantifies the learning potential of each transition by integrating temporal difference error and forecasting accuracy, while the temperature scheduling dynamically adjusts the sampling distribution to balance diverse exploration and high-priority selection throughout training.

\textbf{Priority Metric Design}.
During training, the agent executes actions sampled from the policy network $ \bm{a}_t = \pi_\theta(\bm{s}_t)$ in the environment, producing trajectories that consist of sequential state-action-reward tuples. 
A trajectory  $\tau =(\bm{s}_{ 1},  \ldots,\bm{s}_{ t}, \bm{a}_{ t}, r_{ t}, \bm{s}_{ t+1}, \ldots,  r_{ T})$ comprises a sequence of transitions.
We denote $\zeta_{t} = (\bm{s}_{ t}, \bm{a}_{ t}, r_{ t}, \bm{s}_{ t+1})$ as the transition at time step $t$, where $\zeta_t \in \tau$.
The trajectory can thus be expressed as  $\tau = \zeta_{1:T} = \{\zeta_t\}_{t=1}^T$.

First, we employ the Temporal Difference (TD) error to measure value estimation uncertainty. For each transition $\zeta_t$ in a trajectory $\tau$, the TD error $\delta_t$ is computed as:
\begin{equation}
\delta_t =  r_t + \gamma \cdot \upsilon_{\varphi}(\bm{s}_{t+1}) - \upsilon_{\varphi}(\bm{s}_t),
\label{eq:td_error}
\end{equation}
where $\gamma \in [0,1]$ is the discount factor, and $\upsilon_{\varphi}(\bm{s}_{T+1}) = 0$ by convention.
The magnitude $|\delta_t|$ reflects the discrepancy between the Bellman target $r_t + \gamma  \cdot \upsilon_{\varphi}(\bm{s}_{t+1})$ and the current value estimate $\upsilon_{\varphi}(\bm{s}_t)$.
Transitions with large $|\delta_t|$ are particularly informative, as they reveal regions where the value function is poorly calibrated, making them high-priority candidates for learning.
We use $|\delta_t| \in [0, \infty)$ as a priority metric, where higher values indicate greater learning potential.

Second, we introduce a forecasting error metric to identify transitions where the current architecture struggles with prediction accuracy. 
For each transition $\zeta_t$, we compute the $L_1$ forecasting error as $\epsilon_t = \|\hat{Y}_t - Y_t\|_1$ and apply a transformation:
\begin{equation}
\mathcal{E}_t = 1 - \frac{\alpha}{\alpha + \epsilon_t},
\label{eq:normalized_error}
\end{equation}
where $\alpha > 0$ is the same scaling hyperparameter in \cref{eq:reward} that controls the sensitivity to forecasting errors. 
This formulation bounds $\mathcal{E}_i \in [0, 1)$ while preserving the relative ordering of errors.
When accurate predictions $\epsilon_t \to 0$, $\mathcal{E}_t \to 0$, and when poor predictions $\epsilon_t \to \infty$, $\mathcal{E}_t \to 1$.
By prioritizing transitions with large $\mathcal{E}_t$, we focus learning on actions that currently fail to capture the underlying time series patterns, providing strong feedback for optimization.

We combine the TD error $|\delta_t|$ and forecasting error metric $\mathcal{E}_t$ into a unified priority score:
\begin{equation}
{p}_t = \beta \cdot \frac{|\delta_t|}{\delta_{\max}} + (1-\beta) \cdot \mathcal{E}_t,
\label{eq:priority}
\end{equation}
where $\delta_{\max}$ is the maximum TD error in the experience buffer, used to normalize $|\delta_t|$ to $[0,1]$.
The weighting coefficient $\beta \in [0,1]$ interpolates between value-based prioritization ($\beta=1$ is equivalent to standard prioritized experience replay) and forecasting-based prioritization ($\beta=0$ is equivalent to forecasting-only sampling), allowing flexible adaptation to different learning objectives.

After computing the priority metrics, each transition $\zeta_t$ is stored in the experience buffer $\mathcal{B}$.
Specifically, for a data batch $\mathcal{D}$ containing $|\mathcal{D}|$ time series examples, the agent collects $T$ transitions per example through policy rollouts. 
These transitions are aggregated into the experience buffer as $\mathcal{B} = \{\zeta_m\}_{m=1}^{M}$, where $M = T \times |\mathcal{D}|$ and $m \in \{1, \ldots, M\}$ indexes individual transitions in the buffer.

\textbf{Adaptive Temperature Scheduling}.
To regulate the exploration-exploitation trade-off throughout training, we introduce an adaptive temperature mechanism that modulates the sampling distribution's concentration based on transition priorities and training progress.
Unlike conventional prioritized sampling, which aggressively oversamples high-priority transitions and potentially leads to overfitting on specific transitions, our approach deliberately softens the distribution in regions of uncertainty. This encourages the agent to leverage contextual information from neighboring transitions while retaining priority-guided emphasis.
The base temperature for each transition is defined as:
\begin{equation}
\lambda_m = \lambda_{\min} + {p}_m \cdot (\lambda_{\max} - \lambda_{\min}),
\label{eq:temperature}
\end{equation}
where $\lambda_{\min}$ and $\lambda_{\max}$ are the predefined minimum and maximum temperatures, respectively. 
Higher priorities thus induce larger temperatures, softening the corresponding sampling probabilities and broadening exploration across nearby transitions with comparable importance.

Let $g  \in \{1, \ldots, G_\pi\}$ denote the current epoch for training the agent, where $G_\pi$ is the total number of epochs.
To further prevent premature convergence and enable periodic exploration resurgence, we incorporate a cyclical annealing scheme that modulates temperatures over the course of training:
\begin{equation}
\tilde{\lambda}_{g,m} = \lambda_m \cdot \left( \frac{\lambda_{\min}}{\lambda_m} \right)^{g/G_\pi} \cdot \left[ 1 + \mu \sin\left(2\pi \omega \cdot \frac{g}{G_\pi}\right) \right],
\label{eq:cyclical_temp}
\end{equation}
where $\mu$ controls the oscillation amplitude, and $\omega$ sets the cycle frequency. 
The exponential decay term $(\lambda_{\min}/\lambda_m)^{g/G_\pi}$ gradually anneals the temperature from $\lambda_m$ toward $\lambda_{\min}$ as training progresses, implementing a shift from exploration to exploitation. 
The sinusoidal term introduces periodic temperature increases that temporarily boost exploration, preventing the agent from becoming trapped in local optima. 

The final sampling probability for each transition is obtained via temperature-scaled softmax:
\begin{equation}
\tilde{p}_{g,m} = \frac{\exp\left({p}_m / \tilde{\lambda}_{g,m}\right)}{\sum_{j=1}^{M} \exp\left({p}_j / \tilde{\lambda}_{g,j}\right)}.
\label{eq:sampling_prob}
\end{equation}

In the end, we construct a sampled buffer $\check{\mathcal{B}_g}$ with $\{\zeta_{g, {\check m}}\}_{\check m=1}^{\check{M}}$ by sampling $\check{M}$ transitions from the experience buffer $\mathcal{B}$ according to \cref{eq:sampling_prob}.

The proposed DTS mechanism improves sample efficiency by prioritizing informative transitions, where the forecasting-error component supplies task-specific guidance that complements general RL feedback. In addition, it maintains an appropriate exploration-exploitation balance via adaptive temperature scheduling, which adjusts sampling behavior according to both global training progress and the intrinsic properties of each transition.

\subsection{Co-evolutionary optimization paradigm} \label{sec:Coevolutionary}

\begin{figure}[t]
    \centering
    \includegraphics[width=0.8\linewidth]{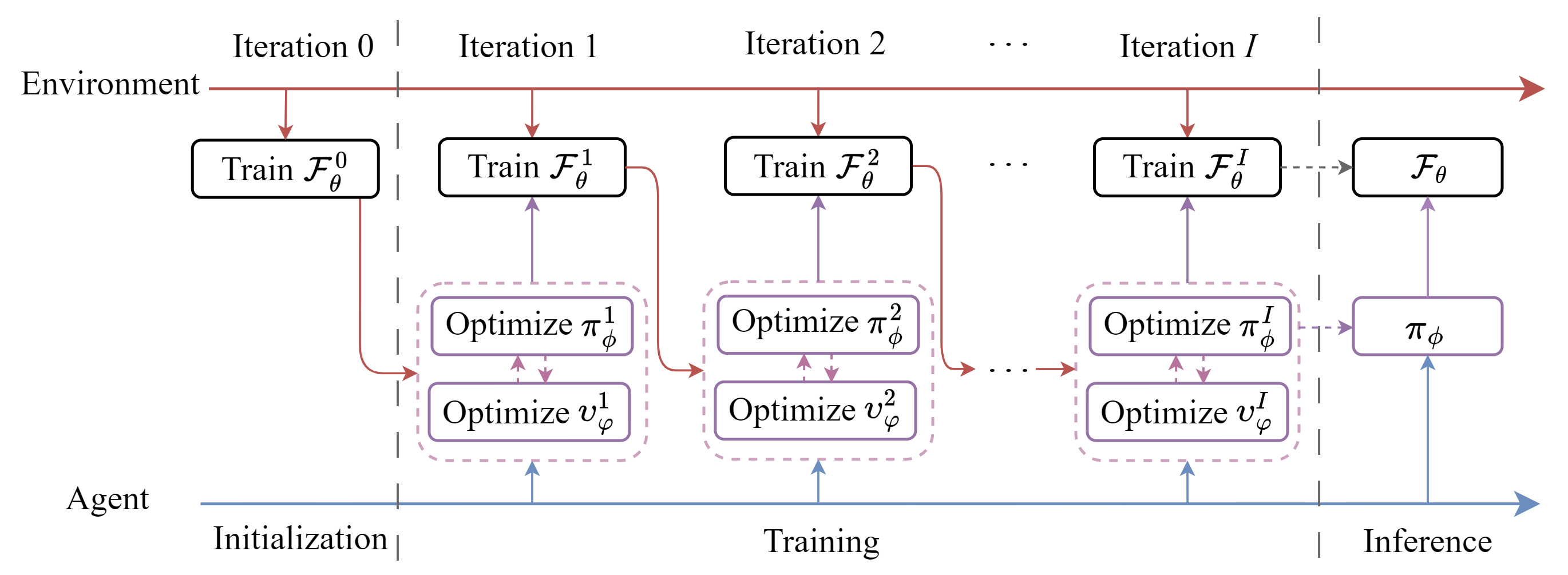}
    \caption{Illustration of asynchronous training and inference. }
    \label{fig:asyntrain}
\end{figure}
RRE-PPO4Pred implements an iterative asynchronous training paradigm to facilitate the joint evolution of the agent in PPO4Pred and RNN environment in the RRE framework.
As illustrated in \cref{fig:asyntrain}, the training procedure consists of $I$ iterations, where each iteration $i \in \{1,2,\ldots,I\}$ involves two sequential stages.
In the first stage, we update the agent, comprising policy and value networks ($\pi_{\phi}^i, \upsilon_{\varphi}^i$), while keeping the RNN environment $\mathcal{F}_{\theta}^{i-1}$ fixed.
In the second stage, we fine-tune the RNN environment $\mathcal{F}_{\theta}^i$, using $\mathcal{F}_{\theta}^{i-1}$ as initialization and the newly trained policy $\pi_{\phi}^i$ for guidance.
Upon completion of all $I$ rounds, the final RNN environment $\mathcal{F}_{\theta}^I$ is evaluated as the predictor, with actions provided by the final policy $\pi_{\phi}^I$.

\subsubsection{Training procedure}
To initialize a well-conditioned environment for policy training, we pre-train the RNN environment $\mathcal{F}_{\theta}^0$ using a conventional baseline configuration.
In this configuration, all input features and output targets are selected, while hidden skip connections are disabled\footnote{Formally, this configuration corresponds to the action trajectory $u_{1:T} = 1$, $k_{1:T} = 0$, and $q_{1:T} = 1$}.

\begin{algorithm}[t]
\caption{Experience Collection}
\label{alg:experience_collection}
\begin{algorithmic}[1]
\Require 
Data batch $\mathcal{D}$, 
old policy network $\pi_{\phi}^{i-1}$, 
value network $\upsilon_\varphi^{i-1}$, 
RNN environment $\mathcal{F}_\theta^{i-1}$,
discount factor $\gamma$, reward sensitivity $\alpha$, threshold $c$,
priority balance coefficient $\beta$.
\Ensure Replay buffer $\mathcal{B}^{i}$.
\State Initial $\mathcal{B}^{i} = \emptyset$.
\For{($X, Y$) in data batch $\mathcal{D}$}
    \For{each transition $ t = 1$ to $T$}
        \State Record the current environment state $\bm{s}_t = \text{concat}(\bm{h}_{t-1}, \bm{x}_t)$ though querying $\mathcal{F}_\theta^{i-1}$.
        \State Generate the action $\bm{a}_t = (u_t, k_t, q_t)$ via the policy network $\pi_\phi^{i-1}(\bm{s}_t)$.
        \State Output prediction $\hat{Y}_t$ with $u_t$ and $k_t$ via \cref{eq:gatea}--\cref{eq:gateh} and \cref{eq:ht}.
        \State Compute reward $r_t$ with the output $\hat{Y}_t$, target $Y_t$, sensitivity $\alpha$, threshold $c$, and $q_t$ via \cref{eq:reward}. 
        \State Record the transited environment state $\bm{s}_{t+1} = \text{concat}(\bm{h}_{t},\bm{x}_{t+1})$.
        \State Store the transition $\zeta_t = (\bm{s}_t, \bm{a}_t, r_t, \bm{s}_{t+1})$ into replay buffer: $\mathcal{B}^i = \mathcal{B}^i \cup \zeta_t$.
    \EndFor
\EndFor


\State \Return Replay buffer $\mathcal{B}^i=\{\zeta_m\}_{m=1}^{M}$.
\end{algorithmic}
\end{algorithm}

\begin{algorithm}[t]
\caption{Dynamical Transition Sampling}
\label{alg:dynamic_sampling}
\begin{algorithmic}[1]
\Require 
Replay buffer $\mathcal{B}^i = \{\zeta_m\}_{m=1}^{M}$,
value network $\upsilon_{\varphi_g}^i$,
current epoch $g$, total epochs $G_\pi$,
mini-batch size $\check{M}$,
discount factor $\gamma$,
priority balance coefficient $\beta$,
reward sensitivity $\alpha$,
temperature range $[\lambda_{\min}, \lambda_{\max}]$,
cyclical parameters $\mu$ and $\omega$.
\Ensure Sampled buffer $\check{\mathcal{B}}_g^i$.

\For{each transition $\zeta_m = (\bm{s}_m, \bm{a}_m, r_m, \bm{s}_{m+1})$ in $\mathcal{B}^i$}
    \State Compute TD error $\delta_m$ with $\upsilon_{\varphi_g}^i$, $r_m$, $\bm{s}_m$, and $\bm{s}_{m+1}$ via \cref{eq:td_error}.
    \State Compute normalized forecasting error $\mathcal{E}_m$ with $\alpha$, $\hat{Y}_m$ and $Y_m$ via \cref{eq:normalized_error}.
\EndFor

\State Compute $\delta_{\max} = \max\{|\delta_m|\}_{m=1}^{M}$.

\For{each transition $\zeta_m$ in $\mathcal{B}^i$}
    \State Compute priority score $p_m$ with $\beta$ $\delta_m$, $\delta_{\max}$, and $\mathcal{E}_m$ via \cref{eq:priority}.
    \State Compute base temperature $\lambda_m$ with $p_m$, $\lambda_{\min}$, and $\lambda_{\max}$ via \cref{eq:temperature}.
    \State Compute adaptive temperature $\tilde{\lambda}_{g,m}$ with $\lambda_m$, $g$, $G_\pi$, $\mu$, and $\omega$ via \cref{eq:cyclical_temp}.
    
\EndFor

\State Compute sampling probability $\tilde{p}_{g,m}$ with $p_m$ and $\tilde{\lambda}_{g,m}$ via \cref{eq:sampling_prob}.

\State Sample $\check{M}$ transitions from $\mathcal{B}^i$ according to distribution $\{\tilde{p}_{g,m}\}_{m=1}^{M}$ to craft $\check{\mathcal{B}}_g^i$.

\State \Return Sampled buffer $\check{\mathcal{B}}_g^i = \{\zeta_{\check m}\}_{\check m=1}^{\check{M}}$.

\end{algorithmic}
\end{algorithm}

In the $i$-th iteration of agent training, where $\pi_{\phi}^i$ and $\upsilon_{\varphi}^i$ are optimized, we leverage the preceding policy $\pi_\phi^{i-1}$ to interact with the RNN environment and accumulate experiences, with $\pi_\phi^{0}$ and $\upsilon_{\varphi}^0$ randomly initialized.
For each time series example $(X, Y)$ in a data batch $\mathcal{D}$, the agent executes the trajectory generation as introduced in \cref{sec:reenc}. 
Specifically, the agent engages with the frozen RNN environment $\mathcal{F}_\theta^{i-1}$ to construct a trajectory $\tau = \{\zeta_{t}\}_{t=1}^{T}$ by sequentially making actions $\bm{a}_t $ at each time step $t$.
For a data batch encompassing multiple time series examples, we aggregate these transitions per example and populate the experience buffer $\mathcal{B}^i = \{\zeta_m\}_{m=1}^{M}$. 
The full experience collection procedure is summarized in \cref{alg:experience_collection}.

Subsequently, we proceed to the agent optimization stage, which entails an inner loop of $G_\pi$ epochs within the current round $i$.
For the $g$-th epoch of this optimization phase, 
let $\pi_{\phi_g}^i$ and $\upsilon_{\varphi_g}^i$ denote the policy network and value network at epoch $g$ of round $i$, respectively.
The agent at round $i$ is initialized by inheriting the weight parameters from the converged agent at round $i-1$, formally expressed as $\pi_{\phi_1}^i \leftarrow \pi_{\phi}^{i-1}$ and $\upsilon_{\varphi_1}^i \leftarrow \upsilon_{\varphi}^{i-1}$.
As illustrated in \cref{alg:dynamic_sampling}, the sampling probability $\tilde{p}_{g,m}^i$ for each transition is calculated through  \cref{eq:td_error}--\cref{eq:sampling_prob} utilizing the current value network $\upsilon_{\varphi_g}^i$, generating a dynamic sampling distribution $\{\tilde{p}_{g,m}^i\}_{m=1}^M$ over the experience buffer $\mathcal{B}^i$.
A sampled buffer $\check{\mathcal{B}}_g^i = \{\zeta_{g, \check m}\}_{\check m=1}^{\check M}$ of  size $\check{M}$ is subsequently instantiated through sampling from this distribution.

For the $\check m$-th transition $\zeta_{\check m}$ in $\check{\mathcal{B}_g^i}$, we compute the advantage $A^i_{g,\check m}$ using Generalized Advantage Estimation (GAE) \citep{schulman2015high} with the GAE parameter set to zero, which reduces to the one-step TD error:
\begin{equation}
{A}_{g,\check m}^i = \delta_{g,\check m}^i = r_{\check m} + \gamma \upsilon_{\varphi_{g}}^i(\bm{s}_{{\check m}+1})-\upsilon_{\varphi_{g}}^i(\bm{s}_{\check m}),
\label{eq:advance}
\end{equation}
where $r_{\check m}$, $\bm{s}_{\check m}$ and $\bm{s}_{\check m+1}$ are components of the transition $\zeta_{\check m}$. 
The policy network $\pi^i_{\phi_g}$ is optimized by maximizing the clipped surrogate objective:
\begin{equation}
   \max \mathcal{L}(\phi_g)=\max_{\phi_g} \mathbb{E}_{\check{\mathcal{B}_g^i}}\left[\min\Bigl(\rho(\phi_g){A}_{g,\check m}^i,\,\operatorname{clip}\bigl(\rho(\phi_g),1-\varepsilon,1+\varepsilon\bigr){A}_{g,\check m}^i\Bigr)\right],
   \label{eq:policy_loss}
\end{equation}
where $\rho(\phi_{g})=\pi_{\phi_{g}}^i(\bm{s}_{\check m})/\pi_{\phi}^{i-1}(\bm{s}_{\check m})$ denotes the probability ratio between the current $g$-th epoch policy network $\pi_{\phi_{g}}^i$ and old policy network $\pi_{\phi}^{i-1}$.  
The parameter $\varepsilon$ denotes the clipping threshold that limits the magnitude of policy updates to promote training stability.
The operator $\mathbb{E}_{\check{\mathcal{B}_g^i}}[\cdot]$ represents the expectation over transitions in $\check{\mathcal{B}_g^i}$.

The learning objective of the value network $\upsilon^i_{\varphi_g}$ is to minimize the discrepancy between its predicted value $\upsilon^i_{\varphi_g}(\bm{s}_{\check m})$ and the actual return $R^i_{g,\check m}$, thereby reducing the mean-squared Bellman error and stabilizing the advantage estimates used in \cref{eq:policy_loss}. 
The expected return $R^i_{g,\check m}$ is computed using temporal difference learning as:
\begin{equation}
{R}_{g,\check m}^i = r_{\check m} + \sum_{l=1}^{T-t} \gamma^l r_{t+l} \approx r_{\check m} + \gamma \upsilon_{\varphi_{g}}^i(\bm{s}_{{\check m}+1}),
\label{eq:return}
\end{equation}
where the cumulative discounted future return $\sum_{l=1}^{T-t} \gamma^l r_{t+l}$ is approximated via bootstrapping from the value estimate $\upsilon_{\varphi_{g}}^i(\bm{s}_{\check m+1})$ at the successor state.
Formally, the value network $\upsilon_{\varphi_g}$ is optimized by minimizing the MSE loss between the predicted and bootstrapped returns:
\begin{equation}
\min \mathcal{L}(\varphi_g) = \min_{\varphi_g} \mathbb{E}_{\check{\mathcal{B}^i_g}}\left[\|\upsilon_{\varphi_g}^i(\bm{s}_{\check m}) - {R}_{g,\check m}^i \|_2^2\right].
\label{eq:critic_loss}
\end{equation}
Here, $\|\cdot\|_2^2$ denotes the squared  $L_2$  norm, which corresponds to the MSE between the predicted value and the bootstrapped return.

At the end of each epoch $g$, the policy and value network parameters are updated via gradient ascent and descent, respectively:
\begin{equation}
\begin{aligned} \label{eq:agentupdate}
\phi_{g+1}   &= \phi_g + \eta_\pi \nabla_{\phi_g} \mathcal{L}(\phi_g), \\
\varphi_{g+1}   &= \varphi_g - \eta_\upsilon \nabla_{\varphi_g} \mathcal{L}(\varphi_g), \\
\end{aligned}
\end{equation}
where $\nabla_{\phi_g} \mathcal{L}(\phi_g)$ and $\nabla_{\varphi_g} \mathcal{L}(\varphi_g)$ denote the gradients of the respective loss functions with respect to the network parameters. 
$\eta_\pi$ and $\eta_\upsilon$ are the learning rates of policy network and value network, respectively.
After completing $G_\pi$ epochs of training, we obtain the final agent for round $i$ as: $\pi^{i}_\phi \leftarrow \pi^{i}_{\phi_{G_\pi}}$ and $\upsilon^{i}_\varphi \leftarrow \upsilon^{i}_{\varphi_{G_\pi}}$.

After that, we step into the environment optimization stage. We fine-tune the RNN environment $\mathcal{F}_{\theta}^i \leftarrow \mathcal{F}_{\theta}^{i-1}$ with Eq.~\ref{eq:gatey}, where the actions  $\{u_{1:T},k_{1:T},q_{1:T}\}$ are produced by the optimized policy network $\pi^{i}_\phi$.
This ensures mutual adaptation between the encoder-only RNN predictor and the policy agent.
We perform this joint agent-environment evolution for $I$ iterations, facilitating interaction between the RNN environment and the agent.
Upon convergence, the final reinforced encoder-only RNN predictor is instantiated as $\mathcal{F}_{\theta} \leftarrow \mathcal{F}_{\theta}^I$ with its corresponding policy $\pi_\phi \leftarrow \pi^{I}_\phi$.
Comprehensive algorithmic details for implementing this procedure with a complete dataset comprising multiple data batches are provided in \cref{alg:training} in the \ref{sec:para}.

\subsubsection{Inference procedure}
During inference, unlike conventional heuristic methods that rely on fixed, manually specified architectural configurations applied uniformly to all input time series, our approach enables a dynamic interaction between the learned policy network $\pi_\phi$ and the RNN-based predictor $\mathcal{F}_\theta$. Instead of following predetermined rules, the agent adaptively selects actions conditioned on the evolving state at each time step.

\begin{algorithm}[t] 
\caption{Model Inference with Trained Policy}
\label{alg:inference}
\begin{algorithmic}[1]
\Require Trained policy network $\pi_\phi$, 
trained RNN $\mathcal{F}_\theta$, unseen input time series $X$.
\Ensure Prediction value $\hat{Y}$.
\State Initialize the hidden state $\h_0 = \bm{0}$.
\For{each $\bm{x}_t$ in the input time series $X$}
    \State Craft the current environment state $\bm{s}_{t} = \text{concat}(\bm{h}_{t-1}, \bm{x}_t)$.
    \State Generate the action $\bm{a}_t = (u_t, k_t, q_t)$ via the policy network $\pi_\phi(\bm{s}_t)$.
    \State Use $\x_t$, $u_t$, $k_t$, $\h_{t-1}$, and $\h_{t-{k_t}-1}$ to produce the hidden state $\h_t$ with $\mathcal{F}_\theta$ via \cref{eq:gateh}.
\EndFor
\State Generate the prediction at the final time step $\hat{Y}_T$ with $\h_T$ via \cref{eq:ht}

\State \Return Final prediction value $\hat{Y}_T$

\end{algorithmic}
\end{algorithm}

Given an unseen input time series ${X}$ from the test set, the inference procedure is executed as illustrated in \cref{alg:inference}.
At each time step $t$, we construct the current state representation $\bm{s}_t$ via concatenating the previous hidden state $\bm{h}_{t-1}$ and the current input $\bm{x}_t$. This state is processed by the trained policy network $\pi_\phi$, yielding a ternary action $\bm{a}_t = (u_t, k_t, q_t)$ that specifies the architectural configurations for time step $t$.
Since the inference stage precludes model training, the output target selection action $q_t$ is discarded, as it serves exclusively to guide training dynamics.
Conversely, the input feature selection action $u_t$ and hidden skip connection action $k_t$ are retained to adaptively modulate the hidden feature representation as defined in \cref{eq:gateh}.
Conditioned on the actions $u_{1:T}$ and $k_{1:T}$, the encoder-only RNN model generates the terminal hidden state $\bm{h}_T$ and produces the output $\hat{Y}_T$ as the final prediction $\hat{Y}$.

Remarkably, the policy network is engaged not only during training but also at each recurrent transition step during inference, where it dynamically modulates the feature representation conditioned on the evolving hidden state and inputs. This stands in sharp contrast to existing heuristic methods that rely on static, predetermined architectures at inference time. By adaptively reconfiguring the predictor through the learned policy, our framework preserves architectural flexibility, enabling it to better accommodate heterogeneous temporal patterns and varying dynamics across diverse time series, thereby improving both forecasting accuracy and robustness on unseen data.
\section{Experimental setup}\label{setup}
\subsection{Datasets}

To evaluate the performance of our approach, we perform extensive experiments on a number of real-world public benchmark datasets, including Traffic, Electricity, ETTh, Weather and ILI datasets. The summary information of these multivariate time series datasets is summarized in \cref{tab:dataset}, which includes  sampling frequency, number of observed steps, number of variables ($D_{in}$), input lag order ($T$), and forecasting horizons ($H$) for each dataset.  Detailed descriptions of each dataset are provided below:
\begin{table}[t]
\centering
\caption{The summary information of time series datasets.}
\label{tab:dataset}
\begin{tabular}{lccccc}
\toprule
Information & Traffic & Electricity & ETTh & Weather & ILI \\ \midrule
Sampling frequency & 1 hour & 1 hour & 1 hour & 10 mins & 1 week \\
Number of observed steps & 17,544 & 26,304 & 17,420 & 52,696 & 966 \\
Number of variables $D_{in}$  & 862 & 321 & 7 & 21 & 7 \\
Input lag order $T$  & 96 & 96 & 96 & 96 & 48 \\
Forecasting horizons $H$ & [24, 48, 96] & [24, 48, 96] & [24, 48, 96] & [24, 48, 96] & [12, 24, 48] \\ \bottomrule
\end{tabular}
\end{table}

\begin{itemize}
    \item The Traffic dataset\footnote{https://pems.dot.ca.gov/}, provided by the California Department of Transportation via the Caltrans Performance Measurement System, contains hourly road occupancy measurements from 862 highway sensors in the San Francisco Bay Area, comprising 17,544 time steps.
    \item The Electricity consumption load dataset\footnote{https://archive.ics.uci.edu/ml/datasets/ElectricityLoadDiagrams20112014} (Electricity) records hourly electricity usage of 321 customers from 2012 to 2014, totaling 26,304 time steps. 
    \item The Electricity Transformers Temperature (ETTh) dataset\footnote{https://github.com/zhouhaoyi/ETDataset} contains hourly records collected over two years from an electricity substation. It includes load and oil temperature measurements across seven variables, comprising 17,420 time steps.
    \item The Weather dataset\footnote{https://www.bgc-jena.mpg.de/wetter}, provided by the Max Planck Institute for Biogeochemistry, comprises four years of meteorological data from the United States. It includes 21 variables such as air temperature and humidity, totaling 52,696 time steps.
    \item The Influenza-Like Illness (ILI) dataset\footnote{https://gis.cdc.gov/grasp/fluview/fluportaldashboard.html}, maintained by the U.S. Centers for Disease Control and Prevention, aggregates weekly patient visit data from 2002 to 2021, comprising 966 time steps.
\end{itemize}

\subsection{Metrics}
To evaluate the predictive performance of all selected methods, following prior work \citep{kong2025deep}, we employ two widely-used statistical measures: Mean Squared Error (MSE) and Mean Absolute Error (MAE). 
MSE quantifies the average of the squared differences between predicted and true values, thereby emphasizing larger errors through the squaring operation. This property makes MSE particularly sensitive to substantial deviations and outliers in model predictions.
MAE calculates the average of the absolute differences between predicted and actual values, providing a more direct metric that penalizes errors linearly.

Note that we perform multivariate-to-univariate forecasting, where multivariate inputs ($D_{in} > 1$) are used to predict a single endogenous target variable ($D_{out}=1$) over the future horizon $H$.
Formally, MSE and MAE are defined as:
\begin{equation}
   MSE=\frac{1}{H} \|\hat{Y} - Y\|_2^2
\end{equation}
\begin{equation}
    MAE=\frac{1}{H} \|\hat{Y} - Y\|_1  
\end{equation}
where $\hat{Y} \in \mathbb{R}^H$ denotes the predicted values and $Y \in \mathbb{R}^H$ denotes the corresponding ground truth values over the forecast horizon $H$.

\subsection{Backbones and Baselines}

To evaluate both the structural generality and improvement effectiveness of the proposed RRE framework and PPO4Pred algorithm, comprehensive experiments are conducted across different RNN backbones and optimization baselines. 
The experiments on RNN backbones examine whether our method can consistently improve performance across diverse RNN architectures, thereby demonstrating its applicability to heterogeneous encoder-only RNNs.
The experiments with optimization baselines evaluate the optimization effectiveness of our PPO4Pred by comparing its performance against a comprehensive suite of baseline optimization strategies, encompassing fixed configurations, heuristic search methods, and alternative reinforcement learning algorithms.

\subsubsection{RNN backbones}
\begin{table}[t]
\centering
\caption{Summary of representative RNN-based architectures used as backbones.}
\label{tab:rnn_variants}
\begin{tabular}{lcl}
\toprule
Model & Abbr. & Key Mechanism \\ 
\midrule
Vanilla RNN \citep{elman1990finding} & RNN & Vanilla recurrent neural network models. \\
Minimal Gated Unit \citep{zhou2016minimal} & MGU & A variant with a single update gate. \\
Gated Recurrent Unit \citep{cho2014learning} & GRU & A variant with reset and update gates. \\
Long Short-Term Memory \citep{hochreiter1997long} & LSTM & A variant with input, forget, and output gates. \\
Independently RNN \citep{Indrnn} & IndRNN & Eliminating inter-neuron connection within a layer. \\
Peephole LSTM \citep{gers2000learning} & phLSTM & Allowing gate units to access the cell state directly. \\
Phased LSTM \citep{neil2016phased} & pLSTM & Introducing a time gate controlled by   oscillation. \\
eXtended LSTM \citep{beck2024xlstm} & xLSTM & Expanding gating and memory mechanisms. \\
\bottomrule
\end{tabular}
\end{table}
\begin{figure}[t]
    \centering
    \begin{subfigure}[b]{0.3\textwidth}
        \centering
        \includegraphics[width=\textwidth]{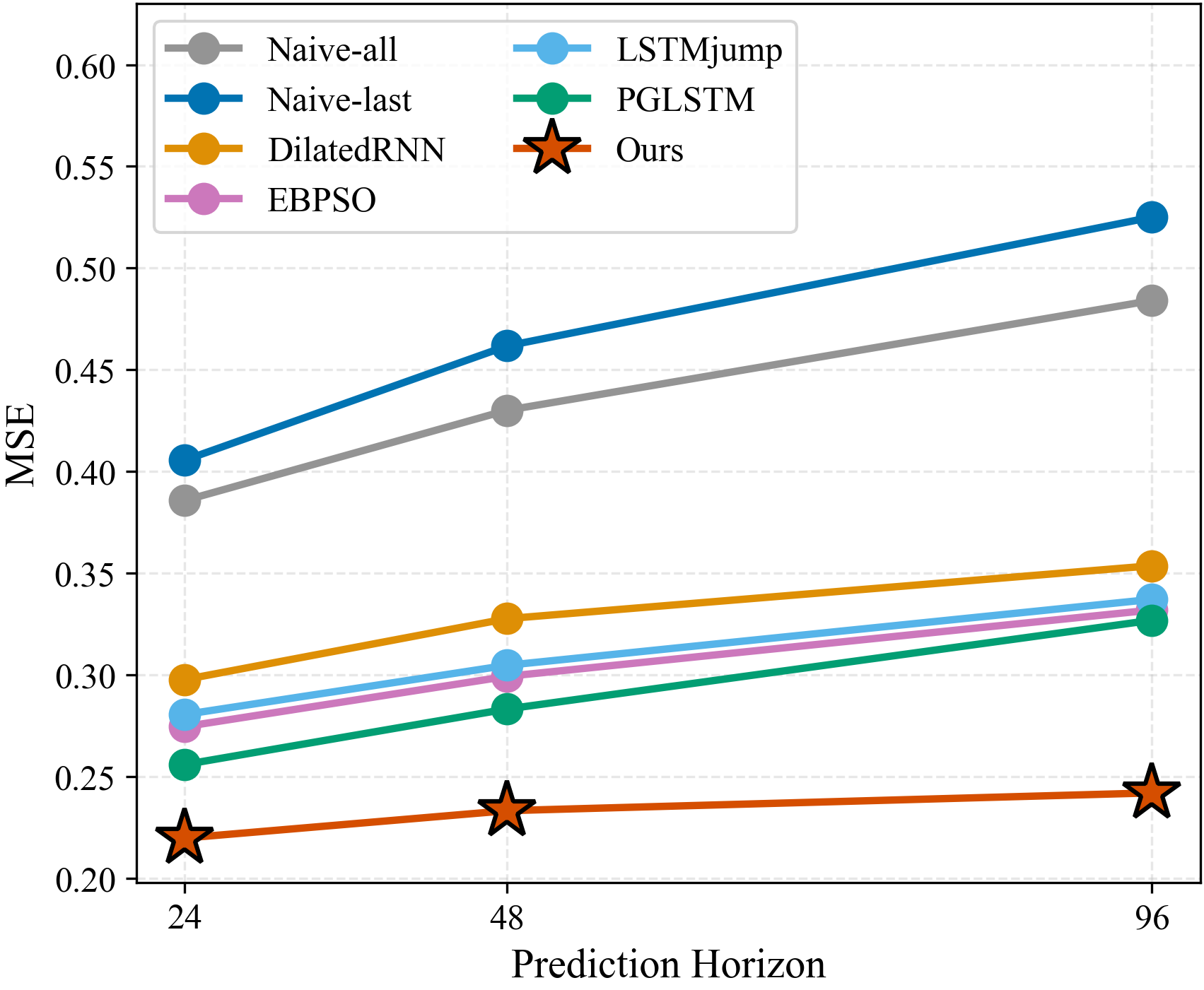} 
        \caption{Traffic  }
    \end{subfigure}
    \begin{subfigure}[b]{0.3\textwidth}
        \centering
        \includegraphics[width=\textwidth]{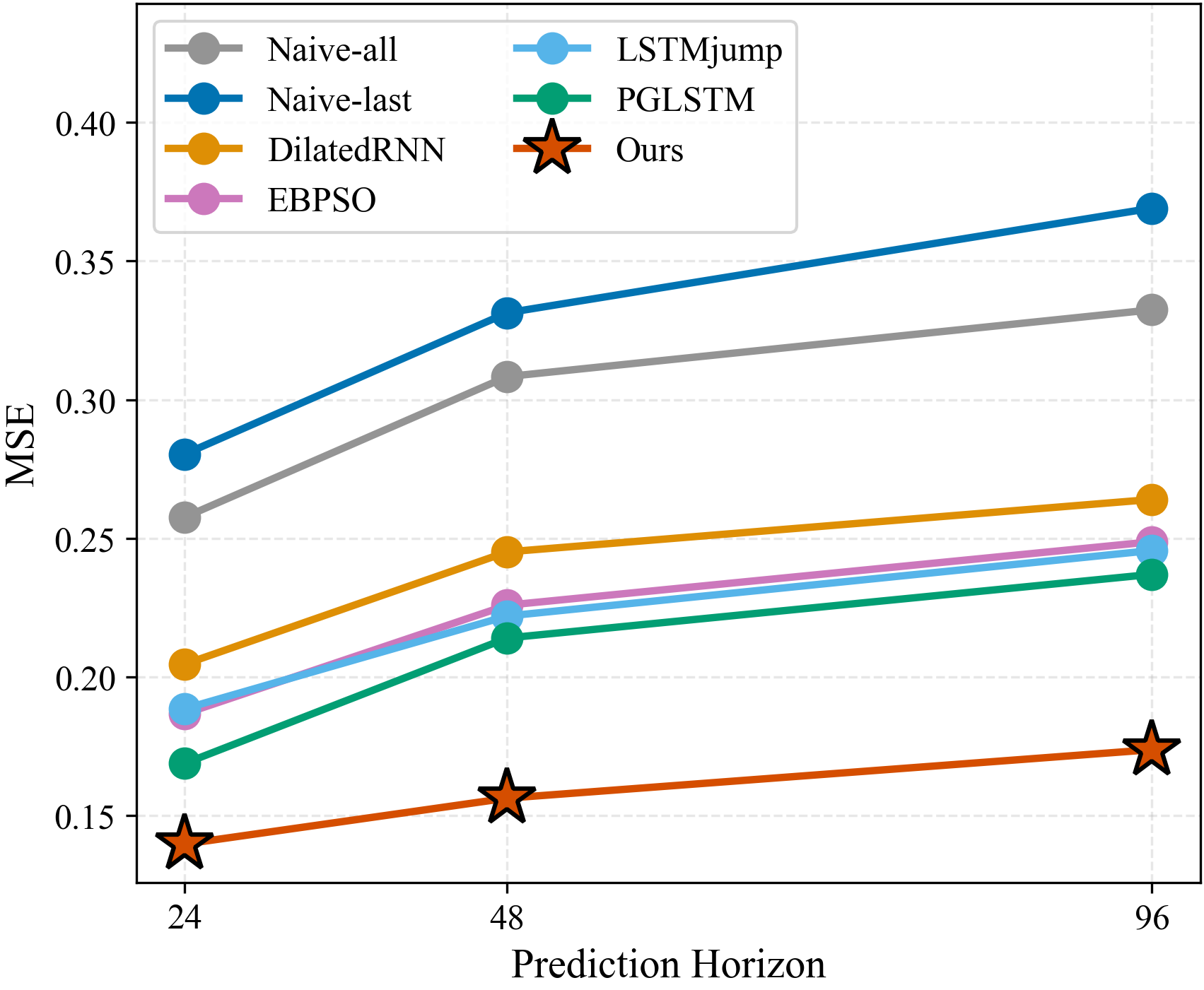} 
        \caption{Electricity  }
    \end{subfigure}
    \begin{subfigure}[b]{0.3\textwidth}
        \centering
        \includegraphics[width=\textwidth]{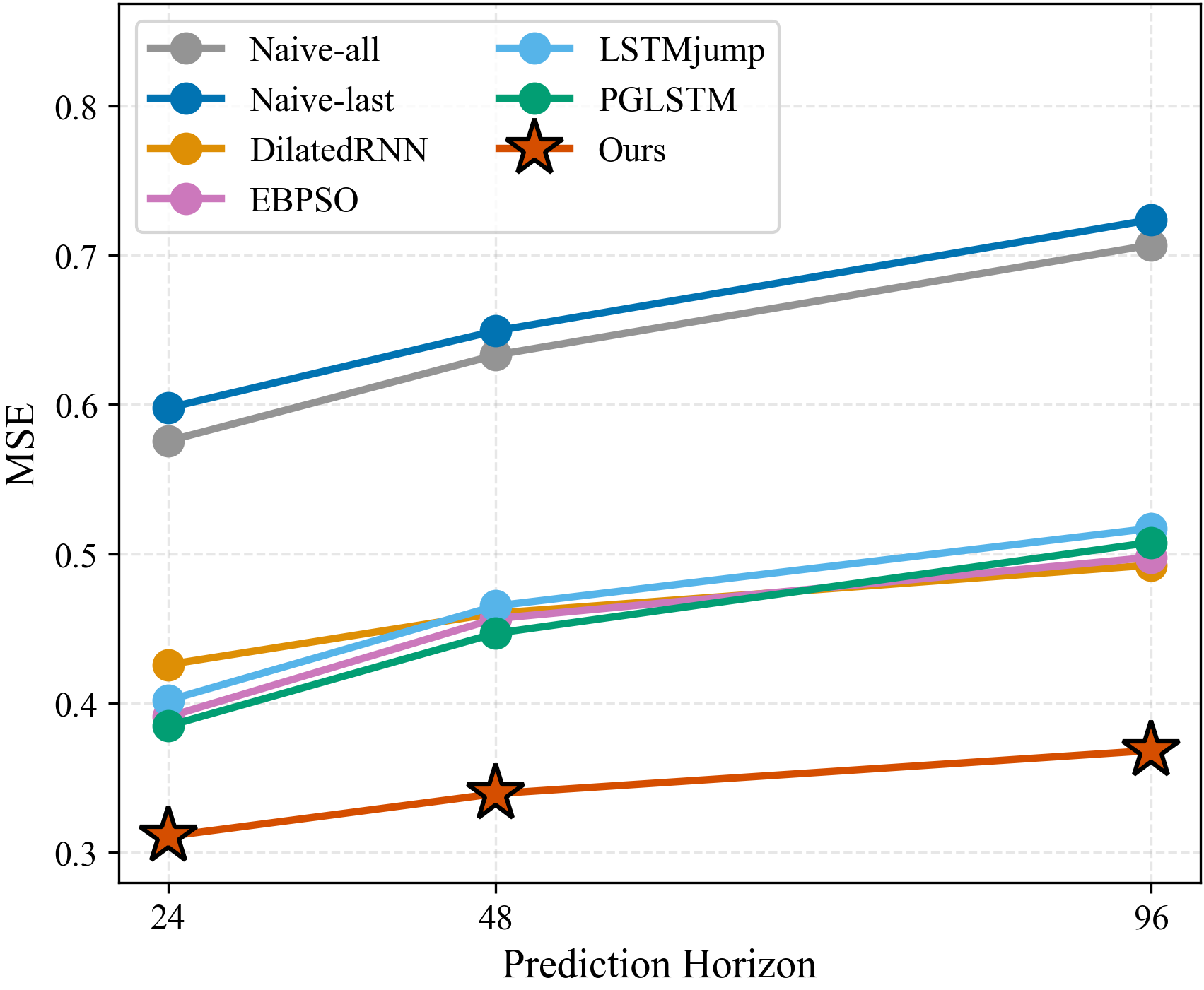} 
        \caption{ETTh  }
    \end{subfigure}
    \hfill 
    \vspace{3mm}
    \begin{subfigure}[b]{0.3\textwidth}
        \centering
        \includegraphics[width=\textwidth]{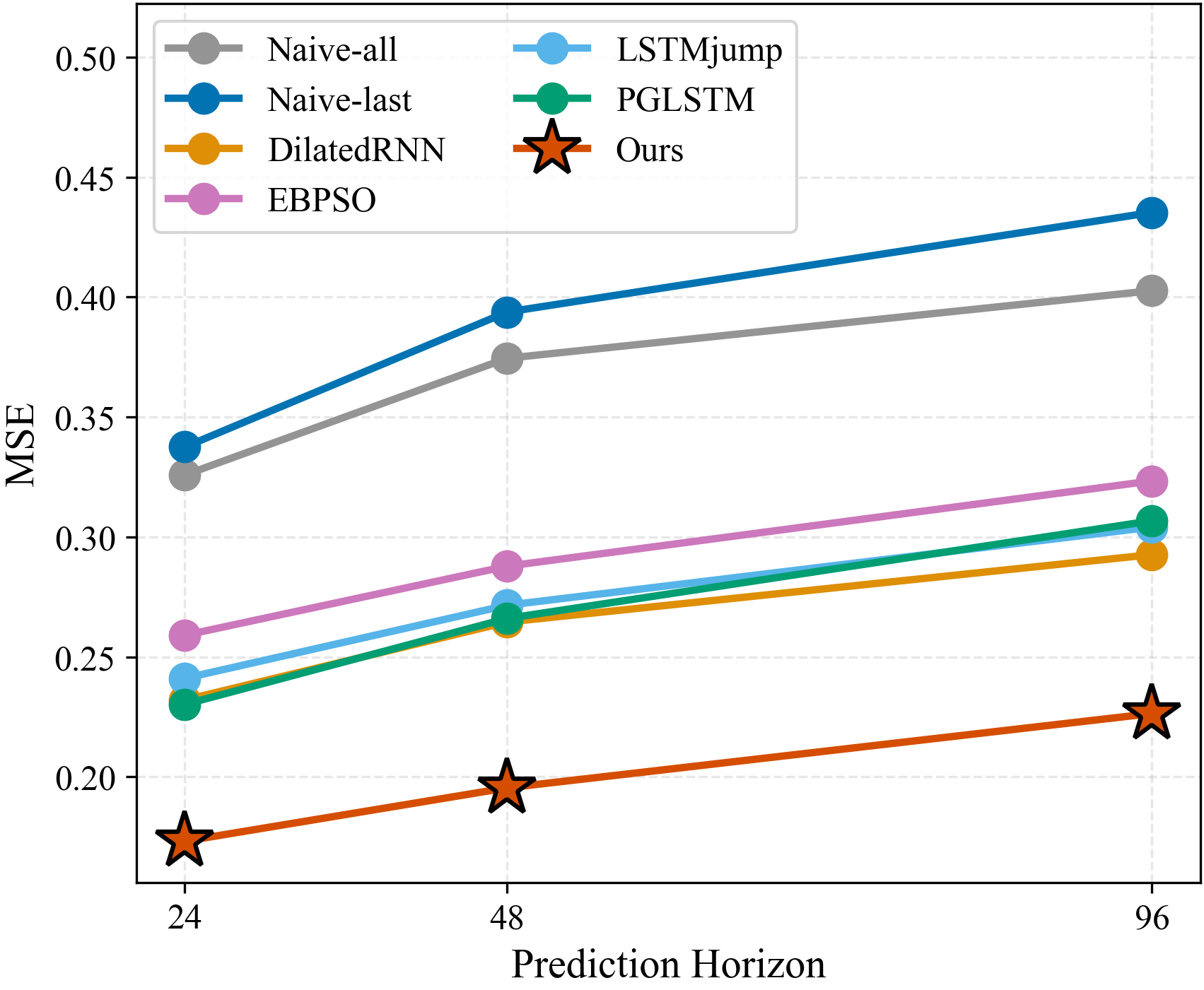} 
        \caption{Weather  }
    \end{subfigure}
    \begin{subfigure}[b]{0.3\textwidth}
        \centering
        \includegraphics[width=\textwidth]{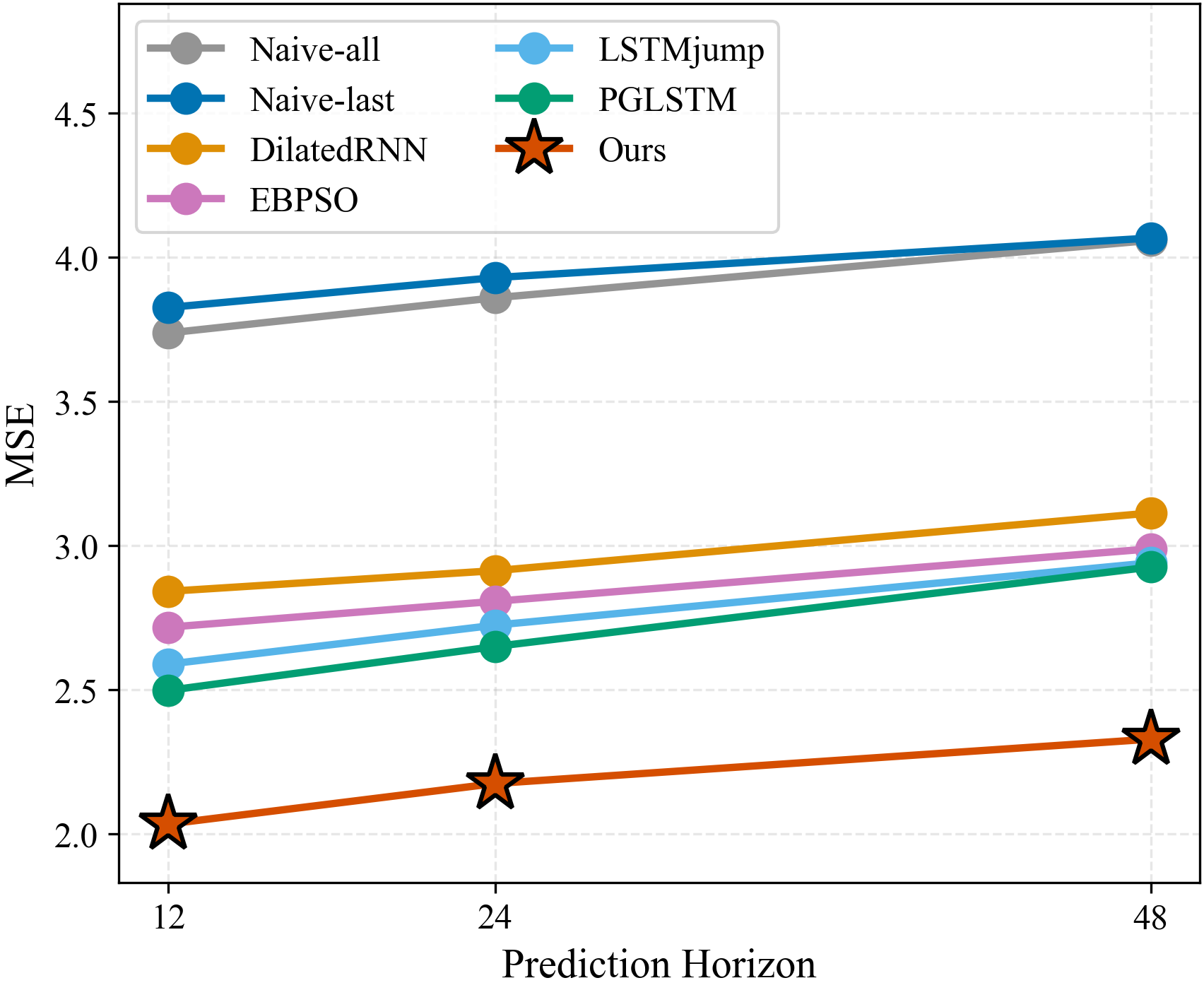} 
        \caption{ILI  }
    \end{subfigure}
    
    \caption{Comparison of MSE averaged across all backbone architectures for the proposed RRE-PPO4Pred method and baseline approaches at varying prediction horizons.}
    \label{fig:horison-mse}
\end{figure}

\begin{table}[t]
\centering
\caption{Average MSE and MAE of our RRE-PPO4Pred and baseline methods on the Traffic dataset.}
\label{tab:traffic-performance}
\resizebox{\textwidth}{!}{%
\setlength{\tabcolsep}{3.5pt} 
\begin{tabular}{@{}cc|ccccccc|ccccccc@{}}
\toprule
\multirow{2}{*}{$H$} &
  \multirow{2}{*}{Backbone} &
  \multicolumn{7}{c|}{MSE} &
  \multicolumn{7}{c}{MAE} \\ \cmidrule(l){3-16} 
 &
   &
  Naive-all &
  Naive-last &
  DilatedRNN &
  EBPSO &
  LSTMjump &
  PGLSTM &
  Ours  &
  Naive-all &
  Naive-last &
  DilatedRNN &
  EBPSO &
  LSTMjump &
  PGLSTM &
  Ours  \\ \midrule
\multirow{8}{*}{24} &
  RNN &
  0.3778 &
  0.3871 &
  0.2806 &
  0.2747 &
  0.2841 &
  0.2561 &
  \textbf{0.2251} &
  0.4168 &
  0.5094 &
  0.4159 &
  0.3568 &
  0.3329 &
  0.3062 &
  \textbf{0.2640} \\
 &
  MGU &
  0.3844 &
  0.4053 &
  0.2850 &
  0.2681 &
  0.2735 &
  0.2569 &
  \textbf{0.2202} &
  0.4797 &
  0.5160 &
  0.4073 &
  0.3346 &
  0.3328 &
  0.3221 &
  \textbf{0.2667} \\
 &
  GRU &
  0.3802 &
  0.3914 &
  0.2781 &
  0.2661 &
  0.2777 &
  0.2590 &
  \textbf{0.2108} &
  0.4589 &
  0.4742 &
  0.4214 &
  0.3191 &
  0.3169 &
  0.2932 &
  \textbf{0.2699} \\
 &
  LSTM &
  0.3999 &
  0.3985 &
  0.2714 &
  0.2696 &
  0.2785 &
  0.2603 &
  \textbf{0.2129} &
  0.4794 &
  0.4777 &
  0.3998 &
  0.3320 &
  0.3148 &
  0.2935 &
  \textbf{0.2582} \\
 &
  IndRNN &
  0.3794 &
  0.4199 &
  0.2774 &
  0.2806 &
  0.2660 &
  0.2599 &
  \textbf{0.2227} &
  0.4236 &
  0.4783 &
  0.4204 &
  0.3443 &
  0.3334 &
  0.3059 &
  \textbf{0.2734} \\
 &
  phLSTM &
  0.4082 &
  0.4620 &
  0.2884 &
  0.2888 &
  0.2924 &
  0.2509 &
  \textbf{0.2321} &
  0.4561 &
  0.4720 &
  0.4118 &
  0.3720 &
  0.3174 &
  0.3114 &
  \textbf{0.2555} \\
 &
  pLSTM &
  0.3775 &
  0.3861 &
  0.2669 &
  0.2828 &
  0.2915 &
  0.2583 &
  \textbf{0.2159} &
  0.4254 &
  0.5107 &
  0.3907 &
  0.3399 &
  0.3391 &
  0.3209 &
  \textbf{0.2657} \\
 &
  xLSTM &
  0.3785 &
  0.3944 &
  0.2737 &
  0.2656 &
  0.2602 &
  0.2457 &
  \textbf{0.2184} &
  0.4162 &
  0.4411 &
  0.3899 &
  0.3615 &
  0.3184 &
  0.2774 &
  \textbf{0.2463} \\ \midrule
\multirow{8}{*}{48} &
  RNN &
  0.4108 &
  0.4079 &
  0.2950 &
  0.2895 &
  0.2944 &
  0.2822 &
  \textbf{0.2249} &
  0.5468 &
  0.5306 &
  0.4272 &
  0.4178 &
  0.3759 &
  0.3468 &
  \textbf{0.2791} \\
 &
  MGU &
  0.4463 &
  0.4392 &
  0.3195 &
  0.2853 &
  0.3101 &
  0.2780 &
  \textbf{0.2397} &
  0.5009 &
  0.5855 &
  0.4335 &
  0.4079 &
  0.3902 &
  0.3640 &
  \textbf{0.2824} \\
 &
  GRU &
  0.4372 &
  0.4218 &
  0.3072 &
  0.2892 &
  0.3091 &
  0.2916 &
  \textbf{0.2280} &
  0.4873 &
  0.5731 &
  0.4478 &
  0.4159 &
  0.3781 &
  0.3662 &
  \textbf{0.2784} \\
 &
  LSTM &
  0.4392 &
  0.4284 &
  0.3126 &
  0.3015 &
  0.2996 &
  0.2784 &
  \textbf{0.2369} &
  0.5638 &
  0.6568 &
  0.4327 &
  0.3848 &
  0.3800 &
  0.3413 &
  \textbf{0.2850} \\
 &
  IndRNN &
  0.4220 &
  0.4478 &
  0.3021 &
  0.3075 &
  0.3135 &
  0.2770 &
  \textbf{0.2376} &
  0.5434 &
  0.6055 &
  0.4518 &
  0.3901 &
  0.3981 &
  0.3715 &
  \textbf{0.2707} \\
 &
  phLSTM &
  0.4385 &
  0.4802 &
  0.3459 &
  0.3155 &
  0.3148 &
  0.2876 &
  \textbf{0.2380} &
  0.5984 &
  0.6556 &
  0.4408 &
  0.4120 &
  0.3915 &
  0.3670 &
  \textbf{0.2723} \\
 &
  pLSTM &
  0.4318 &
  0.4346 &
  0.3258 &
  0.3121 &
  0.3116 &
  0.2894 &
  \textbf{0.2354} &
  0.5520 &
  0.6646 &
  0.4210 &
  0.4067 &
  0.4027 &
  0.3694 &
  \textbf{0.2813} \\
 &
  xLSTM &
  0.4150 &
  0.4215 &
  0.3328 &
  0.2937 &
  0.2846 &
  0.2814 &
  \textbf{0.2252} &
  0.5144 &
  0.5707 &
  0.4116 &
  0.3972 &
  0.3656 &
  0.3429 &
  \textbf{0.2644} \\ \midrule
\multirow{8}{*}{96} &
  RNN &
  0.4525 &
  0.5714 &
  0.3142 &
  0.3429 &
  0.3294 &
  0.3180 &
  \textbf{0.2380} &
  0.5450 &
  0.5698 &
  0.4554 &
  0.4500 &
  0.4118 &
  0.3745 &
  \textbf{0.3090} \\
 &
  MGU &
  0.5005 &
  0.5639 &
  0.3039 &
  0.3278 &
  0.3261 &
  0.3274 &
  \textbf{0.2449} &
  0.5597 &
  0.5998 &
  0.4189 &
  0.4777 &
  0.4175 &
  0.3833 &
  \textbf{0.3000} \\
 &
  GRU &
  0.4718 &
  0.5621 &
  0.3130 &
  0.3408 &
  0.3372 &
  0.3265 &
  \textbf{0.2460} &
  0.4887 &
  0.5324 &
  0.4316 &
  0.4380 &
  0.4008 &
  0.3820 &
  \textbf{0.3197} \\
 &
  LSTM &
  0.4890 &
  0.5172 &
  0.3411 &
  0.3342 &
  0.3427 &
  0.3298 &
  \textbf{0.2446} &
  0.5234 &
  0.6469 &
  0.4662 &
  0.4330 &
  0.4219 &
  0.3799 &
  \textbf{0.3045} \\
 &
  IndRNN &
  0.5000 &
  0.4734 &
  0.3060 &
  0.3461 &
  0.3533 &
  0.3343 &
  \textbf{0.2442} &
  0.5498 &
  0.5896 &
  0.4905 &
  0.4362 &
  0.4086 &
  0.3856 &
  \textbf{0.3065} \\
 &
  phLSTM &
  0.4706 &
  0.4818 &
  0.3167 &
  0.3326 &
  0.3325 &
  0.3275 &
  \textbf{0.2357} &
  0.6196 &
  0.7628 &
  0.5437 &
  0.4605 &
  0.4167 &
  0.3890 &
  \textbf{0.3092} \\
 &
  pLSTM &
  0.4875 &
  0.5024 &
  0.3304 &
  0.3421 &
  0.3266 &
  0.3244 &
  \textbf{0.2395} &
  0.6169 &
  0.6364 &
  0.4705 &
  0.4599 &
  0.4312 &
  0.4117 &
  \textbf{0.2989} \\
 &
  xLSTM &
  0.5007 &
  0.5282 &
  0.3340 &
  0.3437 &
  0.3492 &
  0.3267 &
  \textbf{0.2427} &
  0.5883 &
  0.6332 &
  0.4423 &
  0.4309 &
  0.3612 &
  0.3828 &
  \textbf{0.2827} \\ \midrule
\multicolumn{2}{c|}{Average} &
  0.4333 &
  0.4553 &
  0.3051 &
  0.3042 &
  0.3074 &
  0.2886 &
  \textbf{0.2316} &
  0.5148 &
  0.5705 &
  0.4351 &
  0.3991 &
  0.3747 &
  0.3495 &
  \textbf{0.2810} \\ \cmidrule(l){3-16} 
\multicolumn{2}{c|}{Improvement} &
  \textbf{46.55\%} &
  \textbf{49.14\%} &
  \textbf{24.09\%} &
  \textbf{23.87\%} &
  \textbf{24.66\%} &
  \textbf{19.75\%} &
  - &
  \textbf{45.41\%} &
  \textbf{50.75\%} &
  \textbf{35.42\%} &
  \textbf{29.59\%} &
  \textbf{25.00\%} &
  \textbf{19.60\%} &
  - \\ 
  \bottomrule
\end{tabular}%
}
\end{table}
\begin{table}[t]
\centering
\caption{Average MSE and MAE of our RRE-PPO4Pred  and baseline methods on the Electricity dataset.}
\label{tab:Electricity-performance}
\resizebox{\textwidth}{!}{%
\setlength{\tabcolsep}{3.5pt}
\begin{tabular}{@{}cc|ccccccc|ccccccc@{}}
\toprule
\multirow{2}{*}{$H$} &
  \multirow{2}{*}{Backbone} &
  \multicolumn{7}{c|}{MSE} &
  \multicolumn{7}{c}{MAE} \\ \cmidrule(l){3-16} 
 &
   &
  Naive-all &
  Naive-last &
  DilatedRNN &
  EBPSO &
  LSTMjump &
  PGLSTM &
  Ours  &
  Naive-all &
  Naive-last &
  DilatedRNN &
  EBPSO &
  LSTMjump &
  PGLSTM &
  Ours \\ \midrule
\multirow{8}{*}{24} &
  RNN &
  0.2442 &
  0.2934 &
  0.2272 &
  0.1988 &
  0.1908 &
  0.1735 &
  \textbf{0.1379} &
  0.4117 &
  0.4197 &
  0.2964 &
  0.2992 &
  0.3031 &
  0.2805 &
  \textbf{0.2409} \\
 &
  MGU &
  0.2783 &
  0.2980 &
  0.2213 &
  0.1822 &
  0.1831 &
  0.1704 &
  \textbf{0.1444} &
  0.4168 &
  0.4337 &
  0.2953 &
  0.2933 &
  0.2895 &
  0.2789 &
  \textbf{0.2415} \\
 &
  GRU &
  0.2655 &
  0.2750 &
  0.2310 &
  0.1771 &
  0.1773 &
  0.1604 &
  \textbf{0.1469} &
  0.4097 &
  0.4197 &
  0.3032 &
  0.3061 &
  0.2932 &
  0.2860 &
  \textbf{0.2322} \\
 &
  LSTM &
  0.2763 &
  0.2748 &
  0.2181 &
  0.1809 &
  0.1840 &
  0.1722 &
  \textbf{0.1343} &
  0.4321 &
  0.4289 &
  0.3055 &
  0.3028 &
  0.2937 &
  0.2844 &
  \textbf{0.2368} \\
 &
  IndRNN &
  0.2446 &
  0.2778 &
  0.2309 &
  0.2007 &
  0.1863 &
  0.1669 &
  \textbf{0.1451} &
  0.4150 &
  0.4482 &
  0.2950 &
  0.2988 &
  0.3088 &
  0.2766 &
  \textbf{0.2461} \\
 &
  phLSTM &
  0.2597 &
  0.2694 &
  0.2285 &
  0.1835 &
  0.2031 &
  0.1671 &
  \textbf{0.1348} &
  0.4412 &
  0.4965 &
  0.2931 &
  0.2879 &
  0.3195 &
  0.2911 &
  \textbf{0.2494} \\
 &
  pLSTM &
  0.2464 &
  0.2926 &
  0.2171 &
  0.1984 &
  0.1836 &
  0.1791 &
  \textbf{0.1392} &
  0.4101 &
  0.4151 &
  0.2941 &
  0.2990 &
  0.3097 &
  0.3010 &
  \textbf{0.2398} \\
 &
  xLSTM &
  0.2457 &
  0.2621 &
  0.2224 &
  0.1718 &
  0.1997 &
  0.1604 &
  \textbf{0.1350} &
  0.4106 &
  0.4255 &
  0.2955 &
  0.2871 &
  0.2930 &
  0.2788 &
  \textbf{0.2323} \\ \midrule
\multirow{8}{*}{48} &
  RNN &
  0.3221 &
  0.3185 &
  0.2424 &
  0.2153 &
  0.2145 &
  0.2120 &
  \textbf{0.1613} &
  0.4379 &
  0.4295 &
  0.3391 &
  0.3295 &
  0.3117 &
  0.2998 &
  \textbf{0.2462} \\
 &
  MGU &
  0.2943 &
  0.3502 &
  0.2438 &
  0.2167 &
  0.2284 &
  0.2120 &
  \textbf{0.1615} &
  0.4782 &
  0.4591 &
  0.3159 &
  0.3150 &
  0.3115 &
  0.3165 &
  \textbf{0.2644} \\
 &
  GRU &
  0.2903 &
  0.3427 &
  0.2541 &
  0.2225 &
  0.2135 &
  0.2242 &
  \textbf{0.1563} &
  0.4597 &
  0.4373 &
  0.3249 &
  0.3241 &
  0.3107 &
  0.3084 &
  \textbf{0.2481} \\
 &
  LSTM &
  0.3308 &
  0.3907 &
  0.2460 &
  0.2306 &
  0.2133 &
  0.2091 &
  \textbf{0.1512} &
  0.4667 &
  0.4520 &
  0.3145 &
  0.3103 &
  0.3315 &
  0.2987 &
  \textbf{0.2549} \\
 &
  IndRNN &
  0.3210 &
  0.3594 &
  0.2574 &
  0.2216 &
  0.2210 &
  0.2168 &
  \textbf{0.1522} &
  0.4532 &
  0.4736 &
  0.3265 &
  0.3217 &
  0.3294 &
  0.3151 &
  \textbf{0.2690} \\
 &
  phLSTM &
  0.3526 &
  0.3871 &
  0.2483 &
  0.2326 &
  0.2325 &
  0.2140 &
  \textbf{0.1565} &
  0.4618 &
  0.5013 &
  0.3400 &
  0.3434 &
  0.3418 &
  0.3194 &
  \textbf{0.2632} \\
 &
  pLSTM &
  0.3280 &
  0.3911 &
  0.2378 &
  0.2396 &
  0.2314 &
  0.2141 &
  \textbf{0.1567} &
  0.4528 &
  0.4530 &
  0.3317 &
  0.3257 &
  0.3346 &
  0.3256 &
  \textbf{0.2527} \\
 &
  xLSTM &
  0.3076 &
  0.3410 &
  0.2308 &
  0.2278 &
  0.2209 &
  0.2099 &
  \textbf{0.1538} &
  0.4431 &
  0.4406 &
  0.3076 &
  0.3081 &
  0.3159 &
  0.2896 &
  \textbf{0.2460} \\ \midrule
\multirow{8}{*}{96} &
  RNN &
  0.3226 &
  0.3392 &
  0.2589 &
  0.2457 &
  0.2489 &
  0.2211 &
  \textbf{0.1832} &
  0.4791 &
  0.5957 &
  0.3468 &
  0.3477 &
  0.3600 &
  0.3326 &
  \textbf{0.2616} \\
 &
  MGU &
  0.3326 &
  0.3608 &
  0.2396 &
  0.2337 &
  0.2418 &
  0.2401 &
  \textbf{0.1691} &
  0.5335 &
  0.5885 &
  0.3716 &
  0.3693 &
  0.3469 &
  0.3288 &
  \textbf{0.2701} \\
 &
  GRU &
  0.2908 &
  0.3200 &
  0.2458 &
  0.2488 &
  0.2480 &
  0.2318 &
  \textbf{0.1822} &
  0.4982 &
  0.5838 &
  0.3784 &
  0.3740 &
  0.3679 &
  0.3402 &
  \textbf{0.2698} \\
 &
  LSTM &
  0.3144 &
  0.3801 &
  0.2644 &
  0.2510 &
  0.2368 &
  0.2314 &
  \textbf{0.1769} &
  0.5164 &
  0.5380 &
  0.3732 &
  0.3777 &
  0.3563 &
  0.3460 &
  \textbf{0.2693} \\
 &
  IndRNN &
  0.3208 &
  0.3491 &
  0.2797 &
  0.2529 &
  0.2435 &
  0.2402 &
  \textbf{0.1805} &
  0.5233 &
  0.4944 &
  0.3777 &
  0.3721 &
  0.3615 &
  0.3509 &
  \textbf{0.2648} \\
 &
  phLSTM &
  0.3652 &
  0.4532 &
  0.3068 &
  0.2564 &
  0.2581 &
  0.2414 &
  \textbf{0.1808} &
  0.5005 &
  0.5045 &
  0.3663 &
  0.3656 &
  0.3516 &
  0.3307 &
  \textbf{0.2570} \\
 &
  pLSTM &
  0.3640 &
  0.3755 &
  0.2672 &
  0.2619 &
  0.2459 &
  0.2528 &
  \textbf{0.1723} &
  0.5176 &
  0.5203 &
  0.3626 &
  0.3581 &
  0.3626 &
  0.3324 &
  \textbf{0.2624} \\
 &
  xLSTM &
  0.3488 &
  0.3750 &
  0.2495 &
  0.2397 &
  0.2417 &
  0.2370 &
  \textbf{0.1767} &
  0.5339 &
  0.5487 &
  0.3619 &
  0.3583 &
  0.3612 &
  0.3444 &
  \textbf{0.2607} \\ \midrule
\multicolumn{2}{c|}{Average} &
  0.3028 &
  0.3365 &
  0.2445 &
  0.2204 &
  0.2187 &
  0.2066 &
  \textbf{0.1579} &
  0.4626 &
  0.4795 &
  0.3299 &
  0.3281 &
  0.3277 &
  0.3107 &
  \textbf{0.2533} \\ \cmidrule(l){3-16} 
\multicolumn{2}{c|}{Improvement} &
  \textbf{47.85\% }&
  \textbf{53.08\%} &
  \textbf{35.42\%} &
  \textbf{28.36\%} &
  \textbf{27.80\%} &
  \textbf{23.57\%} &
  - &
  \textbf{45.24\% }&
  \textbf{47.18\%} &
  \textbf{23.22\%} &
  \textbf{22.80\%} &
  \textbf{22.70\%} &
  \textbf{18.47\%} &
  - \\ 
  \bottomrule
\end{tabular}%
}
\end{table}
\begin{table}[t]
\centering
\caption{Average MSE and MAE of our RRE-PPO4Pred and baseline models on the ETTh dataset.}
\label{tab:ETTh-performance}
\resizebox{\textwidth}{!}{%
\setlength{\tabcolsep}{3.5pt}
\begin{tabular}{@{}cc|ccccccc|ccccccc@{}}
\toprule
\multirow{2}{*}{$H$} &
  \multirow{2}{*}{Backbone} &
  \multicolumn{7}{c|}{MSE} &
  \multicolumn{7}{c}{MAE} \\ \cmidrule(l){3-16} 
 &
   &
  Naive-all &
  Naive-last &
  DilatedRNN &
  EBPSO &
  LSTMjump &
  PGLSTM &
  Ours  &
  Naive-all &
  Naive-last &
  DilatedRNN &
  EBPSO &
  LSTMjump &
  PGLSTM &
  Ours  \\ \midrule
\multirow{8}{*}{24} &
  RNN &
  0.5521 &
  0.5819 &
  0.4010 &
  0.3734 &
  0.3949 &
  0.3801 &
  \textbf{0.3096} &
  0.6529 &
  0.6915 &
  0.4817 &
  0.4650 &
  0.4749 &
  0.4449 &
  \textbf{0.3647} \\
 &
  MGU &
  0.5772 &
  0.6090 &
  0.4052 &
  0.4039 &
  0.4041 &
  0.3875 &
  \textbf{0.3016} &
  0.6729 &
  0.7266 &
  0.4869 &
  0.4873 &
  0.4843 &
  0.4666 &
  \textbf{0.3503} \\
 &
  GRU &
  0.6046 &
  0.6111 &
  0.4403 &
  0.4016 &
  0.3947 &
  0.3741 &
  \textbf{0.3059} &
  0.7130 &
  0.7306 &
  0.5208 &
  0.4966 &
  0.4732 &
  0.4512 &
  \textbf{0.3593} \\
 &
  LSTM &
  0.6121 &
  0.6186 &
  0.4229 &
  0.3848 &
  0.4062 &
  0.3979 &
  \textbf{0.3067} &
  0.7212 &
  0.7317 &
  0.5053 &
  0.4576 &
  0.4947 &
  0.4757 &
  \textbf{0.3620} \\
 &
  IndRNN &
  0.5554 &
  0.6286 &
  0.4123 &
  0.4001 &
  0.4100 &
  0.3956 &
  \textbf{0.3147} &
  0.6522 &
  0.7409 &
  0.4886 &
  0.4811 &
  0.4930 &
  0.4671 &
  \textbf{0.3614} \\
 &
  phLSTM &
  0.5840 &
  0.5889 &
  0.4484 &
  0.3802 &
  0.3943 &
  0.4003 &
  \textbf{0.3174} &
  0.6811 &
  0.6949 &
  0.5309 &
  0.4989 &
  0.4791 &
  0.4824 &
  \textbf{0.3666} \\
 &
  pLSTM &
  0.5572 &
  0.5761 &
  0.4604 &
  0.3997 &
  0.4177 &
  0.3795 &
  \textbf{0.3237} &
  0.6579 &
  0.6864 &
  0.5510 &
  0.4942 &
  0.5012 &
  0.4482 &
  \textbf{0.3743} \\
 &
  xLSTM &
  0.5620 &
  0.5675 &
  0.4162 &
  0.3836 &
  0.3931 &
  0.3633 &
  \textbf{0.3075} &
  0.6481 &
  0.6751 &
  0.4860 &
  0.4729 &
  0.4715 &
  0.4203 &
  \textbf{0.3565} \\ \midrule
\multirow{8}{*}{48} &
  RNN &
  0.6003 &
  0.6496 &
  0.4206 &
  0.4631 &
  0.4642 &
  0.4468 &
  \textbf{0.3225} &
  0.6878 &
  0.7477 &
  0.4848 &
  0.5363 &
  0.5395 &
  0.5198 &
  \textbf{0.3716} \\
 &
  MGU &
  0.6148 &
  0.6607 &
  0.4451 &
  0.4573 &
  0.4530 &
  0.4417 &
  \textbf{0.3394} &
  0.7048 &
  0.7674 &
  0.5078 &
  0.5278 &
  0.5338 &
  0.5106 &
  \textbf{0.3820} \\
 &
  GRU &
  0.6456 &
  0.6604 &
  0.4624 &
  0.4491 &
  0.4688 &
  0.4486 &
  \textbf{0.3466} &
  0.7333 &
  0.7544 &
  0.5338 &
  0.5630 &
  0.5518 &
  0.5154 &
  \textbf{0.3865} \\
 &
  LSTM &
  0.6342 &
  0.6267 &
  0.4557 &
  0.4455 &
  0.4605 &
  0.4613 &
  \textbf{0.3449} &
  0.7228 &
  0.7272 &
  0.5179 &
  0.4807 &
  0.5360 &
  0.5320 &
  \textbf{0.3939} \\
 &
  IndRNN &
  0.6265 &
  0.6583 &
  0.4682 &
  0.4647 &
  0.4806 &
  0.4542 &
  \textbf{0.3311} &
  0.7175 &
  0.7647 &
  0.5447 &
  0.5333 &
  0.5562 &
  0.5155 &
  \textbf{0.3776} \\
 &
  phLSTM &
  0.6271 &
  0.6177 &
  0.4929 &
  0.4526 &
  0.4706 &
  0.4385 &
  \textbf{0.3397} &
  0.7114 &
  0.7123 &
  0.5637 &
  0.5328 &
  0.5405 &
  0.5069 &
  \textbf{0.3881} \\
 &
  pLSTM &
  0.6405 &
  0.6154 &
  0.4984 &
  0.4523 &
  0.4557 &
  0.4581 &
  \textbf{0.3536} &
  0.7310 &
  0.7121 &
  0.5706 &
  0.5814 &
  0.5240 &
  0.5259 &
  \textbf{0.3973} \\
 &
  xLSTM &
  0.5962 &
  0.6259 &
  0.4390 &
  0.4686 &
  0.4680 &
  0.4256 &
  \textbf{0.3378} &
  0.6779 &
  0.7191 &
  0.5105 &
  0.5313 &
  0.5420 &
  0.4840 &
  \textbf{0.3746} \\ \midrule
\multirow{8}{*}{96} &
  RNN &
  0.7050 &
  0.7484 &
  0.4714 &
  0.4988 &
  0.5171 &
  0.4906 &
  \textbf{0.3683} &
  0.7778 &
  0.8322 &
  0.5266 &
  0.5561 &
  0.5814 &
  0.5436 &
  \textbf{0.3962} \\
 &
  MGU &
  0.7094 &
  0.6952 &
  0.4692 &
  0.4917 &
  0.5142 &
  0.4995 &
  \textbf{0.3584} &
  0.7770 &
  0.7772 &
  0.5213 &
  0.5528 &
  0.5720 &
  0.5595 &
  \textbf{0.3959} \\
 &
  GRU &
  0.7230 &
  0.7026 &
  0.4979 &
  0.4814 &
  0.5030 &
  0.5081 &
  \textbf{0.3687} &
  0.7963 &
  0.7829 &
  0.5548 &
  0.5937 &
  0.5629 &
  0.5671 &
  \textbf{0.4098} \\
 &
  LSTM &
  0.7237 &
  0.7047 &
  0.4962 &
  0.4837 &
  0.5114 &
  0.5175 &
  \textbf{0.3588} &
  0.7931 &
  0.7773 &
  0.5406 &
  0.5369 &
  0.5677 &
  0.5729 &
  \textbf{0.3911} \\
 &
  IndRNN &
  0.7030 &
  0.7287 &
  0.4967 &
  0.5258 &
  0.5447 &
  0.5290 &
  \textbf{0.3678} &
  0.7707 &
  0.8053 &
  0.5514 &
  0.5867 &
  0.6033 &
  0.5911 &
  \textbf{0.3971} \\
 &
  phLSTM &
  0.6944 &
  0.7809 &
  0.5115 &
  0.5097 &
  0.5181 &
  0.4891 &
  \textbf{0.3863} &
  0.7687 &
  0.7516 &
  0.5689 &
  0.5546 &
  0.5855 &
  0.5388 &
  \textbf{0.4121} \\
 &
  pLSTM &
  0.7209 &
  0.7494 &
  0.5100 &
  0.4982 &
  0.5239 &
  0.5262 &
  \textbf{0.3778} &
  0.7993 &
  0.7709 &
  0.5643 &
  0.5741 &
  0.5833 &
  0.5827 &
  \textbf{0.4055} \\
 &
  xLSTM &
  0.6755 &
  0.6799 &
  0.4869 &
  0.5015 &
  0.5042 &
  0.4924 &
  \textbf{0.3607} &
  0.7391 &
  0.7510 &
  0.5451 &
  0.5473 &
  0.5618 &
  0.5516 &
  \textbf{0.3897} \\ \midrule
\multicolumn{2}{c|}{Average} &
  0.6352 &
  0.6536 &
  0.4595 &
  0.4484 &
  0.4614 &
  0.4464 &
  \textbf{0.3396} &
  0.7212 &
  0.7429 &
  0.5274 &
  0.5268 &
  0.5339 &
  0.5114 &
  \textbf{0.3818} \\ \cmidrule(l){3-16} 
\multicolumn{2}{c|}{Improvement} &
  \textbf{46.54\%} &
  \textbf{48.04\%} &
  \textbf{26.09\%} &
  \textbf{24.27\%} &
  \textbf{26.40\%} &
  \textbf{23.92\%} &
  - &
  \textbf{47.06\%} &
  \textbf{48.61\%} &
  \textbf{27.61\%} &
  \textbf{27.52\%} &
  \textbf{28.49\%} &
  \textbf{25.34\%} &
  - \\ 
  \bottomrule
\end{tabular}%
}
\end{table}

\begin{table}[t]
\centering
\caption{Average MSE and MAE of our RRE-PPO4Pred and baseline methods on the Weather dataset.}
\label{tab:Weather-performance }
\resizebox{\textwidth}{!}{%
\setlength{\tabcolsep}{3.5pt}
\begin{tabular}{@{}cc|ccccccc|ccccccc@{}}
\toprule
\multirow{2}{*}{$H$} &
  \multirow{2}{*}{Backbone} &
  \multicolumn{7}{c|}{MSE} &
  \multicolumn{7}{c}{MAE} \\ \cmidrule(l){3-16} 
 &
   &
  Naive-all &
  Naive-last &
  DilatedRNN &
  EBPSO &
  LSTMjump &
  PGLSTM &
  Ours  &
  Naive-all &
  Naive-last &
  DilatedRNN &
  EBPSO &
  LSTMjump &
  PGLSTM &
  Ours  \\ \midrule
\multirow{8}{*}{24} &
  RNN &
  0.2944 &
  0.3535 &
  0.2301 &
  0.2446 &
  0.2531 &
  0.2396 &
  \textbf{0.1741} &
  0.3974 &
  0.4485 &
  0.3282 &
  0.3532 &
  0.3305 &
  0.3176 &
  \textbf{0.2574} \\
 &
  MGU &
  0.3402 &
  0.3185 &
  0.2458 &
  0.2288 &
  0.2259 &
  0.2214 &
  \textbf{0.1674} &
  0.4455 &
  0.4079 &
  0.3525 &
  0.3354 &
  0.3362 &
  0.3203 &
  \textbf{0.2565} \\
 &
  GRU &
  0.3135 &
  0.3202 &
  0.2507 &
  0.2418 &
  0.2338 &
  0.1921 &
  \textbf{0.1671} &
  0.4233 &
  0.4070 &
  0.3590 &
  0.3448 &
  0.2851 &
  0.2815 &
  \textbf{0.2489} \\
 &
  LSTM &
  0.3459 &
  0.3368 &
  0.2496 &
  0.2386 &
  0.2403 &
  0.2237 &
  \textbf{0.1673} &
  0.4666 &
  0.4312 &
  0.3435 &
  0.3453 &
  0.3452 &
  0.3211 &
  \textbf{0.2528} \\
 &
  IndRNN &
  0.3391 &
  0.3678 &
  0.2566 &
  0.2467 &
  0.2356 &
  0.2294 &
  \textbf{0.1782} &
  0.4639 &
  0.4671 &
  0.3671 &
  0.3517 &
  0.3429 &
  0.3301 &
  \textbf{0.2652} \\
 &
  phLSTM &
  0.3165 &
  0.3184 &
  0.2654 &
  0.2492 &
  0.2685 &
  0.2547 &
  \textbf{0.1902} &
  0.4280 &
  0.4154 &
  0.3652 &
  0.3589 &
  0.3756 &
  0.3375 &
  \textbf{0.2774} \\
 &
  pLSTM &
  0.3484 &
  0.3625 &
  0.2290 &
  0.2366 &
  0.2477 &
  0.2348 &
  \textbf{0.1718} &
  0.4576 &
  0.4680 &
  0.3302 &
  0.3310 &
  0.3455 &
  0.3366 &
  \textbf{0.2620} \\
 &
  xLSTM &
  0.3086 &
  0.3225 &
  0.2292 &
  0.2359 &
  0.2400 &
  0.2261 &
  \textbf{0.1698} &
  0.4124 &
  0.4187 &
  0.3112 &
  0.3382 &
  0.3314 &
  0.3074 &
  \textbf{0.2438} \\ \midrule
\multirow{8}{*}{48} &
  RNN &
  0.3367 &
  0.3847 &
  0.2897 &
  0.2746 &
  0.2622 &
  0.2565 &
  \textbf{0.2002} &
  0.4373 &
  0.4648 &
  0.3871 &
  0.3752 &
  0.3427 &
  0.3391 &
  \textbf{0.2865} \\
 &
  MGU &
  0.4324 &
  0.3612 &
  0.2567 &
  0.2756 &
  0.2686 &
  0.2496 &
  \textbf{0.1791} &
  0.5415 &
  0.4454 &
  0.3454 &
  0.3755 &
  0.3439 &
  0.3589 &
  \textbf{0.2605} \\
 &
  GRU &
  0.3285 &
  0.3388 &
  0.2558 &
  0.2636 &
  0.2916 &
  0.2826 &
  \textbf{0.1979} &
  0.4247 &
  0.4439 &
  0.3403 &
  0.3726 &
  0.3583 &
  0.3302 &
  \textbf{0.2895} \\
 &
  LSTM &
  0.3885 &
  0.3583 &
  0.2647 &
  0.2544 &
  0.2751 &
  0.2705 &
  \textbf{0.1884} &
  0.5028 &
  0.4462 &
  0.3635 &
  0.3471 &
  0.3662 &
  0.3317 &
  \textbf{0.2784} \\
 &
  IndRNN &
  0.3841 &
  0.3837 &
  0.2663 &
  0.2694 &
  0.2761 &
  0.2742 &
  \textbf{0.1953} &
  0.4880 &
  0.4701 &
  0.3558 &
  0.3722 &
  0.3722 &
  0.3396 &
  \textbf{0.2844} \\
 &
  phLSTM &
  0.3792 &
  0.3796 &
  0.2812 &
  0.2837 &
  0.2748 &
  0.2791 &
  \textbf{0.2064} &
  0.4758 &
  0.4724 &
  0.3629 &
  0.3796 &
  0.3723 &
  0.3367 &
  \textbf{0.2812} \\
 &
  pLSTM &
  0.4013 &
  0.4231 &
  0.2632 &
  0.2709 &
  0.2703 &
  0.2691 &
  \textbf{0.2125} &
  0.5036 &
  0.5178 &
  0.3437 &
  0.3594 &
  0.3560 &
  0.3568 &
  \textbf{0.3056} \\
 &
  xLSTM &
  0.3454 &
  0.3606 &
  0.2391 &
  0.2505 &
  0.2618 &
  0.2469 &
  \textbf{0.1839} &
  0.4440 &
  0.4400 &
  0.3215 &
  0.3494 &
  0.3287 &
  0.3270 &
  \textbf{0.2552} \\ \midrule
\multirow{8}{*}{96} &
  RNN &
  0.4384 &
  0.4625 &
  0.3176 &
  0.3110 &
  0.2981 &
  0.3013 &
  \textbf{0.2412} &
  0.5119 &
  0.5361 &
  0.3904 &
  0.3762 &
  0.3655 &
  0.3904 &
  \textbf{0.2896} \\
 &
  MGU &
  0.4340 &
  0.4176 &
  0.3179 &
  0.2981 &
  0.2948 &
  0.2934 &
  \textbf{0.2407} &
  0.5052 &
  0.5034 &
  0.3919 &
  0.3763 &
  0.3743 &
  0.3914 &
  \textbf{0.2786} \\
 &
  GRU &
  0.3322 &
  0.3663 &
  0.2873 &
  0.3097 &
  0.3298 &
  0.3316 &
  \textbf{0.2355} &
  0.4049 &
  0.4523 &
  0.3530 &
  0.3953 &
  0.3994 &
  0.3622 &
  \textbf{0.2987} \\
 &
  LSTM &
  0.4337 &
  0.4482 &
  0.2809 &
  0.3125 &
  0.3062 &
  0.3107 &
  \textbf{0.2129} &
  0.5120 &
  0.4902 &
  0.3604 &
  0.3847 &
  0.3864 &
  0.3918 &
  \textbf{0.3074} \\
 &
  IndRNN &
  0.3997 &
  0.4601 &
  0.2815 &
  0.3213 &
  0.3074 &
  0.3137 &
  \textbf{0.2298} &
  0.4855 &
  0.5458 &
  0.3498 &
  0.3879 &
  0.3787 &
  0.3856 &
  \textbf{0.3006} \\
 &
  phLSTM &
  0.3838 &
  0.4532 &
  0.3053 &
  0.3298 &
  0.3073 &
  0.3171 &
  \textbf{0.2317} &
  0.4555 &
  0.5214 &
  0.3804 &
  0.4235 &
  0.3880 &
  0.3461 &
  \textbf{0.2944} \\
 &
  pLSTM &
  0.4323 &
  0.4410 &
  0.2772 &
  0.3297 &
  0.3033 &
  0.2987 &
  \textbf{0.2182} &
  0.5005 &
  0.5239 &
  0.3572 &
  0.4045 &
  0.3750 &
  0.3837 &
  \textbf{0.3156} \\
 &
  xLSTM &
  0.3672 &
  0.4333 &
  0.2736 &
  0.2948 &
  0.2861 &
  0.2877 &
  \textbf{0.1927} &
  0.4518 &
  0.5193 &
  0.3435 &
  0.3877 &
  0.3499 &
  0.3527 &
  \textbf{0.2715} \\ \midrule
\multicolumn{2}{c|}{Average} &
  0.3677 &
  0.3822 &
  0.2664 &
  0.2734 &
  0.2733 &
  0.2677 &
  \textbf{0.1984} &
  0.4642 &
  0.4690 &
  0.3539 &
  0.3677 &
  0.3563 &
  0.3452 &
  \textbf{0.2776} \\ \cmidrule(l){3-16} 
\multicolumn{2}{c|}{Improvement} &
  \textbf{46.05\%} &
  \textbf{48.09\%} &
  \textbf{25.53\%} &
  \textbf{27.43\%} &
  \textbf{27.40\%} &
  \textbf{25.89\%} &
  - &
  \textbf{40.20\%} &
  \textbf{40.82\%} &
  \textbf{21.57\%} &
  \textbf{24.52\%} &
  \textbf{22.08\%} &
  \textbf{19.60\%} &
  - \\ 
  \bottomrule
\end{tabular}%
}
\end{table}

\begin{table}[t]
\centering
\caption{Average MSE and MAE of our RRE-PPO4Pred and baseline methods on the ILI dataset.}
\label{tab:ILI-performance}
\resizebox{\textwidth}{!}{%
\setlength{\tabcolsep}{3.5pt}
\begin{tabular}{@{}cc|ccccccc|ccccccc@{}}
\toprule
\multirow{2}{*}{$H$} &
  \multirow{2}{*}{Backbone} &
  \multicolumn{7}{c|}{MSE} &
  \multicolumn{7}{c}{MAE} \\ \cmidrule(l){3-16} 
 &
   &
  Naive-all &
  Naive-last &
  DilatedRNN &
  EBPSO &
  LSTMjump &
  PGLSTM &
  Ours  &
  Naive-all &
  Naive-last &
  DilatedRNN &
  EBPSO &
  LSTMjump &
  PGLSTM &
  Ours  \\ \midrule
\multirow{8}{*}{12} &
  RNN &
  3.5493 &
  3.7089 &
  2.7232 &
  2.6249 &
  2.5176 &
  2.4549 &
  \textbf{2.0618} &
  1.7768 &
  1.8593 &
  1.3633 &
  1.3151 &
  1.2742 &
  1.1648 &
  \textbf{0.9366} \\
 &
  MGU &
  3.6203 &
  3.9187 &
  2.7880 &
  2.7088 &
  2.6264 &
  2.3748 &
  \textbf{2.1034} &
  1.8144 &
  1.9633 &
  1.3975 &
  1.3551 &
  1.3326 &
  1.2140 &
  \textbf{0.9751} \\
 &
  GRU &
  3.7342 &
  3.7681 &
  2.7967 &
  2.6498 &
  2.5557 &
  2.4075 &
  \textbf{2.1072} &
  1.8691 &
  1.8887 &
  1.3986 &
  1.3257 &
  1.2837 &
  1.1508 &
  \textbf{0.9650} \\
 &
  LSTM &
  3.8082 &
  3.6981 &
  2.7355 &
  2.5795 &
  2.6523 &
  2.5163 &
  \textbf{1.9536} &
  1.9062 &
  1.8499 &
  1.3712 &
  1.2943 &
  1.3317 &
  1.2019 &
  \textbf{0.8959} \\
 &
  IndRNN &
  3.6330 &
  3.9465 &
  2.9202 &
  2.8686 &
  2.5251 &
  2.5550 &
  \textbf{2.0732} &
  1.8195 &
  1.9752 &
  1.4639 &
  1.4351 &
  1.2826 &
  1.1676 &
  \textbf{0.9550} \\
 &
  phLSTM &
  3.9479 &
  3.9443 &
  3.1138 &
  2.9215 &
  2.6682 &
  2.5793 &
  \textbf{2.0434} &
  1.9740 &
  1.9768 &
  1.5578 &
  1.4653 &
  1.3365 &
  1.2123 &
  \textbf{0.9353} \\
 &
  pLSTM &
  3.9236 &
  3.9943 &
  2.9151 &
  2.7848 &
  2.6297 &
  2.6196 &
  \textbf{1.9853} &
  1.9636 &
  1.9984 &
  1.4581 &
  1.3951 &
  1.3334 &
  1.1860 &
  \textbf{0.9015} \\
 &
  xLSTM &
  3.6790 &
  3.6246 &
  2.7349 &
  2.5948 &
  2.5324 &
  2.4712 &
  \textbf{1.9429} &
  1.8423 &
  1.8146 &
  1.3702 &
  1.3023 &
  1.2691 &
  1.1359 &
  \textbf{0.9021} \\ \midrule
\multirow{8}{*}{24} &
  RNN &
  3.8180 &
  3.8992 &
  2.8312 &
  2.7573 &
  2.7611 &
  2.6297 &
  \textbf{2.2374} &
  2.0142 &
  2.0545 &
  1.4922 &
  1.4546 &
  1.4539 &
  1.3039 &
  \textbf{1.0716} \\
 &
  MGU &
  3.7246 &
  3.9232 &
  2.7175 &
  2.8181 &
  2.7237 &
  2.6009 &
  \textbf{2.2749} &
  1.9603 &
  2.0695 &
  1.4342 &
  1.4880 &
  1.4350 &
  1.3488 &
  \textbf{1.0979} \\
 &
  GRU &
  3.8524 &
  3.7907 &
  2.4683 &
  2.7604 &
  2.6093 &
  2.6087 &
  \textbf{2.1150} &
  2.0288 &
  1.9959 &
  1.3041 &
  1.4534 &
  1.3823 &
  1.2444 &
  \textbf{1.0054} \\
 &
  LSTM &
  3.8919 &
  3.8378 &
  2.8225 &
  2.8377 &
  2.7370 &
  2.5984 &
  \textbf{2.0167} &
  2.0488 &
  2.0231 &
  1.4893 &
  1.4972 &
  1.4055 &
  1.2518 &
  \textbf{0.9663} \\
 &
  IndRNN &
  3.7271 &
  3.9673 &
  2.6403 &
  2.6875 &
  2.8555 &
  2.7027 &
  \textbf{1.9912} &
  1.9639 &
  2.0882 &
  1.3922 &
  1.4156 &
  1.4228 &
  1.3028 &
  \textbf{0.9755} \\
 &
  phLSTM &
  4.0572 &
  3.9740 &
  3.0070 &
  2.8756 &
  2.7383 &
  2.7409 &
  \textbf{2.4847} &
  2.1389 &
  2.0926 &
  1.5866 &
  1.5166 &
  1.4545 &
  1.2777 &
  \textbf{1.2013} \\
 &
  pLSTM &
  4.0688 &
  3.9718 &
  2.8506 &
  2.9331 &
  2.7412 &
  2.7449 &
  \textbf{2.2223} &
  2.1465 &
  2.0915 &
  1.5043 &
  1.5439 &
  1.4642 &
  1.3063 &
  \textbf{1.0687} \\
 &
  xLSTM &
  3.7355 &
  3.7830 &
  2.8455 &
  2.7835 &
  2.6251 &
  2.5733 &
  \textbf{2.0552} &
  1.9713 &
  1.9932 &
  1.4997 &
  1.4701 &
  1.3902 &
  1.2751 &
  \textbf{1.0805} \\ \midrule
\multirow{8}{*}{48} &
  RNN &
  3.9956 &
  4.1089 &
  2.7302 &
  2.9133 &
  2.8041 &
  2.8201 &
  \textbf{2.3958} &
  2.2233 &
  2.2845 &
  1.5191 &
  1.6212 &
  1.5834 &
  1.4187 &
  \textbf{1.2153} \\
 &
  MGU &
  4.3145 &
  4.2513 &
  3.0289 &
  2.9806 &
  2.9667 &
  2.8576 &
  \textbf{2.2778} &
  2.3975 &
  2.3642 &
  1.6860 &
  1.6602 &
  1.6746 &
  1.4557 &
  \textbf{1.1513} \\
 &
  GRU &
  3.9792 &
  4.0439 &
  2.8843 &
  3.0398 &
  2.8613 &
  2.8017 &
  \textbf{2.2779} &
  2.2148 &
  2.2495 &
  1.6046 &
  1.6913 &
  1.5954 &
  1.5534 &
  \textbf{1.1471} \\
 &
  LSTM &
  4.1105 &
  4.0951 &
  3.0058 &
  3.0517 &
  2.9190 &
  2.9276 &
  \textbf{2.5619} &
  2.2872 &
  2.2787 &
  1.6741 &
  1.6968 &
  1.6278 &
  1.5924 &
  \textbf{1.2968} \\
 &
  IndRNN &
  4.0335 &
  4.0182 &
  3.0907 &
  3.0219 &
  3.1871 &
  2.9941 &
  \textbf{2.2225} &
  2.2434 &
  2.2374 &
  1.7191 &
  1.6838 &
  1.6865 &
  1.5089 &
  \textbf{1.1167} \\
 &
  phLSTM &
  4.1356 &
  4.1689 &
  2.9460 &
  3.0563 &
  2.9915 &
  3.0373 &
  \textbf{2.2940} &
  2.2981 &
  2.3167 &
  1.6367 &
  1.7004 &
  1.6877 &
  1.5243 &
  \textbf{1.1621} \\
 &
  pLSTM &
  4.1876 &
  4.0254 &
  3.0641 &
  3.0070 &
  2.9843 &
  3.0832 &
  \textbf{2.4081} &
  2.2159 &
  2.2413 &
  1.7024 &
  1.6715 &
  1.6586 &
  1.5678 &
  \textbf{1.2193} \\
 &
  xLSTM &
  3.9482 &
  3.8198 &
  2.8719 &
  2.8371 &
  2.8058 &
  2.8818 &
  \textbf{2.1893} &
  2.1956 &
  2.1249 &
  1.5982 &
  1.5778 &
  1.5666 &
  1.4659 &
  \textbf{1.1572} \\ \midrule
\multicolumn{2}{c|}{Average} &
  3.8948 &
  3.9284 &
  2.8555 &
  2.8372 &
  2.7508 &
  2.6909 &
  \textbf{2.1790} &
  2.0548 &
  2.0763 &
  1.5093 &
  1.5013 &
  1.4555 &
  1.3263 &
  \textbf{1.0583} \\ \cmidrule(l){3-16} 
\multicolumn{2}{c|}{Improvement} &
  \textbf{44.06\%} &
  \textbf{44.54\%} &
  \textbf{23.70\%} &
  \textbf{23.21\%} &
  \textbf{20.78\%} &
  \textbf{19.04\%} &
  - &
  \textbf{48.49\%} &
  \textbf{49.03\%} &
  \textbf{29.88\%} &
  \textbf{29.49\%} &
  \textbf{27.29\%} &
  \textbf{20.19\%} &
  - \\ 
  \bottomrule
\end{tabular}%
}
\end{table}

\begin{figure}[t]
    \centering
    \begin{subfigure}[b]{0.3\textwidth}
        \centering
        \includegraphics[width=\textwidth]{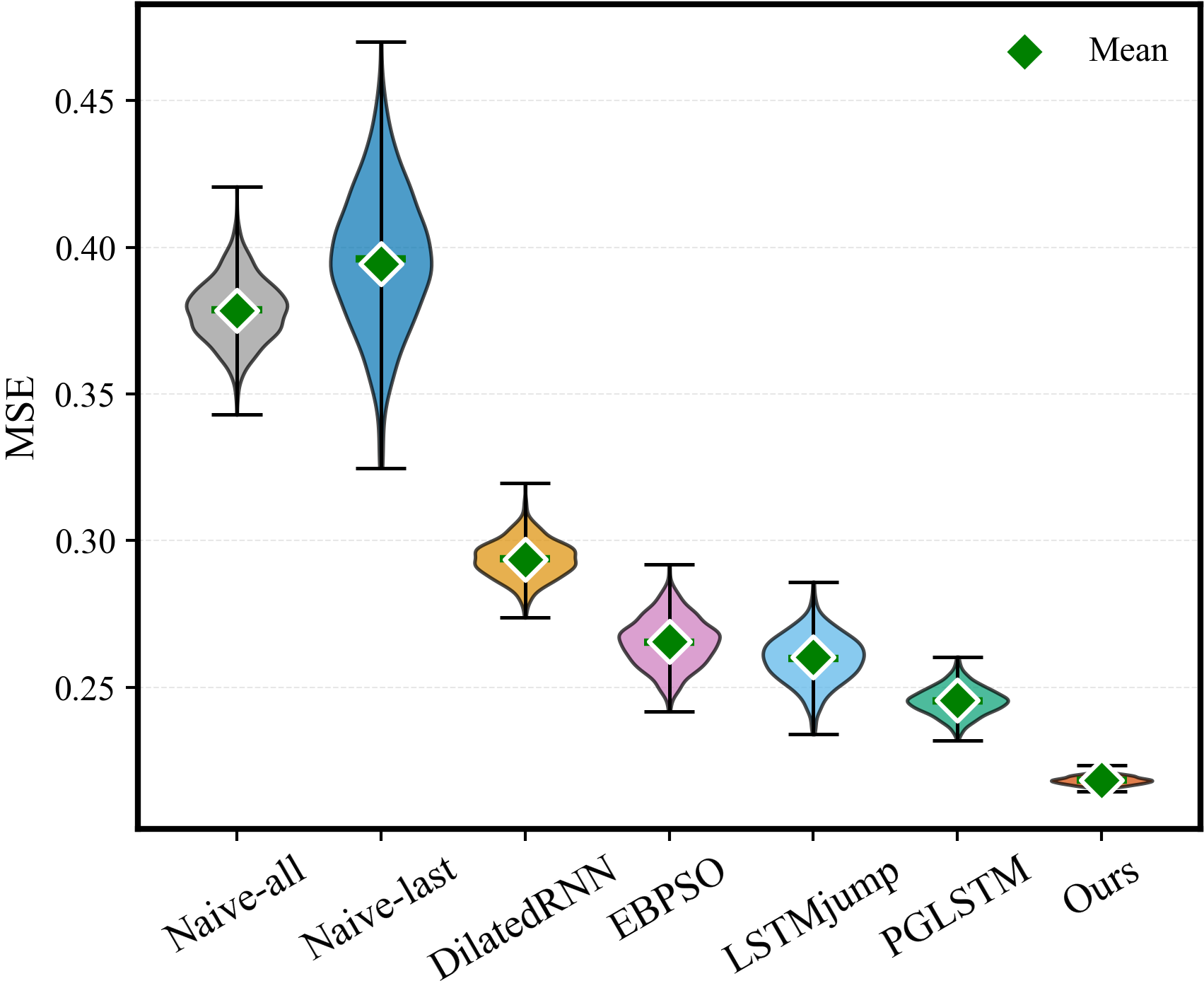} 
        \caption{Traffic,  $H = 24$}
        \label{fig:traffic-violin}
    \end{subfigure}
    \hfill 
    \begin{subfigure}[b]{0.3\textwidth}
        \centering
        \includegraphics[width=\textwidth]{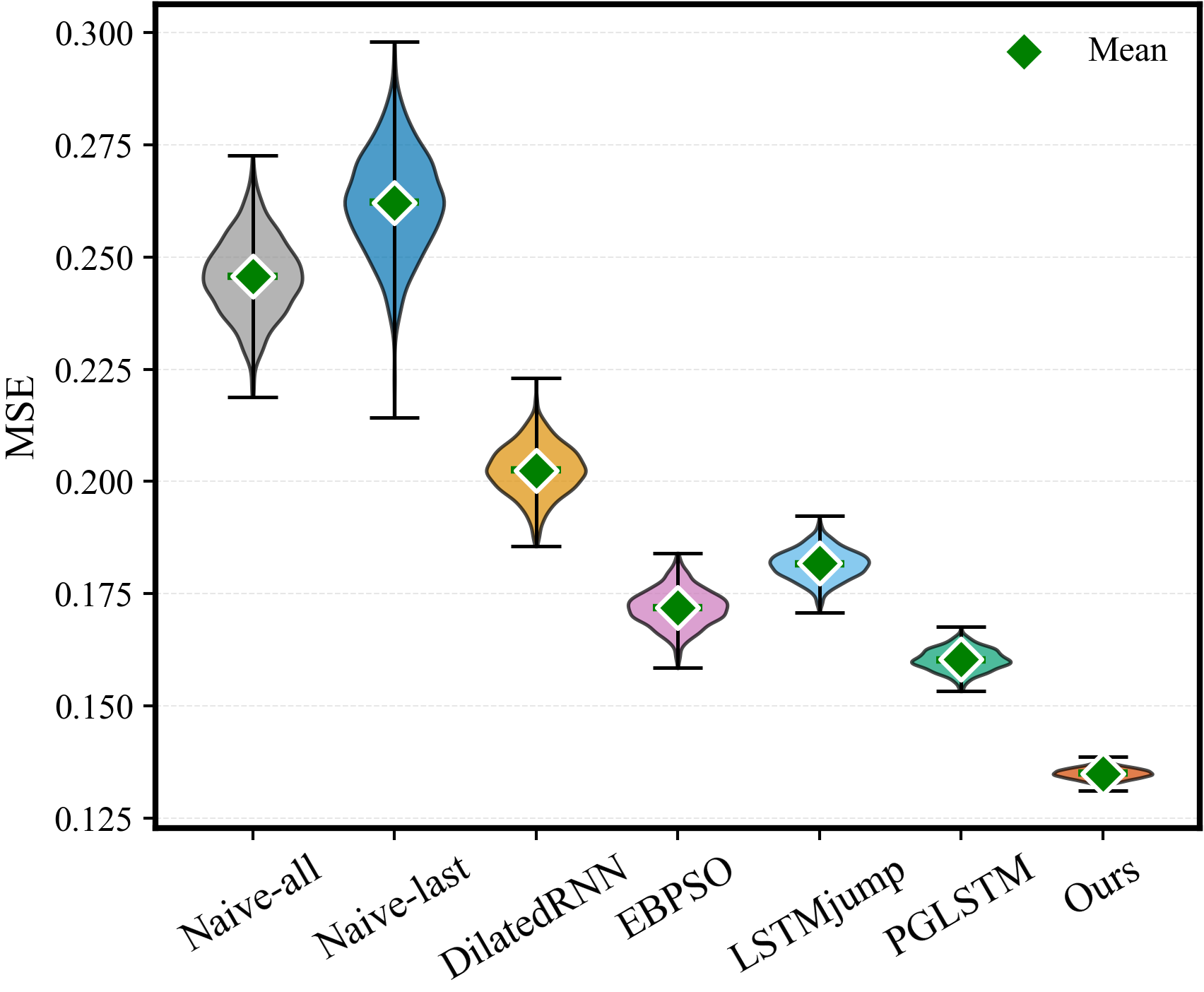} 
        \caption{Electricity,  $H = 24$}
        \label{fig:Electricity-violin}
    \end{subfigure}
    \hfill
        \begin{subfigure}[b]{0.3\textwidth}
        \centering
        \includegraphics[width=\textwidth]{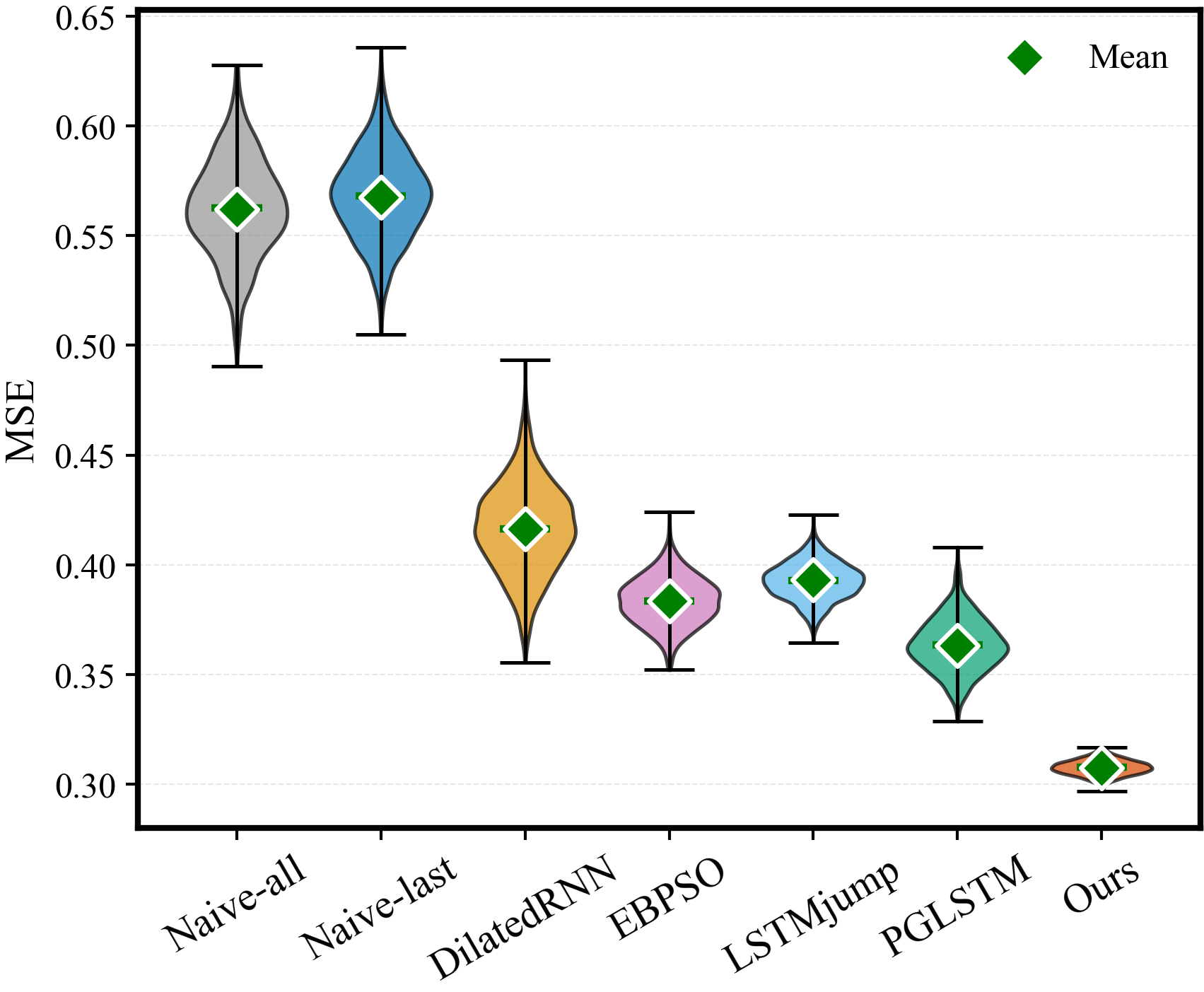} 
        \caption{ETTh, $H = 24$}
        \label{fig:violin_ETTh}
    \end{subfigure}
    \hfill 
    \vspace{3mm}
    \begin{subfigure}[b]{0.3\textwidth}
        \centering
        \includegraphics[width=\textwidth]{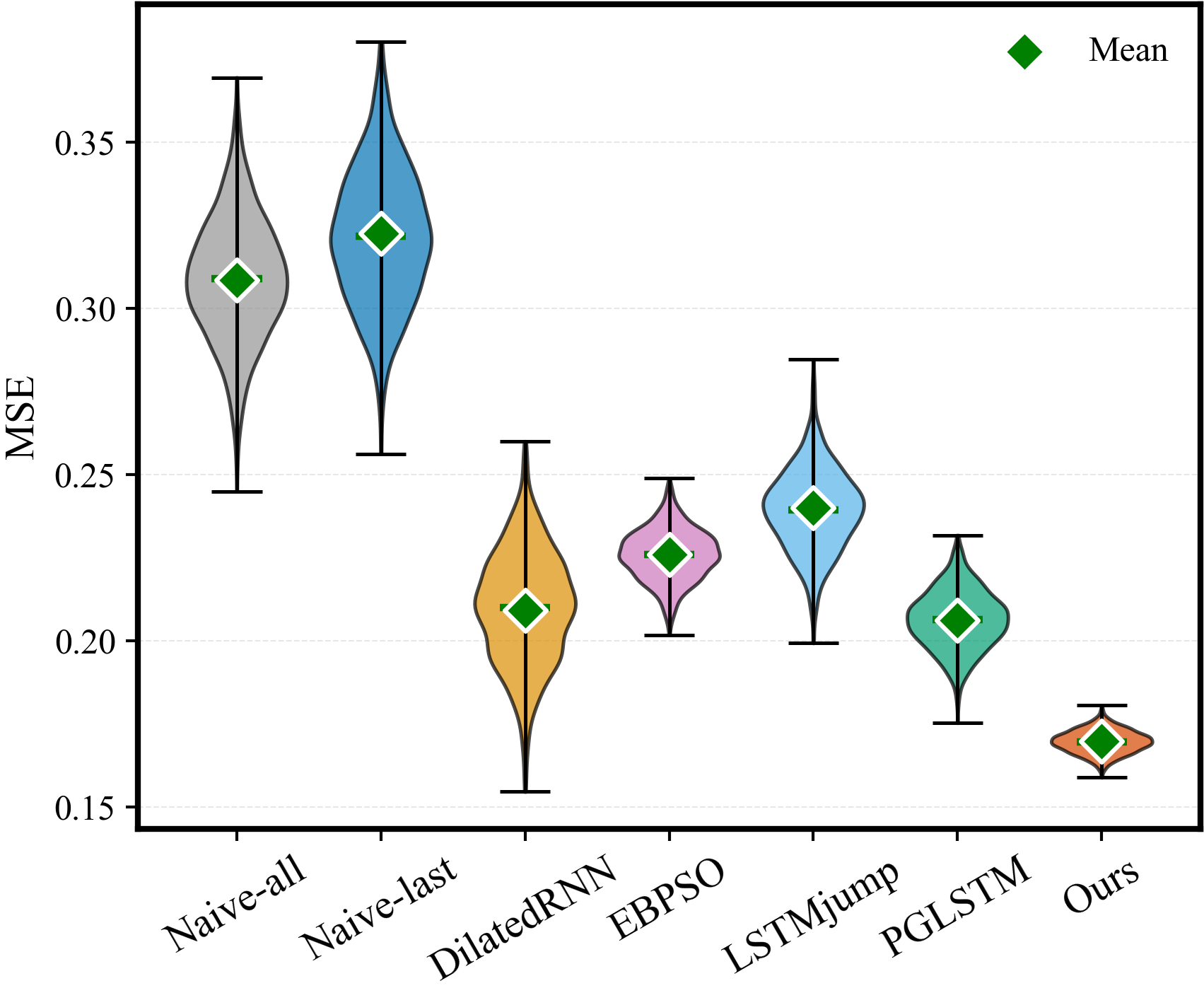} 
        \caption{Weather, $H = 24$ }
        \label{fig:violin_Weather}
    \end{subfigure}
    \begin{subfigure}[b]{0.3\textwidth}
        \centering
        \includegraphics[width=\textwidth]{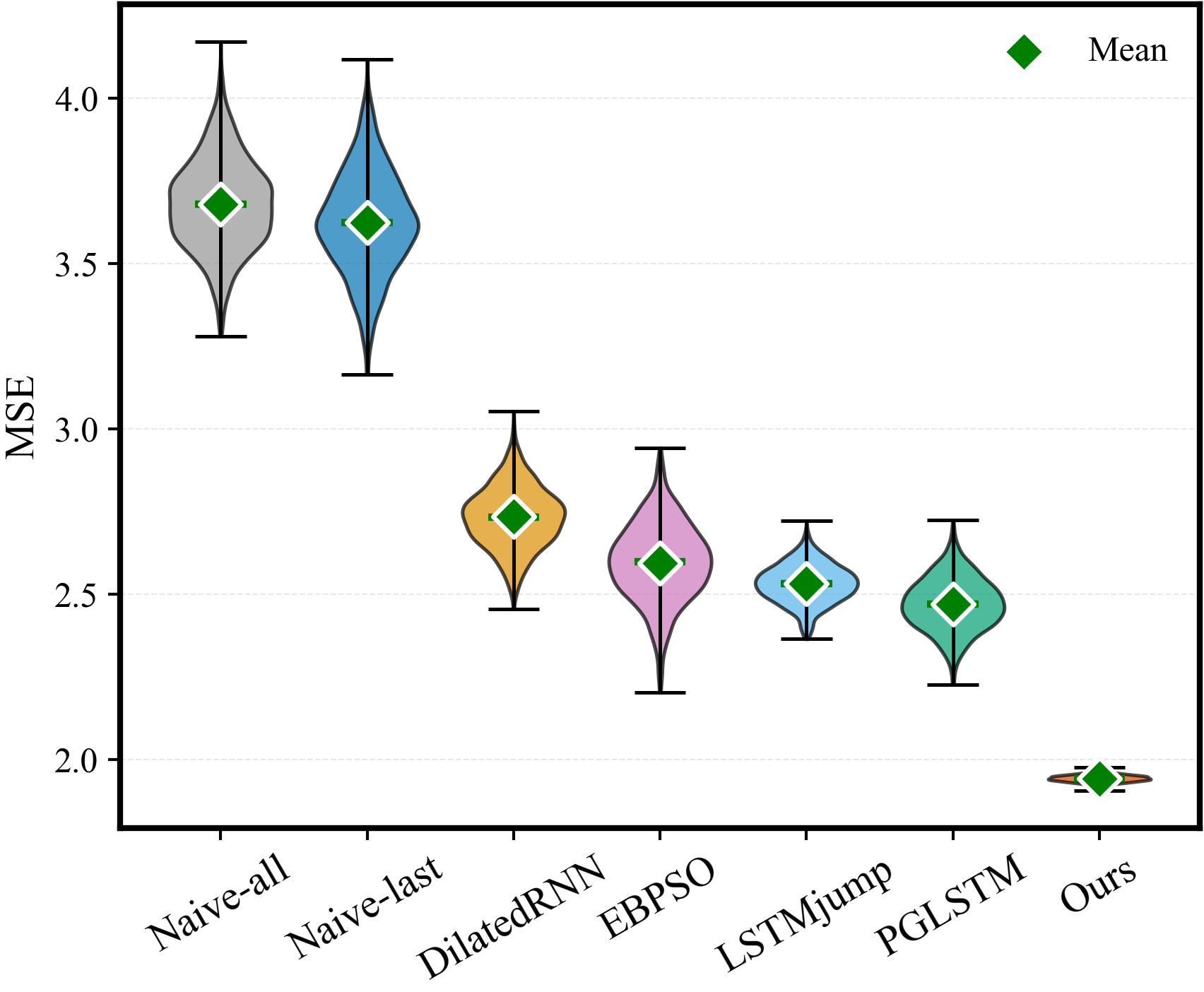} 
        \caption{ILI, $H=12$ }
        \label{fig:violin_ILI}
    \end{subfigure}
    
    \caption{Violin plots of MSE for the proposed RRE-PPO4Pred method and baseline methods with xLSTM backbone  at horizon $H = 12$ for ILI dataset and horizon $H = 24$ for the remaining datasets.}
    \label{fig:violin-mse}
\end{figure}

\begin{figure}[t]
    \centering
    \begin{subfigure}[b]{0.32\textwidth}
        \centering
        \includegraphics[width=\textwidth]{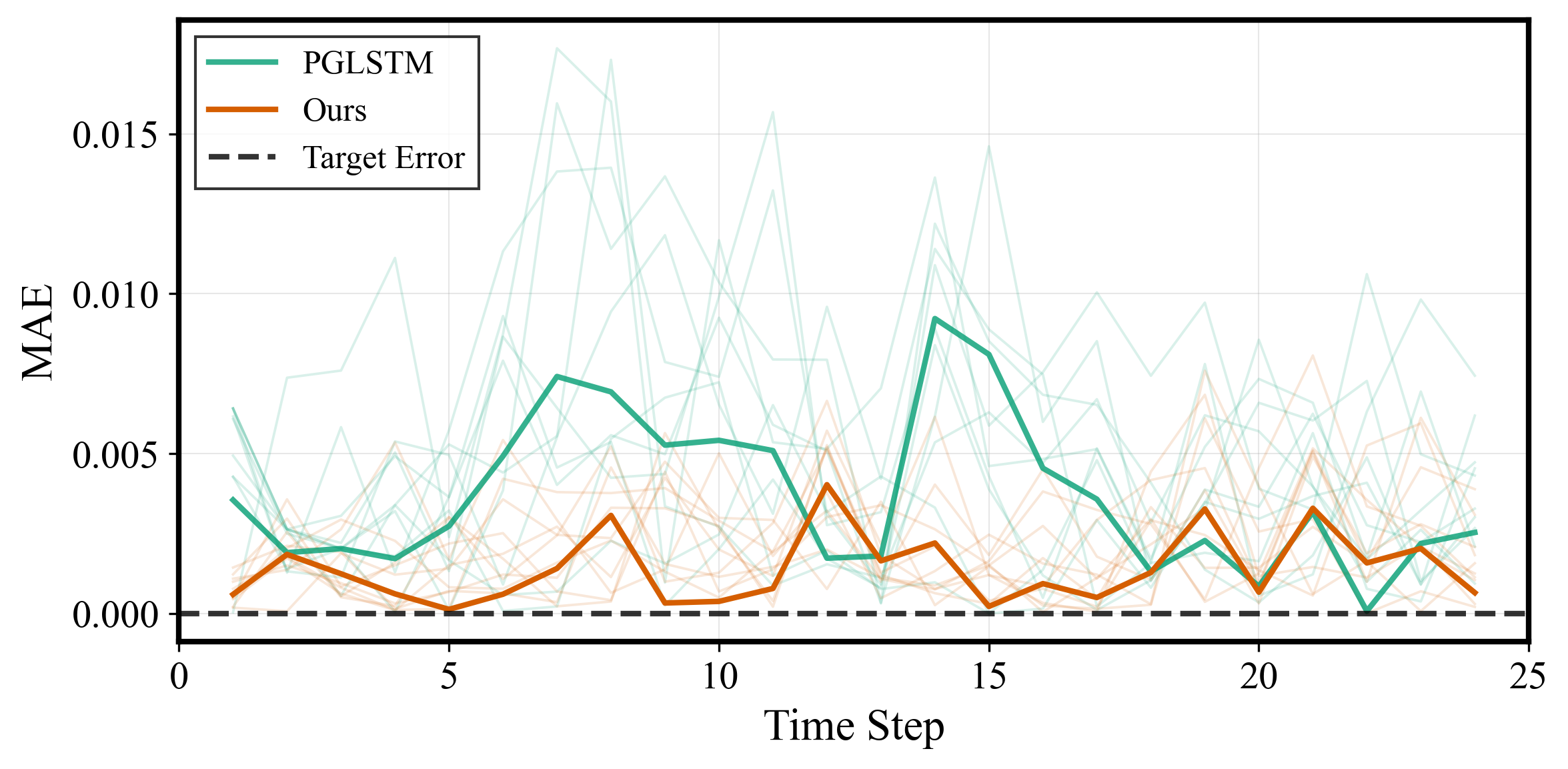} 
        \caption{Traffic  }
        \label{fig:errors_traffic}
    \end{subfigure}
    \begin{subfigure}[b]{0.32\textwidth}
        \centering
        \includegraphics[width=\textwidth]{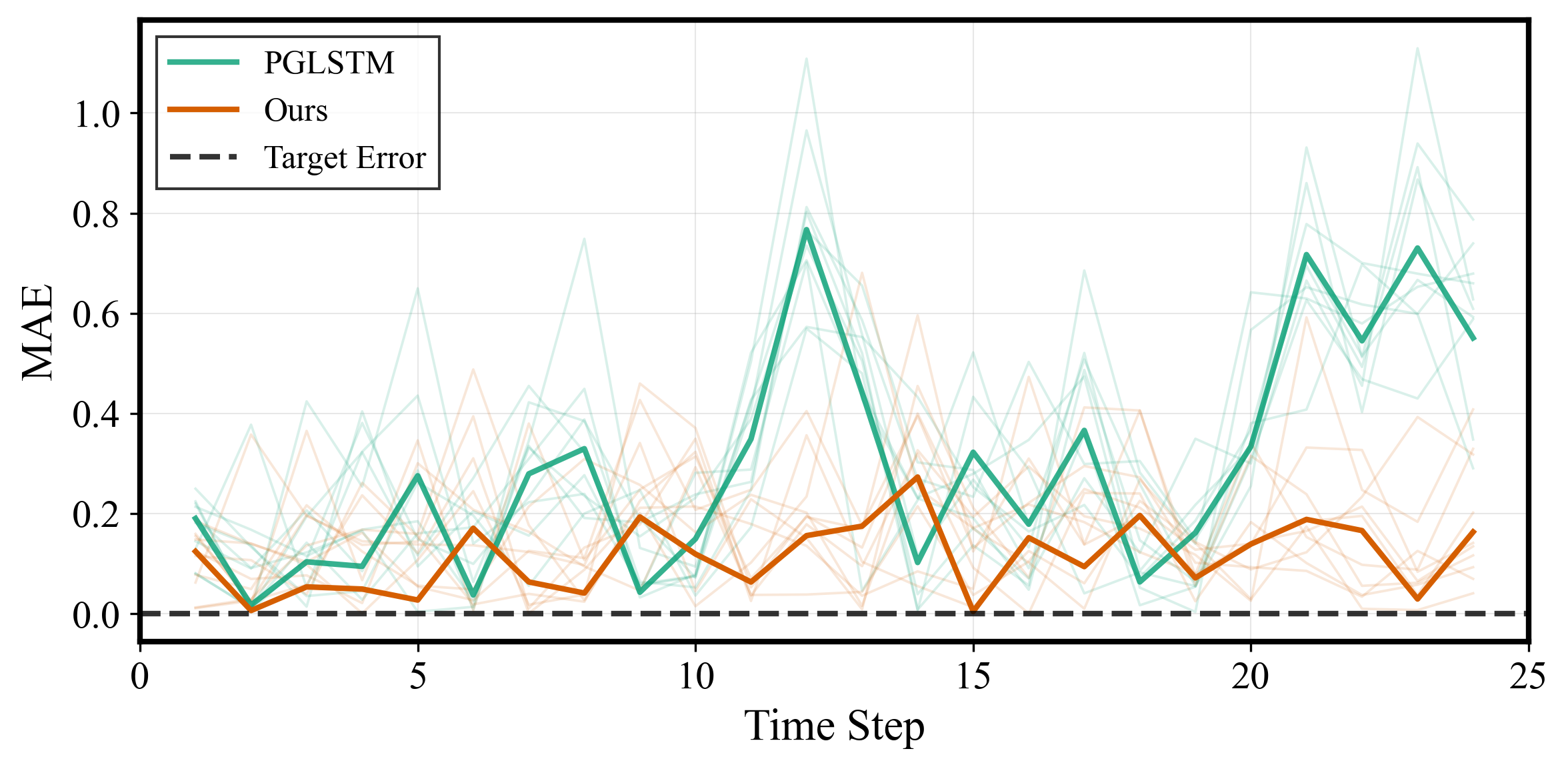} 
        \caption{Electricity }
        \label{fig:errors_Electricity}
    \end{subfigure}
        \begin{subfigure}[b]{0.32\textwidth}
        \centering
        \includegraphics[width=\textwidth]{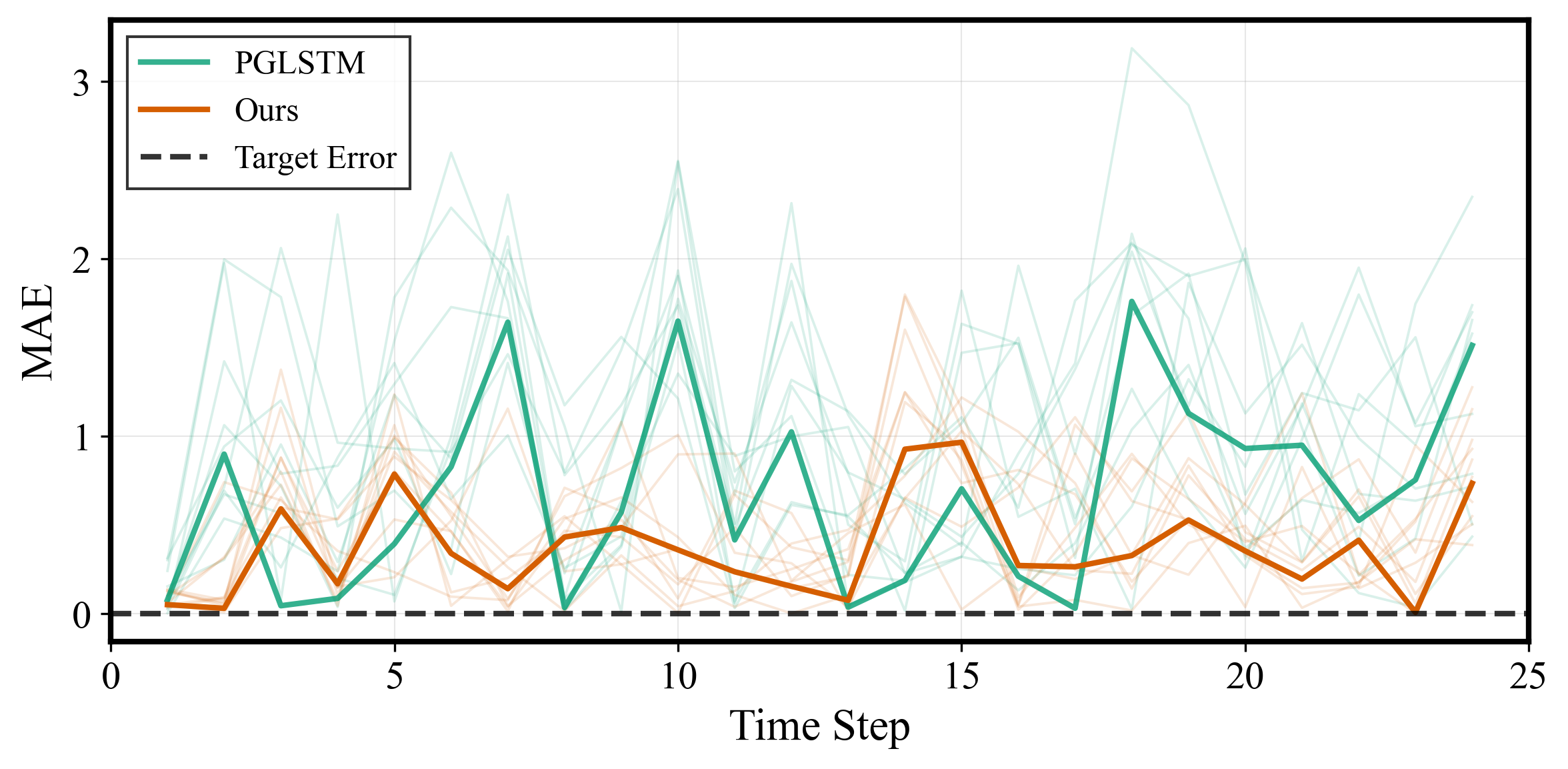} 
        \caption{ETTh }
        \label{fig:errors_ETTh}
    \end{subfigure}
        \hfill 
    \vspace{3mm}
    \begin{subfigure}[b]{0.32\textwidth}
        \centering
        \includegraphics[width=\textwidth]{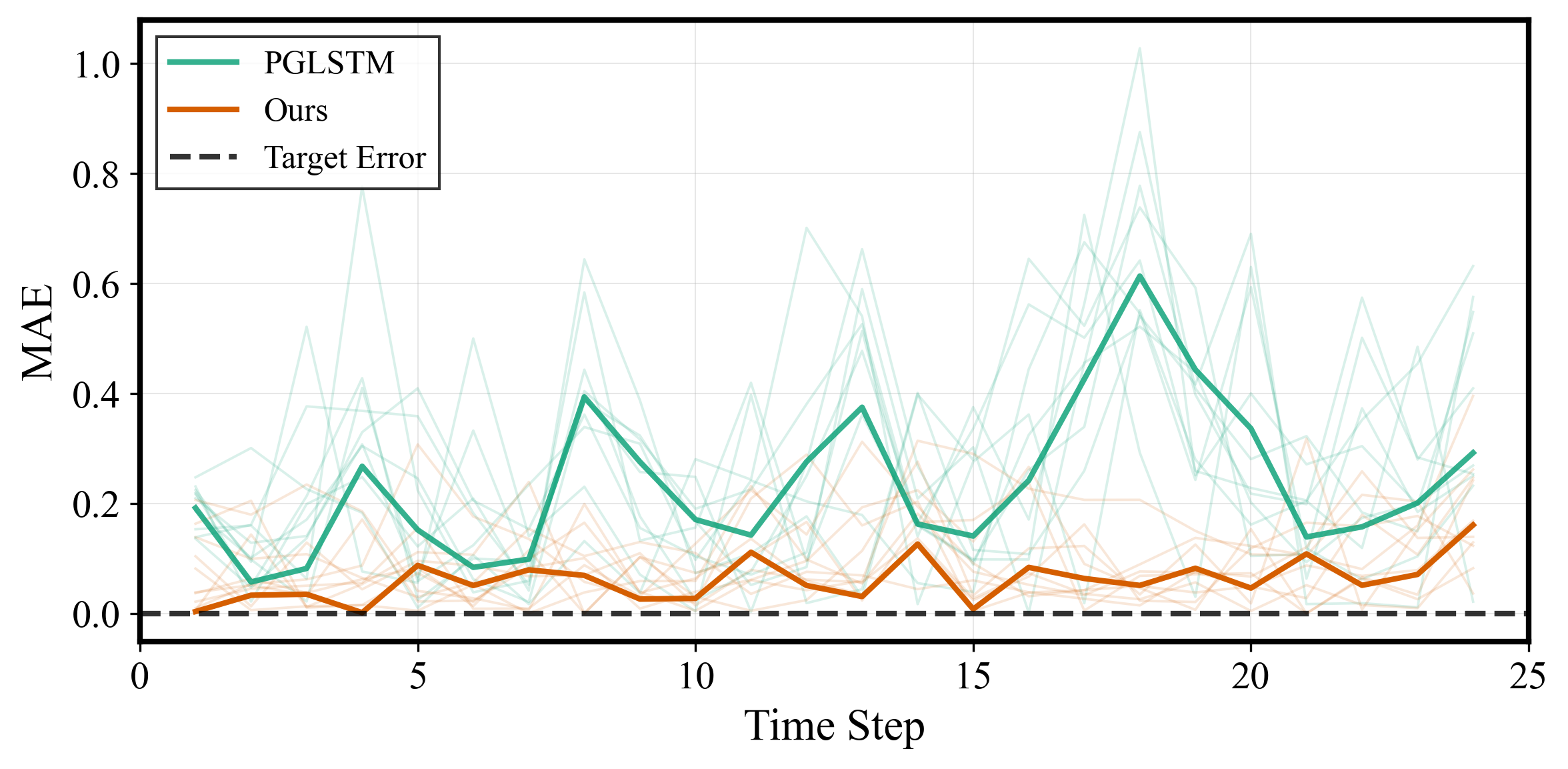} 
        \caption{Weather }
        \label{fig:errors_Weather}
    \end{subfigure}
    \begin{subfigure}[b]{0.32\textwidth}
        \centering
        \includegraphics[width=\textwidth]{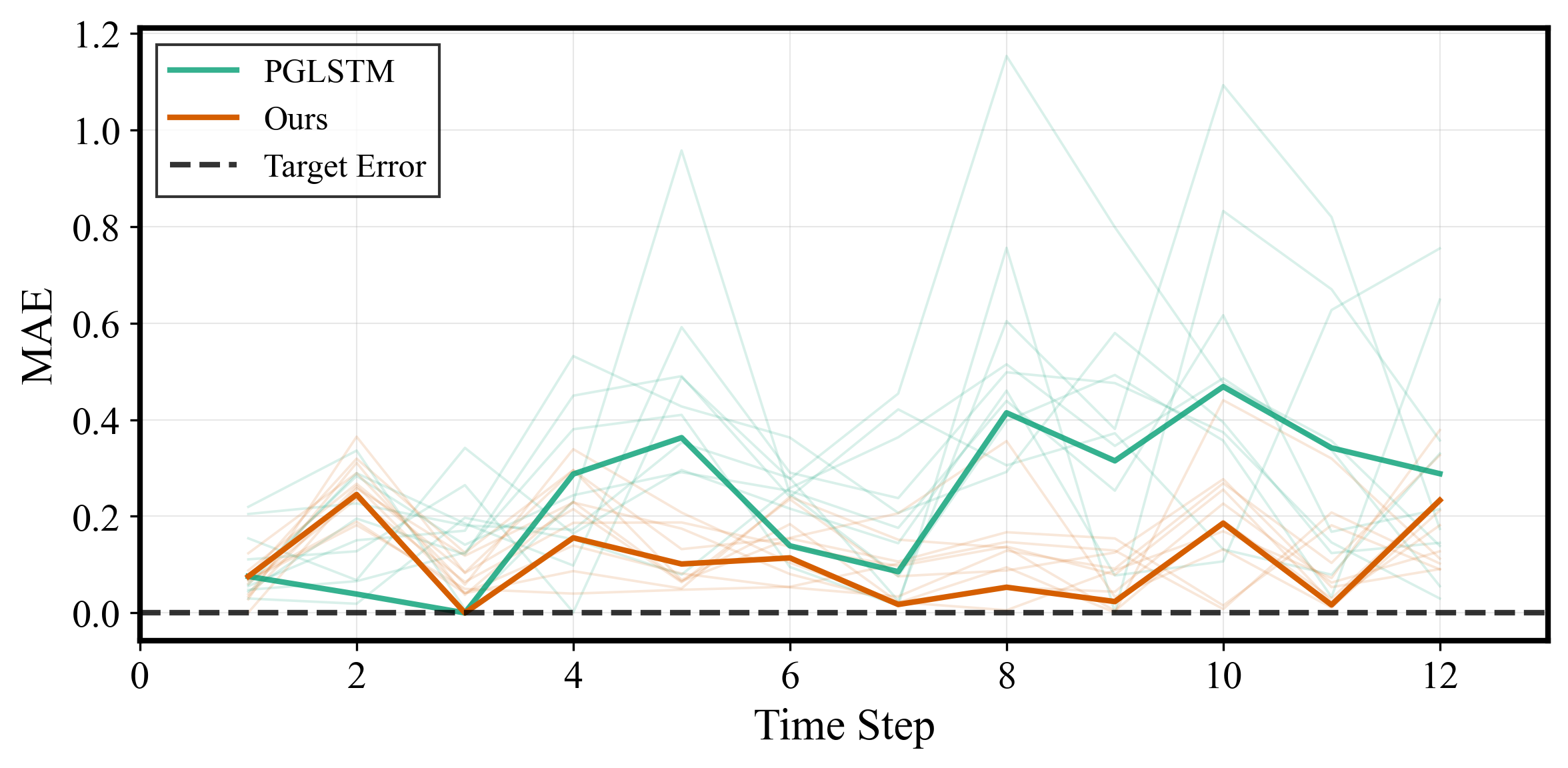} 
        \caption{ILI }
        \label{fig:errors_ILI}
    \end{subfigure}
    
    \caption{Per-step MAE performance of RRE-PPO4Pred and the best baseline method with the xLSTM backbone at horizon 12 for the ILI dataset and horizon 24 for the remaining datasets. Semi-transparent lines represent individual executions, while bold lines indicate averaged MAE of 10 independent runs.}
    \label{fig:errors}
\end{figure}

\begin{figure}[htbp]
    \centering
    \begin{subfigure}[b]{0.32\textwidth}
        \centering
        \includegraphics[width=\textwidth]{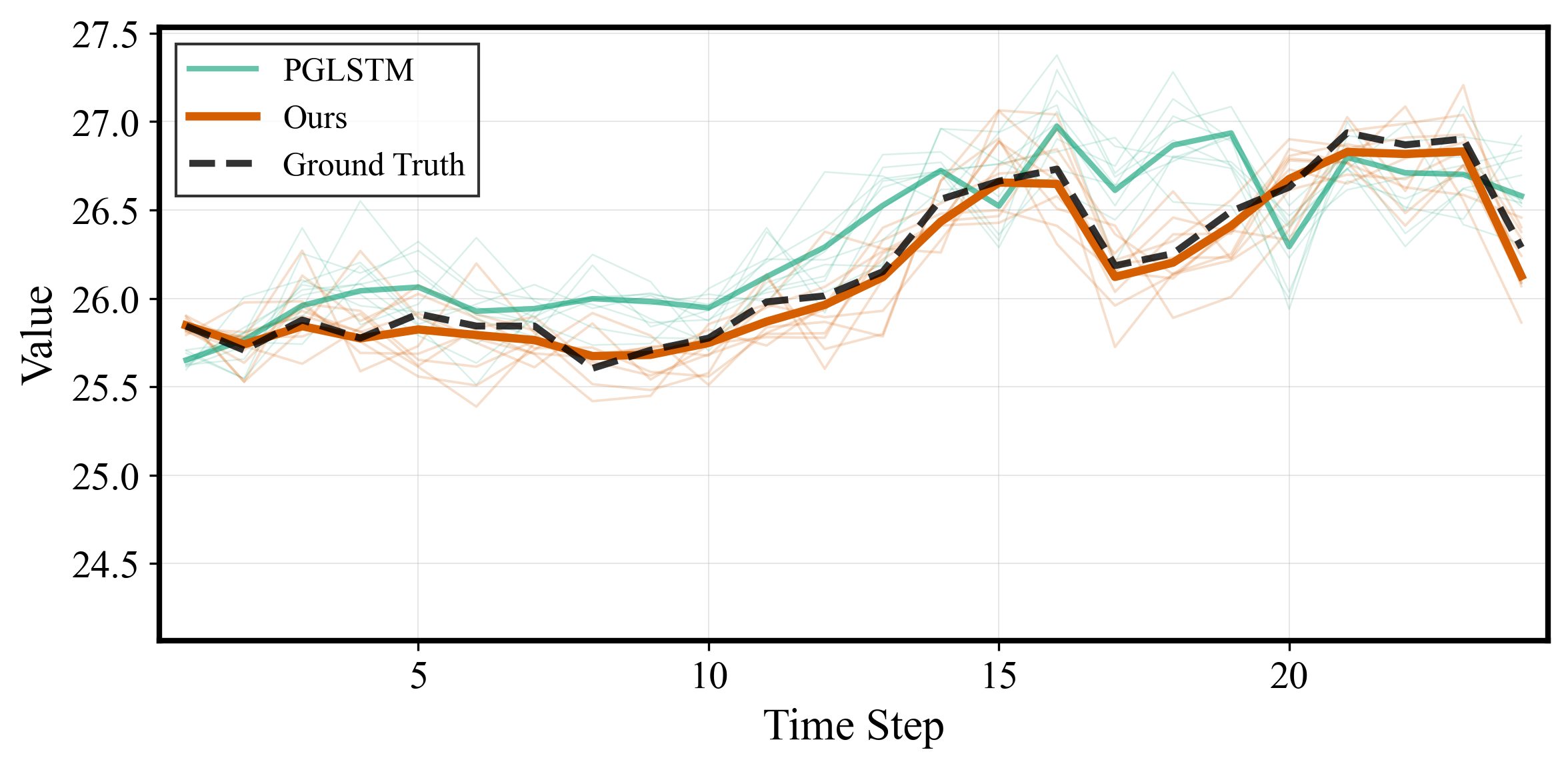} 
        \caption{$H$=24  }
        \label{fig:Weather_24}
    \end{subfigure}
    \begin{subfigure}[b]{0.32\textwidth}
        \centering
        \includegraphics[width=\textwidth]{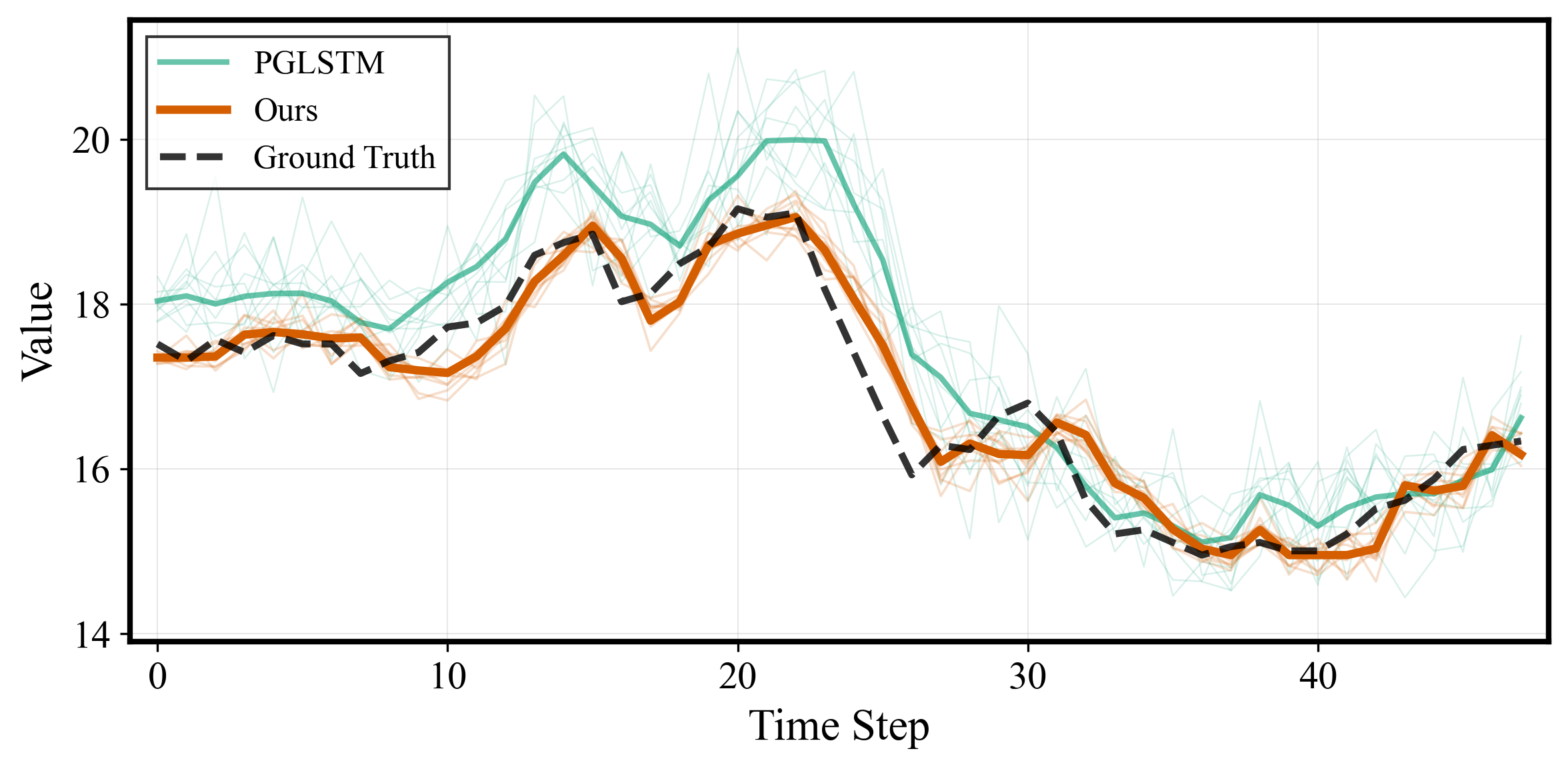} 
        \caption{$H$=48 }
        \label{fig:Weather_48}
    \end{subfigure}
        \begin{subfigure}[b]{0.32\textwidth}
        \centering
        \includegraphics[width=\textwidth]{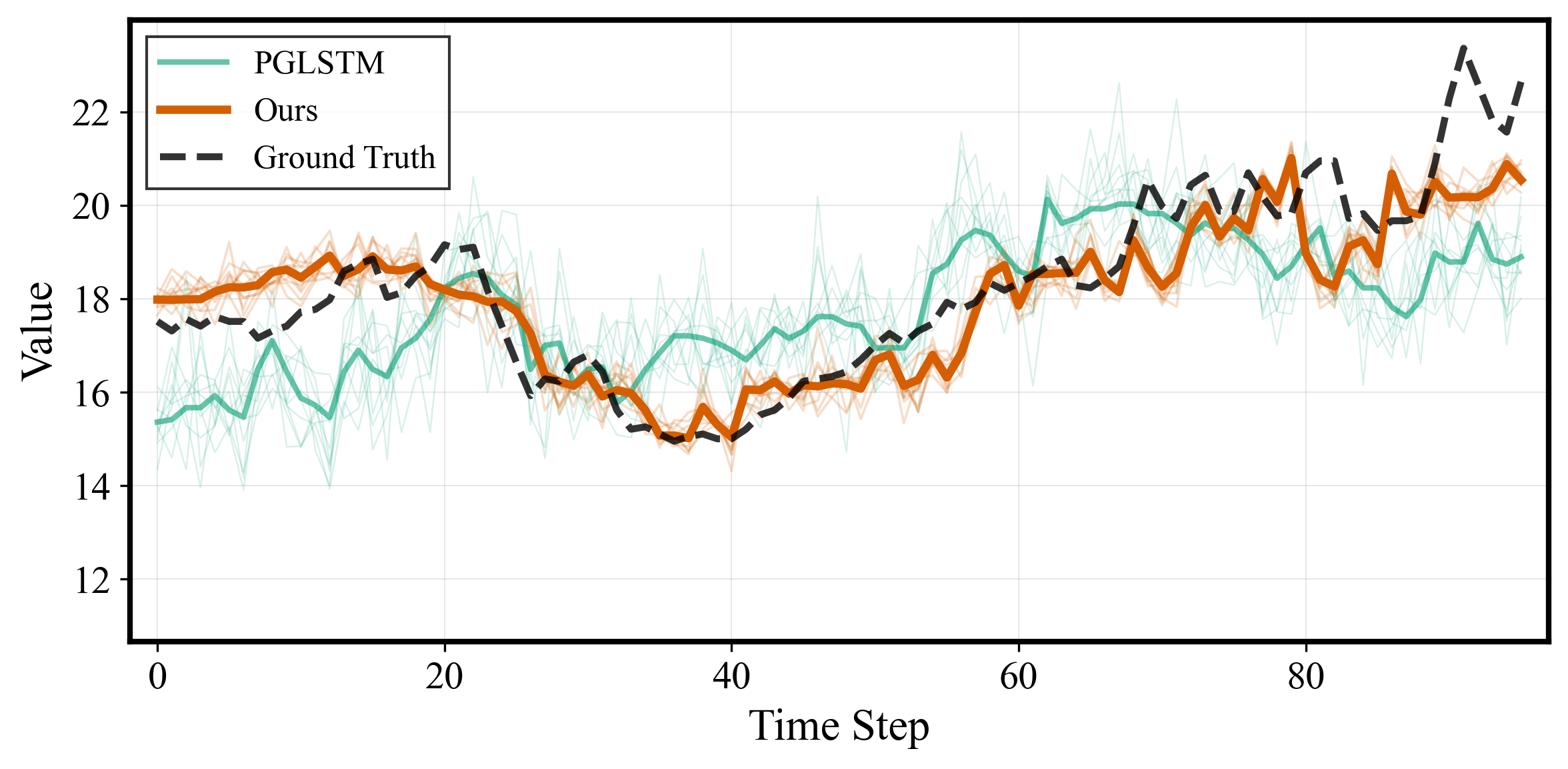} 
        \caption{$H$=96 }
        \label{fig:Weather_96}
    \end{subfigure}

    \caption{Illustration of prediction curves by the proposed RRE-PPO4Pred method and the best baseline with the xLSTM backbone on the Weather dataset. Each semi-transparent line represents an individual execution, while the bold lines denote the averaged predictions of all runs.}
    \label{fig:slices}
\end{figure}

To comprehensively evaluate the generality of our method across heterogeneous RNN architectures, we implement RRE-PPO4Pred and baselines across a diverse set of RNN backbones.
This experimental design is intended to demonstrate that our approach can integrate seamlessly with different RNN variants without requiring architecture-specific modifications.
The selected RNN backbone models and their key mechanisms are summarized in \cref{tab:rnn_variants}.
These backbones are categorized into two groups: (1) basic RNNs, comprising vanilla RNN \citep{elman1990finding}, MGU \citep{zhou2016minimal}, GRU \citep{cho2014learning}, and LSTM \citep{hochreiter1997long}; and (2) advanced RNNs, including IndRNN \citep{Indrnn}, phLSTM \citep{gers2000learning}, pLSTM \citep{neil2016phased}, and xLSTM \citep{beck2024xlstm}, which incorporate specialized mechanisms for enhanced gradient propagation, temporal modeling, or architectural scalability. 
This setting enables systematic evaluation of whether the efficacy of our approach derives from its intrinsic decision optimization advantages rather than from specific structural properties of particular RNN variants.

\subsubsection{Baseline methods}

To evaluate the superiority of our RRE-PPO4Pred against the related prior works, we conduct comparative analysis against several representative baseline methods spanning fixed, heuristic, and reinforcement learning categories.
The utilized baseline methods are summarized as follows:

\begin{itemize}
\item Naive-all: A conventional encoder-only RNN method that utilizes outputs from all time steps for supervised training, representing the standard configuration without selective supervision.
\item Naive-last: A conventional encoder-only RNN method that uses the final hidden state as its output. This method serves as a basic baseline with a fixed output target selection policy.
\item DilatedRNN \citep{chang2017dilated}: A fixed recurrent method with  predetermined dilated hidden skip connections\footnote{https://github.com/code-terminator/DilatedRNN}, representing a fixed hidden skip connection strategy.
\item EBPSO \citep{tijjani2024enhanced}: An Enhanced Binary Particle Swarm Optimization approach for input feature selection, representing a dynamic heuristic strategy for feature selection.
\item LSTMjump \citep{yu2017learning}: A Policy Gradient-based method that employs PG approach \citep{sutton2000policy} to skip certain input time steps, representing an RL-based approach for input feature selection.
\item PGLSTM \citep{weerakody2023policy}: A PG-based method that applies the PG algorithm to dynamically add extra hidden state connections between different time steps, representing an RL-based policy for hidden skip connections.
\end{itemize}

Note that Naive-last and Naive-all use vanilla configurations that utilize the complete input time series without feature selection and employ conventional sequential hidden state transitions without skip connection mechanisms.
Furthermore, while DilatedRNN, LSTMjump, and PGLSTM are originally proposed for specific RNN architectures, we adapt their respective methodologies to heterogeneous RNN backbones while maintaining their originally designed action spaces, thereby ensuring consistent evaluation across all architectures.

\subsection{Implementation details}

In all experiments, each dataset is scaled to $[-1, 1 ]$ by the min-max method, and is split into training set, validation set, and testing set with a ratio of 7:1:2. 
All methods are trained on the training set using the Adam optimizer, with the maximum number of training epochs set to 50 for agent networks and 20 for the RNN environment network. 
An early stopping mechanism is employed through cross-validation on the validation set with a patience of 6 epochs to save the best performing RNN model across all methods.
During evaluation on the testing set, predictions are transformed back to the original scale and then evaluated using the MSE and MAE metrics. 
All experiments are run 10 times with distinct random seeds on an NVIDIA RTX A4500 20GB GPU, and the mean performance across runs is reported.

For configuring the prediction tasks,
our experimental setup follows the general configuration strategy \citep{nie2022time} with moderate adjustments. Specifically, since the ILI data size is smaller than others, we set the input lag order $T$ to $48$ for the ILI dataset and $T=96$ for the others.
Correspondingly, the prediction horizons $H$ are $\{12, 24, 48\}$ for ILI and $\{24, 48, 96\}$ for the remaining datasets, enabling evaluation across multiple forecasting ranges. 
This setting provides a practical balance between short-term and long-term forecasting evaluation while preserving comparability with existing RNN-based benchmarks and enabling fair assessment of our proposed method's effectiveness across diverse temporal scales.

We select the hyperparameters for our method and all baseline methods via random search \citep{bergstra2012random} to ensure fair comparison. 
For each combination of RNN backbone and horizon setting, independent hyperparameter searches are conducted to identify the optimal learning rate and hidden dimension for both our RRE-PPO4Pred method and all baselines.
For all adopted RNN backbones, the learning rate $\eta_\theta$ of the RNN predictor is sampled from a log-uniform distribution Log-U $[10^{-5}, 10^{-1}]$.
The hidden dimension of RNN backbones $\mathcal{F}_\theta$ is sampled from a discrete uniform distribution over $\{10, 20, \ldots, 200\}$.
Other method-specific hyperparameters of the baselines are configured according to their original papers.
For our RRE-PPO4Pred, we search over the same learning rate space for agent ($\eta_\pi$ and $\eta_\upsilon$) and the RNN predictor, and further implementation details are provided in the \ref{sec:para}.
To ensure reproducibility, the code and complete hyperparameter configurations are provided in \url{https://github.com/Laixin233/-Reinforced-Encoder}.

\section{Experimental analysis}\label{experiment}

\subsection{Main results\label{sec:mainresults}}

To evaluate the predictive performance of RRE-PPO4Pred and all baseline methods, we report the comparative results in terms of MSE and MAE metrics across five real-world time series datasets at multiple forecasting horizons.
The averaged MSE and MAE values for all methods are presented in \cref{tab:traffic-performance}--\cref{tab:ILI-performance}, where lower values indicate superior forecasting accuracy. The best results are highlighted in bold.
We further calculate two comprehensive evaluation metrics, reported in the last two rows of \cref{tab:traffic-performance}--\cref{tab:ILI-performance}. The ``Average'' denotes the mean error across all three forecasting horizons and eight adopted RNN backbones for each metric. The ``Improvement'' represents the relative improvement rate of our method compared to the baseline, computed with the average value as $(\text{baseline} - \text{ours}) / \text{baseline}$. 

Specifically, compared to the static RNN baselines (Naive-last, Naive-all, and DilatedRNN), our RRE-PPO4Pred achieves MSE improvements ranging from 23.7\% to 53.08\% and MAE improvements from 21.57\% to 50.75\%.
Compared to the heuristic-based EBPSO method, our approach achieves MSE improvements of 23.21\%--28.36\% and MAE improvements of 22.80\%--29.59\%.
Compared with RL-based methods (PGLSTM and LSTMjump), our method still achieves significant improvements, by 19.04\%--27.80\% in MSE and 18.47\%--28.49\% in MAE.
This demonstrates consistent superiority of our method across all baseline categories.

To provide a more intuitive comparison, \cref{fig:horison-mse} presents the MSE results obtained by averaging performance across all backbones for our RRE-PPO4Pred method and other baselines on five datasets. For brevity, the corresponding MAE results are provided in \ref{sec:deti}. 
The comparison results between our RRE-PPO4Pred and the selected baselines 
yield the following conclusions:

\begin{itemize}

\item Our RRE-PPO4Pred consistently outperforms all baselines across all experimental conditions. For average MSE, it surpasses the best baseline by 19.75\%, 23.57\%, 23.92\%, 25.53\%, and 19.04\% on the Traffic, Electricity, ETTh, Weather, and ILI datasets, respectively. For average MAE, improvements of 19.60\%, 18.47\%,  25.34\%, 19.6\%, and 20.19\% over the best baseline are achieved. This consistent superiority demonstrates the effectiveness and generalizability of the proposed approach.

\item In the comparison between static strategies (Naive-last, Naive-all, and DilatedRNN), DilatedRNN demonstrates superior performance over Naive-last and Naive-all through its predetermined dilated skip connections, highlighting the necessity of incorporating hidden skip connections in encoder-only RNNs to capture long-range temporal dependencies, even with fixed patterns.

\item Comparing the static skip connection method (DilatedRNN) with the dynamic RL-based approach (PGLSTM), PGLSTM achieves superior performance in most cases, with a noticeably smaller enclosed area in the radar chart. This indicates that RL-based dynamic skip connections, which adapt to dataset-specific temporal patterns, are more effective than fixed dilation patterns.

\item Comparing the heuristic-based input feature selection approach (EBPSO) with the RL-based approach (LSTMjump), these two methods show comparable performance across different datasets. Neither approach shows consistent superiority, underscoring the challenge of input feature selection and the necessity for more effective optimization algorithm.

\item In comparison between the RL-based baselines (LSTMjump and PGLSTM), PGLSTM shows superior performance over LSTMjump. Although both methods leverage RL techniques, PGLSTM optimizes hidden state skip connections while LSTMjump optimizes input features. This comparative result suggests that skip connection optimization provides greater individual benefits than input selection alone. However, our RRE-PPO4Pred, which jointly optimizes both mechanisms along with output targeting, achieves substantially greater improvements than either approach, demonstrating the value of unified optimization.

\end{itemize}


The violin plots in \cref{fig:violin-mse} display the mean and variance of MSE for the proposed RRE-PPO4Pred method and baseline methods with xLSTM backbone at horizon 12 for ILI and horizon 24 for other datasets. Notably, our method consistently attains both the lowest mean and substantially reduced variance across all datasets, indicating superior prediction stability compared to baseline approaches.  These observations are further confirmed by the MAE results presented in \ref{sec:deti}. 

To comprehensively evaluate the predictive performance of our proposed RRE-PPO4Pred method, \cref{fig:errors} presents the comparative per-step MAE performance of RRE-PPO4Pred and the best baseline method (PGLSTM) with xLSTM backbone across different datasets.
The results reveal that our method exhibits more stable error patterns across time steps, with prediction errors remaining consistently close to target thresholds throughout the forecasting horizon.

Furthermore, \cref{fig:slices} illustrates the prediction trajectories of RRE-PPO4Pred and PGLSTM with xLSTM backbone for $H=24$, $H=48$, and $H=96$ to visualize this advantage across different horizons.
As shown in \cref{fig:slices}, our method smoothly tracks ground truth patterns across multiple runs, while PGLSTM predictions exhibit greater divergence from target values and higher variability across executions, especially during periods of rapid value changes. 
Additional comparisons with other baseline methods are provided in \ref{sec:deti}.
These results highlight that our method achieves superior and more stable performance across different forecasting horizons, making it highly suitable for practical multi-step-ahead prediction tasks.




\subsection{Ablation study}

To justify the individual contributions of the proposed RRE framework and PPO4Pred algorithm, we conduct an ablation study with the variants described in \cref{tab:variant}.
We design TA-PSO as a heuristic-based variant that employs the PSO algorithm \citep{kennedy1995particle} to optimize our designed Ternary Action (TA) from \cref{sec:stateandaction}.
Additionally, we develop three RL-based ablative variants, RRE-PG, RRE-DQN, and RRE-PPO,  which  retain the RRE framework but replace the PPO4Pred algorithm with alternative RL algorithms. 
RRE-PG uses vanilla Policy Gradient \citep{sutton2000policy}, a foundational policy-based RL method without value function estimation.
RRE-DQN uses Deep Q-Network \citep{mnih2015human}, a value-based RL method that learns action-value functions.
RRE-PPO uses standard PPO \citep{schulman2017proximal} without our prediction-oriented modifications.
These RL-based variants isolate the contribution of our PPO4Pred algorithm by maintaining the same action space and framework while varying only the optimization method.

\begin{table}[t]
\centering
\caption{Component configurations of our RRE-PPO4Pred method and its ablative methods.}
\label{tab:variant}
\begin{tabular}{lccc}
\toprule
\multicolumn{1}{l}{Method} &  Ternary Action & RecEnc & PPO4Pred \\ \midrule
TA-PSO                   & \cmark  & \xmark   & \xmark   \\
RRE-PG                   & \cmark  & \cmark  & \xmark   \\
RRE-DQN                  & \cmark  & \cmark  & \xmark   \\
RRE-PPO               & \cmark  & \cmark  & \xmark   \\
RRE-PPO4Pred    & \cmark  & \cmark  & \cmark  \\ \bottomrule
\end{tabular}%
\end{table}
\begin{table}[t]
\centering
\caption{Average MSE and MAE across all RNN backbones with corresponding improvements of our RRE-PPO4Pred over ablation variants.}
\label{tab:ablation}
\resizebox{\textwidth}{!}{
\begin{tabular}{ll|ccccc|ccccc}
\toprule
\multirow{2}{*}{Dataset} & \multirow{2}{*}{Metric} & 
  \multicolumn{5}{c|}{MSE} &
  \multicolumn{5}{c}{MAE} \\
\cmidrule(lr){3-7} \cmidrule(lr){8-12}
 & & TA-PSO & RRE-PG & RRE-DQN & RRE-PPO & Ours & TA-PSO & RRE-PG & RRE-DQN & RRE-PPO & Ours \\ 
\midrule
\multirow{2}{*}{Traffic} 
& Average & 0.3088 & 0.2494 & 0.2440 & 0.2459 & \textbf{0.2316} & 0.4062 & 0.3088 & 0.3021 & 0.2972 & \textbf{0.2810} \\
& Improvement & 24.99\% & 7.14\% & 5.08\% & 5.79\% & - & 30.81\% & 9.01\% & 6.98\% & 5.45\% & - \\
\midrule
\multirow{2}{*}{Electricity} 
& Average & 0.2250 & 0.1754 & 0.1699 & 0.1674 & \textbf{0.1579} & 0.3363 & 0.2799 & 0.2677 & 0.2689 & \textbf{0.2533} \\
& Improvement & 29.85\% & 10.00\% & 7.07\% & 5.71\% & - & 24.69\% & 9.52\% & 5.40\% & 5.81\% & - \\
\midrule
\multirow{2}{*}{ETTh} 
& Average & 0.4743 & 0.3740 & 0.3578 & 0.3635 & \textbf{0.3396} & 0.5406 & 0.4268 & 0.4084 & 0.4159 & \textbf{0.3818} \\
& Improvement & 28.40\% & 9.21\% & 5.10\% & 6.58\% & - & 29.37\% & 10.54\% & 6.50\% & 8.18\% & - \\
\midrule
\multirow{2}{*}{Weather} 
& Average & 0.3149 & 0.2293 & 0.2166 & 0.2180 & \textbf{0.1984} & 0.3882 & 0.2950 & 0.2894 & 0.2908 & \textbf{0.2776}  \\
& Improvement & 37.01\% & 13.49\% & 8.43\% & 9.01\% & -  & 28.50\% & 5.90\% & 4.10\% & 4.54\% & - \\
\midrule
\multirow{2}{*}{ILI} 
& Average & 2.9101 & 2.5155 & 2.4430 & 2.3931 & \textbf{2.1790} & 1.5394 & 1.1996 & 1.1865 & 1.1416 & \textbf{1.0583} \\
& Improvement & 25.12\% & 13.38\% & 10.81\% & 8.95\% & - & 31.25\% & 11.78\% & 10.80\% & 7.30\% & - \\
\bottomrule
\end{tabular}
}
\end{table}

To comprehensively evaluate the effectiveness of each component, all ablative variants are tested across all selected RNN backbones, forecasting horizons, and five benchmark datasets. The average MSE and MAE, along with the relative improvement of RRE-PPO4Pred over each variant, are presented in \cref{tab:ablation}, where the best results are highlighted in bold. Detailed results of all ablative methods with different backbones for each forecasting horizon are provided in \ref{sec:deti}.  
These results lead to the following conclusions:

\begin{itemize}
\item  The proposed RRE-PPO4Pred consistently achieves the best average performance among all variants on both metrics over the five datasets and three forecasting horizons. The improvements over ablation variants range from 5.08\% to 37.01 \% in MSE and 4.10\% to 31.25\% in MAE, with the largest gains observed over the heuristic baseline (TA-PSO) and substantial gains even over strong RL baselines. These results confirm that jointly integrating all proposed components is critical for achieving superior forecasting performance.

\item In comparison between TA-PSO and RL-based ablative methods (RRE-PG, RRE-DQN, RRE-PPO), all RL-based ablative methods  outperform TA-PSO in terms of average MSE and MAE across all datasets.
Even the simplest RL variant (RRE-PG) achieves substantial improvements over TA-PSO.
This confirms the necessity of reinforcement learning and the RRE framework for effectively optimizing the complex ternary action space  beyond what heuristic search strategy can achieve.

\item Among the three adopted RL algorithms, RRE-PG shows inferior performance compared to RRE-DQN and RRE-PPO, while RRE-DQN achieves comparable accuracy to RRE-PPO in most cases.
This performance hierarchy demonstrates that the complexity of our optimization problem requires sophisticated RL algorithms with advanced modifications, which justifies the necessity of employing a strong RL algorithm within our designed RRE framework, rather than relying on basic policy-based methods.

\item Compared with RRE-PPO, our RRE-PPO4Pred consistently outperforms it across all five datasets, achieving additional improvements of 5.71\%--8.95\% in MSE and 4.54\%--8.18\% in MAE. 
These consistent gains demonstrate the effectiveness of our prediction-oriented enhancements to standard PPO algorithm for enhancing encoder-only RNN-based predictors.

\end{itemize}

\subsection{Hyperparameter sensitivity and convergence analysis}

\begin{figure}[t]
    \centering
    \begin{subfigure}[b]{0.32\textwidth}
        \centering
        \includegraphics[width=\textwidth]{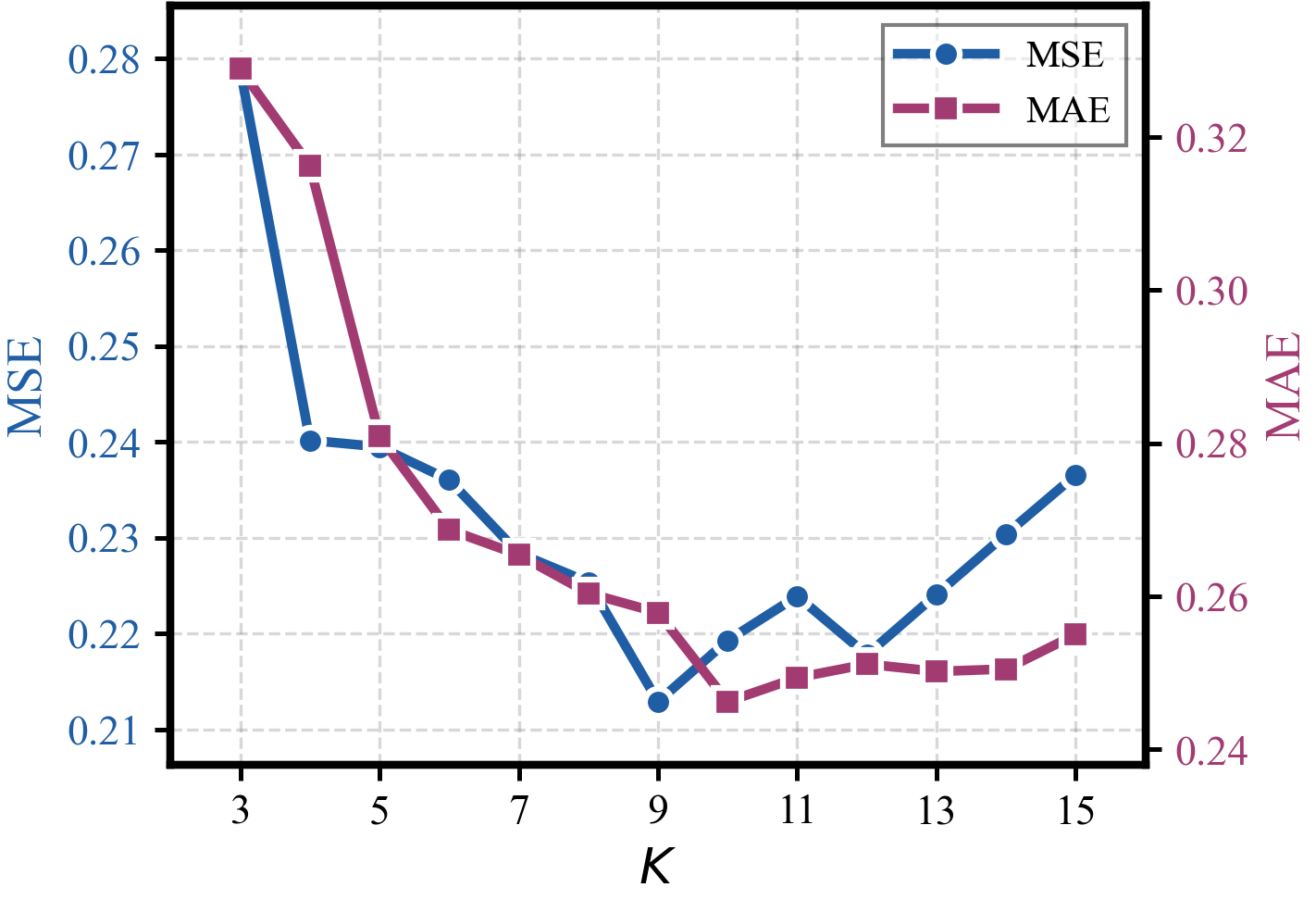} 
        \caption{Traffic }
        \label{fig:k-traffic}
    \end{subfigure}
    \begin{subfigure}[b]{0.32\textwidth}
        \centering
        \includegraphics[width=\textwidth]{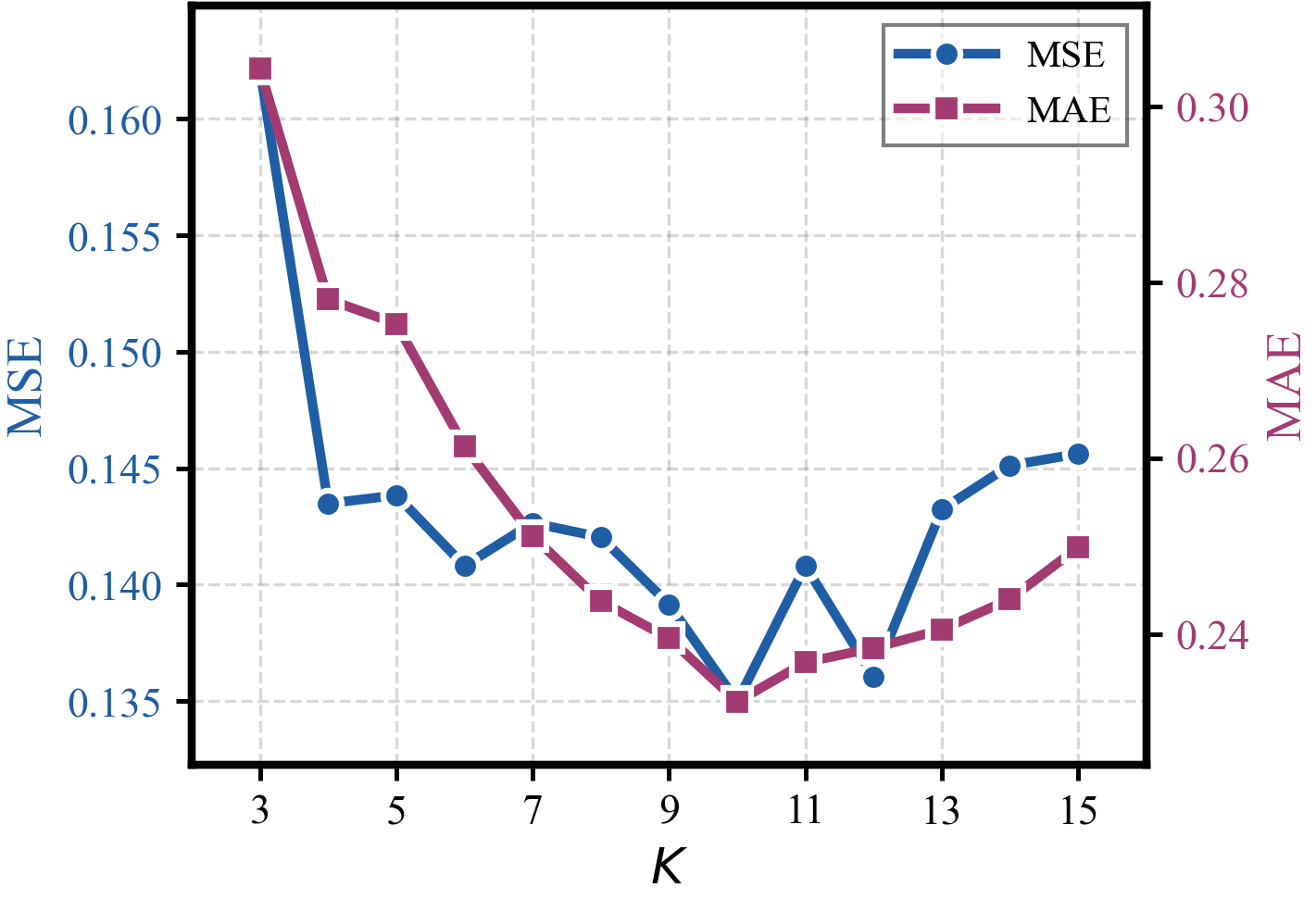} 
        \caption{Electricity }
        \label{fig:k-Electricity}
    \end{subfigure}
    \begin{subfigure}[b]{0.32\textwidth}
    \centering
    \includegraphics[width=\textwidth]{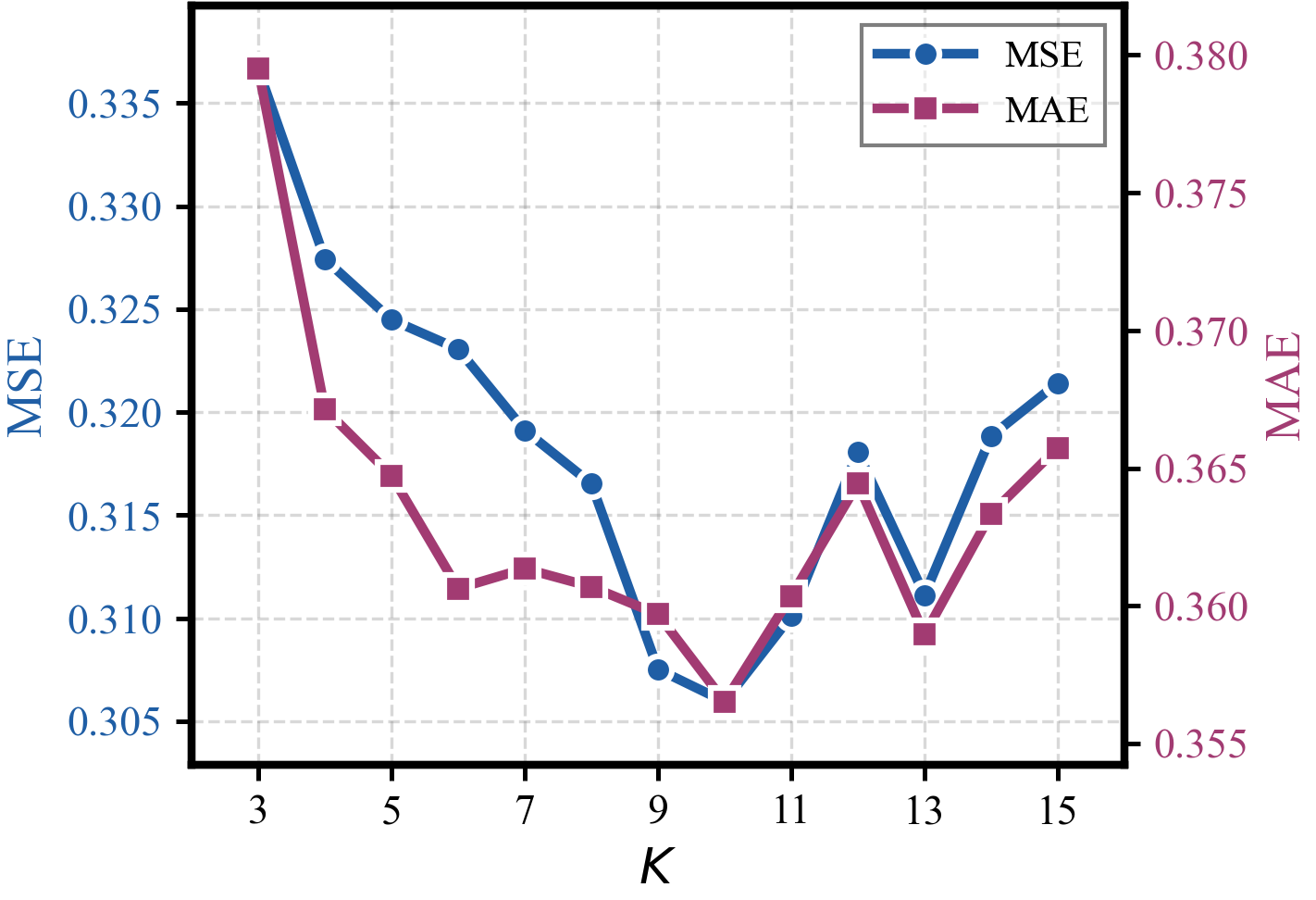} 
    \caption{ETTh }
    \label{fig:k-etth}
    \end{subfigure}
     \hfill 
    \vspace{3mm}
    \begin{subfigure}[b]{0.32\textwidth}
    \centering
    \includegraphics[width=\textwidth]{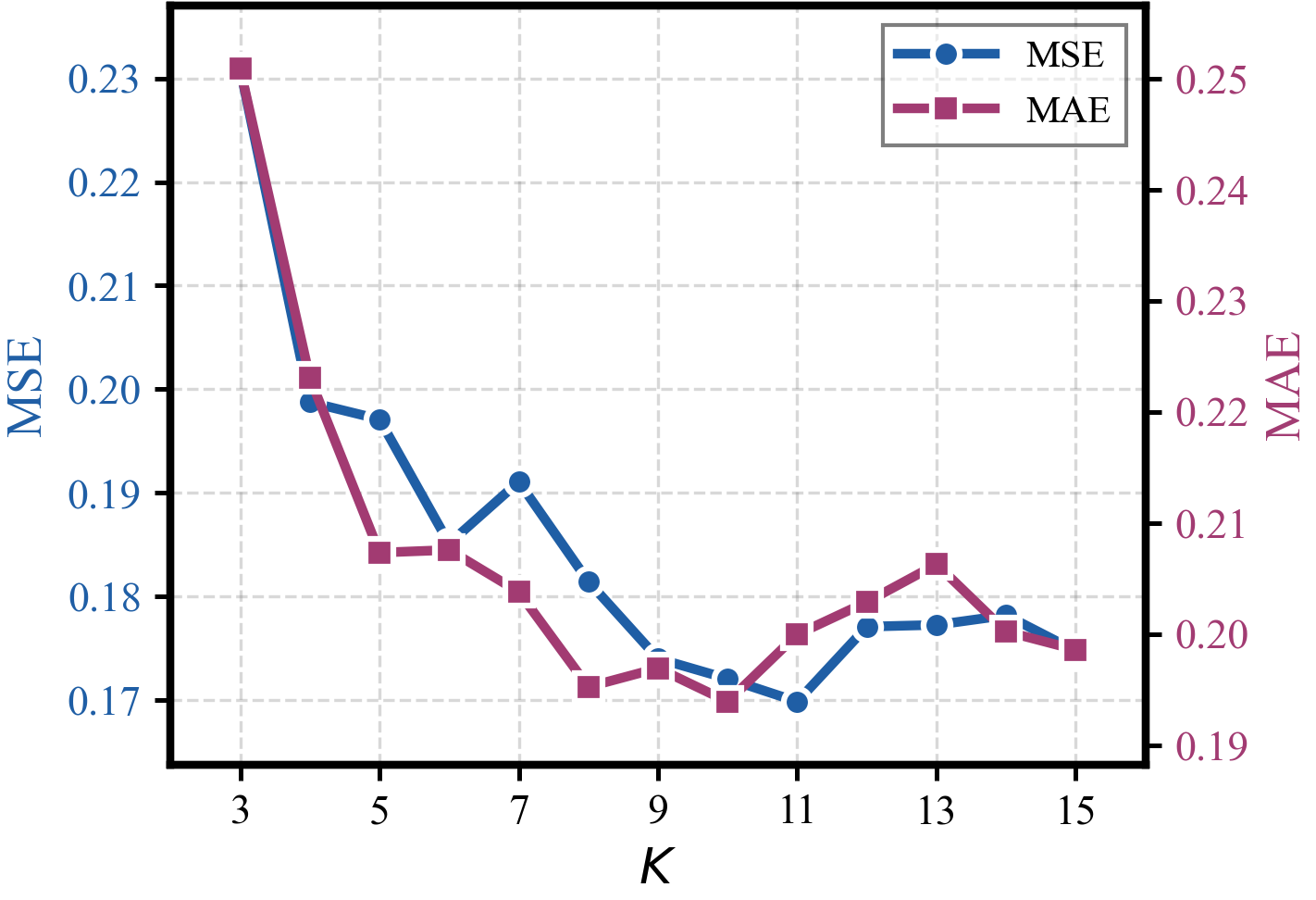}
    \caption{Weather }
    \label{fig:k-weather dataset}
    \end{subfigure}
    \begin{subfigure}[b]{0.32\textwidth}
    \centering
    \includegraphics[width=\textwidth]{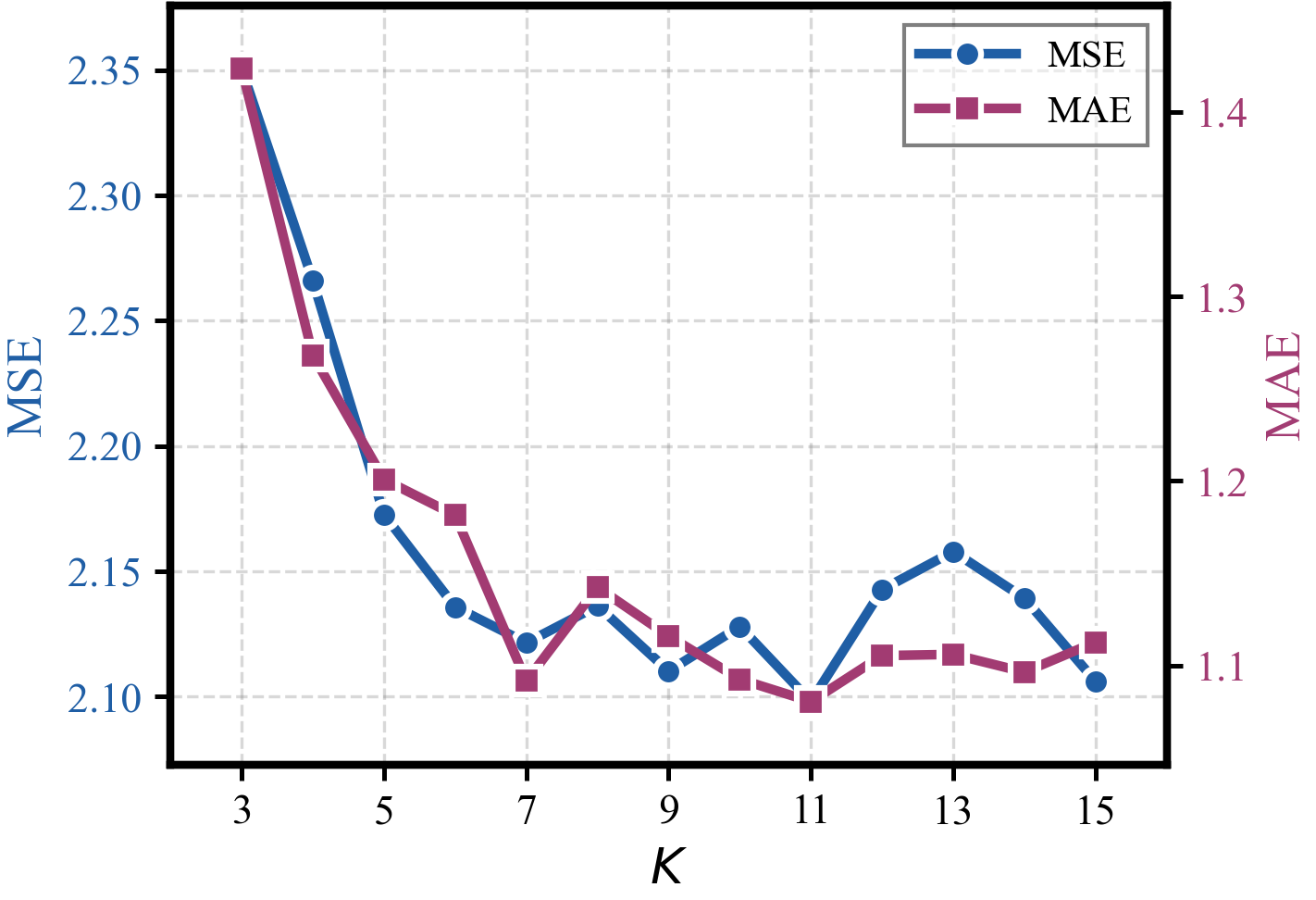} 
    \caption{ILI }
    \label{fig:k-ili}
    \end{subfigure}
    \caption{Performance sensitivity to historical state buffer size $K$ with xLSTM backbone. MSE (left y-axis, black) and MAE (right y-axis, red) are evaluated as $K$ ranges from 3 to 15 for RRE-PPO4Pred method at prediction horizon $H=24$.}
    \label{fig:k}
\end{figure}
In this section, we conduct a sensitivity analysis of the skip window size $K$ (defined in \cref{sec:stateandaction}) to investigate its impact on prediction performance. The skip window size $K$ determines the maximum temporal distance for skip connections in the encoder, controlling the range of historical hidden states that can be accessed at each time step.
\cref{fig:k} illustrates the performance of our RRE-PPO4Pred method with xLSTM backbone across five benchmark datasets as $K$ varies from 3 to 15 under the prediction horizon of $H=24$. Each subplot employs a dual-axis design: the left vertical axis represents MSE, the right vertical axis represents MAE, and the horizontal axis indicates the $K$ values tested.

\begin{figure}[t]
    \centering
    \begin{subfigure}[b]{0.32\textwidth}
        \centering
        \includegraphics[width=\textwidth]{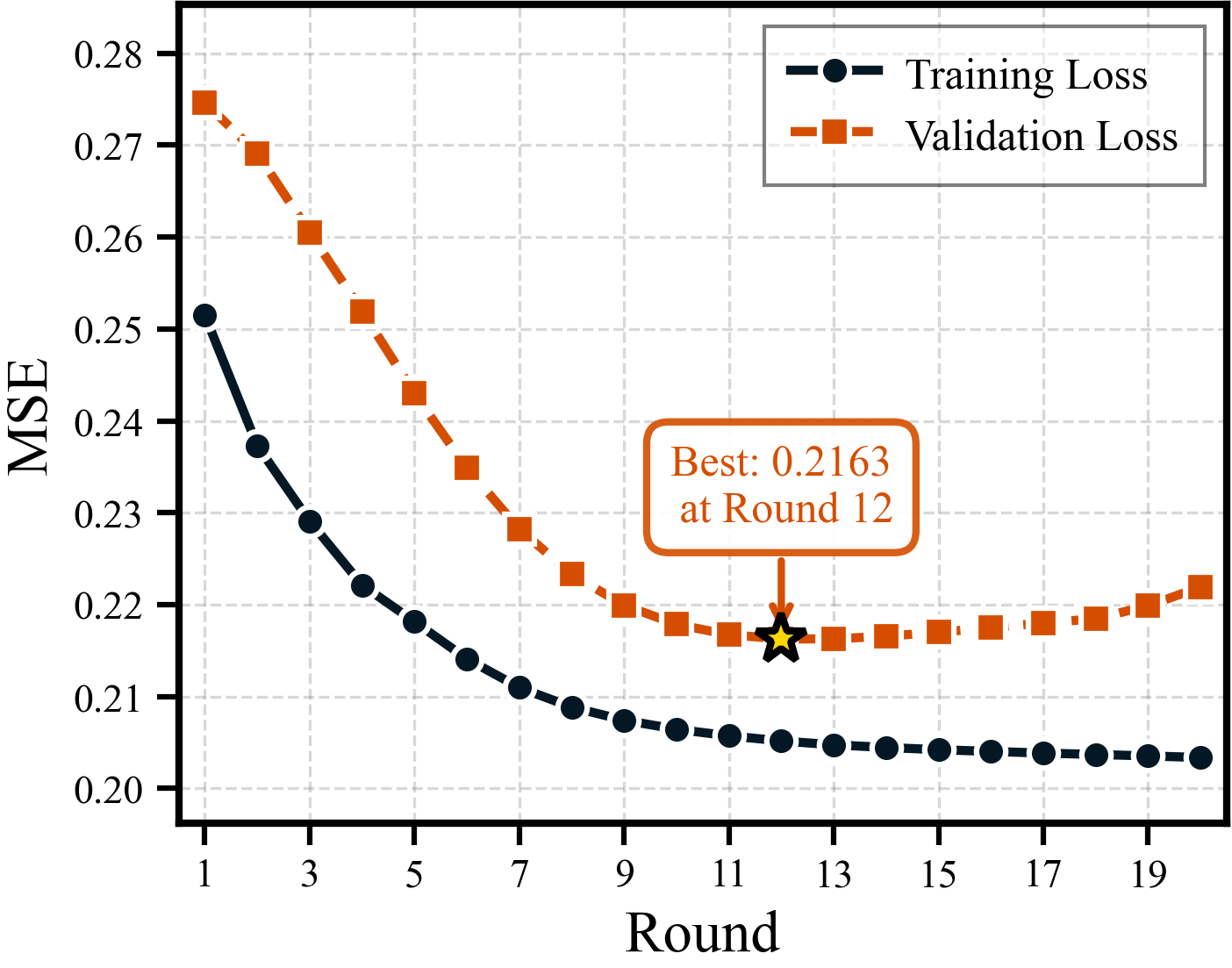} 
        \caption{Traffic }
        \label{fig:I-traffic}
    \end{subfigure}
    \begin{subfigure}[b]{0.32\textwidth}
        \centering
        \includegraphics[width=\textwidth]{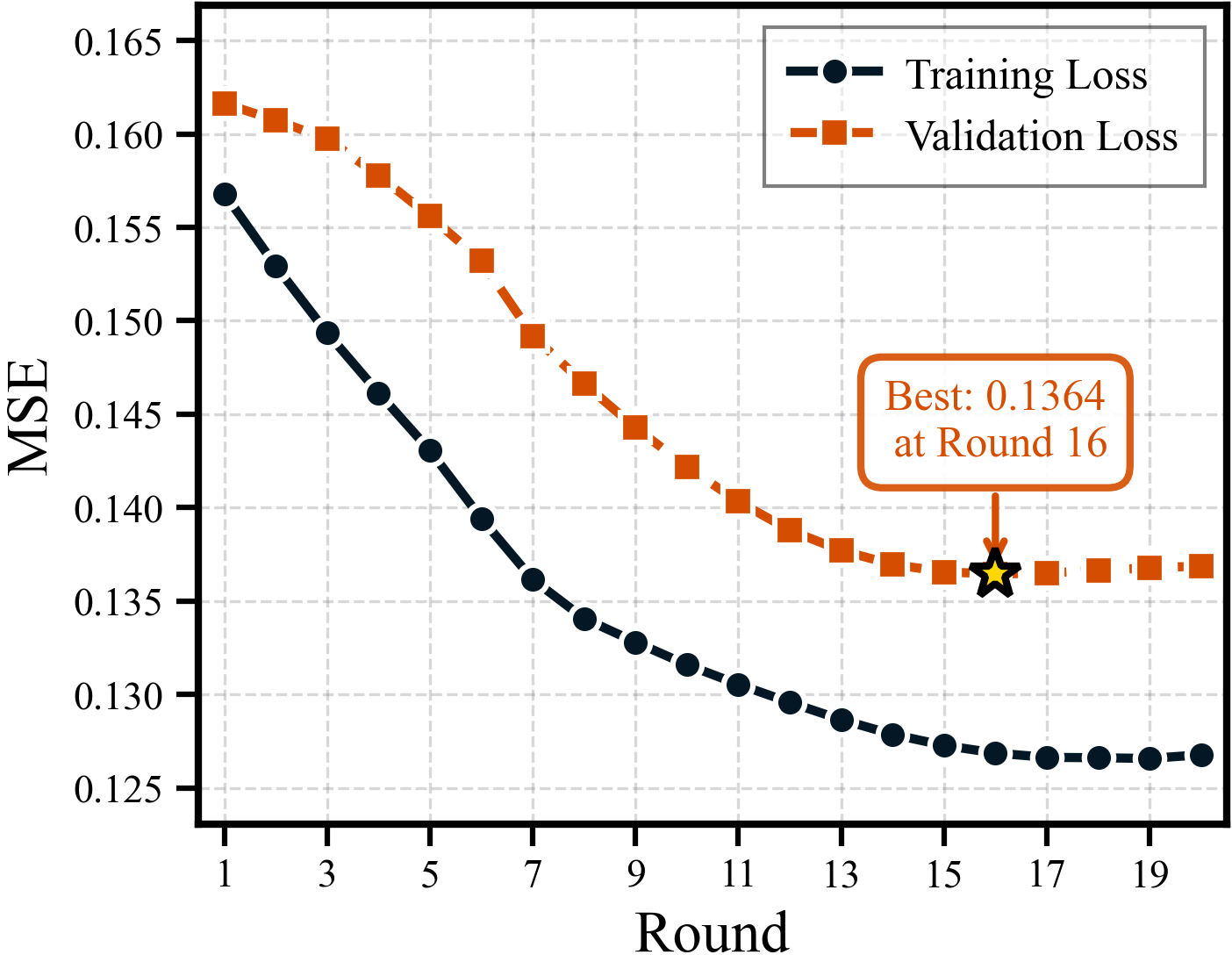} 
        \caption{Electricity }
        \label{fig:I-Electricity}
    \end{subfigure}
    \begin{subfigure}[b]{0.32\textwidth}
    \centering
    \includegraphics[width=\textwidth]{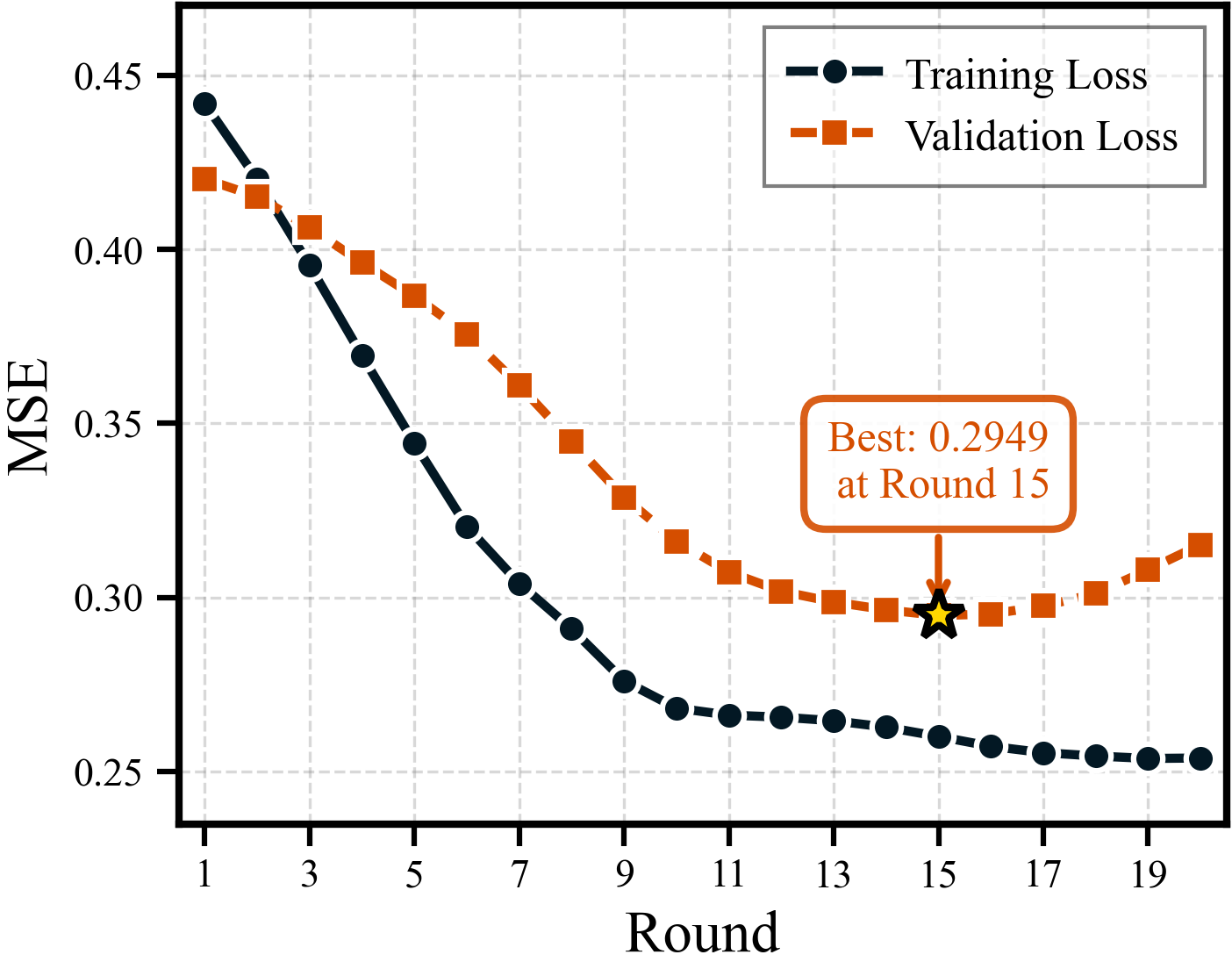} 
    \caption{ETTh }
    \label{fig:I-etth}
    \end{subfigure}

    \vspace{3mm}
    \begin{subfigure}[b]{0.32\textwidth}
    \centering
    \includegraphics[width=\textwidth]{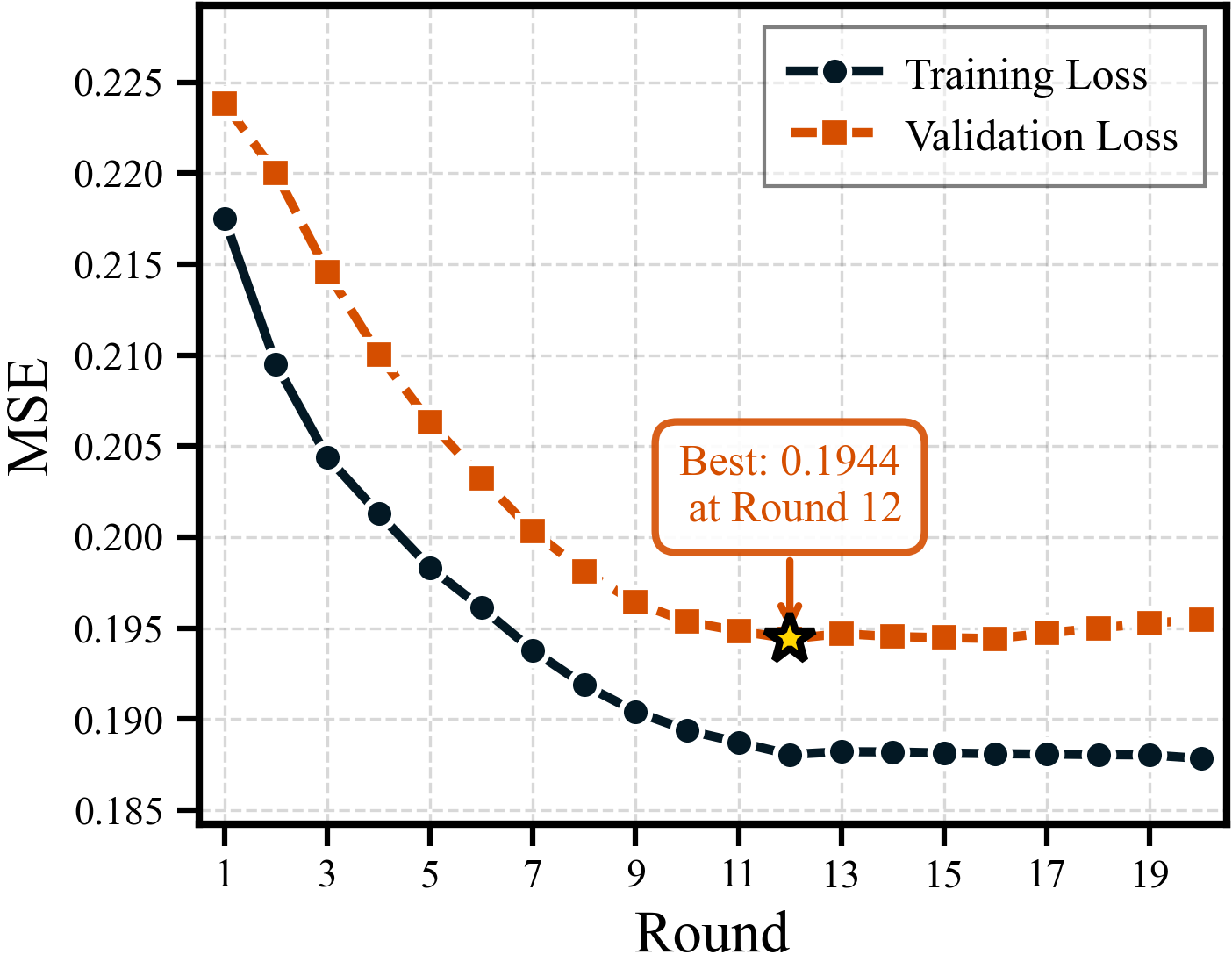}
    \caption{Weather }
    \label{fig:I-weather dataset}
    \end{subfigure}
    \begin{subfigure}[b]{0.32\textwidth}
    \centering
    \includegraphics[width=\textwidth]{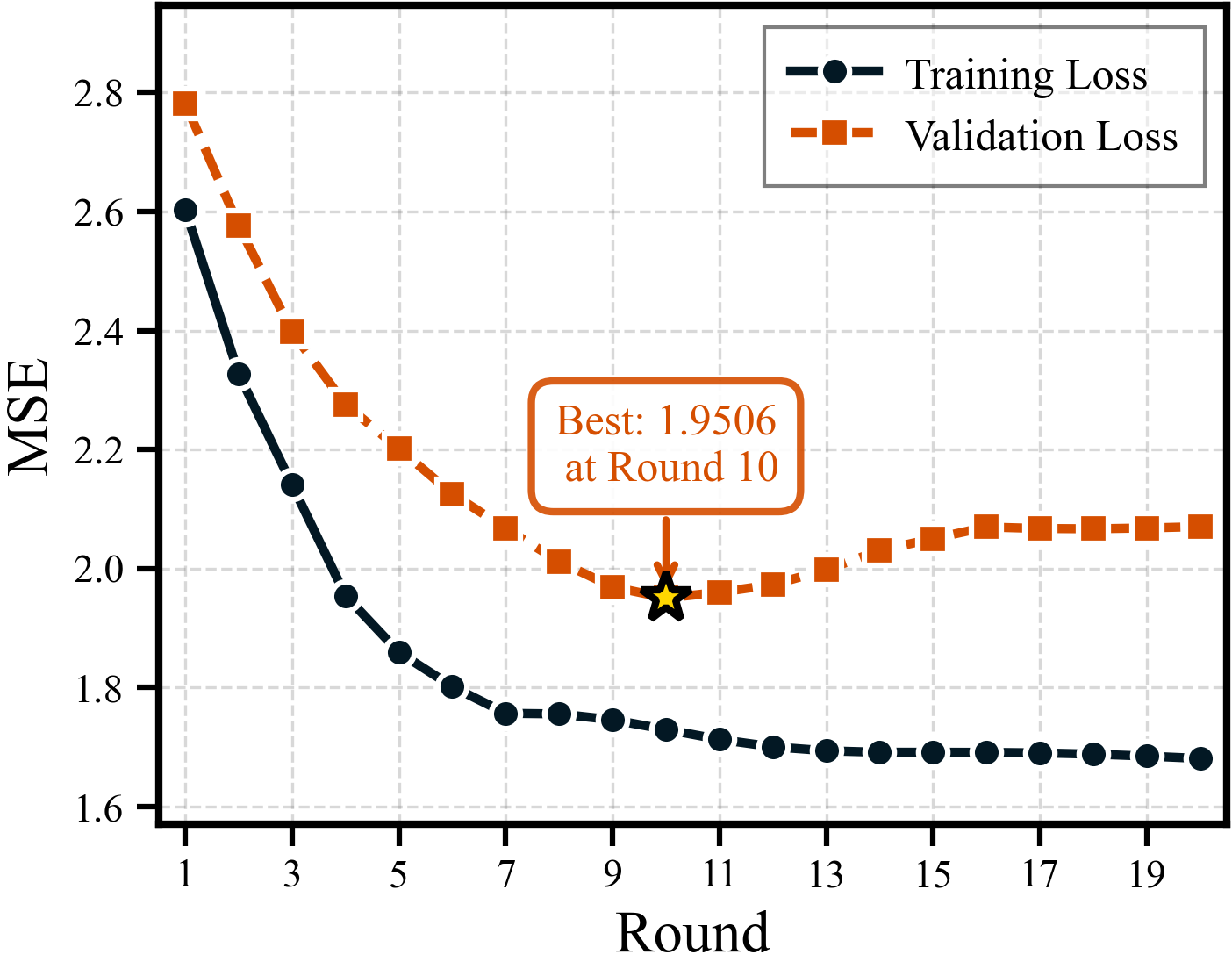} 
    \caption{ILI }
    \label{fig:I-ili}
    \end{subfigure}
    \caption{Training and validation loss convergence across 20 asynchronous training rounds. Black curves represent training MSE loss, and red curves represent validation MSE loss.}
    \label{fig:I}
\end{figure}

As shown in \cref{fig:k}, both MSE and MAE consistently decrease as $K$ increases from 3 to 8 across all five datasets, indicating that incorporating more historical hidden states through skip connections enhances the method's ability to capture long-range temporal dependencies. 
Beyond $K=8$ the performance stabilizes or shows only minor fluctuations beyond $K=10$ across most datasets, suggesting that excessively large skip window sizes provide diminishing returns. This saturation behavior indicates that the most relevant temporal dependencies for forecasting are captured within a window of approximately 8--10 time steps, and extending the window further does not significantly improve the model's ability to extract useful information.
 Based on these observations, a skip window size of $K=8$ to $K=10$ achieves a favorable balance between prediction performance and computational efficiency across most datasets. This range captures the essential temporal dependencies while avoiding the increased computational cost and potential overfitting associated with larger window sizes, which is adopted in this study.

As described in \cref{sec:Coevolutionary}, our training procedure adopts an asynchronous optimization strategy that alternates between agent training and predictor training across multiple rounds. To demonstrate the convergence behavior of this co-evolutionary process, we visualize the training and validation loss curves across 20 rounds of asynchronous training on five benchmark datasets  with xLSTM backbone and $H=24$, as presented in \cref{fig:I}. The black curves represent training loss (MSE), while the red curves denote validation loss, both plotted with the training round index. Each round corresponds to one complete cycle of updating the agent followed by finetuning the predictor.

The results in \cref{fig:I} demonstrate clear convergence patterns across all five datasets. 
Both training and validation losses exhibit consistent downward trends across all datasets, validating the effectiveness of our asynchronous optimization strategy. Most datasets achieve convergence within 8--10 rounds, with the loss curves gradually stabilizing after approximately 10--15 rounds, suggesting that the architecture-predictor coevolution has reached near-optimal configurations.
Notably, the observed divergence between training and validation curves in later rounds (typically after round 10--12) indicates the onset of overfitting, highlighting both the necessity and effectiveness of our early stopping mechanism.
By continuously monitoring validation performance, our method can automatically terminate training at the point where generalization capability peaks depending on dataset characteristics.

\begin{table}[!t]
\centering
\caption{Computational efficiency comparison: training time (hours), number of parameters (millions), and T/P ratio of baselines and our RRE-PPO4Pred method across eight RNN backbones on the ETTh dataset at horizon $H$=24.}
\label{tab:efficiency}
\resizebox{0.85\textwidth}{!}{
\begin{threeparttable}
\begin{tabular}{cl|cccccccc}
\toprule
Metric & Method & RNN & MGU & GRU & LSTM & IndRNN & phLSTM & pLSTM & xLSTM \\
\midrule
\multirowcell{7}{Time\\ (h)}
 & Naive-all   & 0.11 & 0.12 & 0.12 & 0.12 & 0.12 & 0.13 & 0.12 & 0.13 \\
 & Naive-last  & 0.11 & 0.11 & 0.11 & 0.12 & 0.11 & 0.12 & 0.11 & 0.13 \\
 & EBPSO       & 1.15 & 1.22 & 1.32 & 1.46 & 1.19 & 1.38 & 1.26 & 1.61 \\
 & LSTMjump    & 0.43 & 0.46 & 0.52 & 0.56 & 0.47 & 0.59 & 0.43 & 0.63 \\
 & PGLSTM      & 0.47 & 0.52 & 0.54 & 0.60 & 0.50 & 0.66 & 0.49 & 0.69 \\
 & Ours        & 0.86 & 0.93 & 0.96 & 0.99 & 0.90 & 1.03 & 0.95 & 1.08 \\
\midrule
\multirowcell{7}{Parameters\\ (M)}
 & Naive-all   & 0.03 & 0.06 & 0.10 & 0.13 & 0.03 & 0.13 & 0.13 & 0.18 \\
 & Naive-last  & 0.03 & 0.06 & 0.10 & 0.13 & 0.03 & 0.13 & 0.13 & 0.18 \\
 & EBPSO       & 0.03 & 0.06 & 0.10 & 0.13 & 0.03 & 0.13 & 0.13 & 0.18 \\
 & LSTMjump    & 0.12 & 0.15 & 0.19 & 0.22 & 0.12 & 0.22 & 0.22 & 0.27 \\
 & PGLSTM      & 0.15 & 0.18 & 0.22 & 0.25 & 0.15 & 0.25 & 0.25 & 0.30 \\
 & Ours        & 0.87 & 0.90 & 0.94 & 0.97 & 0.87 & 0.97 & 0.97 & 1.02 \\
\midrule
\multirowcell{7}{T/P Ratio\\ (h/M)}
 & Naive-all   & 3.78 & 2.00 & 1.21 & 0.95 & 3.93 & 1.00 & 0.91 & 0.74 \\
 & Naive-last  & 3.59 & 1.90 & 1.15 & \textbf{0.91} & 3.77 & \textbf{0.96} & \textbf{0.87} & \textbf{0.71} \\
 & EBPSO       & 38.31 & 20.38 & 13.21 & 11.26 & 39.60 & 10.61 & 9.68 & 8.96 \\
 & LSTMjump    & 3.59 & 3.09 & 2.73 & 2.53 & 3.90 & 2.70 & 1.96 & 2.34 \\
 & PGLSTM      & 3.15 & 2.90 & 2.47 & 2.41 & 3.32 & 2.64 & 1.94 & 2.30 \\
 & Ours        & \textbf{0.99} & \textbf{1.04} & \textbf{1.02} & 1.02 & \textbf{1.03} & 1.06 & 0.98 & 1.06 \\
\bottomrule
\end{tabular}
\begin{tablenotes}
\footnotesize
\item T/P Ratio is calculated as $\frac{\text{Method}_\text{time}}{\text{Method}_\text{params}}$, measuring the training time cost per parameter. A smaller value indicates higher computational efficiency, as the method requires less training time per parameter.
\end{tablenotes}
\end{threeparttable}
}
\end{table}

\begin{figure}[t]
    \centering
    \begin{subfigure}[b]{0.32\textwidth}
        \centering
        \includegraphics[width=\textwidth]{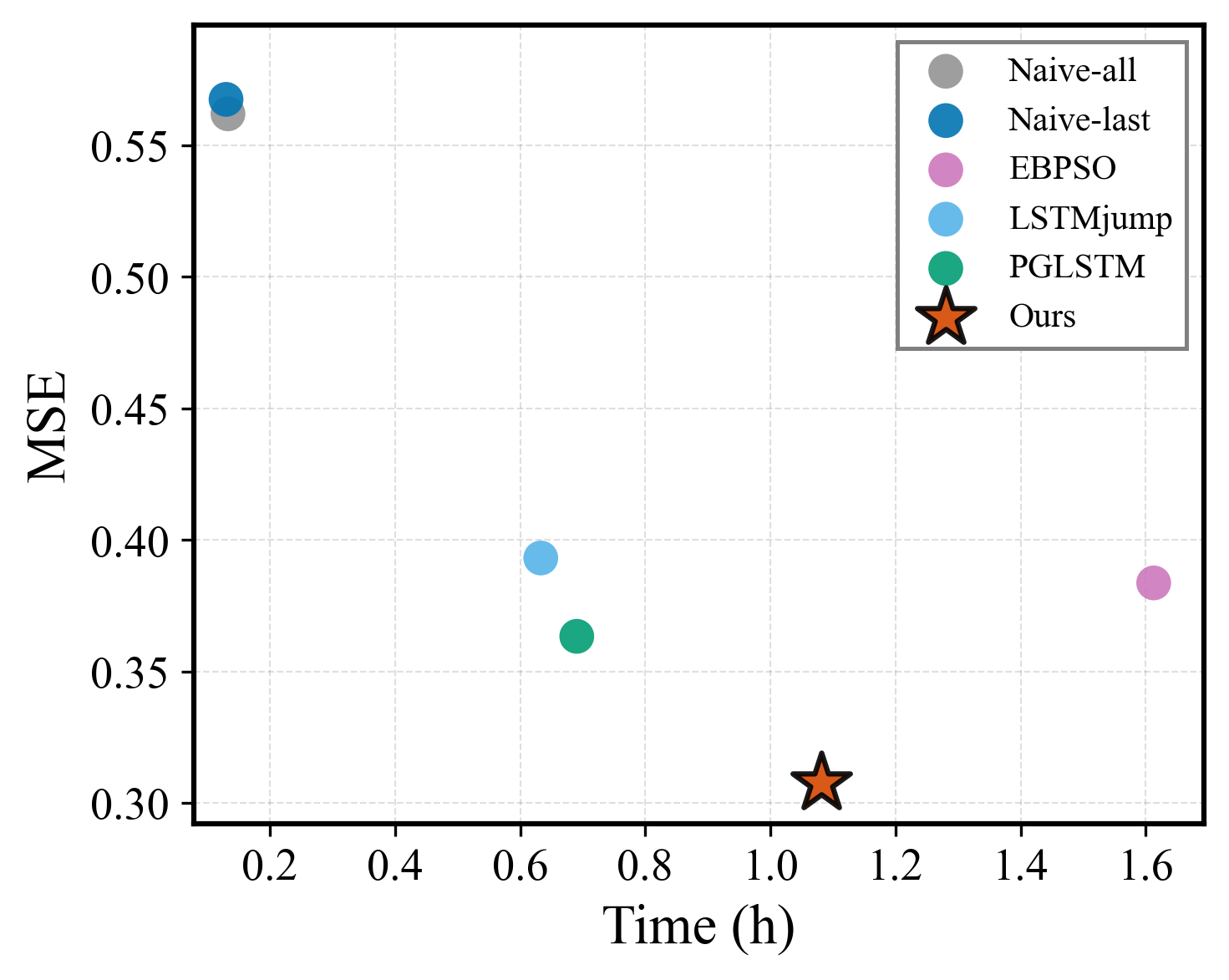}
        \label{fig:time}
    \end{subfigure}
    \hfill 
    \begin{subfigure}[b]{0.32\textwidth}
        \centering
        \includegraphics[width=\textwidth]{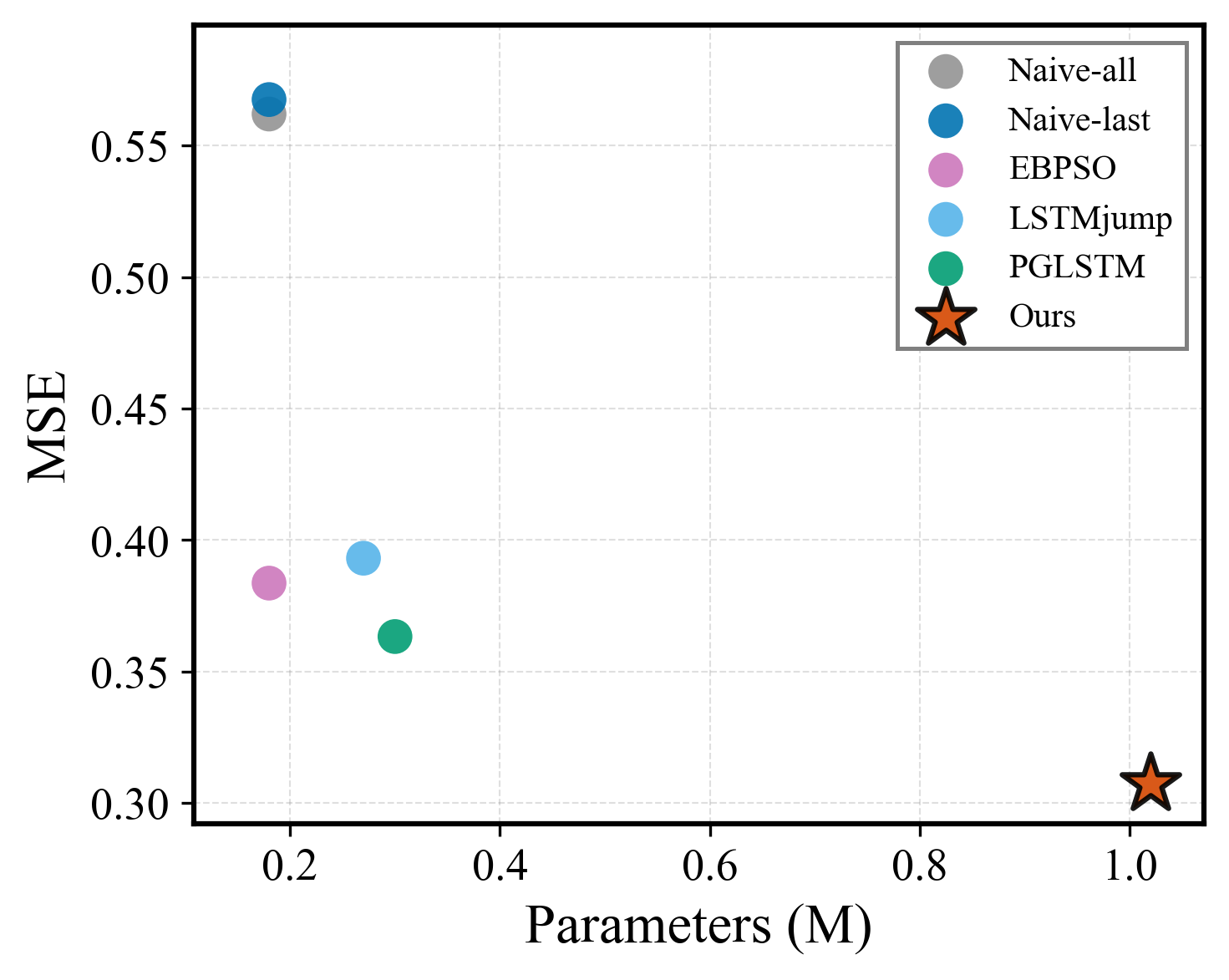} 
        \label{fig:para}
    \end{subfigure}
    \begin{subfigure}[b]{0.32\textwidth}
        \centering
        \includegraphics[width=\textwidth]{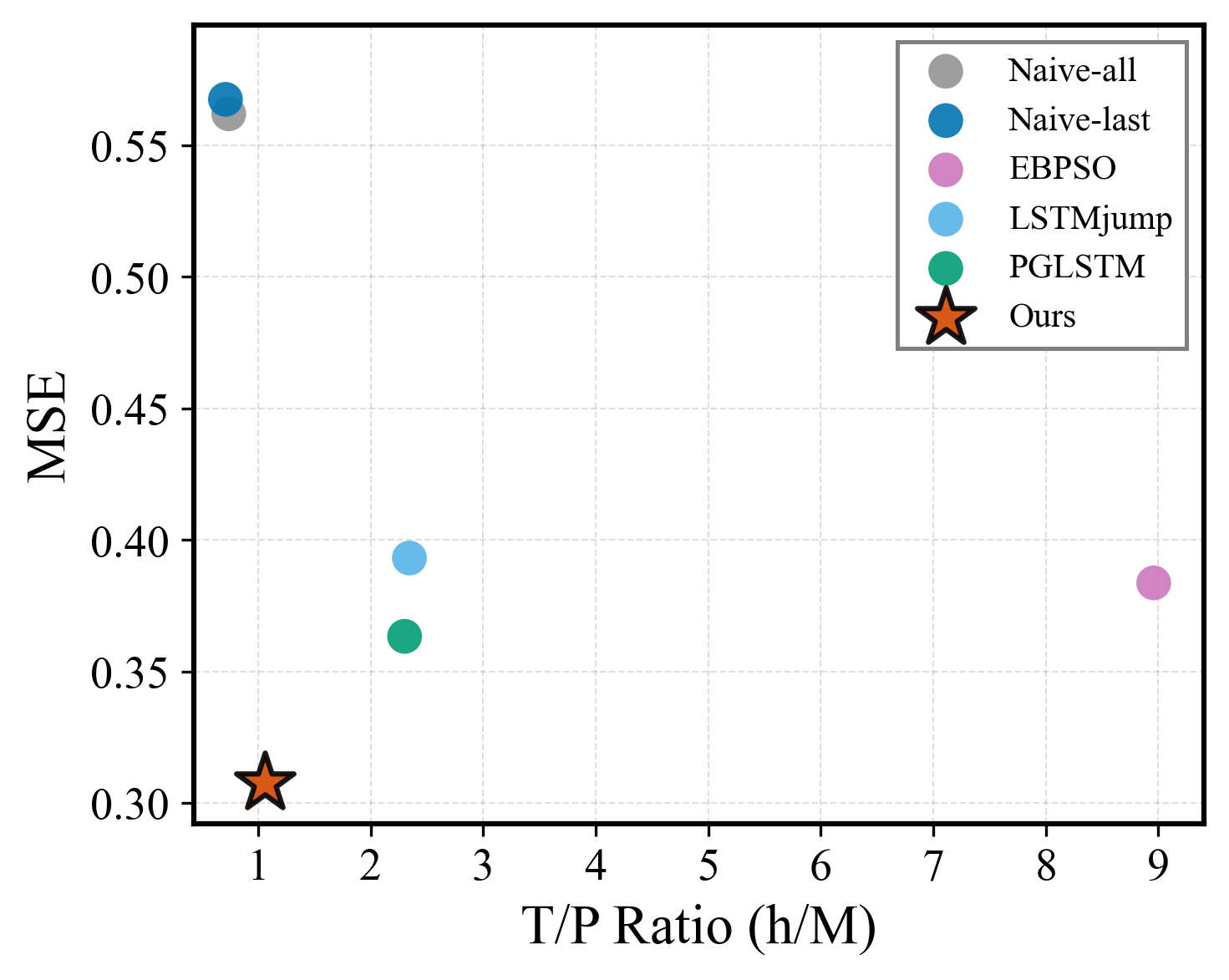} 
        \label{fig:ratio}
    \end{subfigure}
    \vspace{-5mm}
    \caption{Visualization of the trade-off between prediction accuracy and training time as well as parameter quantity across different methods with xLSTM backbone at horizon $H=24$ on the ETTh dataset. \label{fig:computation} }
    
\end{figure}

\subsection{Computational efficiency analysis}

To illustrate the computational costs of our RRE-PPO4Pred, we measure both training time (in hours) and number of parameters (in millions) across different RNN backbones on the ETTh dataset with a prediction horizon of 24.
To jointly assess the efficiency of the training time relative to model size, we introduce a efficiency metric, T/P (Time-to-Parameter) Ratio, as shown in \cref{tab:efficiency}, where $\text{Method}_\text{time}$ and $\text{Method}_\text{params}$ denote the training runtime and number of parameters for a given method, respectively.
Notably, EBPSO optimizes the RNN models without additional agent, so its parameter overhead is identical to that of the Naive baselines.
However, due to its heuristic search procedure, which performs repeated population-level evaluations and updates, where each generation requires training and evaluating multiple candidate solutions,  EBPSO consistently requires the longest training time across all backbones.

Compared to LSTMjump and PGLSTM, our method incurs a moderately higher training time due to the asynchronous update mechanism, which alternates between predictor and agent update in an asynchronous manner, whereas both LSTMjump and PGLSTM update the predictor and agent synchronously.
Additionally, our method has a larger parameter overhead compared to LSTMjump and PGLSTM. This is due to our Transformer-based PPO agent, which includes multi-head attention mechanisms and feed-forward layers, whereas LSTMjump and PGLSTM use simple MLP-based agent.

However, despite these overheads, our method achieves favorable computational efficiency as measured by the T/P ratio.
On the LSTM families (LSTM, phLSTM, pLSTM, and xLSTM), our RRE-PPO4Pred achieves a T/P ratio comparable to the Naive-all and Naive-last baselines, indicating that the training time increase is proportional to the parameter increase.
More impressively, our method consistently achieves the lowest T/P ratio across other RNN backbones (RNN, MGU, GRU, and IndRNN).
This suggests the effectiveness of our proposed dynamic transition sampling strategy, which intelligently selects training samples based on their informativeness.
By focusing computational resources on high-value transitions, our method reduces training time growth relative to parameter increase, resulting in superior overall computational efficiency despite the larger model size.

We further visualize the trade-off between the accuracy gain and computational cost in \cref{fig:computation}.
Compared to the baseline methods, our RRE-PPO4Pred method consistently achieves lower MSE.
As there is no free lunch in model optimization, the improved accuracy of our method comes at the cost of  increased training time and parameters.
However, as modern GPUs typically have 6--24 GB of memory, this parameter overhead, approximately 0.82 GB GPU memory for training and 0.48 GB for inference\footnote{All GPU memory measurements are recorded with PyTorch v2.5.0 and CUDA v12.1 on NVIDIA RTX A4500 GPU.}, can be acceptable in many practical applications.
Moreover, our training time remains significantly lower than that of EBPSO, and our T/P ratio is comparable to Naive-all and Naive-last baselines, demonstrating a favorable balance between accuracy and efficiency.

\subsection{Performance comparison with Transformer-based methods}

\begin{table}[t]
\centering
\caption{Average MSE and MAE performance of Transformer-based models and xLSTM with our RRE-PPO4Pred method across prediction horizons on the ETTh dataset. Standard deviations are shown in bracket.}
\label{tab:transformer}
\resizebox{\textwidth}{!}{%
\begin{tabular}{c|cccc|cccc}
\toprule
\multirow{2}{*}{$H$} & \multicolumn{4}{c|}{MSE}                     & \multicolumn{4}{c}{MAE}                             \\ \cmidrule(l){2-9} 
                     & iTransformer      & PatchTST          & xLSTM  & xLSTM+Ours                 & iTransformer      & PatchTST          & xLSTM  & xLSTM + Ours               \\ \midrule
24                   & 0.3540 (7.02E-04) & 0.3195 (4.83E-04) & 0.5620 (2.59E-03)
 & \textbf{0.3075 (3.54E-03)} & 0.3855 (1.05E-03) & 0.3701 (6.80E-04) & 0.6481 (8.40E-04)
 & \textbf{0.3565 (2.64E-03)} \\
48                   & 0.3790 (6.99E-04) & 0.3477 (2.35E-04) & 0.5962 (2.85E-03) & \textbf{0.3378 (2.14E-03)} & 0.4062 (6.21E-04) & 0.3925 (4.39E-04) & 0.6779 (4.06E-03) & \textbf{0.3746 (8.52E-04)} \\
96                   & 0.3921 (4.81E-03) & 0.3746 (7.24E-04) & 0.6755 (9.93E-04)& \textbf{0.3607 (5.61E-04)} & 0.4159 (9.31E-04) & 0.4086 (8.60E-04) & 0.7391 (6.40E-03) & \textbf{0.3897 (1.34E-03)} \\
 \bottomrule
\end{tabular}%
}
\end{table}

While the previous experiments focused on comparisons with RNN-based methods and ablation studies, we further examine whether our proposed RRE-PPO4Pred method can enable RNN-based models to compete with or outperform modern Transformer architectures, which represent recent advances in time series forecasting. 

Specifically, we include two representative Transformer-based methods, PatchTST \citep{nie2022time} and iTransformer \citep{liu2023itransformer}. 
PatchTST applies patching and channel-independence mechanisms to achieve state-of-the-art performance on various benchmarks, and iTransformer \citep{liu2023itransformer} modifies the traditional Transformer-based forecasting model by applying attention across variates rather than time steps.
Thus, we perform the experiments on the multivariate dataset ETTh, and multiple forecasting horizons are selected to evaluate performance across both short-term and long-term prediction scenarios.
According to previous results in \cref{tab:ETTh-performance}, we instantiate our RRE-PPO4Pred with xLSTM backbone, which shows a moderate performance among the selected recurrent architectures, avoiding cherry-picking the best-performing backbone and providing a more conservative evaluation of our method's effectiveness. For further comparison, we include a vanilla xLSTM model optimized using the Naive-all baseline, which represents the standard xLSTM without policy-based modifications.

As shown in \cref{tab:transformer}, 
the vanilla xLSTM model (trained by Naive-all) underperforms both iTransformer and PatchTST across all prediction horizons, confirming the conventional wisdom that RNN-based architectures struggle to compete with modern Transformer-based methods on time series forecasting tasks. However, after integrating our RRE-PPO4Pred approach, the enhanced xLSTM (xLSTM+Ours) consistently outperforms both Transformer-based baselines across all prediction horizons.
For example, at the horizon $H=96$, our method achieves an MSE of 0.3607 compared to 0.3921 for iTransformer and 0.3746 for PatchTST, representing improvements of 8.0\% and 3.7\%, respectively. 
These results demonstrate that RRE-PPO4Pred substantially enhances the predictive capabilities of RNN-based models. 
Remarkably, our approach enables RNN-based models to not merely match but actually surpass state-of-the-art Transformer-based methods for time series forecasting.
This suggests that the architectural inductive biases of RNNs, when properly optimized, remain highly competitive in current forecasting field.

\subsection{Visualization analysis}

\begin{figure}[t]
    \centering
    \begin{subfigure}[b]{0.32\textwidth}
        \centering
        \includegraphics[width=\textwidth]{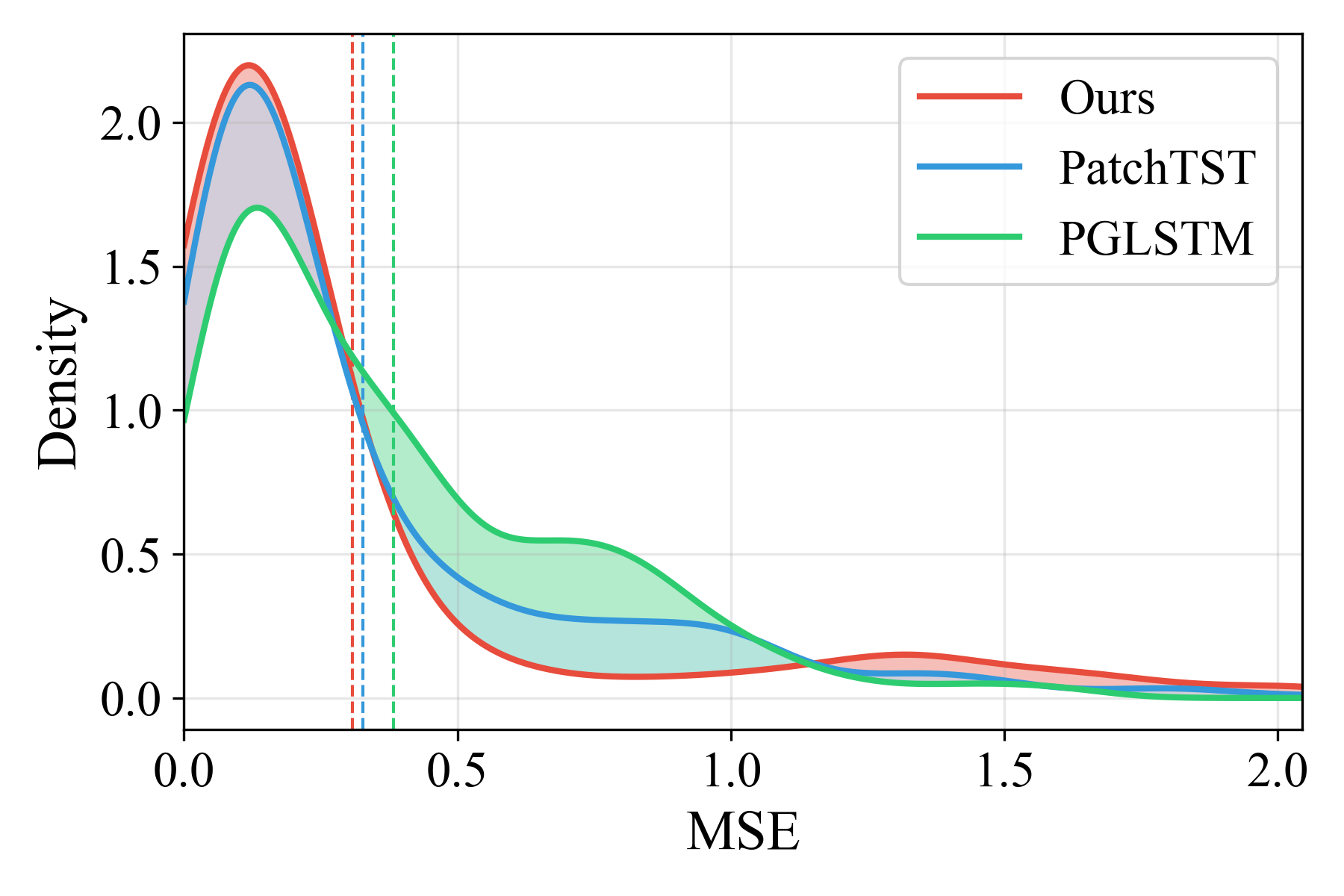} 
        \caption{MSE density, horizon $H=24$}
        \label{fig:density24mse}
    \end{subfigure}
    \hfill 
        \begin{subfigure}[b]{0.32\textwidth}
        \centering
        \includegraphics[width=\textwidth]{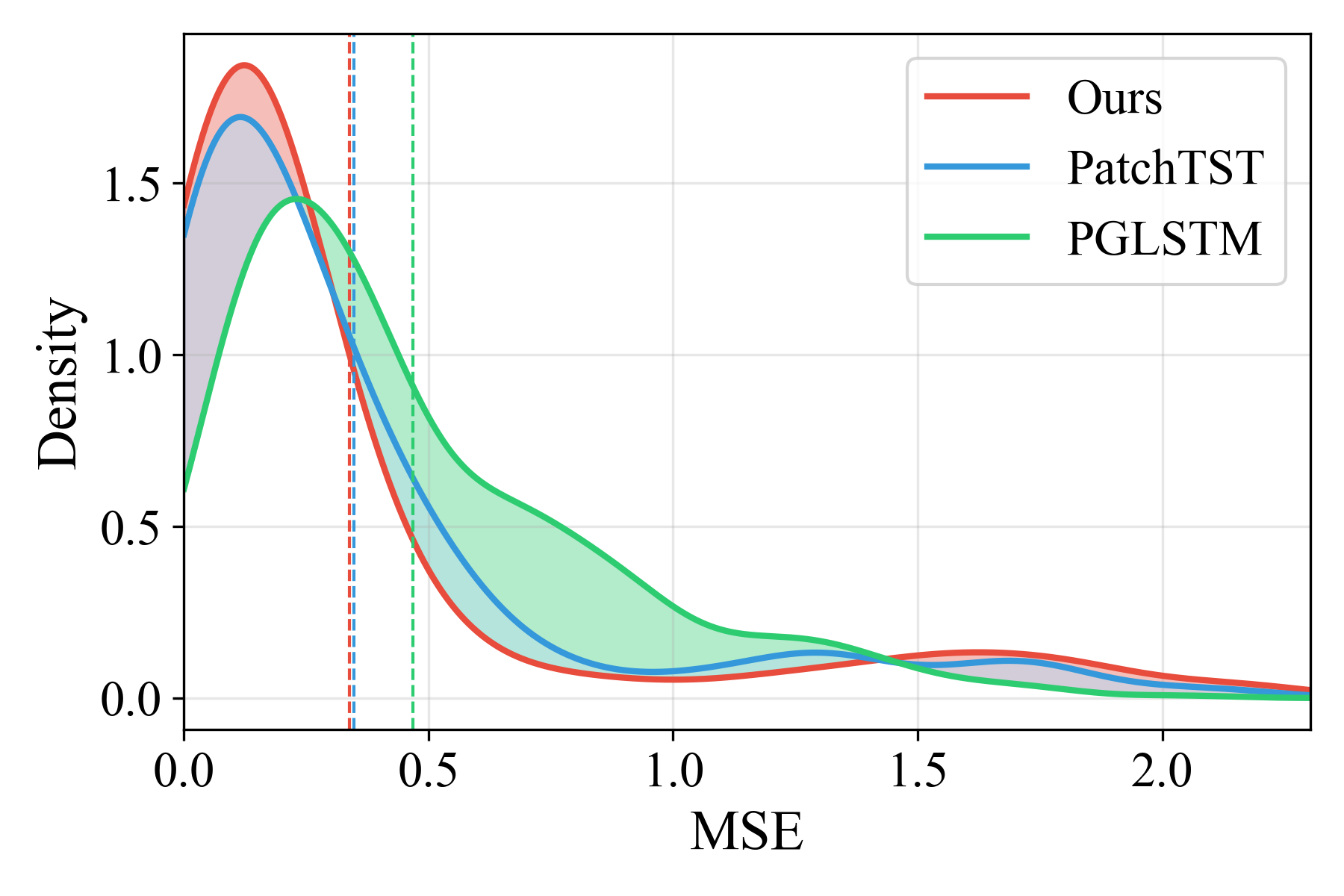} 
        \caption{MSE density, horizon $H=48$}
        \label{fig:density48mse}
    \end{subfigure}
    \hfill
         \begin{subfigure}[b]{0.32\textwidth}
        \centering
        \includegraphics[width=\textwidth]{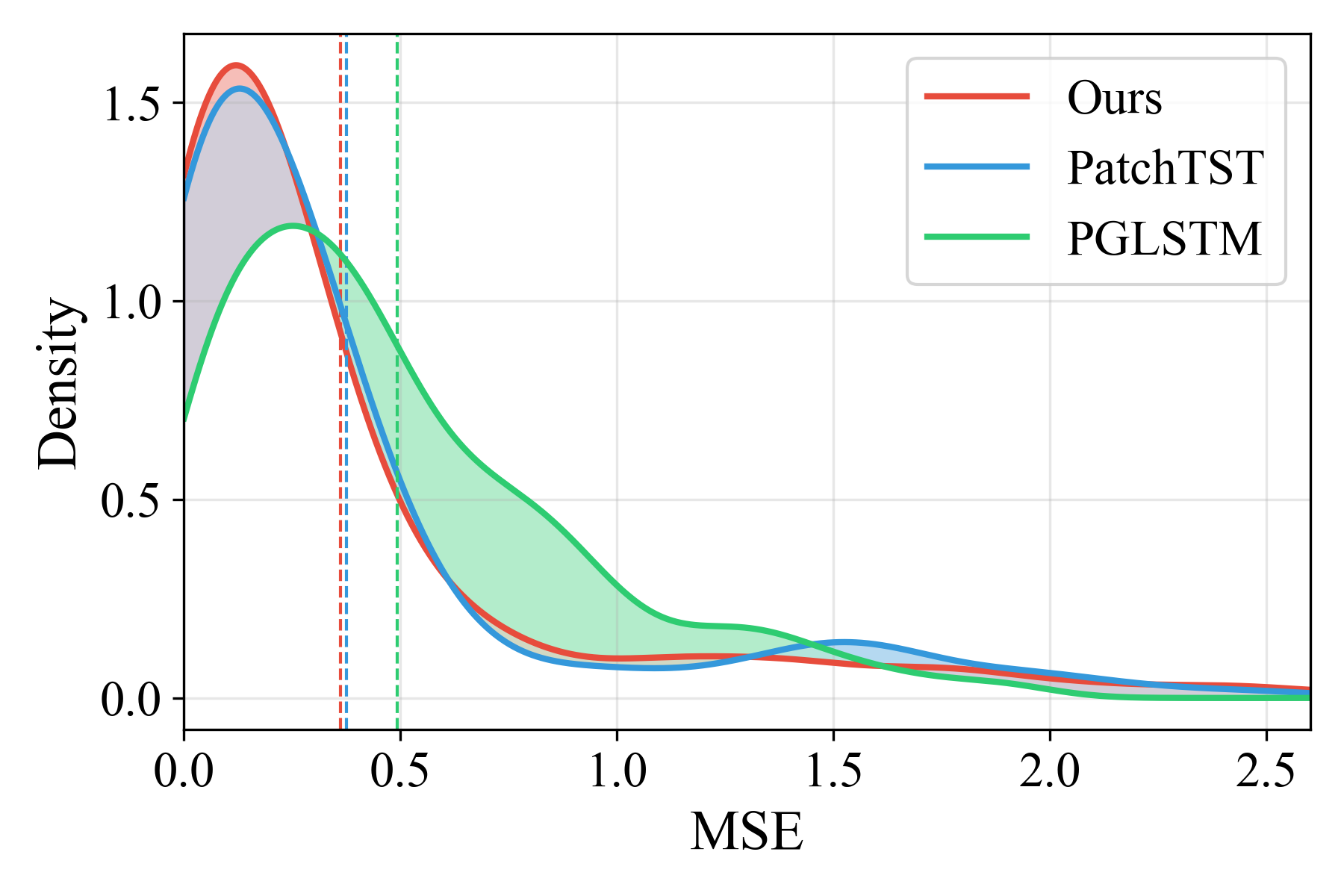} 
        \caption{MSE density, horizon $H=96$}
        \label{fig:density96mse}
    \end{subfigure}
    \hfill
    \vspace{1mm}
    \begin{subfigure}[b]{0.32\textwidth}
        \centering
        \includegraphics[width=\textwidth]{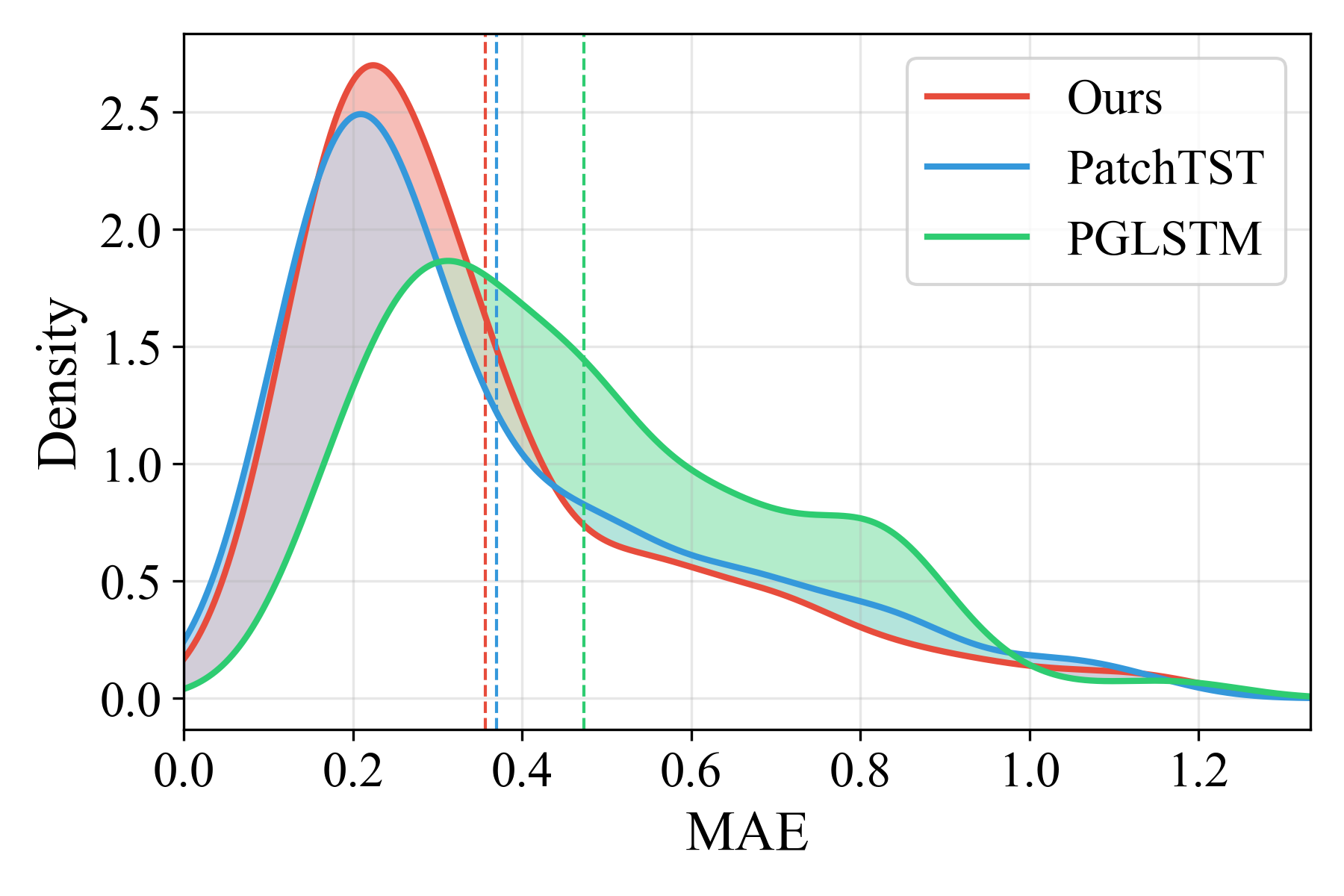} 
        \caption{ MAE density, horizon $H=24$}
        \label{fig:density24mae}
    \end{subfigure}
    \hfill
    \begin{subfigure}[b]{0.32\textwidth}
    \centering
    \includegraphics[width=\textwidth]{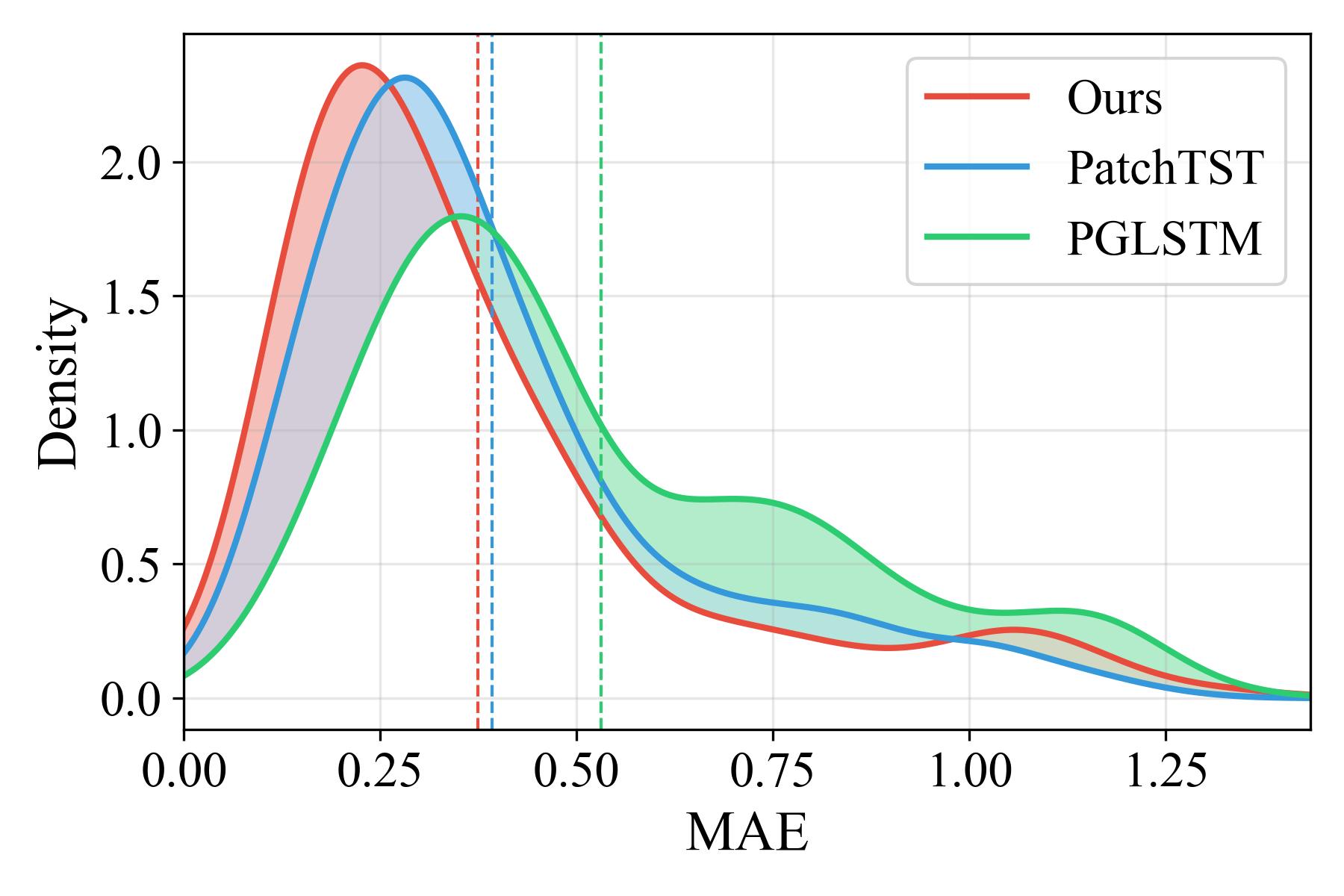} 
    \caption{MAE density, horizon $H=48$}
    \label{fig:density48mae}
\end{subfigure}
    \hfill
    \begin{subfigure}[b]{0.32\textwidth}
    \centering
    \includegraphics[width=\textwidth]{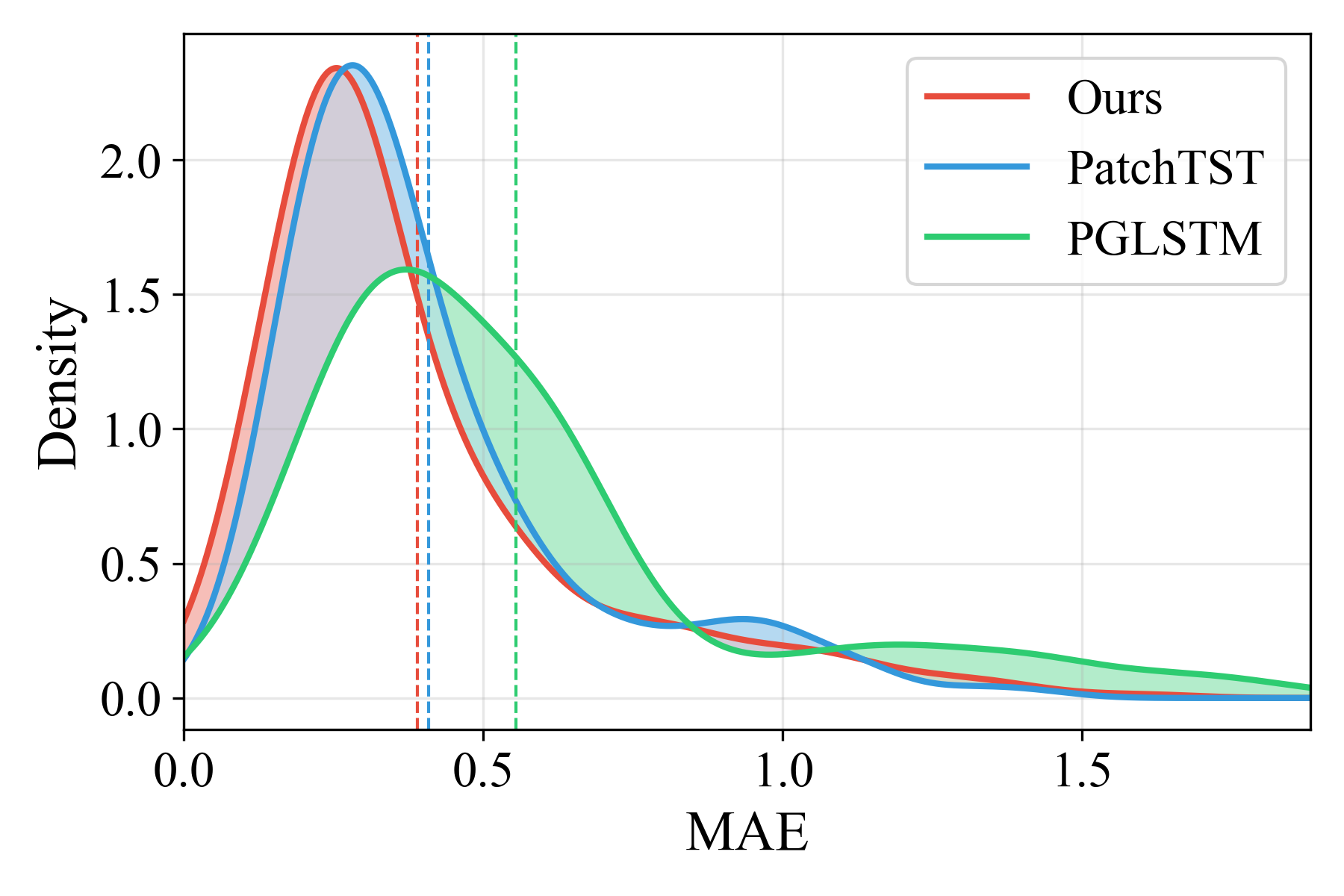} 
    \caption{MAE density, horizon $H=96$}
    \label{fig:density96mae}
\end{subfigure}
    \caption{Kernel density estimation of prediction error distributions for our RRE-PPO4Pred, PatchTST, and PGLSTM on the ETTh dataset, where RRE-PPO4Pred and PGLSTM are equipped with xLSTM backbone. Solid curves represent the probability density of MSE and MAE values, with vertical dashed lines indicating the mean error for each method. Shaded regions highlight the density differences between methods. \label{fig:density}}
    
\end{figure}

To investigate the robustness and reliability of the proposed RRE-PPO4Pred method, we analyze the prediction error distributions of the unseen time series instances on the ETTh testing set. We compare three representative methods: xLSTM+RRE-PPO4Pred, PatchTST (the best-performing Transformer-based model in \cref{tab:transformer}), and PGLSTM (the strongest RNN-based baseline in \cref{tab:ETTh-performance}). \cref{fig:density} presents the Kernel Density Estimation (KDE) curves of MSE and MAE for each method, with vertical dashed lines indicating the error expectation.

Across all subplots in \cref{fig:density}, our method substantially reduces the occurrence of high-error instances, and consistently exhibits the most concentrated error distribution with the lowest error expectation.
This distribution indicates fewer catastrophic prediction failures, which is crucial for reliable deployment in real-world applications.
These findings demonstrate that our method achieves superior robustness and generalization capability compared to strong baseline methods, effectively addressing the challenge of instance-level variations in time series forecasting.

\begin{figure}[t]
    \centering
    \begin{subfigure}[b]{0.86\textwidth}
        \centering
        \includegraphics[width=\textwidth]{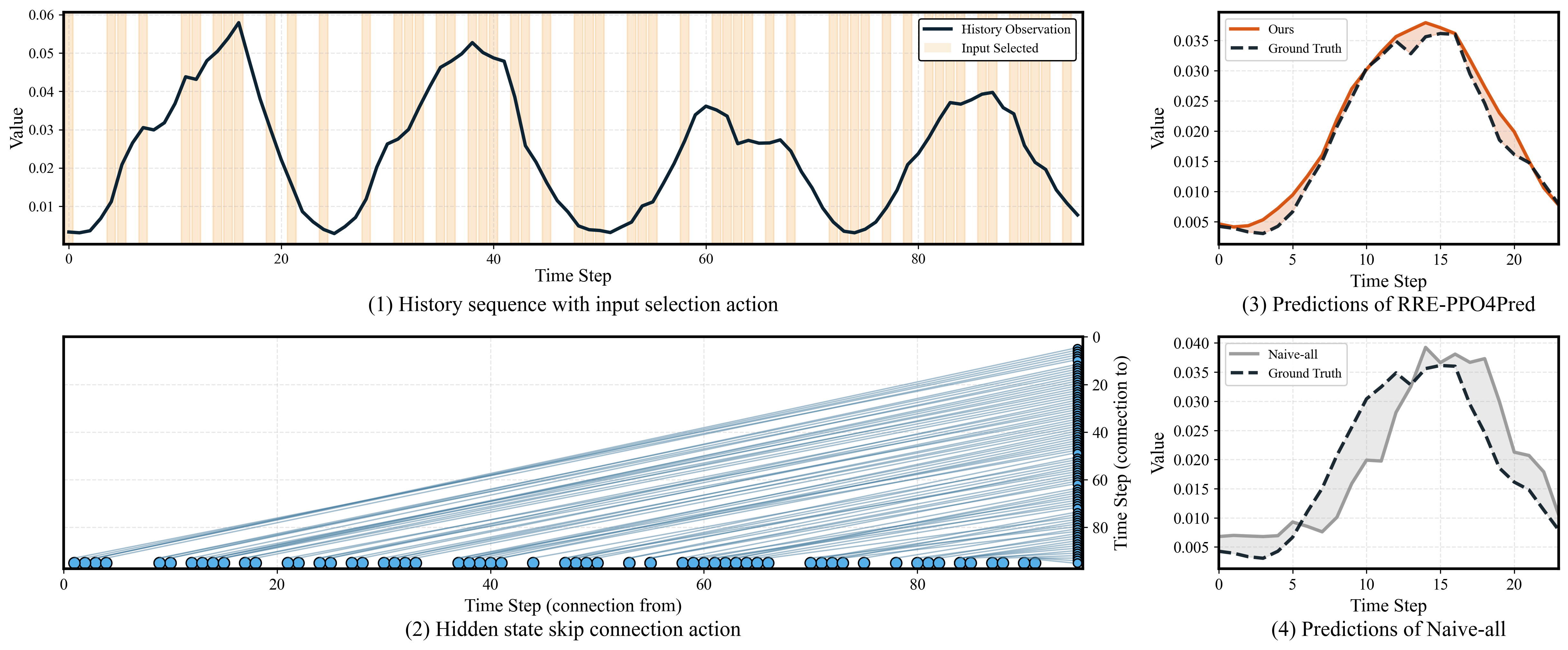} 
        \caption{Prediction example 1}
        \label{fig:pred1}
    \end{subfigure}
    \vspace{1mm}
    \begin{subfigure}[b]{0.86\textwidth}
        \centering
        \includegraphics[width=\textwidth]{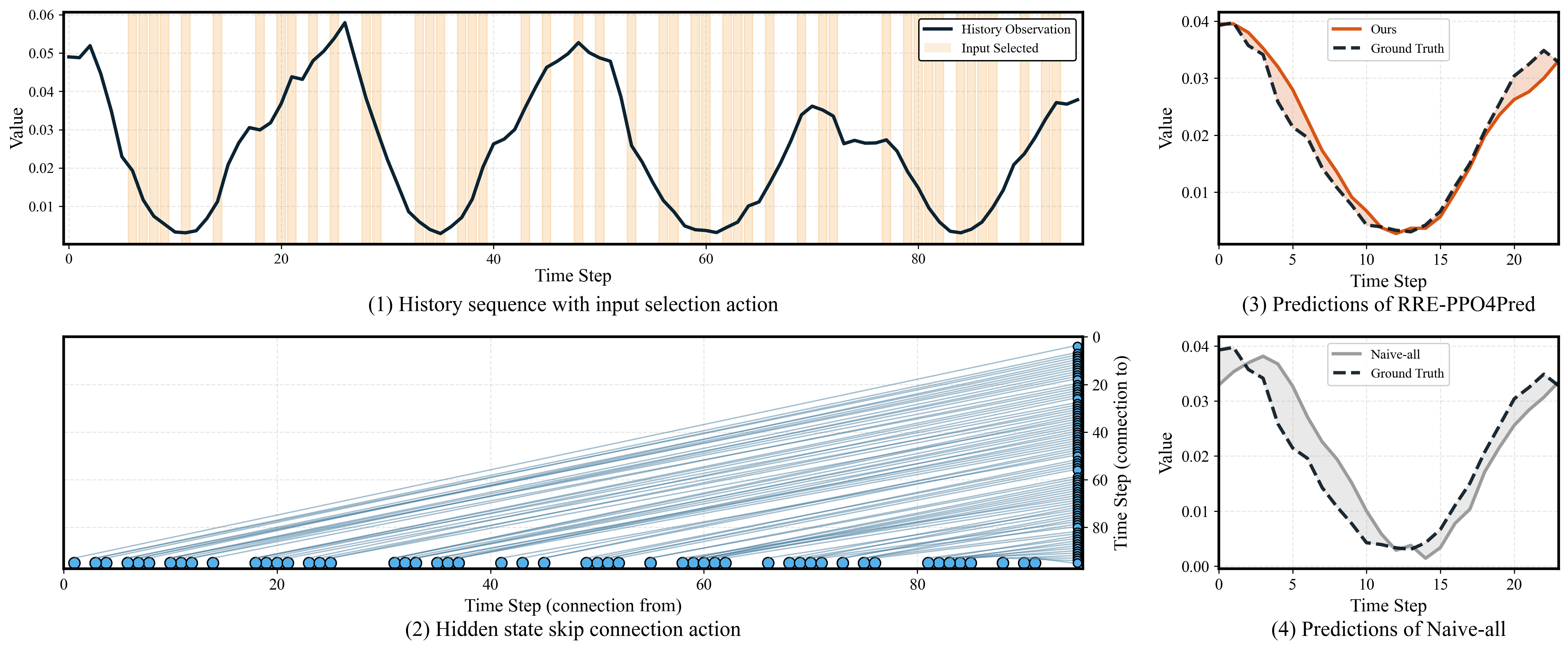}
        \caption{Prediction example 2}
        \label{fig:pred2}
    \end{subfigure}

    \caption{Visualization of two prediction example with our policy behavior on the Traffic testing data set using xLSTM backbone and a 24-step horizon. The selected time steps ($u_t=1$) in the historical sequence are highlighted in red. The hidden state skip connection pattern shows which historical hidden state (x-axis) is connected to future time step (y-axis) via action $k_t$. }
    \label{fig:predcase}
\end{figure}

To further demonstrate the interpretability and effectiveness of our proposed method, we visualize the learned policy behavior (input selection and skip connection actions) and prediction results on two representative unseen examples from the Traffic dataset in \cref{fig:predcase}.

As shown in \cref{fig:pred1}, when forecasting a peak traffic flow pattern, our method exhibits intelligent behavior. Panel 1 shows that the policy selectively attends to relevant historical input steps, particularly those preceding similar peak patterns in the historical sequence. Panel 2 reveals that the policy establishes skip connections from the hidden states at these relevant time steps, enabling the encoder to directly access informative representations from critical moments. As a result, panel 3 shows that our method accurately captures the peak magnitude and timing. In contrast, panel 4 shows that the Naive-all baseline, which uses all inputs uniformly without selective mechanism, fails to capture the peak accurately. 
This demonstrates the effectiveness of our learned policy in recognizing critical patterns.

As shown in \cref{fig:pred2}, when forecasting a sudden drop pattern, our method demonstrates remarkable adaptivity by attending to significantly different regions compared to \cref{fig:pred1}. The policy now focuses on historical time steps that exhibit similar downward trends or volatility, rather than the peak-related patterns from example 1. The skip connections are also established at different temporal locations, reflecting the policy's ability to dynamically adjust its skip mechanism based on the specific characteristics of each input pattern. As a result, our method consistently achieves accurate predictions across both cases, demonstrating its adaptivity to diverse temporal patterns.
\section{Conclusion}\label{conclusion}
In this study, a novel Reinforced Recurrent Encoder (RRE) framework with Prediction-oriented Proximal Policy Optimization (PPO4Pred) is proposed to enhance RNN models for time series forecasting. 
The proposed RRE framework dynamically optimizes the RNN-based predictor through joint decisions on input feature selection, hidden skip connection, and output target selection. To support this adaptive mechanism, we develop the PPO4Pred algorithm, which is tailored for enhancing the forecasting models with a Transformer-based agent and dynamic transition sampling.
Finally, we further establish a co-evolutionary optimization paradigm, facilitating the joint evolution of agent in PPO4Pred and RNN model in RRE framework. 
Extensive experiments conducted on multiple real-world industrial datasets demonstrate that
our RRE-PPO4Pred provides an effective, superior, and robust method to improve the RNN-based predictors, which not merely outperforms the existing RNN optimization baselines but also surpasses state-of-the-art Transformer-based forecasting models, thus providing an advanced time series predictor in engineering informatics.

Regarding limitations, RRE-PPO4Pred entails additional computational cost during improving RNN models. However, once trained, the agent remains active during inference and can dynamically adapt the RNN to unseen patterns without retraining, which may reduce retraining frequency in real-time applications. 
Future work will pursue a better trade-off between performance gain and added overhead through more efficient approaches, such as knowledge distillation, transfer learning, and lightweight Transformers.

\bibliography{refs}

@inproceedings{yu2017learning,
  title={Learning to skim text},
  author={Yu, Adams Wei and Lee, Hongrae and Le, Quoc},
  booktitle={Proceedings of the 55th Annual Meeting of the Association for Computational Linguistics},
volume={1},
  pages={1880--1890},
  year={2017}
}

@inproceedings{schulman2015high,
  title={High-dimensional continuous control using generalized advantage estimation},
  author={Schulman, John and Moritz, Philipp and Levine, Sergey and Jordan, Michael and Abbeel, Pieter},
  booktitle    = {Proceedings of the 4th International Conference on Learning Representations},
  year         = {2016},
 
}

@article{barrera2022rainfall,
  title={Rainfall prediction: A comparative analysis of modern machine learning algorithms for time-series forecasting},
  author={Barrera-Animas, Ari Yair and Oyedele, Lukumon O and Bilal, Muhammad and Akinosho, Taofeek Dolapo and Delgado, Juan Manuel Davila and Akanbi, Lukman Adewale},
  journal={Machine Learning with Applications},
  volume={7},
  pages={100204},
  year={2022},
  publisher={Elsevier}
}

@article{schaffer2021interrupted,
  title={{Interrupted time series analysis using autoregressive integrated moving average (ARIMA) models: A guide for evaluating large-scale health interventions}},
  author={Schaffer, Andrea L and Dobbins, Timothy A and Pearson, Sallie Anne},
  journal={BMC Medical Research Methodology},
  volume={21},
  pages={1--12},
  year={2021},
  publisher={Springer}
}

@article{sutskever2014sequence,
  title={Sequence to sequence learning with neural networks},
  author={Sutskever, Ilya and Vinyals, Oriol and Le, Quoc V},
  journal={Advances in Neural Information Processing Systems},
  volume={27},
  year={2014}
}

@inproceedings{see2017get,
  title={Get to the point: Summarization with pointer-generator networks},
  author={See, Abigail and Liu, Peter J and Manning, Christopher D},
  booktitle={Proceedings of the 55th Annual Meeting of the Association for Computational Linguistics},
    volume={1},
  pages={1073--1083},
  year={2017}
}

@article{hewamalage2021recurrent,
  title={Recurrent neural networks for time series forecasting: Current status and future directions},
  author={Hewamalage, Hansika and Bergmeir, Christoph and Bandara, Kasun},
  journal={International Journal of Forecasting},
  volume={37},
  number={1},
  pages={388--427},
  year={2021},
  publisher={Elsevier}
}

@article{schulman2017proximal,
  title={Proximal policy optimization algorithms},
  author={Schulman, John and Wolski, Filip and Dhariwal, Prafulla and Radford, Alec and Klimov, Oleg},
  journal={arXiv preprint arXiv:1707.06347},
  year={2017}
}

@inproceedings{Indrnn,
  author={Li, Shuai and Li, Wanqing and Cook, Chris and Zhu, Ce and Gao, Yanbo},
  booktitle={Proceedings of the IEEE/CVF Conference on Computer Vision and Pattern Recognition}, 
  title={{Independently recurrent neural network (IndRNN): Building a longer and deeper RNN}}, 
  year={2018},
  volume={},
  number={},
  pages={5457-5466}
 }

@article{tijjani2024enhanced,
  title={An enhanced particle swarm optimization with position update for optimal feature selection},
  author={Tijjani, Sani and Ab Wahab, Mohd Nadhir and Noor, Mohd Halim Mohd},
  journal={Expert Systems with Applications},
  volume={247},
  pages={123337},
  year={2024},
  publisher={Elsevier}
}

@inproceedings{liu2023itransformer,
  title={{iTransformer: Inverted transformers are effective for time series forecasting}},
  author={Liu, Yong and Hu, Tengge and Zhang, Haoran and Wu, Haixu and Wang, Shiyu and Ma, Lintao and Long, Mingsheng},
  booktitle={Proceedings of the Twelfth International Conference on Learning Representations},
  year={2023}
}

@inproceedings{nie2022time,
  title={{A time series is worth 64 words: Long-term forecasting with Transformers}},
  author={Nie, Yuqi and Nguyen, Nam H and Sinthong, Phanwadee and Kalagnanam, Jayant},
  booktitle={Proceedings of the Eleventh International Conference on Learning Representations},
  year={2022}
}

@article{bergstra2012random,
  title={Random search for hyper-parameter optimization},
  author={Bergstra, James and Bengio, Yoshua},
  journal={The Journal of Machine Learning Research},
  volume={13},
  number={1},
  pages={281--305},
  year={2012},
  publisher={JMLR. org}
}

@article{gers2000learning,
  title={{Learning to forget: Continual prediction with LSTM} },
  author={Gers, Felix A and Schmidhuber, J{\"u}rgen and Cummins, Fred},
  journal={Neural Computation},
  volume={12},
  number={10},
  pages={2451--2471},
  year={2000},
  publisher={MIT press}
}

@article{zhou2016minimal,
  title={Minimal gated unit for recurrent neural networks},
  author={Zhou, Guo Bing and Wu, Jianxin and Zhang, Chen Lin and Zhou, Zhi Hua},
  journal={International Journal of Automation and Computing},
  volume={13},
  number={3},
  pages={226--234},
  year={2016},
  publisher={Springer}
}

@article{sherstinsky2020fundamentals,
  title={{Fundamentals of recurrent neural network (RNN) and long short-term memory (LSTM) network}},
  author={Sherstinsky, Alex},
  journal={Physica D: Nonlinear Phenomena},
  volume={404},
  pages={132306},
  year={2020},
  publisher={Elsevier}
}

@article{xue2015survey,
  title={A survey on evolutionary computation approaches to feature selection},
  author={Xue, Bing and Zhang, Mengjie and Browne, Will N and Yao, Xin},
  journal={IEEE Transactions on Evolutionary Computation},
  volume={20},
  number={4},
  pages={606--626},
  year={2015},
  publisher={IEEE}
}

@article{ren2023mafsids,
  title={{MAFSIDS: A reinforcement learning-based intrusion detection model for multi-agent feature selection networks}},
  author={Ren, Kezhou and Zeng, Yifan and Zhong, Yuanfu and Sheng, Biao and Zhang, Yingchao},
  journal={Journal of Big Data},
  volume={10},
  number={1},
  pages={137},
  year={2023},
  publisher={Springer}
}

@inproceedings{fan2020autofs,
  title={{AutoFS: Automated feature selection via diversity-aware interactive reinforcement learning}},
  author={Fan, Wei and Liu, Kunpeng and Liu, Hao and Wang, Pengyang and Ge, Yong and Fu, Yanjie},
  booktitle={Proceedings of the IEEE International Conference on Data Mining},
  pages={1008--1013},
  year={2020},
  organization={IEEE}
}

@inproceedings{zhuang2023survey,
  title={A survey on efficient training of transformers},
  author={Zhuang, Bohan and Liu, Jing and Pan, Zizheng and He, Haoyu and Weng, Yuetian and Shen, Chunhua},
  booktitle={Proceedings of the Thirty-Second International Joint Conference on Artificial Intelligence},
  pages={6823--6831},
  year={2023}
}

@article{li2020federated,
  title={{Federated learning: Challenges, methods, and future directions}},
  author={Li, Tian and Sahu, Anit Kumar and Talwalkar, Ameet and Smith, Virginia},
  journal={IEEE Signal Processing Magazine},
  volume={37},
  number={3},
  pages={50--60},
  year={2020},
  publisher={IEEE}
}

@article{wang2018deep,
  title={{Deep learning for smart manufacturing: Methods and applications}},
  author={Wang, Jinjiang and Ma, Yulin and Zhang, Laibin and Gao, Robert X and Wu, Dazhong},
  journal={Journal of Manufacturing Systems},
  volume={48},
  pages={144--156},
  year={2018},
  publisher={Elsevier}
}

@article{hochreiter1997long,
  title={Long short-term memory},
  author={Hochreiter, Sepp and Schmidhuber, J{\"u}rgen},
  journal={Neural Computation},
  volume={9},
  number={8},
  pages={1735--1780},
  year={1997},
  publisher={MIT press}
}

@inproceedings{kennedy1995particle,
  title={Particle swarm optimization},
  author={Kennedy, James and Eberhart, Russell},
  booktitle={Proceedings of the International Conference on Neural Networks},
  volume={4},
  pages={1942--1948},
  year={1995},
  organization={IEEE}
}

@article{sutton2000policy,
  title={Policy gradient methods for reinforcement learning with function approximation},
  author={Sutton, R. S. and McAllester, D. A. and Singh, S. P. and Mansour, Y.},
  journal={Advances in Neural Information Processing Systems},
  volume={12},
  pages={1057--1063},
  year={2000}
}

@article{pabuccu2024feature,
  title={Feature selection with annealing for forecasting financial time series},
  author={Pabuccu, Hakan and Barbu, Adrian},
  journal={Financial Innovation},
  volume={10},
  number={1},
  pages={87},
  year={2024},
  publisher={Springer}
}

@article{mnih2015human,
  title={Human-level control through deep reinforcement learning},
  author={Mnih, V. and Kavukcuoglu, K. and Silver, D. and others},
  journal={Nature},
  volume={518},
  pages={529--533},
  year={2015}
}

@article{beck2024xlstm,
  title={{xLSTM: Extended long short-term memory}},
  author={Beck, Maximilian and P{\"o}ppel, Korbinian and Spanring, Markus and Auer, Andreas and Prudnikova, Oleksandra and Kopp, Michael and Klambauer, G{\"u}nter and Brandstetter, Johannes and Hochreiter, Sepp},
  journal={Advances in Neural Information Processing Systems},
  volume={37},
  pages={107547--107603},
  year={2024}
}

@inproceedings{koutnik2014clockwork,
  title={{A clockwork RNN}},
  author={Koutnik, Jan and Greff, Klaus and Gomez, Faustino and Schmidhuber, Juergen},
  booktitle={Proceedings of the International Conference on Machine Learning},
  pages={1863--1871},
  year={2014},
  organization={PMLR}
}

@article{elman1990finding,
  title={Finding structure in time},
  author={Elman, Jeffrey L},
  journal={Cognitive Science},
  volume={14},
  number={2},
  pages={179--211},
  year={1990},
  publisher={Wiley Online Library}
}

@article{huang2021novel,
  title={{A novel model based on DA-RNN network and skip gated recurrent neural network for periodic time series forecasting}},
  author={Huang, Bingqing and Zheng, Haonan and Guo, Xinbo and Yang, Yi and Liu, Ximing},
  journal={Sustainability},
  volume={14},
  number={1},
  pages={326},
  year={2021},
  publisher={MDPI}
}

@article{kong2025deep,
  title={Deep learning for time series forecasting: A survey},
  author={Kong, Xiangjie and Chen, Zhenghao and Liu, Weiyao and Ning, Kaili and Zhang, Lechao and Muhammad Marier, Syauqie and Liu, Yichen and Chen, Yuhao and Xia, Feng},
  journal={International Journal of Machine Learning and Cybernetics},
  pages={1--34},
  year={2025},
  publisher={Springer}
}

@article{ermagun2018spatiotemporal,
  title={{Spatiotemporal traffic forecasting: Review and proposed directions}},
  author={Ermagun, Alireza and Levinson, David},
  journal={Transport Reviews},
  volume={38},
  number={6},
  pages={786--814},
  year={2018},
  publisher={Taylor \& Francis}
}

@inproceedings{cho2014learning,
  title = {{Learning phrase representations using RNN encoder-decoder for statistical machine translation}},
  author = {Cho, Kyunghyun and Van Merri{\"e}nboer, Bart and Gulcehre, Caglar and Bahdanau, Dzmitry and Bougares, Fethi and Schwenk, Holger and Bengio, Yoshua},
  booktitle = {Proceedings of the 2014 Conference on Empirical Methods in Natural Language Processing},
  pages = {1724--1734},
  year = {2014}
}

@inproceedings{huang2019leap,
  title        = {{Leap-LSTM: Enhancing long short-term memory for text categorization}},
author={Huang, Ting and Shen, Gehui and Deng, Zhi Hong},
  booktitle    = {Proceedings of the Twenty-Eighth International Joint Conference on
                  Artificial Intelligence},
  pages        = {5017--5023},
  year         = {2019},
}

@article{tsungnanlearning,
  title={{Learning long-term dependencies is not as difficult with NARX recurrent neural networks.}},
  author={Tsungnan, L and others},
  journal={University of Maryland at College Park},
    year={1995},
  pages={23}
}

@book{box2015time,
  title={Time series analysis: Forecasting and control},
  author={Box, George EP and Jenkins, Gwilym M and Reinsel, Gregory C and Ljung, Greta M},
  year={2015},
  publisher={John Wiley \& Sons}
}

@inproceedings{krueger2016zoneout,
  author={Krueger, David and Maharaj, Tegan and Kram{\'a}r, J{\'a}nos and Pezeshki, Mohammad and Ballas, Nicolas and Ke, Nan Rosemary and Goyal, Anirudh and Bengio, Yoshua and Courville, Aaron and Pal, Chris},
  title        = {{Zoneout: Regularizing RNNs by randomly preserving hidden activations}},
  booktitle    = {Proceedings of the 5th International Conference on Learning Representations},
  year         = {2017}

}

@article{wang2019optimizing,
  title={Optimizing echo state network with backtracking search optimization algorithm for time series forecasting},
  author={Wang, Zhigang and Zeng, Yu-Rong and Wang, Sirui and Wang, Lin},
  journal={Engineering Applications of Artificial Intelligence},
  volume={81},
  pages={117--132},
  year={2019},
  publisher={Elsevier}
}

@inproceedings{ranzato2015sequence,
  title = {Sequence level training with recurrent neural networks},
  author = {Ranzato, Marc’Aurelio and Chopra, Sumit and Auli, Michael and Zaremba, Wojciech},
  booktitle = {Proceedings of the 4th International Conference on Learning Representations},
  year = {2016}
}

@inproceedings{kong2025unlocking,
  title={{Unlocking the power of LSTM for long term time series forecasting}},
  author={Kong, Yaxuan and Wang, Zepu and Nie, Yuqi and Zhou, Tian and Zohren, Stefan and Liang, Yuxuan and Sun, Peng and Wen, Qingsong},
  booktitle={Proceedings of the AAAI Conference on Artificial Intelligence},
  volume={39},
  pages={11968--11976},
  year={2025}
}

@article{sima2025enhancing,
  title={Enhancing echo state network with reservoir state selection for time series forecasting},
  author={Sima, Qi and Bao, Yukun and Zhang, Xinze and He, Kun and Lai, Xin},
  journal={Neurocomputing},
  pages={131283},
  year={2025},
  publisher={Elsevier}
}

@inproceedings{oreshkin2019n,
  title={{N-BEATS: Neural basis expansion analysis for interpretable time series forecasting}},
  author={Oreshkin, Boris N and Carpov, Dmitri and Chapados, Nicolas and Bengio, Yoshua},
  booktitle = {Proceedings of the International Conference on Learning Representations},
  year = {2020}
}

@inproceedings{yoon2018invase,
  title={{INVASE: Instance-wise variable selection using neural networks}},
  author={Yoon, Jinsung and Jordon, James and Van der Schaar, Mihaela},
  booktitle={Proceedings of the International Conference on Learning Representations},
  year={2018}
}

@article{gebremeskel2021long,
  title={{Long-term evolution of energy and electricity demand forecasting: The case of Ethiopia}},
  author={Gebremeskel, Dawit Habtu and Ahlgren, Erik O and Beyene, Getachew Bekele},
  journal={Energy Strategy Reviews},
  volume={36},
  pages={100671},
  year={2021},
  publisher={Elsevier}
}

@inproceedings{zhang2016architectural,
  title={Architectural complexity measures of recurrent neural networks},
  author={Zhang, Saizheng and Wu, Yuhuai and Che, Tong and Lin, Zhouhan and Memisevic, Roland and Salakhutdinov, Ruslan and Bengio, Yoshua},
  booktitle={Proceedings of the 30th International Conference on Neural Information Processing Systems},
  pages={1830--1838},
  year={2016}
}

@inproceedings{irie2016lstm,
  title={{LSTM, GRU, highway and a bit of attention: An empirical overview for language modeling in speech recognition}},
  author={Irie, Kazuki and T{\"u}ske, Zolt{\'a}n and Alkhouli, Tamer and Schl{\"u}ter, Ralf and Ney, Hermann and others},
  booktitle={Interspeech},
  pages={3519--3523},
  year={2016}
}

@inproceedings{campos2017skip,
  title = {{Skip RNN: Learning to skip state updates in recurrent neural networks}},
  author = {Campos, Víctor and Jou, Brendan and Giró-i-Nieto, Xavier and Torres, Jordi and Chang, Shih Fu},
  booktitle = {Proceedings of the International Conference on Learning Representations},
  year = {2018}
}

@inproceedings{gui2019long,
  title={Long short-term memory with dynamic skip connections},
  author={Gui, Tao and Zhang, Qi and Zhao, Lujun and Lin, Yaosong and Peng, Minlong and Gong, Jingjing and Huang, Xuanjing},
  booktitle={Proceedings of the AAAI Conference on Artificial Intelligence},
  volume={33},
  pages={6481--6488},
  year={2019}
}

@article{ileberi2022machine,
  title={{A machine learning based credit card fraud detection using the GA algorithm for feature selection}},
  author={Ileberi, Emmanuel and Sun, Yanxia and Wang, Zenghui},
  journal={Journal of Big Data},
  volume={9},
  number={1},
  pages={24},
  year={2022},
  publisher={Springer}
}

@article{nadimi2022enhanced,
  title={{Enhanced whale optimization algorithm for medical feature selection: A COVID-19 case study}},
  author={Nadimi-Shahraki, Mohammad H and Zamani, Hoda and Mirjalili, Seyedali},
  journal={Computers in Biology and Medicine},
  volume={148},
  pages={105858},
  year={2022},
  publisher={Elsevier}
}

@article{liu2021automated,
  title={Automated feature selection: A reinforcement learning perspective},
  author={Liu, Kunpeng and Fu, Yanjie and Wu, Le and Li, Xiaolin and Aggarwal, Charu and Xiong, Hui},
  journal={IEEE Transactions on Knowledge and Data Engineering},
  volume={35},
  number={3},
  pages={2272--2284},
  year={2021},
  publisher={IEEE}
}

@article{weerakody2023policy,
  title={{Policy gradient empowered LSTM with dynamic skips for irregular time series data}},
  author={Weerakody, Philip B and Wong, Kok Wai and Wang, Guanjin},
  journal={Applied Soft Computing},
  volume={142},
  pages={110314},
  year={2023},
  publisher={Elsevier}
}

@article{chang2017dilated,
  title={Dilated recurrent neural networks},
  author={Chang, Shiyu and Zhang, Yang and Han, Wei and Yu, Mo and Guo, Xiaoxiao and Tan, Wei and Cui, Xiaodong and Witbrock, Michael and Hasegawa-Johnson, Mark A and Huang, Thomas S},
  journal={Advances in Neural Information Processing Systems},
  volume={30},
  year={2017}
}

@inproceedings{hansen2019neural,
title={{Neural speed reading with structural-jump-LSTM}},
  author={Hansen, Christian and Hansen, Casper and Alstrup, Stephen and Simonsen, Jakob Grue and Lioma, Christina},
  booktitle    = {Proceedings of the 7th International Conference on Learning Representations},
  year         = {2019}
}

@article{neil2016phased,
  title={{Phased LSTM: Accelerating recurrent network training for long or event-based sequences}},
  author={Neil, Daniel and Pfeiffer, Michael and Liu, Shih-Chii},
  journal={Advances in Neural Information Processing Systems},
  volume={29},
  year={2016}
}


\section*{Declaration of competing interest}
The authors declare that they have no known competing interests that could have appeared to influence the work reported in this paper.

\section*{Acknowledgments}
This work was supported by the National Natural Science Foundation of China under Grant 71931005, the China Postdoctoral Science Foundation under Grant 2024M761027, and the Hubei Provincial Natural Science Foundation of China under Grant 2025AFB110.

\newpage
\appendix
\section{Illustration of the proposed RRE framework with the PPO4Pred algorithm} \label{sec:para}
In this appendix, we illustrate the procedures for implementing the proposed RRE framework with our PPO4Pred method on a full training set. Let $\mathbb{D}$ denote the full training set. We follow the standard preprocessing procedure to divide $\mathbb{D}$ into multiple data batches $\mathcal{D}$. Then, we proceed with $I$ rounds of iterative asynchronous training to enhance both the RNN-based predictor and the agent, as shown in \cref{alg:training}.

For details of the parameters used in our proposed approach, we employ an iterative asynchronous training with $I = 20$ total training rounds. 
Within each round $i$, the agent policy network is trained for $G_\pi = 50$ epochs while the RNN environment model undergoes $G_\mathcal{F} = 20$ training epochs. 
The size of replay buffer $M$ within each data batch $\mathcal{D}$ is set to $M = |\mathcal{D}| \times T$, where $|\mathcal{D}|$ denotes the batch size of $\mathcal{D}$.
Then, we sample $\check{M}$ transitions from the replay buffer $\mathcal{B}^i$ to accelerate the agent training.
We set the batch size of $|\mathcal{D}| = 32$ for the ILI dataset and $|\mathcal{D}| = 256$ for the other four datasets.
Correspondingly, we sample $\check{M} = 64$ transitions for and $\check{M} = 128$ for the four other datasets.
The discount factor is set as $\gamma = 0.95$ for estimating TD error in \cref{eq:td_error}, advantage score in \cref{eq:advance}, and actual return value in \cref{eq:return}.
The PPO clipping range in \cref{eq:policy_loss} is configured with $\varepsilon = 0.2$ to constrain policy updates and ensure training stability. 

The dynamic transition sampling mechanism is governed by several key parameters that balance exploration and exploitation. 
The reward threshold in \cref{eq:reward} is set as $c = 0.5 $ to identify high-value transitions.
The reward sensitivity parameter in \cref{eq:reward} and \cref{eq:normalized_error} is configured as $\alpha = 1.0$.
The priority balance coefficient in \cref{eq:priority} is set as $\beta = 0.5$, providing equal weight between reward-based and model-uncertainty-based sampling priorities.  
Temperature scheduling in \cref{eq:temperature} ranges from $\lambda_{\min}=0.1$ to $\lambda_{\max}=2.0$, modulated by cyclical variations in \cref{eq:cyclical_temp} with amplitude $\mu=0.2$ and frequency $\omega=2$ to promote diverse sampling patterns across training phases. 

To ensure fair comparisons, we set the data split ratio for each dataset chronologically at 7:1:2 for training, validation, and testing, respectively. To prevent overfitting and ensure robust convergence, we implement comprehensive early stopping criteria throughout the training pipeline. During pretraining, the RNN environment model is evaluated using validation MSE with a patience of 10 epochs, while the iterative training phase monitors validation return with a patience of 6 epochs for $G_\mathcal{F}$. A minimum improvement threshold of 1E-3 is required to reset the patience counter, and the best performing model checkpoint is preserved based on validation metrics. 
Across the total of $I$ training rounds, early stopping is also employed to terminate the iteration process with a patience of 5 epochs.

Both the policy network and value network employ Transformer encoders with identical architectural configurations but separate parameters. Each encoder consists of 3 layers with a hidden dimension $D_e$ of 256. The multi-head self-attention mechanism in each layer utilizes 8 attention heads, while the feedforward network has a dimension of 1024. To enhance training stability and prevent overfitting, we apply a dropout rate of 0.1 throughout the network.

\begin{breakablealgorithm}
\caption{Training Procedure of RRE Framework with PPO4Pred}
\label{alg:training}
\begin{algorithmic}[1]
\Require   
Training set $\mathbb{D}$, Total rounds of asynchronous training $I$, agent training epochs $G_\pi$, RNN training epochs $G_\mathcal{F}$, mini-batch size $\check{M}$, discount factor $\gamma$, 
clipping range $\varepsilon$, learning rates $\eta_{\phi}, \eta_{\varphi}, \eta_{\theta}$.
\Ensure RNN-based predictor $\mathcal{F}_\theta$ and policy network $\pi_\phi$.

\State Pre-train RNN environment $\mathcal{F}_\theta^0$ on traing set $\mathbb{D}$ with conventional configuration.


\For{ $i=1$ to $I$}
    \State Initialize the collection $\mathbb{B} = \emptyset$.
    \For{each data batch $\mathcal{D}$ in traing set $\mathbb{D}$}
        \State Execute \cref{alg:experience_collection} to obtain experience buffer $\mathcal{B}^i = \{\zeta_m\}_{m=1}^{M}$.
        \State Store the data batch and the corresponding experience buffer as $\mathbb{B} = \mathbb{B} \cup (\mathcal{D}, \mathcal{B}^i)$.
    \EndFor
    \State{Initialize the agent as $\pi_{\phi_1}^i \leftarrow \pi_{\phi}^{i-1}$ and $\upsilon_{\varphi_1}^i \leftarrow \upsilon_{\varphi}^{i-1}$.}
    \For{$g=1 $ to $G_\pi$}
    \For{each batch $(\mathcal{B}^i, \mathcal{D})$ in the collection $\mathbb{B}$}
        \State Execute \cref{alg:dynamic_sampling} to craft sampled buffer $\check{\mathcal{B}}_g^i = \{\zeta_{\check m}\}_{{\check m}=1}^{\check{M}}$ from $\mathcal{B}^i$.
        \For{each $\zeta_{\check m}$ in $\check{\mathcal{B}}_g^i$}
        \State Use $\upsilon_{\varphi_{g}}^i$ to compute $A_{g,{\check m}}^i$ and $R_{g,{\check m}}^i$ with $\gamma$, $\bm s_{\check m+1}$, $\bm s_{\check m}$ and $r_{\check m}$ via \cref{eq:advance} and \cref{eq:return}.  
        \EndFor
        \State Use $\pi_{\phi}^{i-1}$, $\pi_{\phi_g}^{i}$, $\{\zeta_{\check m}\}_{{\check m}=1}^{\check{M}}$, and $\{A_{g,{\check m}}^i\}_{\check m=1}^{\check{M}}$ to compute policy loss $\mathcal{L}(\phi_{g})$ via \cref{eq:policy_loss}.
        \State Use $\upsilon_{\varphi_{g}}^i$, $\{\zeta_{\check m}\}_{{\check m}=1}^{\check{M}}$, and $\{R_{g,{\check m}}^i\}_{\check m=1}^{\check{M}}$ to compute value loss $\mathcal{L}(\varphi_{g})$ via \cref{eq:critic_loss}.
        \State Update weight parameters $\phi_{g+1} \leftarrow \phi_{g}$, $\varphi_{g+1} \leftarrow \varphi_{g}$ with $\eta_{\phi}$, $\eta_{\varphi}$ via \cref{eq:agentupdate}.   
    \EndFor
\EndFor
\State Update policy network $\pi^i_\phi \leftarrow \pi^i_{\phi_{G_\pi}}$ and value network $\upsilon^i_\varphi \leftarrow \upsilon^i_{\varphi_{G_\pi}}$.

    \State{Initialize RNN as $\mathcal{F}_{\theta_1}^{i} \leftarrow  \mathcal{F}_{\theta}^{i-1}$.}
    \For{$g=1 $ to $G_\mathcal{F}$}
    \For{each data batch $\mathcal{D}$ in training set $\mathbb{D}$}
    
    \For{each ($X,Y$) in $\mathcal{D}$ }
        \State Generate actions $\bm{a}_{1:T}$ via trained policy network $\pi_{\phi}^i$.
         \State Compute the RNN loss $\mathcal{L}(\theta_g)$ with $X$, $Y$, $\bm{a}_{1:T}$, and $\theta_g$ via \cref{eq:gateh}--\cref{eq:gatey}.        
    \EndFor
    
    \State Compute the average loss of the data batch $\mathcal{D}$ as:$\mathcal{L}(\theta_g) \leftarrow \mathbb{E}_{\mathcal{D}} [\mathcal{L}(\theta_g)]$.
    \State Update weight parameters $\theta_{g+1} = \theta_{g} - \eta_{\theta} \nabla_\theta \mathcal{L}(\theta_g)$.
    
    \EndFor 
\EndFor
    \State Update RNN environment network $\mathcal{F}_{\theta}^i \leftarrow \mathcal{F}_{\theta_{G_\mathcal{F}}}^{i}$.
\EndFor
\State \Return The final RNN $\mathcal{F}^I_\theta$ and policy network $\pi^I_\phi$.

\end{algorithmic}
\end{breakablealgorithm}

\section{Detailed experimental results} \label{sec:deti}
This section provides comprehensive experimental results to supplement the main findings presented in the paper. We present detailed performance comparisons across different methods and provide ablation study results.

\begin{figure}[t]
    \centering
    \begin{subfigure}[b]{0.32\textwidth}
        \centering
        \includegraphics[width=\textwidth]{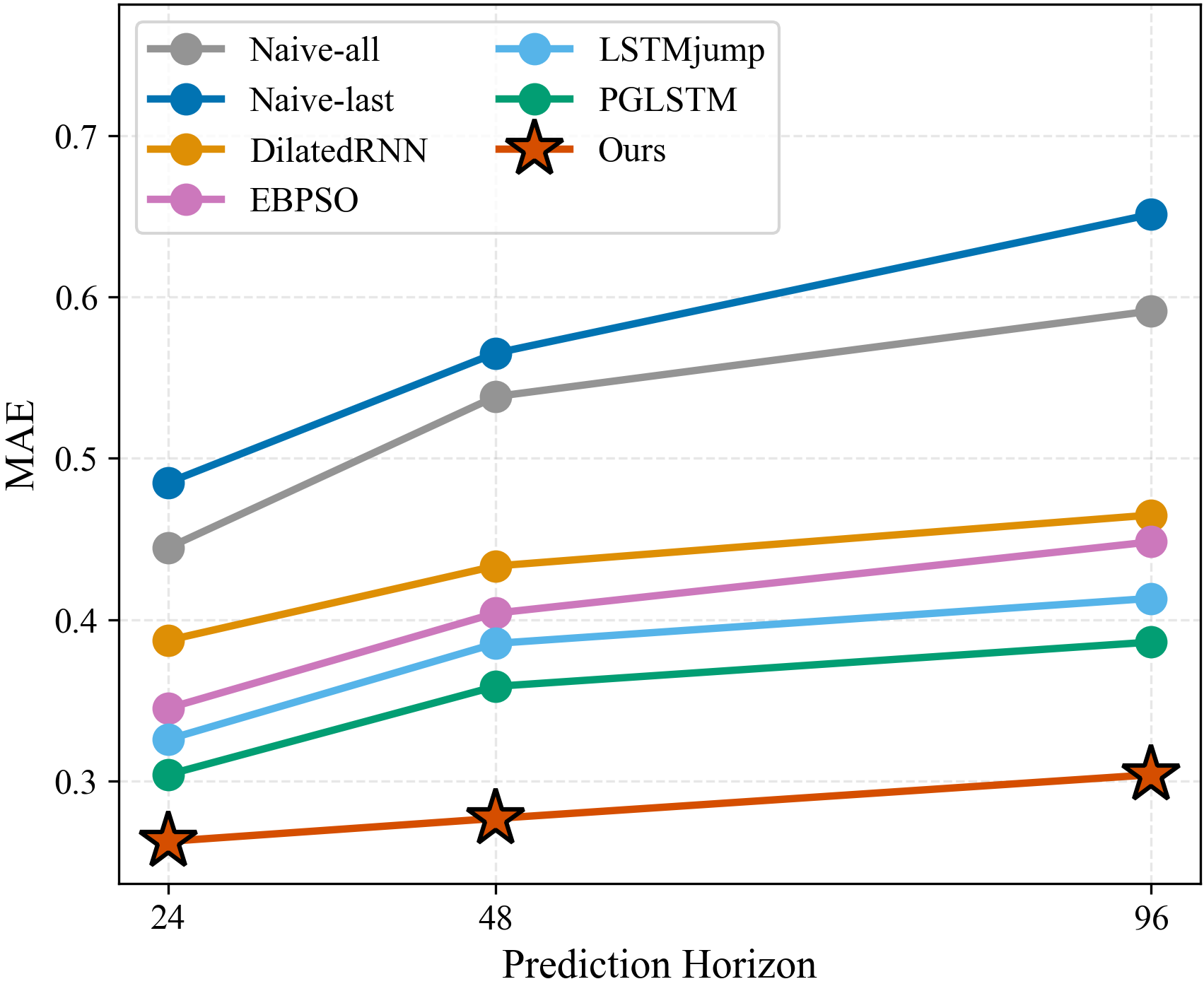} 
        \caption{Traffic  }
    \end{subfigure}
    \begin{subfigure}[b]{0.32\textwidth}
        \centering
        \includegraphics[width=\textwidth]{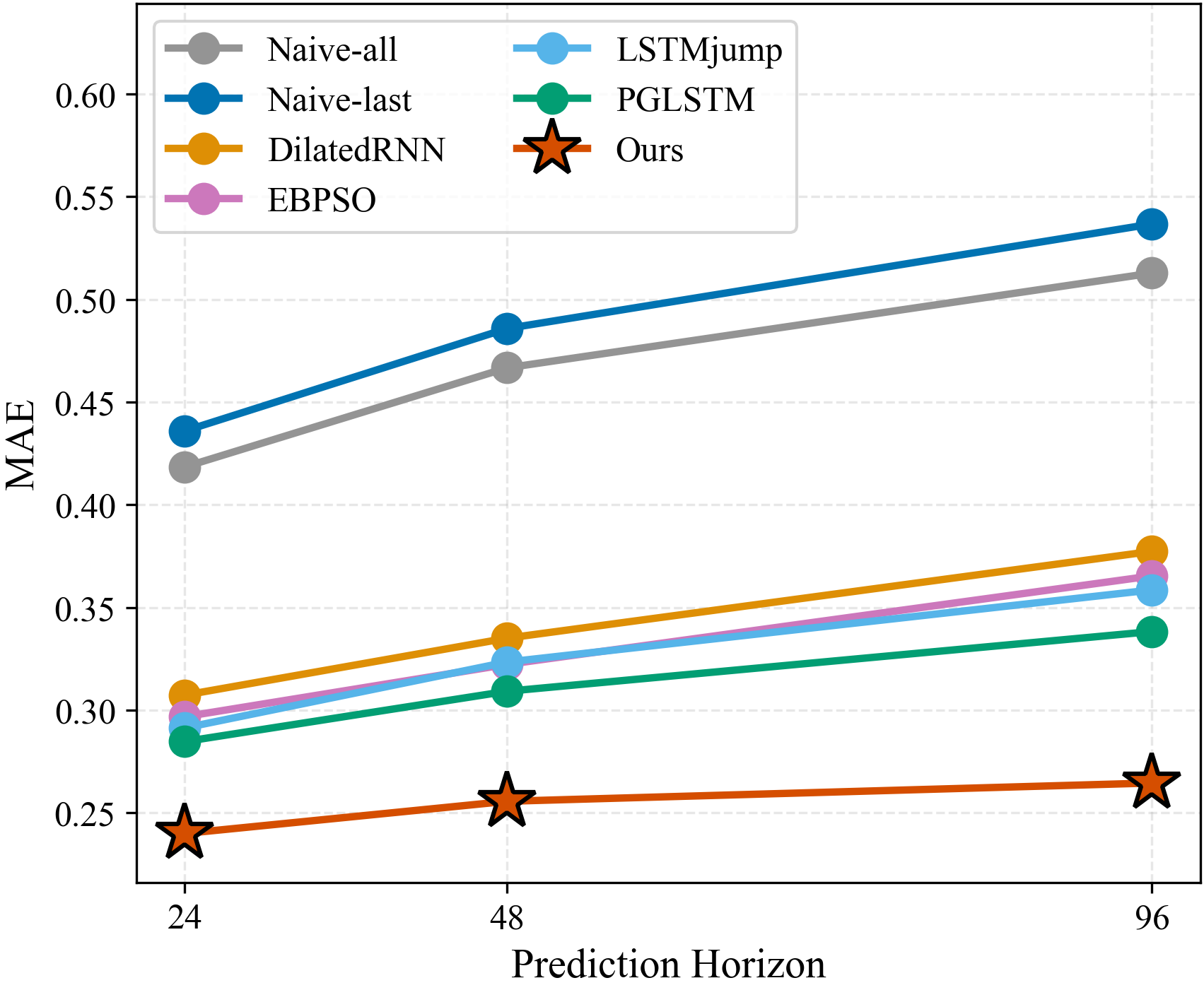} 
        \caption{Electricity  }
    \end{subfigure}
    \begin{subfigure}[b]{0.32\textwidth}
        \centering
        \includegraphics[width=\textwidth]{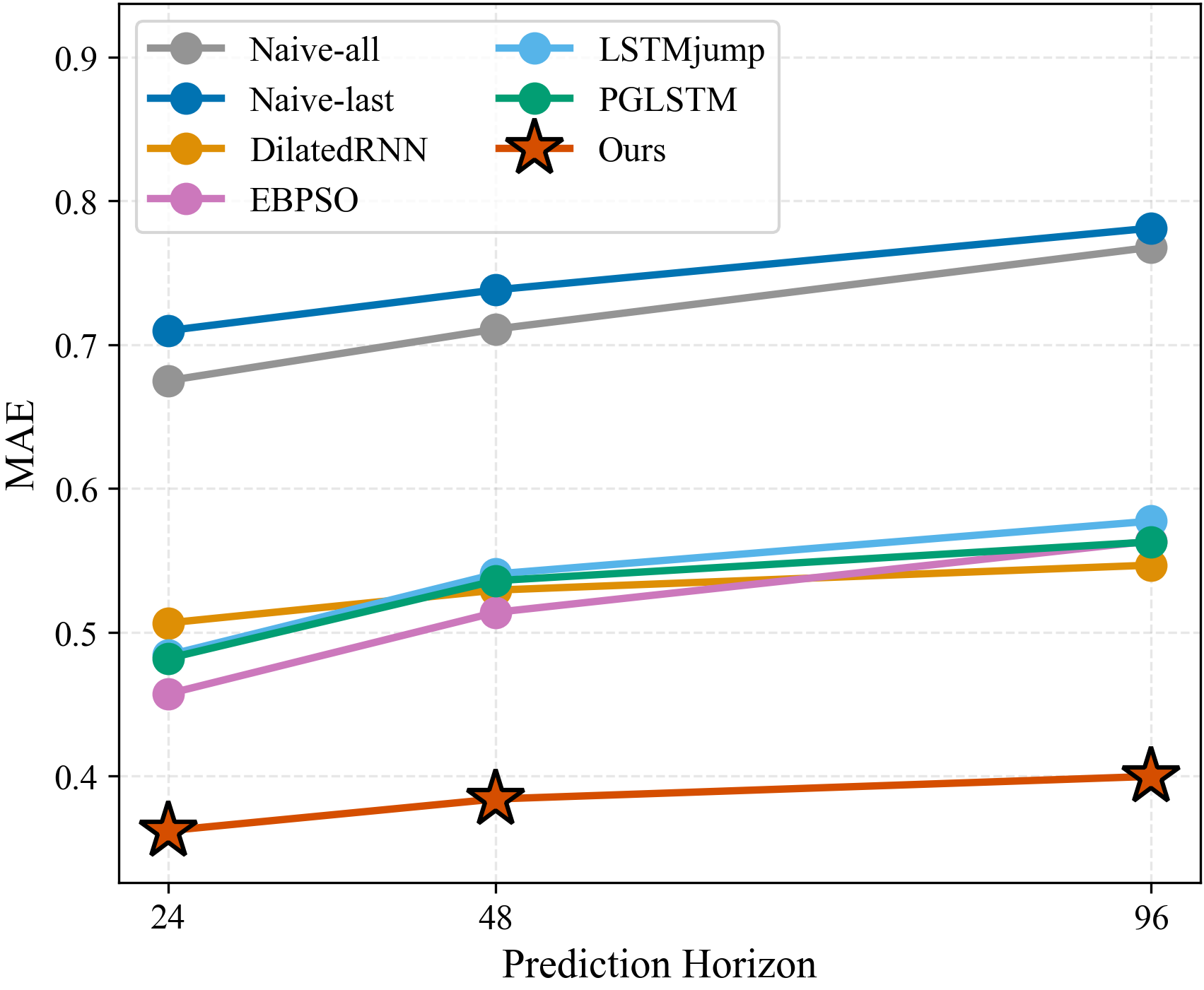} 
        \caption{ETTh  }
    \end{subfigure}
    \hfill 
    \vspace{3mm}
    \begin{subfigure}[b]{0.32\textwidth}
        \centering
        \includegraphics[width=\textwidth]{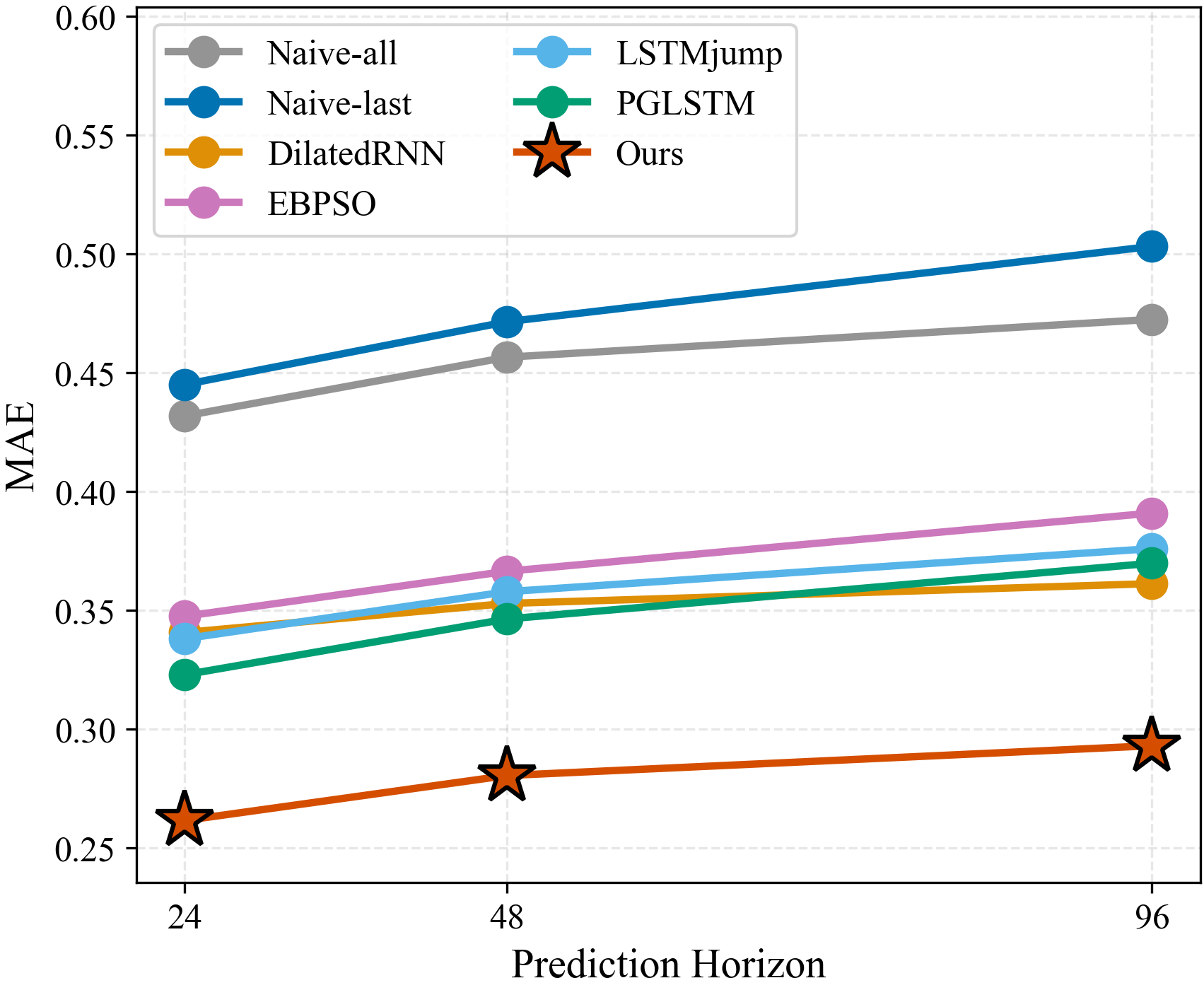} 
        \caption{Weather  }
    \end{subfigure}
    \begin{subfigure}[b]{0.32\textwidth}
        \centering
        \includegraphics[width=\textwidth]{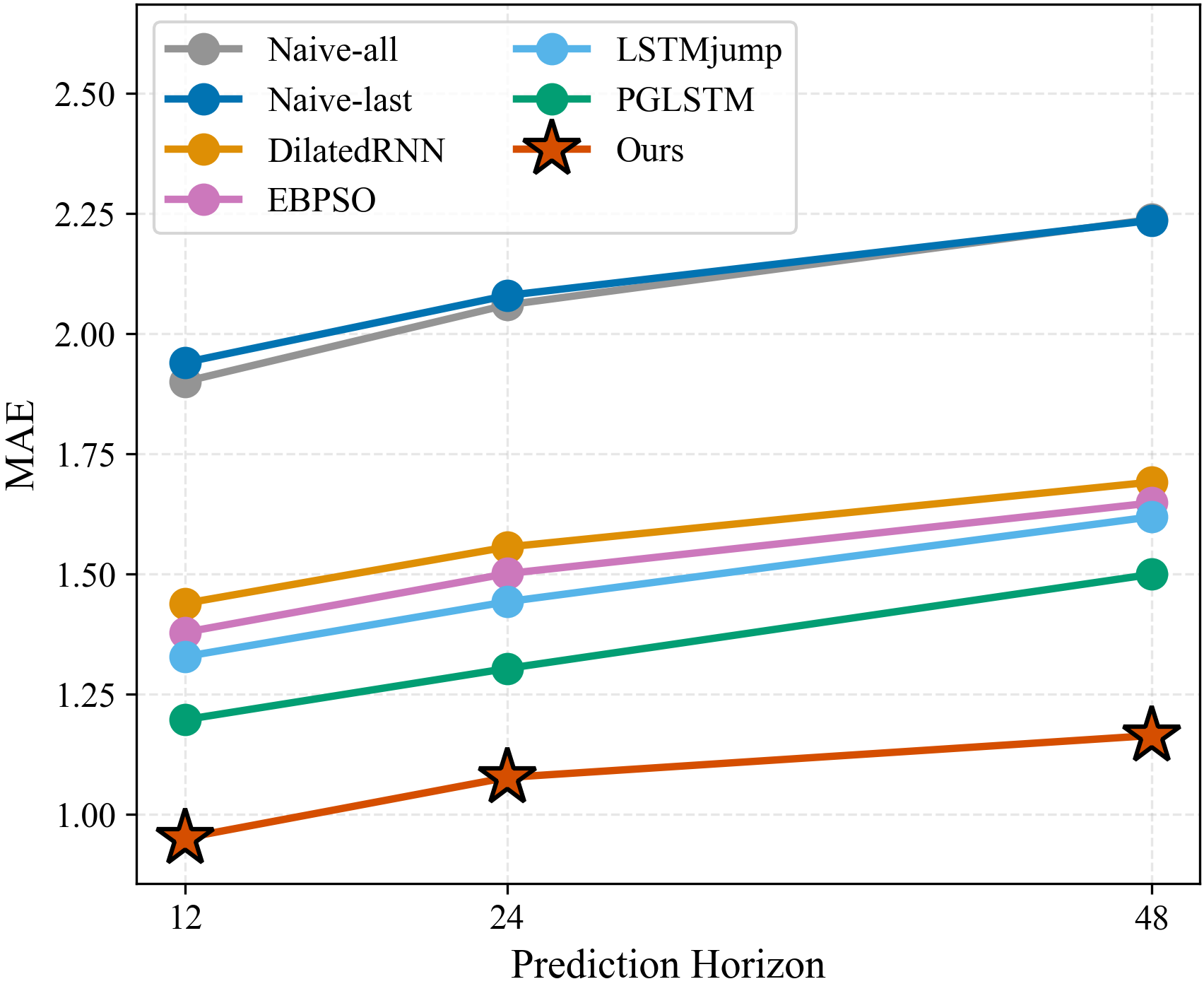} 
        \caption{ILI  }
    \end{subfigure}
    
    \caption{Comparison of MAE averaged across all backbone architectures for the proposed RRE-PPO4Pred method and baseline approaches at varying prediction horizons.}
    \label{fig:horison-mae}
\end{figure}

\begin{figure}[htbp]
    \centering
    \begin{subfigure}[b]{0.32\textwidth}
        \centering
        \includegraphics[width=\textwidth]{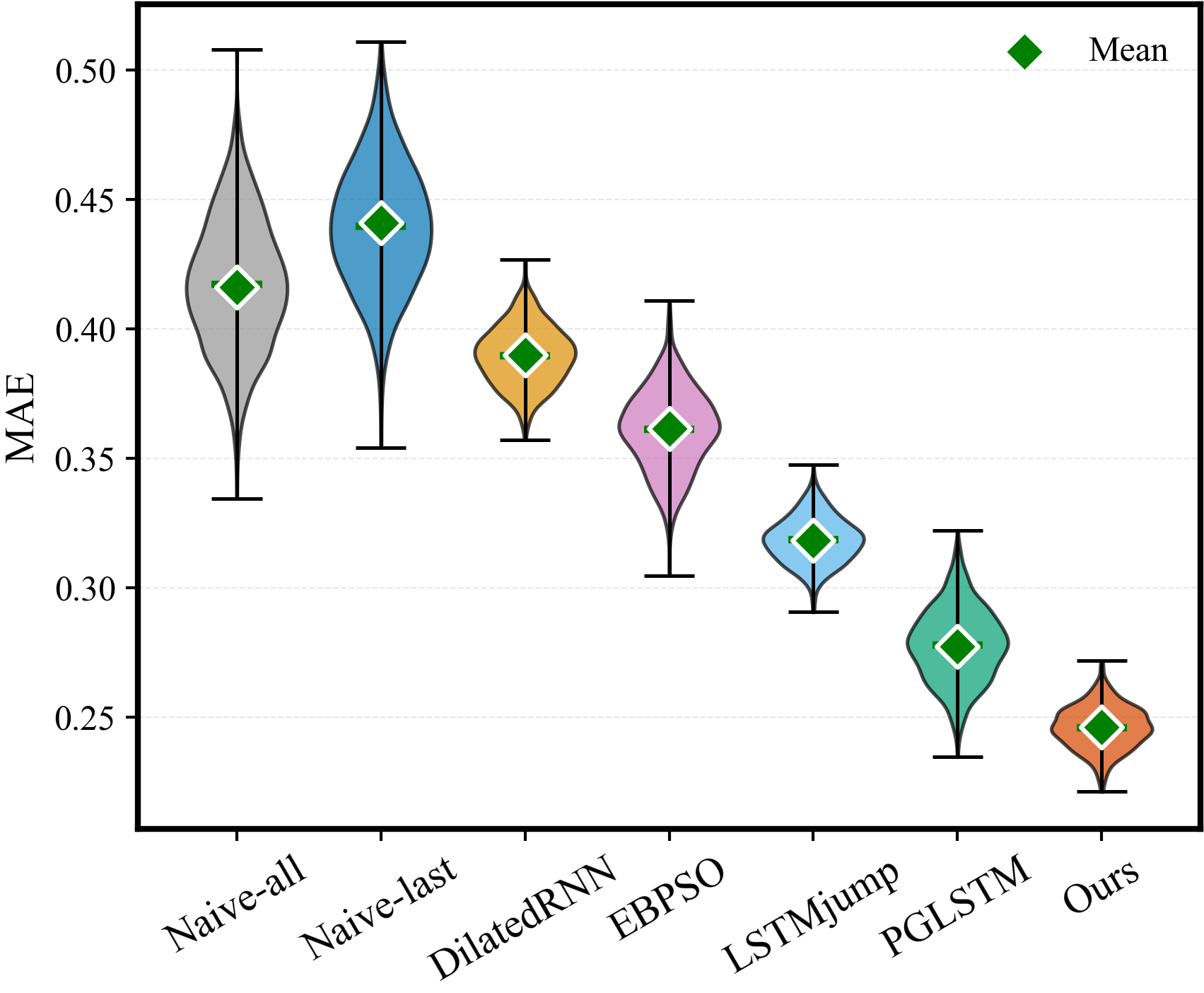} 
        \caption{Traffic  }
    \end{subfigure}
    \begin{subfigure}[b]{0.32\textwidth}
        \centering
        \includegraphics[width=\textwidth]{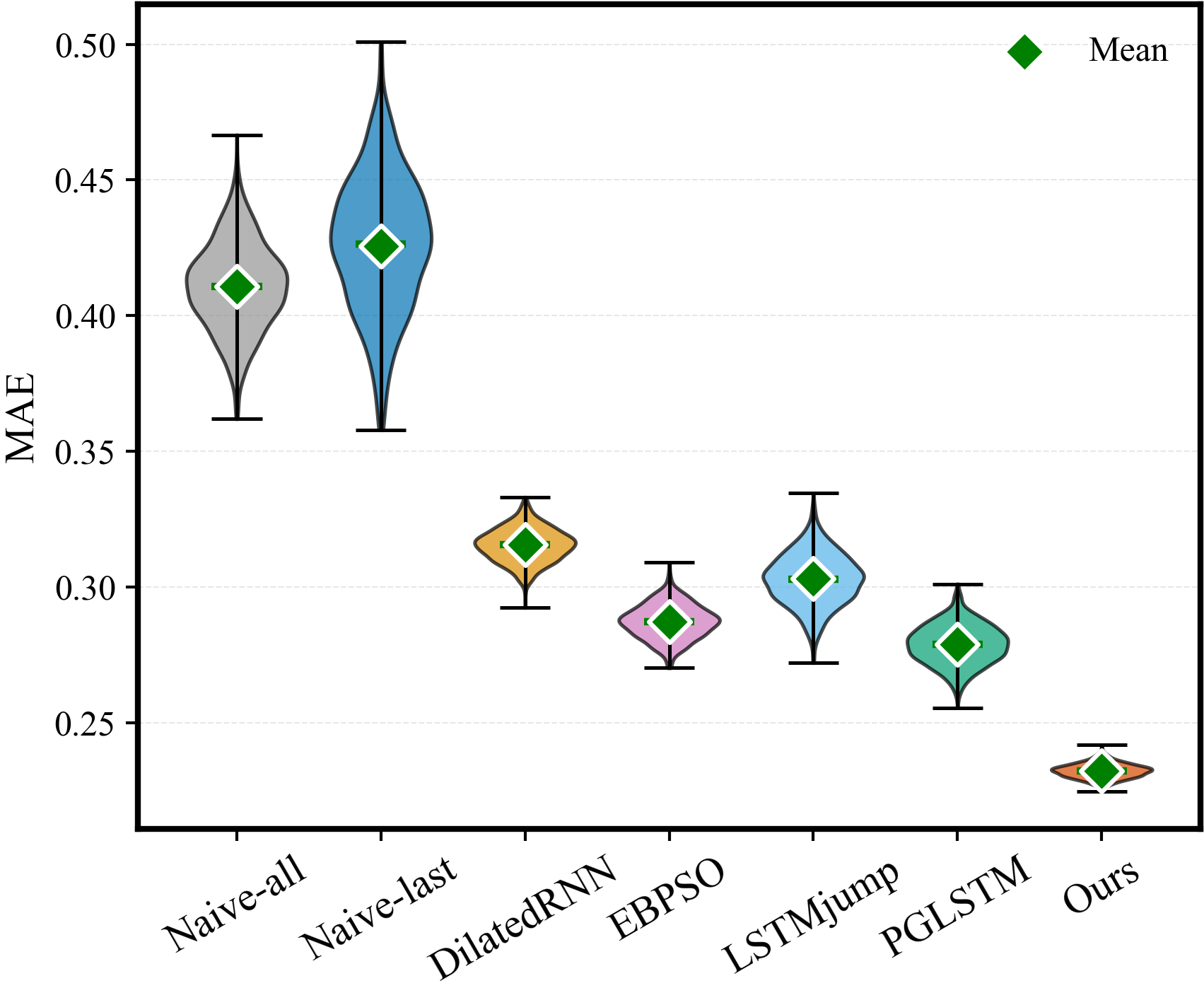} 
        \caption{ Electricity  }
    \end{subfigure}
        \begin{subfigure}[b]{0.32\textwidth}
        \centering
        \includegraphics[width=\textwidth]{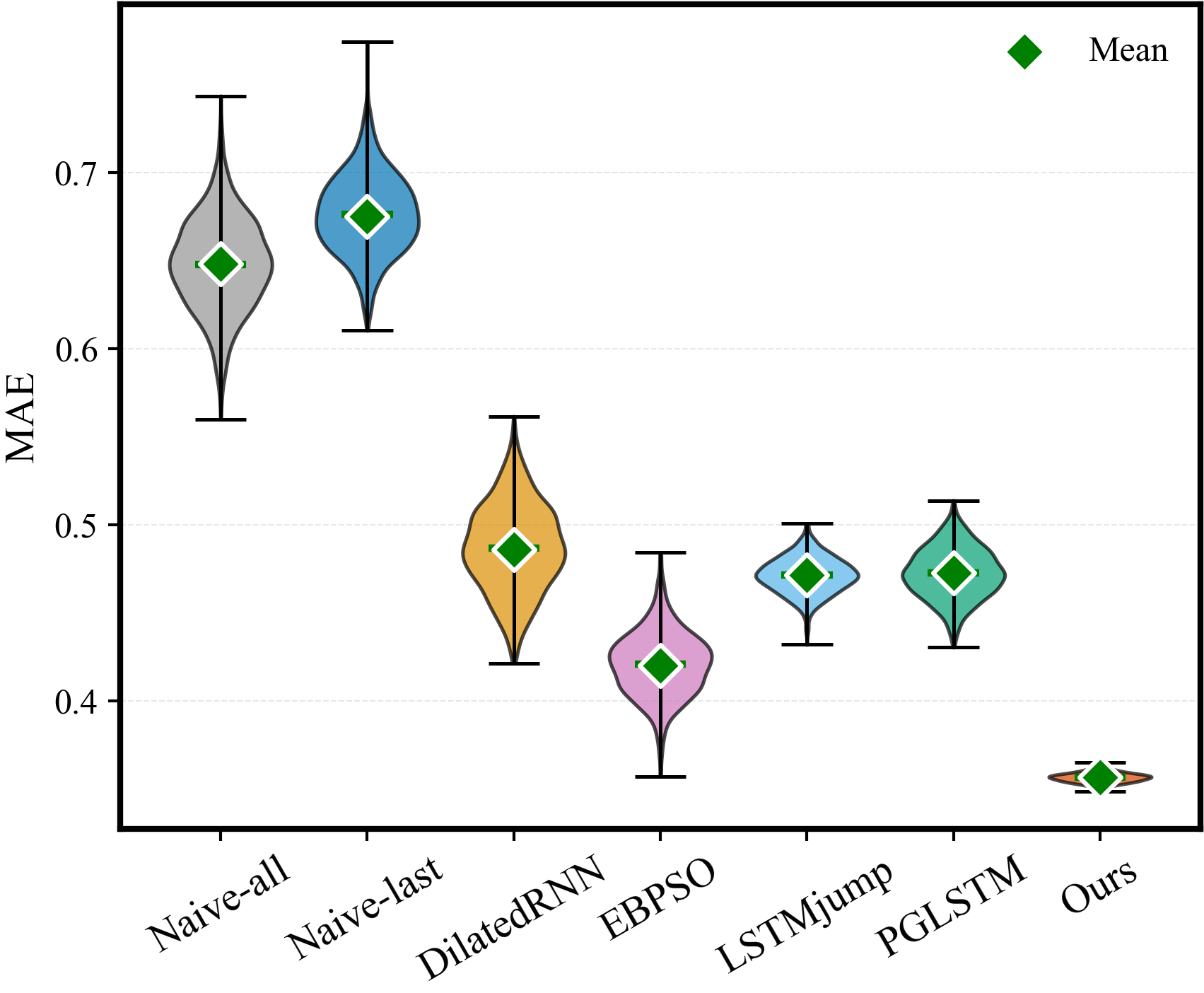} 
        \caption{ETTh }
    \end{subfigure}
    \hfill 
    \vspace{2mm}
    \begin{subfigure}[b]{0.32\textwidth}
        \centering
        \includegraphics[width=\textwidth]{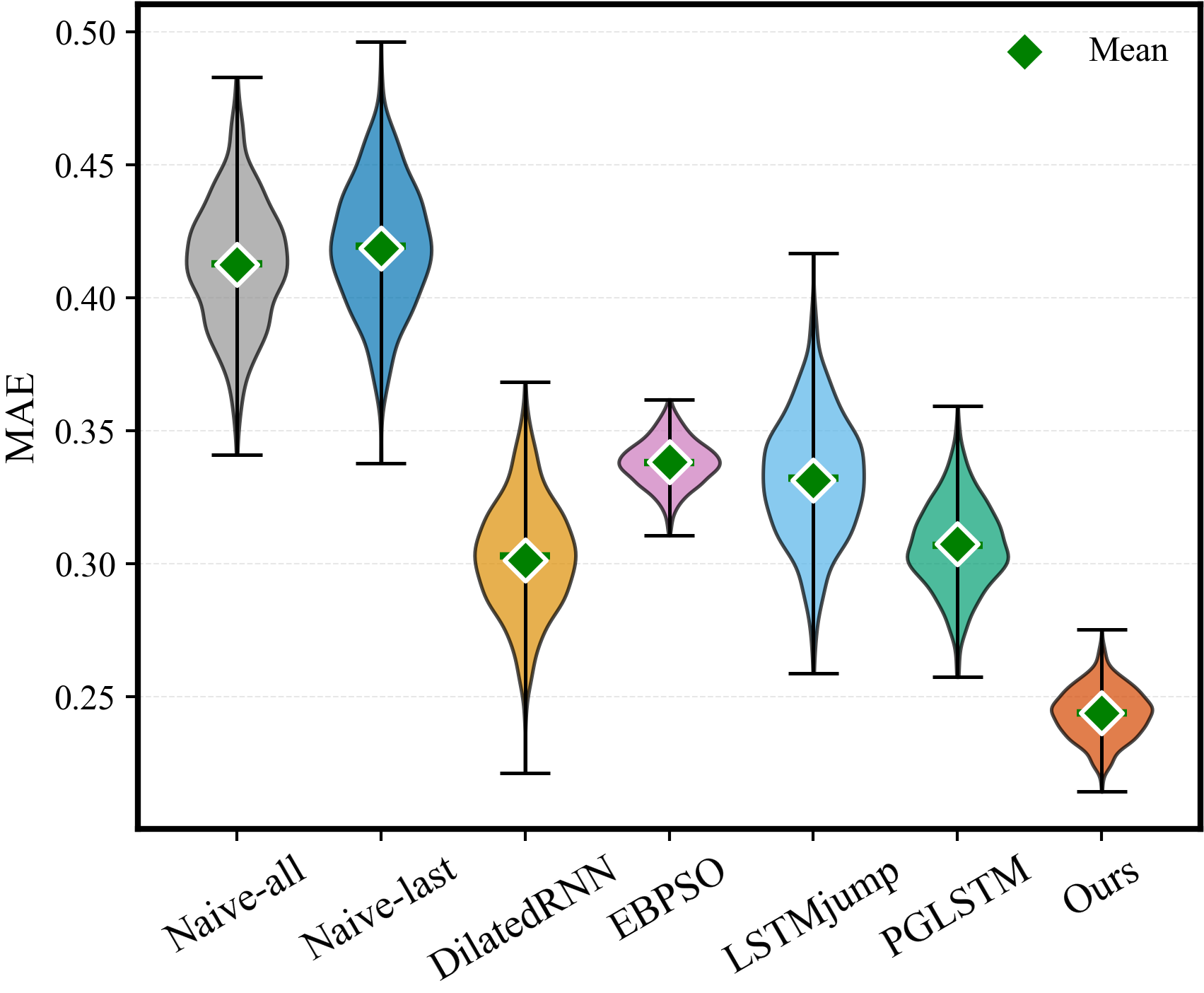} 
        \caption{Weather }
    \end{subfigure}
    \begin{subfigure}[b]{0.32\textwidth}
        \centering
        \includegraphics[width=\textwidth]{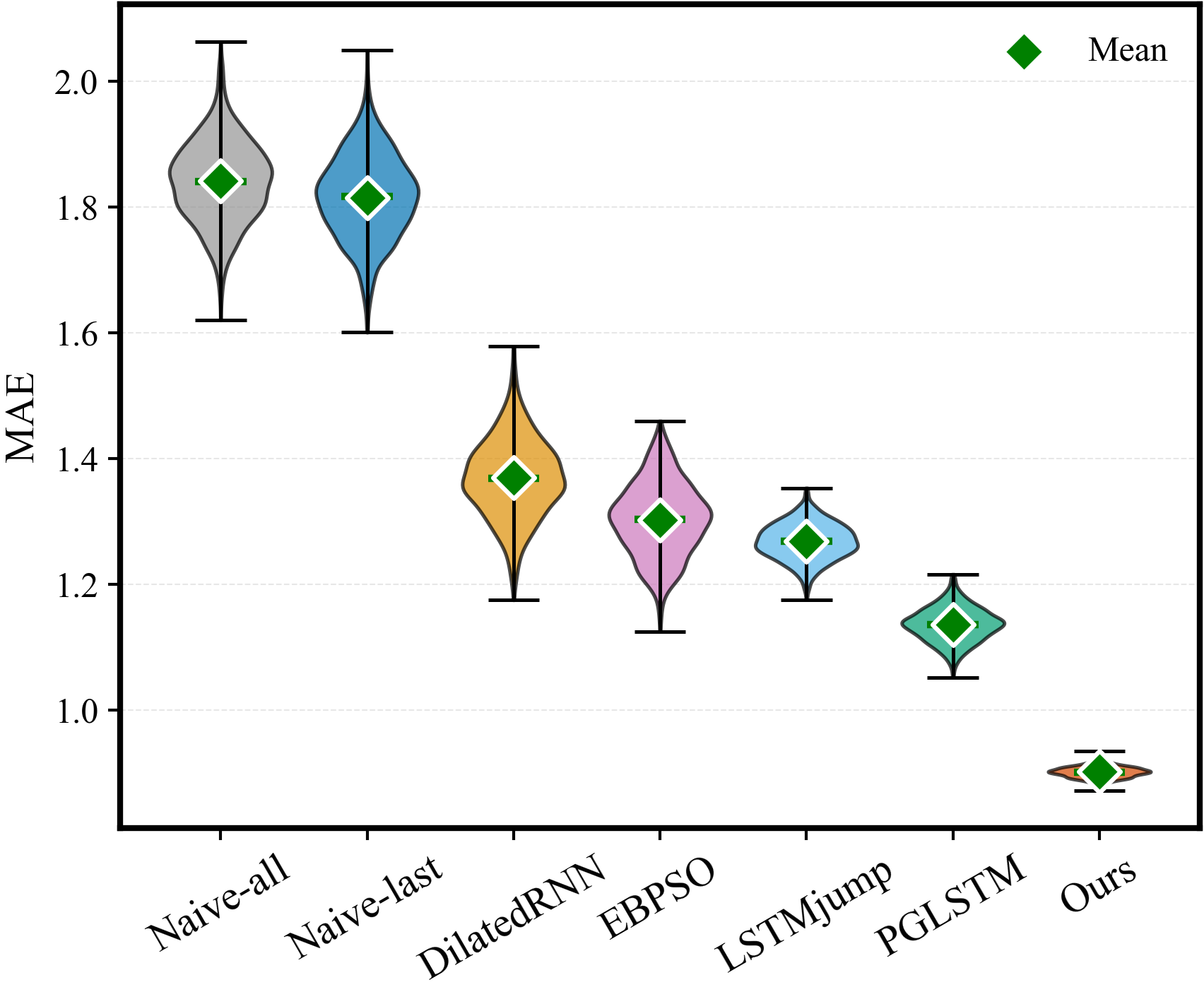} 
        \caption{ILI }
    \end{subfigure}
    
    \caption{Violin plots of MAE performance for RRE-PPO4Pred and baseline methods with xLSTM backbone  at horizon 12 for ILI dataset and horizon 24 for the remaining datasets}
    \label{fig:violin-mae}
\end{figure}

\begin{figure}[htbp]
    \centering
    \begin{subfigure}[b]{0.32\textwidth}
        \centering
        \includegraphics[width=\textwidth]{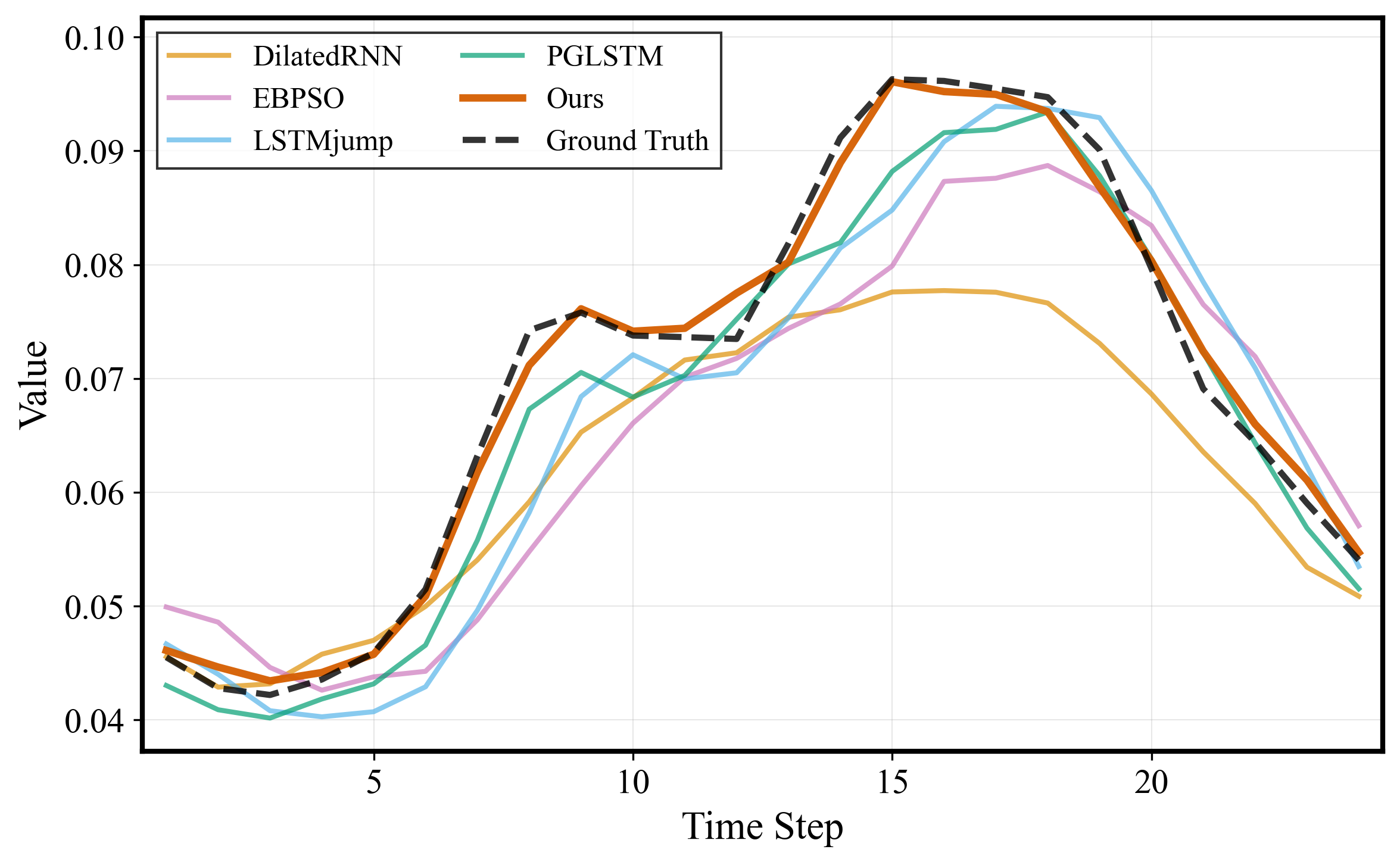} 
        \caption{Traffic  }
        \label{fig:slices_traffic}
    \end{subfigure}
    \begin{subfigure}[b]{0.32\textwidth}
        \centering
        \includegraphics[width=\textwidth]{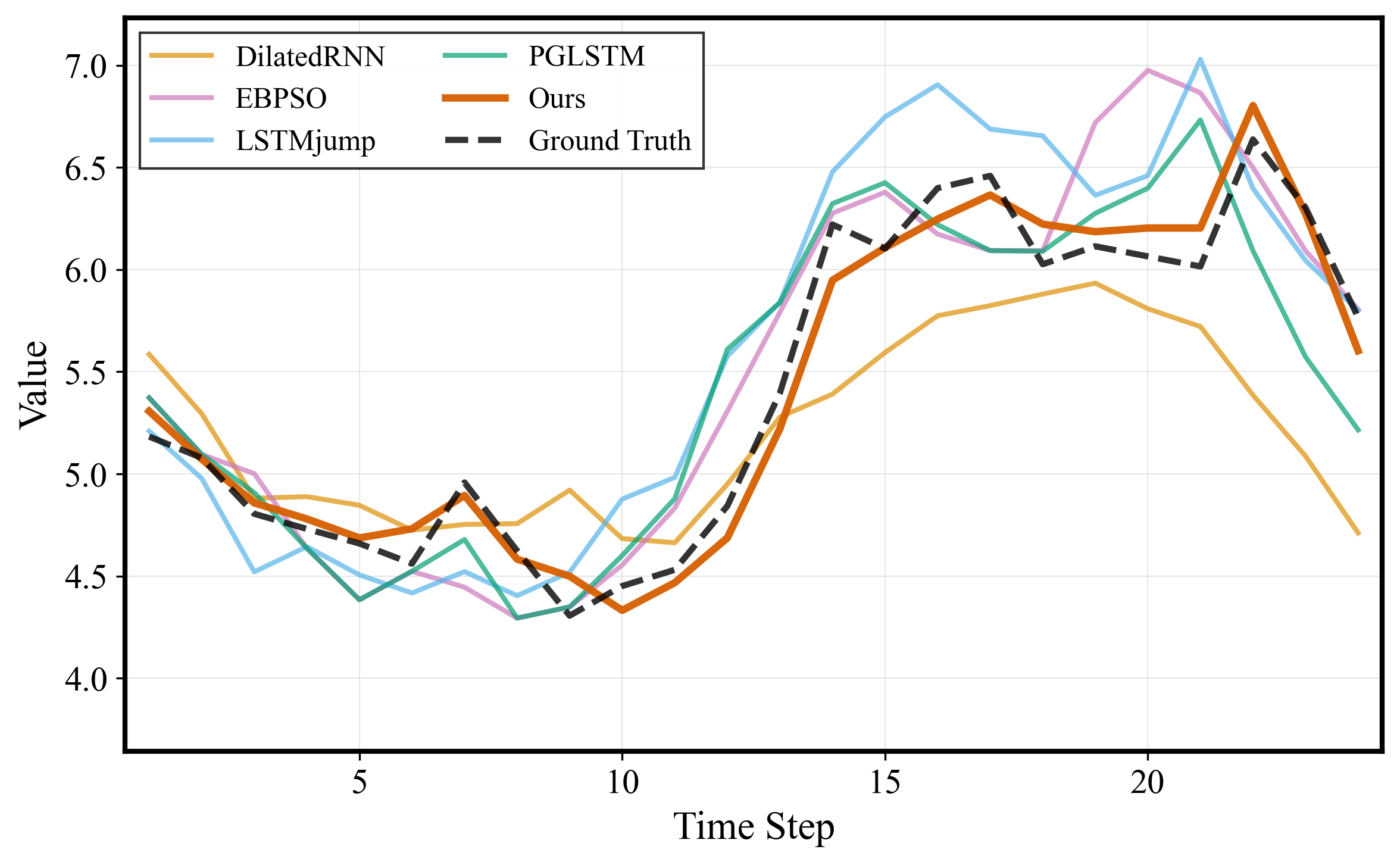} 
        \caption{Electricity }
        \label{fig:slices_Electricity}
    \end{subfigure}
        \begin{subfigure}[b]{0.32\textwidth}
        \centering
        \includegraphics[width=\textwidth]{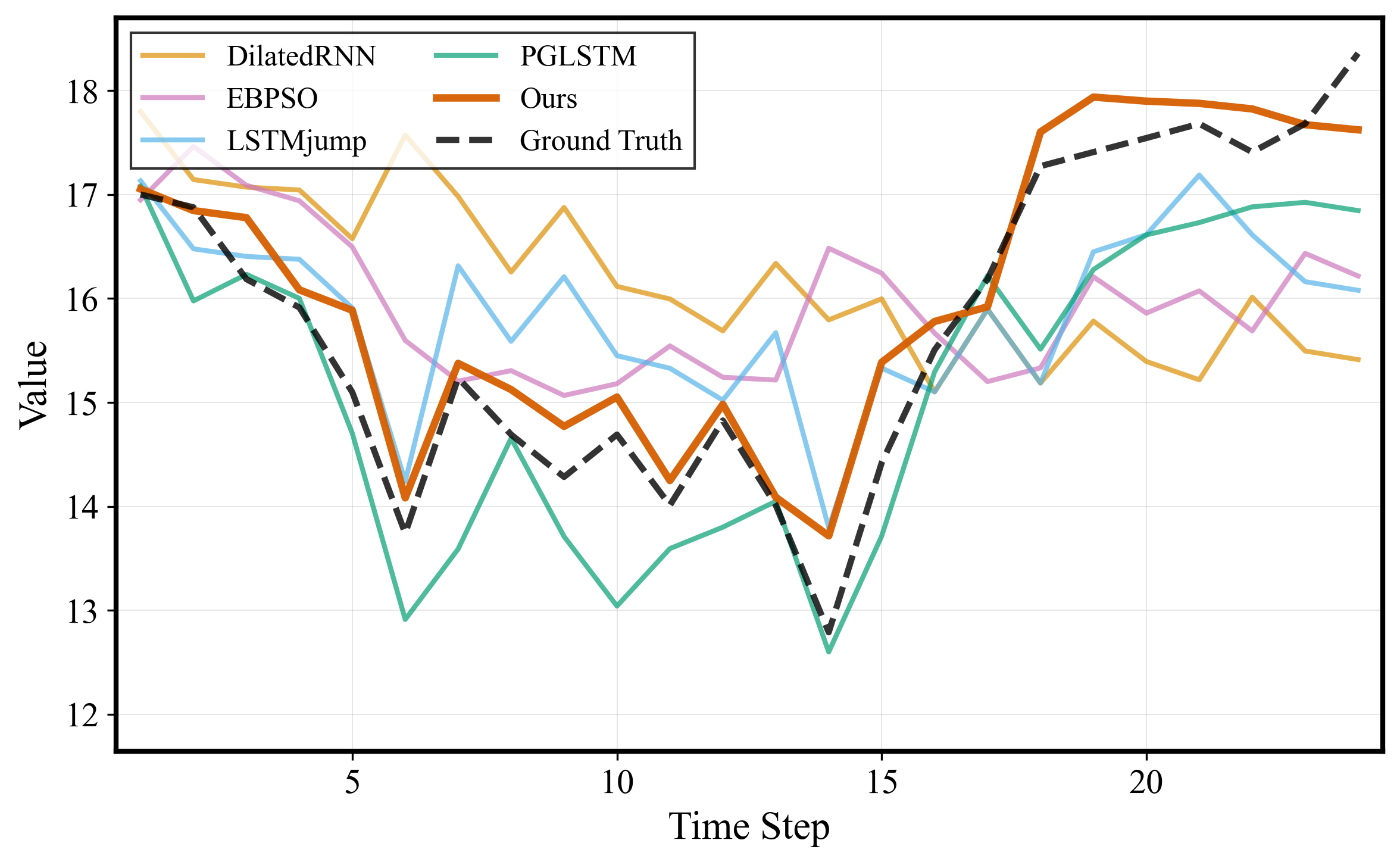} 
        \caption{ETTh }
        \label{fig:slices_ETTh}
    \end{subfigure}
        \hfill 
    \vspace{3mm}
    \begin{subfigure}[b]{0.32\textwidth}
        \centering
        \includegraphics[width=\textwidth]{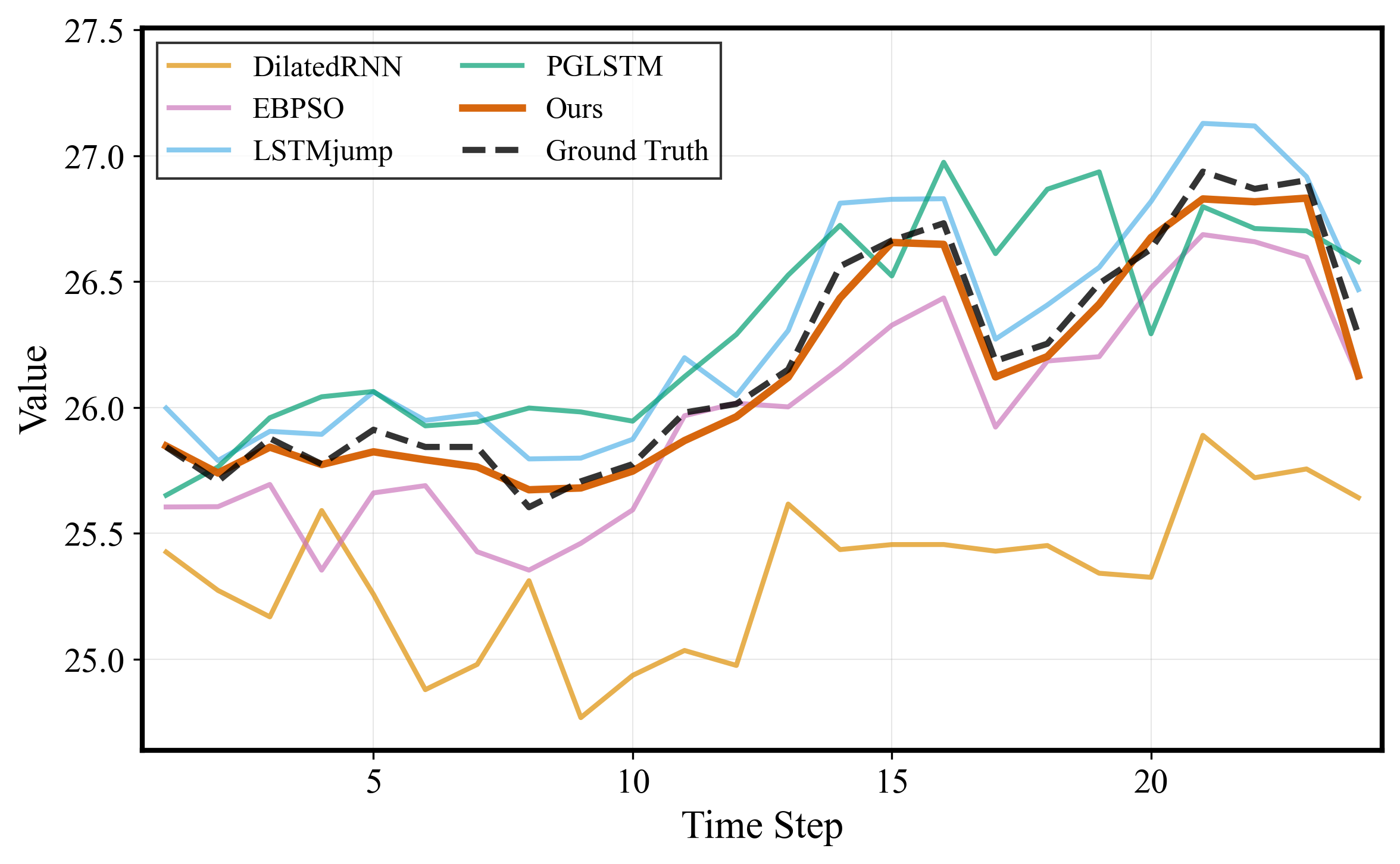} 
        \caption{Weather }
        \label{fig:slices_Weather}
    \end{subfigure}
    \begin{subfigure}[b]{0.32\textwidth}
        \centering
        \includegraphics[width=\textwidth]{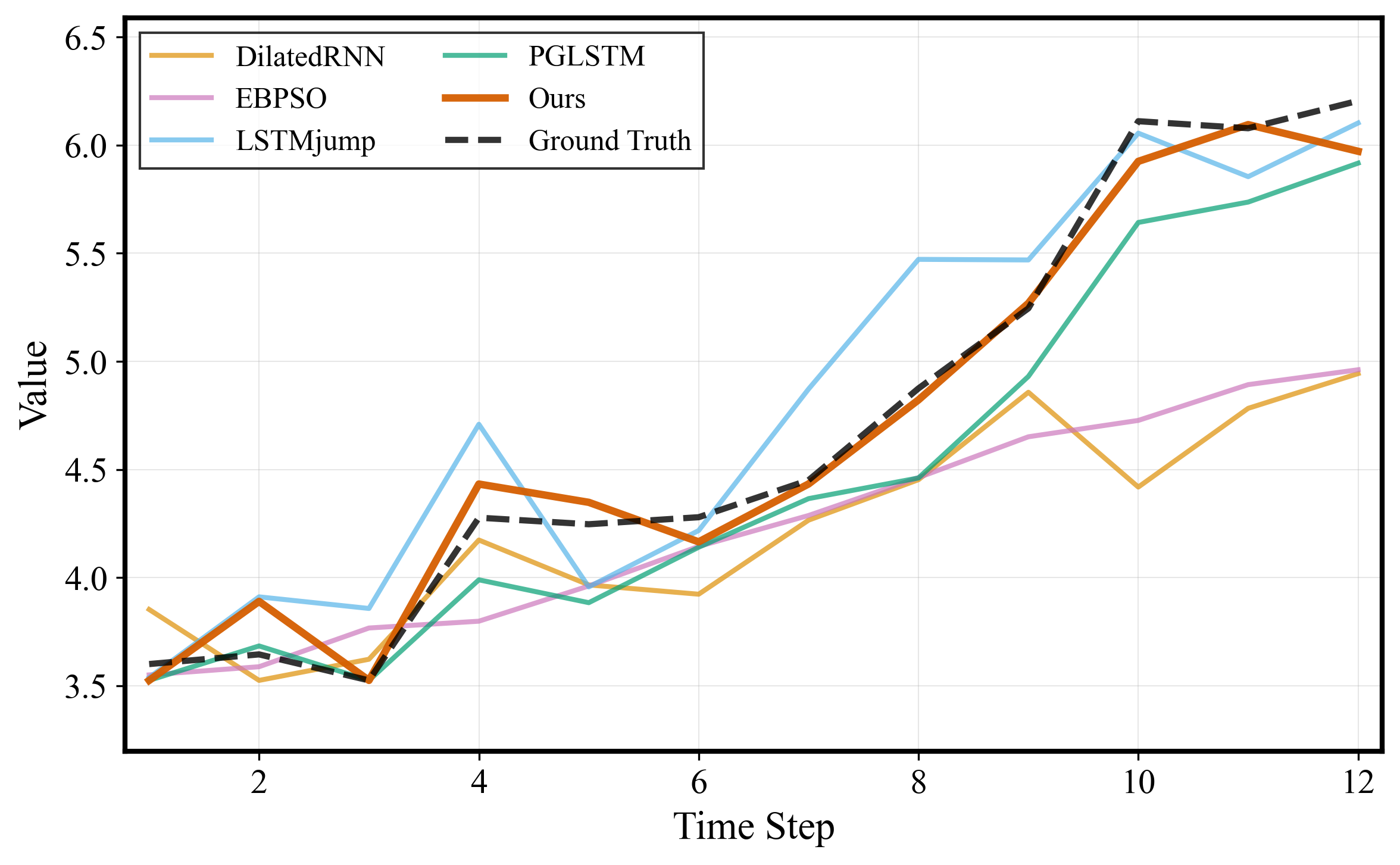} 
        \caption{ILI }
        \label{fig:slices_ILI}
    \end{subfigure}
    
    \caption{Illustration of prediction curves by RRE-PPO4Pred and baseline methods employing the xLSTM backbone at horizon 12 for the ILI dataset and horizon 24 for the remaining datasets. }
    \label{fig:slices-all}
\end{figure}

The MAE results obtained by averaging performance across all backbone architectures for our RRE-PPO4Pred method and baseline methods at varying prediction horizons on five datasets are presented in \cref{fig:horison-mae}. As shown in the figures, our method consistently outperforms all baseline approaches across different horizons, with the performance gap becoming more pronounced as the prediction horizon increases.
The violin plots, as showed in \cref{fig:violin-mae}, display the mean and variance of MAE for RRE-PPO4Pred and baseline methods with xLSTM backbone at horizon 12 for ILI and horizon 24 for other datasets. The narrower distribution and lower mean values of our method indicate not only superior average performance but also greater stability compared to baseline methods.
The prediction fitting curves are illustrated in \cref{fig:slices-all}, depicting our RRE-PPO4Pred method alongside all baseline methods utilizing the xLSTM backbone. The forecast horizons are set to 12 for the ILI dataset and 24 for the remaining datasets. 
The results demonstrate that RRE-PPO4Pred consistently achieves superior prediction accuracy across all five datasets, with prediction curves closely tracking the ground truth and outperforming all baseline methods.

\begin{figure}[t]
    \centering
    \includegraphics[width=0.85\linewidth]{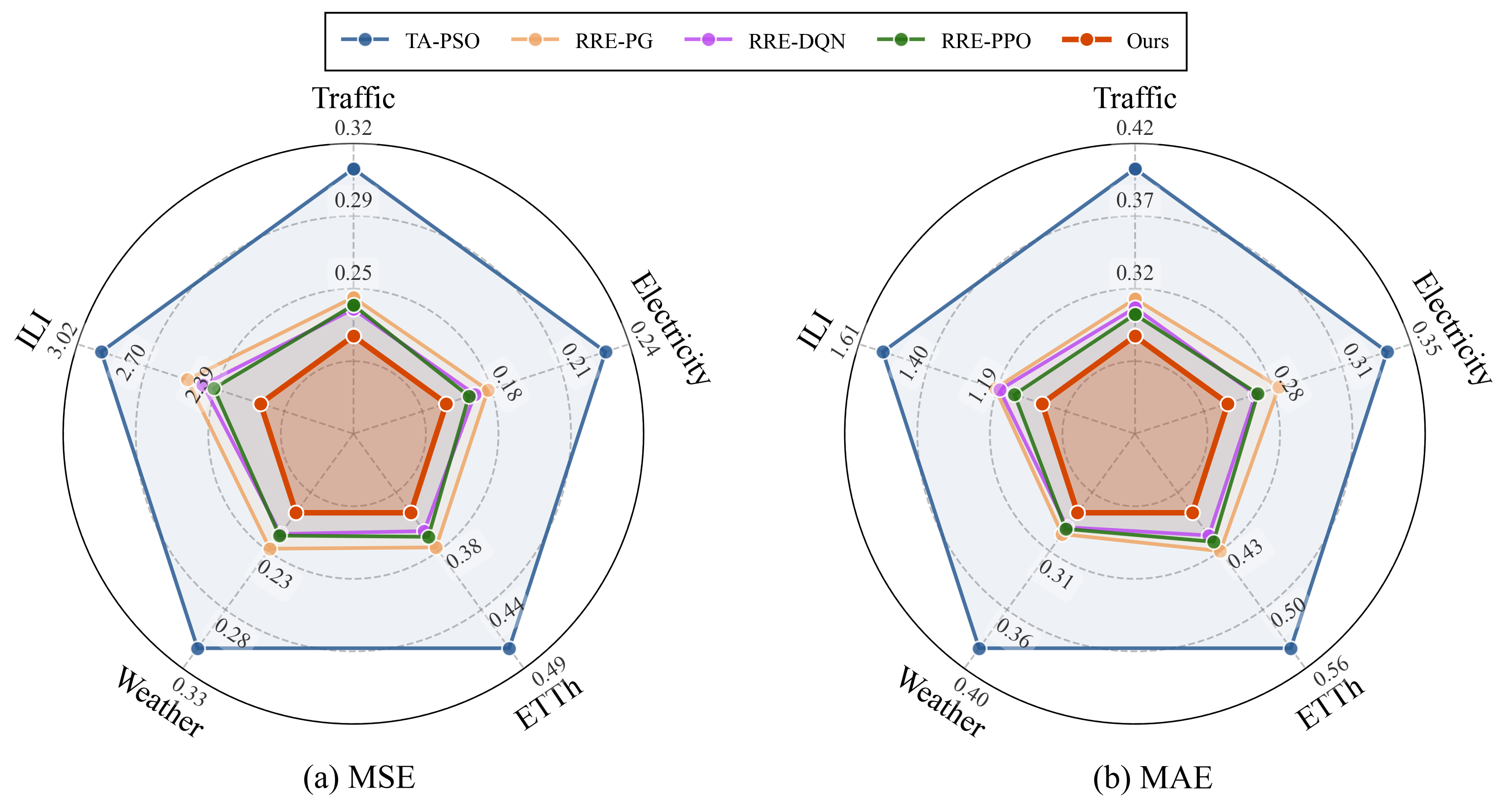}
    \caption{Radar charts of MSE and MAE averaged over all horizons for RRE-PPO4Pred and its ablative methods.}
    \label{fig:radar-ablation}
\end{figure}

The detailed ablation study results for each forecasting horizon across all five datasets are presented in Tables \ref{tab:ablation-traffic}--\ref{tab:ablation-ILI}. These tables provide comprehensive MSE and MAE values for RRE-PPO4Pred and all ablative variants (TA-PSO, RRE-PG, RRE-DQN, and RRE-PPO) at different prediction horizons.
The ``Count'' indicates the number of times each method achieves the best performance across all horizon-backbone settings on each dataset. 
Furthermore, radar charts of the average MSE and MAE for RRE-PPO4Pred and the ablative methods on five datasets are provided in \cref{fig:radar-ablation}.
The results clearly demonstrate that each component of our framework contributes to the overall performance improvement, with the full RRE-PPO4Pred model achieving the best results in most cases.

\begin{table}[t]
\centering
\caption{Average MSE and MAE performance of our RRE-PPO4Pred method  and its ablation variant methods over different backbones and different horizons on Traffic data.}
\label{tab:ablation-traffic}
\resizebox{\textwidth}{!}{%
\begin{tabular}{@{}cc|ccccc|ccccc@{}}
\toprule
\multirow{2}{*}{$H$} &
  \multirow{2}{*}{Backbone} &
  \multicolumn{5}{c|}{MSE} &
  \multicolumn{5}{c}{MAE} \\ \cmidrule(l){3-12} 
 &
   &
  TA-PSO &
  RRE-PG &
  RRE-DQN &
  RRE-PPO &
  Ours  &
  TA-PSO &
  RRE-PG &
  RRE-DQN &
  RRE-PPO &
  Ours  \\ \midrule
\multirow{8}{*}{24} &
  RNN &
  0.2752 &
  0.2313 &
  \textbf{0.2222} &
  0.2247 &
  0.2251 &
  0.3301 &
  0.2757 &
  0.2679 &
  0.2771 &
  \textbf{0.2640} \\
 &
  MGU &
  0.2706 &
  0.2238 &
  0.2221 &
  0.2234 &
  \textbf{0.2202} &
  0.3501 &
  0.2777 &
  0.2768 &
  0.2819 &
  \textbf{0.2667} \\
 &
  GRU &
  0.2785 &
  0.2239 &
  0.2173 &
  0.2226 &
  \textbf{0.2108} &
  0.3348 &
  0.2799 &
  \textbf{0.2681} &
  0.2691 &
  0.2699 \\
 &
  LSTM &
  0.2826 &
  0.2210 &
  0.2217 &
  0.2239 &
  \textbf{0.2129} &
  0.3583 &
  0.2864 &
  0.2640 &
  0.2698 &
  \textbf{0.2582} \\
 &
  IndRNN &
  0.2730 &
  0.2334 &
  \textbf{0.2218} &
  0.2314 &
  0.2227 &
  0.3592 &
  0.2973 &
  0.3088 &
  0.2897 &
  \textbf{0.2734} \\
 &
  phLSTM &
  0.2684 &
  0.2333 &
  0.2328 &
  \textbf{0.2320} &
  0.2321 &
  0.3754 &
  0.2757 &
  0.2888 &
  0.2748 &
  \textbf{0.2555} \\
 &
  pLSTM &
  0.2732 &
  0.2279 &
  0.2198 &
  0.2217 &
  \textbf{0.2159} &
  0.3802 &
  0.2894 &
  0.2795 &
  0.2769 &
  \textbf{0.2657} \\
 &
  xLSTM &
  0.2657 &
  0.2206 &
  0.2207 &
  0.2216 &
  \textbf{0.2184} &
  0.3240 &
  0.2848 &
  0.2630 &
  0.2676 &
  \textbf{0.2463} \\ \midrule
\multirow{8}{*}{48} &
  RNN &
  0.3137 &
  0.2447 &
  0.2542 &
  0.2533 &
  \textbf{0.2249} &
  0.4372 &
  0.3150 &
  0.3013 &
  0.2902 &
  \textbf{0.2791} \\
 &
  MGU &
  0.2960 &
  0.2532 &
  \textbf{0.2313} &
  0.2477 &
  0.2397 &
  0.4181 &
  0.3092 &
  0.2873 &
  0.2936 &
  \textbf{0.2824} \\
 &
  GRU &
  0.3044 &
  0.2638 &
  0.2388 &
  0.2409 &
  \textbf{0.2280} &
  0.4286 &
  0.2995 &
  0.2805 &
  0.2861 &
  \textbf{0.2784} \\
 &
  LSTM &
  0.2968 &
  0.2438 &
  0.2461 &
  0.2453 &
  \textbf{0.2369} &
  0.3835 &
  0.2944 &
  \textbf{0.2743} &
  0.2864 &
  0.2850 \\
 &
  IndRNN &
  0.3057 &
  0.2550 &
  0.2506 &
  0.2471 &
  \textbf{0.2376} &
  0.4070 &
  0.3117 &
  0.3098 &
  0.2841 &
  \textbf{0.2707} \\
 &
  phLSTM &
  0.3227 &
  0.2578 &
  0.2505 &
  0.2470 &
  \textbf{0.2380} &
  0.4404 &
  0.3221 &
  0.3261 &
  0.3060 &
  \textbf{0.2723} \\
 &
  pLSTM &
  0.3077 &
  0.2511 &
  0.2528 &
  0.2518 &
  \textbf{0.2354} &
  0.4047 &
  0.3132 &
  0.3018 &
  0.2935 &
  \textbf{0.2813} \\
 &
  xLSTM &
  0.2965 &
  0.2506 &
  0.2540 &
  0.2410 &
  \textbf{0.2252} &
  0.3902 &
  0.3144 &
  0.2924 &
  0.3056 &
  \textbf{0.2644} \\ \midrule
\multirow{8}{*}{96} &
  RNN &
  0.3321 &
  0.2596 &
  0.2640 &
  0.2592 &
  \textbf{0.2380} &
  0.4609 &
  0.3316 &
  0.3170 &
  0.3207 &
  \textbf{0.3090} \\
 &
  MGU &
  0.3504 &
  0.2600 &
  \textbf{0.2418} &
  0.2618 &
  0.2449 &
  0.4677 &
  0.3200 &
  \textbf{0.2998} &
  0.3169 &
  0.3000 \\
 &
  GRU &
  0.3557 &
  0.2708 &
  0.2567 &
  0.2711 &
  \textbf{0.2460} &
  0.4465 &
  0.3319 &
  0.3220 &
  0.3213 &
  \textbf{0.3197} \\
 &
  LSTM &
  0.3569 &
  0.2699 &
  0.2731 &
  0.2691 &
  \textbf{0.2446} &
  0.4592 &
  0.3222 &
  0.3294 &
  0.3247 &
  \textbf{0.3045} \\
 &
  IndRNN &
  0.3572 &
  0.2750 &
  0.2641 &
  0.2677 &
  \textbf{0.2442} &
  0.4550 &
  0.3417 &
  0.3372 &
  0.3162 &
  \textbf{0.3065} \\
 &
  phLSTM &
  0.3467 &
  0.2752 &
  0.2665 &
  0.2625 &
  \textbf{0.2357} &
  0.4432 &
  0.3651 &
  0.3822 &
  0.3561 &
  \textbf{0.3092} \\
 &
  pLSTM &
  0.3437 &
  0.2802 &
  0.2768 &
  0.2729 &
  \textbf{0.2395} &
  0.4714 &
  0.3381 &
  0.3418 &
  0.3128 &
  \textbf{0.2989} \\
 &
  xLSTM &
  0.3381 &
  0.2706 &
  0.2573 &
  0.2617 &
  \textbf{0.2427} &
  0.4222 &
  0.3148 &
  0.3307 &
  0.3116 &
  \textbf{0.2827} \\ \midrule
\multicolumn{2}{c|}{Average} &
  0.3088 &
  0.2494 &
  0.2440 &
  0.2459 &
  \textbf{0.2316} &
  0.4062 &
  0.3088 &
  0.3021 &
  0.2972 &
  \textbf{0.2810} \\ \midrule
\multicolumn{2}{c|}{Improvement} &
  \textbf{24.99\%} &
  \textbf{7.14\%} &
  \textbf{5.08\% }&
  \textbf{5.79\% }&
  - &
  \textbf{30.81\%} &
  \textbf{9.01\% }&
  \textbf{6.98\% }&
 \textbf{ 5.45\%} &
  - \\ \midrule
\multicolumn{2}{c|}{Count} &
  0 &
  0 &
  4 &
  1 &
  \textbf{19} &
  0 &
  0 &
  3 &
  0 &
  \textbf{21} \\ \bottomrule
\end{tabular}%
}
\end{table}

\begin{table}[t]
\centering
\caption{Average MSE and MAE performance of our RRE-PPO4Pred method  and its ablation variant methods over different backbones and different horizons on Electricity data.}
\label{tab:ablation-Electricity}
\resizebox{\textwidth}{!}{%
\begin{tabular}{@{}cc|ccccc|ccccc@{}}
\toprule
\multirow{2}{*}{$H$} &
  \multirow{2}{*}{Backbone} &
  \multicolumn{5}{c|}{MSE} &
  \multicolumn{5}{c}{MAE} \\ \cmidrule(l){3-12} 
 &
   &
  TA-PSO &
  RRE-PG &
  RRE-DQN &
  RRE-PPO &
  Ours  &
  TA-PSO &
  RRE-PG &
  RRE-DQN &
  RRE-PPO &
  Ours  \\ \midrule
\multirow{8}{*}{24} &
  RNN &
  0.1810 &
  0.1545 &
  0.1447 &
  0.1526 &
  \textbf{0.1379} &
  0.3037 &
  0.2551 &
  0.2494 &
  0.2517 &
  \textbf{0.2409} \\
 &
  MGU &
  0.1897 &
  0.1515 &
  \textbf{0.1427} &
  0.1517 &
  0.1444 &
  0.3031 &
  0.2522 &
  \textbf{0.2404} &
  0.2455 &
  0.2415 \\
 &
  GRU &
  0.1812 &
  0.1553 &
  \textbf{0.1419} &
  0.1475 &
  0.1469 &
  0.3041 &
  0.2571 &
  \textbf{0.2318} &
  0.2483 &
  0.2322 \\
 &
  LSTM &
  0.1958 &
  0.1605 &
  0.1577 &
  0.1470 &
  \textbf{0.1343} &
  0.3117 &
  0.2524 &
  0.2497 &
  0.2520 &
  \textbf{0.2368} \\
 &
  IndRNN &
  0.1950 &
  0.1673 &
  0.1677 &
  0.1577 &
  \textbf{0.1451} &
  0.3042 &
  0.2689 &
  0.2604 &
  0.2521 &
  \textbf{0.2461} \\
 &
  phLSTM &
  0.2061 &
  0.1543 &
  0.1553 &
  0.1544 &
  \textbf{0.1348} &
  0.2951 &
  0.2720 &
  0.2563 &
  0.2595 &
  \textbf{0.2494} \\
 &
  pLSTM &
  0.2107 &
  0.1619 &
  0.1504 &
  0.1526 &
  \textbf{0.1392} &
  0.3082 &
  0.2614 &
  0.2428 &
  0.2442 &
  \textbf{0.2398} \\
 &
  xLSTM &
  0.1765 &
  0.1591 &
  0.1428 &
  0.1479 &
  \textbf{0.1350} &
  0.2902 &
  0.2543 &
  0.2479 &
  0.2509 &
  \textbf{0.2323} \\ \midrule
\multirow{8}{*}{48} &
  RNN &
  0.2445 &
  0.1821 &
  0.1709 &
  0.1668 &
  \textbf{0.1613} &
  0.3263 &
  0.2781 &
  0.2713 &
  0.2769 &
  \textbf{0.2462} \\
 &
  MGU &
  0.2327 &
  0.1752 &
  0.1652 &
  0.1673 &
  \textbf{0.1615} &
  0.3282 &
  0.2817 &
  0.2745 &
  0.2693 &
  \textbf{0.2644} \\
 &
  GRU &
  0.2407 &
  0.1719 &
  \textbf{0.1504} &
  0.1636 &
  0.1563 &
  0.3392 &
  0.2753 &
  \textbf{0.2434} &
  0.2663 &
  0.2481 \\
 &
  LSTM &
  0.2136 &
  0.1654 &
  0.1648 &
  0.1550 &
  \textbf{0.1512} &
  0.3227 &
  0.2691 &
  0.2650 &
  0.2738 &
  \textbf{0.2549} \\
 &
  IndRNN &
  0.2313 &
  0.1790 &
  0.1760 &
  0.1629 &
  \textbf{0.1522} &
  0.3225 &
  0.2915 &
  0.2785 &
  0.2738 &
  \textbf{0.2690} \\
 &
  phLSTM &
  0.2434 &
  0.1817 &
  0.1824 &
  0.1734 &
  \textbf{0.1565} &
  0.3399 &
  0.2860 &
  0.2791 &
  0.2722 &
  \textbf{0.2632} \\
 &
  pLSTM &
  0.2247 &
  0.1789 &
  0.1701 &
  0.1666 &
  \textbf{0.1567} &
  0.3386 &
  0.2807 &
  0.2775 &
  0.2727 &
  \textbf{0.2527} \\
 &
  xLSTM &
  0.2182 &
  0.1812 &
  0.1655 &
  0.1698 &
  \textbf{0.1538} &
  0.3218 &
  0.2870 &
  0.2742 &
  0.2630 &
  \textbf{0.2460} \\ \midrule
\multirow{8}{*}{96} &
  RNN &
  0.2526 &
  0.1913 &
  0.1832 &
  0.1835 &
  \textbf{0.1832} &
  0.3611 &
  0.2884 &
  0.2821 &
  0.2755 &
  \textbf{0.2616} \\
 &
  MGU &
  0.2578 &
  0.1820 &
  0.1836 &
  0.1891 &
  \textbf{0.1691} &
  0.3713 &
  0.2843 &
  0.2852 &
  0.2829 &
  \textbf{0.2701} \\
 &
  GRU &
  0.2502 &
  0.1906 &
  0.1820 &
  \textbf{0.1812} &
  0.1822 &
  0.3812 &
  0.3041 &
  0.2743 &
  0.2883 &
  \textbf{0.2698} \\
 &
  LSTM &
  0.2543 &
  0.1847 &
  0.1851 &
  0.1826 &
  \textbf{0.1769} &
  0.3831 &
  0.2936 &
  0.2938 &
  0.2888 &
  \textbf{0.2693} \\
 &
  IndRNN &
  0.2564 &
  0.1934 &
  0.1902 &
  0.1809 &
  \textbf{0.1805} &
  0.3865 &
  0.3015 &
  0.2833 &
  0.2919 &
  \textbf{0.2648} \\
 &
  phLSTM &
  0.2457 &
  0.2089 &
  0.2158 &
  0.2040 &
  \textbf{0.1808} &
  0.3754 &
  0.3124 &
  0.2859 &
  0.2759 &
  \textbf{0.2570} \\
 &
  pLSTM &
  0.2625 &
  0.1911 &
  0.1969 &
  0.1801 &
  \textbf{0.1723} &
  0.3769 &
  0.3163 &
  0.2986 &
  0.2962 &
  \textbf{0.2624} \\
 &
  xLSTM &
  0.2359 &
  0.1875 &
  0.1916 &
  0.1799 &
  \textbf{0.1767} &
  0.3774 &
  0.2950 &
  0.2807 &
  0.2826 &
  \textbf{0.2607} \\ \midrule
\multicolumn{2}{c|}{Average} &
  0.2250 &
  0.1754 &
  0.1699 &
  0.1674 &
  \textbf{0.1579} &
  0.3363 &
  0.2799 &
  0.2677 &
  0.2689 &
  \textbf{0.2533} \\ \midrule
\multicolumn{2}{c|}{Improvement} &
  \textbf{29.85\%} &
  \textbf{10.00\%} &
  \textbf{7.07\% }&
  \textbf{5.71\%} &
  - &
  \textbf{24.69\%} &
  \textbf{9.52\%} &
  \textbf{5.40\%} &
  \textbf{5.81\% }&
  - \\ \midrule
\multicolumn{2}{c|}{Count} &
  0 &
  0 &
  3 &
  1 &
  \textbf{20} &
  0 &
  0 &
  3 &
  0 &
  \textbf{21} \\ \bottomrule
\end{tabular}%
}
\end{table}
\begin{table}[t]
\centering
\caption{Average MSE and MAE performance of our method  and other baseline methods over different backbone on ETTh data.}
\label{tab:ablation-ETTh}
\resizebox{\textwidth}{!}{%
\begin{tabular}{@{}cc|ccccc|ccccc@{}}
\toprule
\multirow{2}{*}{$H$} &
  \multirow{2}{*}{Backbone} &
  \multicolumn{5}{c|}{MSE} &
  \multicolumn{5}{c}{MAE} \\ \cmidrule(l){3-12} 
 &
   &
  TA-PSO &
  RRE-PG &
  RRE-DQN &
  RRE-PPO &
  Ours  &
  TA-PSO &
  RRE-PG &
  RRE-DQN &
  RRE-PPO &
  Ours  \\ \midrule
\multirow{8}{*}{24} &
  RNN &
  0.3946 &
  0.3256 &
  0.3113 &
  0.3214 &
  \textbf{0.3096} &
  0.4642 &
  0.3874 &
  0.3669 &
  0.3775 &
  \textbf{0.3647} \\
 &
  MGU &
  0.4129 &
  0.3122 &
  0.3261 &
  0.3156 &
  \textbf{0.3016} &
  0.4837 &
  0.3798 &
  0.3813 &
  0.3743 &
  \textbf{0.3503} \\
 &
  GRU &
  0.3960 &
  0.3246 &
  0.3303 &
  0.3343 &
  \textbf{0.3059} &
  0.4639 &
  0.3869 &
  0.3881 &
  0.3901 &
  \textbf{0.3593} \\
 &
  LSTM &
  0.4099 &
  0.3241 &
  0.3157 &
  0.3159 &
  \textbf{0.3067} &
  0.4864 &
  0.3901 &
  \textbf{0.3612} &
  0.3731 &
  0.3620 \\
 &
  IndRNN &
  0.4372 &
  0.3418 &
  0.3367 &
  0.3273 &
  \textbf{0.3147} &
  0.5166 &
  0.3920 &
  0.3978 &
  0.3909 &
  \textbf{0.3614} \\
 &
  phLSTM &
  0.4235 &
  0.3511 &
  0.3297 &
  0.3269 &
  \textbf{0.3174} &
  0.4971 &
  0.4032 &
  0.3708 &
  0.3914 &
  \textbf{0.3666} \\
 &
  pLSTM &
  0.3961 &
  0.3434 &
  0.3395 &
  0.3388 &
  \textbf{0.3237} &
  0.4674 &
  0.4026 &
  0.3875 &
  0.3969 &
  \textbf{0.3743} \\
 &
  xLSTM &
  0.4023 &
  0.3161 &
  0.3265 &
  0.3216 &
  \textbf{0.3075} &
  0.4716 &
  0.3729 &
  0.3858 &
  0.3769 &
  \textbf{0.3565} \\ \midrule
\multirow{8}{*}{48} &
  RNN &
  0.4790 &
  0.3609 &
  0.3434 &
  0.3392 &
  \textbf{0.3225} &
  0.5470 &
  0.4140 &
  0.3925 &
  0.3899 &
  \textbf{0.3716} \\
 &
  MGU &
  0.4812 &
  0.3723 &
  0.3651 &
  0.3817 &
  \textbf{0.3394} &
  0.5572 &
  0.4298 &
  0.4209 &
  0.4398 &
  \textbf{0.3820} \\
 &
  GRU &
  0.4836 &
  0.3657 &
  \textbf{0.3421} &
  0.3678 &
  0.3466 &
  0.5497 &
  0.4185 &
  0.4238 &
  0.4293 &
  \textbf{0.3865} \\
 &
  LSTM &
  0.4666 &
  0.3594 &
  0.3613 &
  0.3556 &
  \textbf{0.3449} &
  0.5429 &
  0.4142 &
  0.4172 &
  0.4084 &
  \textbf{0.3939} \\
 &
  IndRNN &
  0.4616 &
  0.3824 &
  0.3829 &
  0.3804 &
  \textbf{0.3311} &
  0.5331 &
  0.4364 &
  0.3944 &
  0.4396 &
  \textbf{0.3776} \\
 &
  phLSTM &
  0.4746 &
  0.3545 &
  0.3683 &
  0.3502 &
  \textbf{0.3397} &
  0.5448 &
  0.4101 &
  \textbf{0.3841} &
  0.3998 &
  0.3881 \\
 &
  pLSTM &
  0.4921 &
  0.3930 &
  0.3674 &
  0.3750 &
  \textbf{0.3536} &
  0.5578 &
  0.4508 &
  \textbf{0.3912} &
  0.4371 &
  0.3973 \\
 &
  xLSTM &
  0.4564 &
  0.3885 &
  0.3461 &
  0.3548 &
  \textbf{0.3378} &
  0.5194 &
  0.4462 &
  0.4154 &
  0.4055 &
  \textbf{0.3746} \\ \midrule
\multirow{8}{*}{96} &
  RNN &
  0.5281 &
  0.4187 &
  0.3992 &
  0.4063 &
  \textbf{0.3683} &
  0.5914 &
  0.4617 &
  0.4390 &
  0.4476 &
  \textbf{0.3962} \\
 &
  MGU &
  0.5346 &
  0.4443 &
  0.3725 &
  0.3969 &
  \textbf{0.3584} &
  0.5949 &
  0.4902 &
  0.4859 &
  0.4447 &
  \textbf{0.3959} \\
 &
  GRU &
  0.5530 &
  0.3970 &
  0.3899 &
  0.3908 &
  \textbf{0.3687} &
  0.6103 &
  0.4426 &
  0.4408 &
  0.4300 &
  \textbf{0.4098} \\
 &
  LSTM &
  0.5636 &
  0.3828 &
  0.3876 &
  0.3888 &
  \textbf{0.3588} &
  0.6163 &
  0.4232 &
  0.4337 &
  0.4244 &
  \textbf{0.3911} \\
 &
  IndRNN &
  0.5461 &
  0.3952 &
  0.4032 &
  0.3866 &
  \textbf{0.3678} &
  0.5990 &
  0.4384 &
  0.4365 &
  0.4264 &
  \textbf{0.3971} \\
 &
  phLSTM &
  0.5413 &
  0.4344 &
  \textbf{0.3753} &
  0.3937 &
  0.3863 &
  0.5958 &
  0.4731 &
  0.4226 &
  0.4391 &
  \textbf{0.4121} \\
 &
  pLSTM &
  0.5331 &
  0.4575 &
  0.3902 &
  0.4659 &
  \textbf{0.3778} &
  0.5946 &
  0.5057 &
  0.4345 &
  0.5151 &
  \textbf{0.4055} \\
 &
  xLSTM &
  0.5150 &
  0.4311 &
  0.3873 &
  0.3882 &
  \textbf{0.3607} &
  0.5702 &
  0.4741 &
  0.4295 &
  0.4329 &
  \textbf{0.3897} \\ \midrule
\multicolumn{2}{c|}{Average} &
  0.4743 &
  0.3740 &
  0.3578 &
  0.3635 &
  \textbf{0.3396} &
  0.5406 &
  0.4268 &
  0.4084 &
  0.4159 &
  \textbf{0.3818} \\ \midrule
\multicolumn{2}{c|}{Improvement} &
  \textbf{28.40\%} &
  \textbf{9.21\%} &
  \textbf{5.10\%} &
  \textbf{6.58\%} &
  - &
  \textbf{29.37\%} &
  \textbf{10.54\%} &
  \textbf{6.50\%} &
  \textbf{8.18\%} &
  - \\ \midrule
\multicolumn{2}{c|}{Count} &
  0 &
  0 &
  2 &
  0 &
  \textbf{22} &
  0 &
  0 &
  3 &
  0 &
  \textbf{21} \\ \bottomrule
\end{tabular}%
}
\end{table}

\begin{table}[t]
\centering
\caption{Average MSE and MAE performance of our RRE-PPO4Pred method  and its ablation variant methods over different backbones and different horizons on Weather data.}
\label{tab:ablation-Weather}
\resizebox{\textwidth}{!}{%
\begin{tabular}{@{}cc|ccccc|ccccc@{}}
\toprule
\multirow{2}{*}{$H$} &
  \multirow{2}{*}{Backbone} &
  \multicolumn{5}{c|}{MSE} &
  \multicolumn{5}{c}{MAE} \\ \cmidrule(l){3-12} 
 &
   &
  TA-PSO &
  RRE-PG &
  RRE-DQN &
  RRE-PPO &
  Ours  &
  PSO-TA &
  PG-TA &
  DQN-TA &
  PPO-TA &
  DTAS  \\ \midrule
\multirow{8}{*}{24} &
  RNN &
  0.2849 &
  0.1881 &
  0.1768 &
  0.1841 &
  \textbf{0.1741} &
  0.3608 &
  0.2695 &
  0.2582 &
  0.2592 &
  \textbf{0.2574} \\
 &
  MGU &
  0.2707 &
  0.1996 &
  0.1788 &
  0.1747 &
  \textbf{0.1674} &
  0.3485 &
  0.2610 &
  0.2606 &
  0.2915 &
  \textbf{0.2565} \\
 &
  GRU &
  0.2817 &
  0.1948 &
  0.1850 &
  0.1920 &
  \textbf{0.1671} &
  0.3607 &
  0.2671 &
  0.2692 &
  0.2877 &
  \textbf{0.2489} \\
 &
  LSTM &
  0.2699 &
  0.1871 &
  0.1740 &
  0.1735 &
  \textbf{0.1673} &
  0.3564 &
  0.2621 &
  0.2842 &
  0.2579 &
  \textbf{0.2528} \\
 &
  IndRNN &
  0.2825 &
  0.2038 &
  0.1965 &
  0.1979 &
  \textbf{0.1782} &
  0.3690 &
  0.2719 &
  0.2718 &
  0.2821 &
  \textbf{0.2652} \\
 &
  phLSTM &
  0.2863 &
  0.1959 &
  0.1911 &
  0.1906 &
  \textbf{0.1902} &
  0.3794 &
  0.2772 &
  \textbf{0.2662} &
  0.2701 &
  0.2774 \\
 &
  pLSTM &
  0.2725 &
  0.2158 &
  0.2057 &
  0.2028 &
  \textbf{0.1718} &
  0.3609 &
  0.2872 &
  0.2833 &
  0.2646 &
  \textbf{0.2620} \\
 &
  xLSTM &
  0.2702 &
  0.1967 &
  0.1880 &
  0.1891 &
  \textbf{0.1698} &
  0.3352 &
  0.2540 &
  0.2543 &
  0.2629 &
  \textbf{0.2438} \\ \midrule
\multirow{8}{*}{48} &
  RNN &
  0.3155 &
  0.2108 &
  \textbf{0.1951} &
  0.2189 &
  0.2002 &
  0.3804 &
  0.2824 &
  \textbf{0.2716} &
  0.2864 &
  0.2865 \\
 &
  MGU &
  0.3120 &
  0.2273 &
  0.2009 &
  0.1958 &
  \textbf{0.1791} &
  0.3971 &
  0.2993 &
  0.2776 &
  0.2889 &
  \textbf{0.2605} \\
 &
  GRU &
  0.3229 &
  0.2117 &
  \textbf{0.1938} &
  0.2008 &
  0.1979 &
  0.3887 &
  0.2880 &
  \textbf{0.2717} &
  0.2940 &
  0.2895 \\
 &
  LSTM &
  0.3003 &
  0.2240 &
  0.2149 &
  0.2200 &
  \textbf{0.1884} &
  0.3673 &
  0.2860 &
  0.2910 &
  0.3005 &
  \textbf{0.2784} \\
 &
  IndRNN &
  0.3218 &
  0.2303 &
  0.2216 &
  0.2287 &
  \textbf{0.1953} &
  0.3851 &
  0.2903 &
  0.2906 &
  0.3087 &
  \textbf{0.2844} \\
 &
  phLSTM &
  0.3277 &
  0.2204 &
  \textbf{0.2057} &
  0.2135 &
  0.2064 &
  0.3988 &
  0.3092 &
  0.2856 &
  0.2821 &
  \textbf{0.2812} \\
 &
  pLSTM &
  0.2979 &
  0.2480 &
  0.2194 &
  0.2143 &
  \textbf{0.2125} &
  0.3907 &
  0.3073 &
  \textbf{0.2849} &
  0.3162 &
  0.3056 \\
 &
  xLSTM &
  0.3018 &
  0.2409 &
  0.2429 &
  0.2421 &
  \textbf{0.1839} &
  0.3621 &
  0.2950 &
  0.2764 &
  0.2765 &
  \textbf{0.2552} \\ \midrule
\multirow{8}{*}{96} &
  RNN &
  0.3420 &
  0.2476 &
  0.2419 &
  \textbf{0.2411} &
  0.2412 &
  0.4183 &
  0.3221 &
  \textbf{0.3031} &
  0.2924 &
  0.2896 \\
 &
  MGU &
  0.3374 &
  0.2461 &
  0.2437 &
  0.2442 &
  \textbf{0.2407} &
  0.4032 &
  0.3199 &
  0.3080 &
  0.2929 &
  \textbf{0.2786} \\
 &
  GRU &
  0.3453 &
  0.2550 &
  0.2461 &
  0.2394 &
  \textbf{0.2355} &
  0.4175 &
  0.3189 &
  0.3207 &
  0.3074 &
  \textbf{0.2987} \\
 &
  LSTM &
  0.3449 &
  0.2642 &
  0.2517 &
  0.2536 &
  \textbf{0.2129} &
  0.4196 &
  0.3195 &
  0.3296 &
  0.3206 &
  \textbf{0.3074} \\
 &
  IndRNN &
  0.3544 &
  0.2925 &
  0.2510 &
  0.2470 &
  \textbf{0.2298} &
  0.4281 &
  0.3379 &
  0.3252 &
  0.3107 &
  \textbf{0.3006} \\
 &
  phLSTM &
  0.3789 &
  0.2735 &
  0.2700 &
  0.2612 &
  \textbf{0.2317} &
  0.4422 &
  0.3147 &
  0.3255 &
  0.2993 &
  \textbf{0.2944} \\
 &
  pLSTM &
  0.3773 &
  0.2775 &
  0.2741 &
  0.2801 &
  \textbf{0.2182} &
  0.4470 &
  0.3317 &
  0.3398 &
  0.3482 &
  \textbf{0.3156} \\
 &
  xLSTM &
  0.3589 &
  0.2516 &
  0.2304 &
  0.2267 &
  \textbf{0.1927} &
  0.4006 &
  0.3074 &
  0.2974 &
  0.2779 &
  \textbf{0.2715} \\ \midrule
\multicolumn{2}{c|}{Average} &
  0.3149 &
  0.2293 &
  0.2166 &
  0.2180 &
  \textbf{0.1984} &
  0.3882 &
  0.2950 &
  0.2894 &
  0.2908 &
  \textbf{0.2776} \\ \midrule
\multicolumn{2}{c|}{Improvement} &
  \textbf{37.01\%} &
  \textbf{13.49\%} &
  \textbf{8.43\% }&
  \textbf{9.01\%} &
  - &
  \textbf{28.50\%} &
  \textbf{5.90\% }&
  \textbf{4.10\%} &
  \textbf{4.54\%} &
  - \\ \midrule
\multicolumn{2}{c|}{Count} &
  0 &
  0 &
  3 &
  1 &
  \textbf{20} &
  0 &
  0 &
  4 &
  0 &
  \textbf{20} \\ \bottomrule
\end{tabular}%
}
\end{table}

\begin{table}[t]
\centering
\caption{Average MSE and MAE performance of our RRE-PPO4Pred method  and its ablation variant methods over different backbones and different horizons on ILI data.}
\label{tab:ablation-ILI}
\resizebox{\textwidth}{!}{%
\begin{tabular}{@{}cc|ccccc|ccccc@{}}
\toprule
\multirow{2}{*}{$H$} &
  \multirow{2}{*}{Backbone} &
  \multicolumn{5}{c|}{MSE} &
  \multicolumn{5}{c}{MAE} \\ \cmidrule(l){3-12} 
 &
   &
  TA-PSO &
  RRE-PG &
  RRE-DQN &
  RRE-PPO &
  Ours  &
  TA-PSO &
  RRE-PG &
  RRE-DQN &
  RRE-PPO &
  Ours \\ \midrule
\multirow{8}{*}{12} &
  RNN &
  2.7473 &
  2.3437 &
  2.1283 &
  2.1456 &
  \textbf{2.0618} &
  1.3744 &
  1.0552 &
  0.9376 &
  0.9446 &
  \textbf{0.9366} \\
 &
  MGU &
  2.6592 &
  2.5335 &
  2.4477 &
  2.2992 &
  \textbf{2.1034} &
  1.3327 &
  1.1424 &
  1.0809 &
  1.1026 &
  \textbf{0.9751} \\
 &
  GRU &
  2.7418 &
  2.2215 &
  2.1478 &
  2.2728 &
  \textbf{2.1072} &
  1.3731 &
  1.0030 &
  \textbf{0.9464} &
  1.0454 &
  0.9650 \\
 &
  LSTM &
  2.7874 &
  2.3270 &
  2.2667 &
  2.0031 &
  \textbf{1.9536} &
  1.3949 &
  1.0507 &
  0.9983 &
  0.9854 &
  \textbf{0.8959} \\
 &
  IndRNN &
  2.6951 &
  2.3564 &
  \textbf{2.0645} &
  2.0912 &
  2.0732 &
  1.3524 &
  1.0609 &
  1.0758 &
  0.9742 &
  \textbf{0.9550} \\
 &
  phLSTM &
  2.9618 &
  2.3424 &
  2.2615 &
  2.2780 &
  \textbf{2.0434} &
  1.4854 &
  1.0577 &
  0.9985 &
  1.0941 &
  \textbf{0.9353} \\
 &
  pLSTM &
  2.9460 &
  2.1346 &
  2.0941 &
  2.1874 &
  \textbf{1.9853} &
  1.4761 &
  0.9606 &
  0.9254 &
  0.9664 &
  \textbf{0.9015} \\
 &
  xLSTM &
  2.6380 &
  2.0411 &
  1.9686 &
  1.9816 &
  \textbf{1.9429} &
  1.3237 &
  0.9229 &
  \textbf{0.8676} &
  0.8737 &
  0.9021 \\ \midrule
\multirow{8}{*}{24} &
  RNN &
  2.9389 &
  2.4864 &
  \textbf{2.1924} &
  2.3411 &
  2.2374 &
  1.5483 &
  1.1799 &
  1.0717 &
  1.0859 &
  \textbf{1.0716} \\
 &
  MGU &
  2.8906 &
  2.6026 &
  2.5371 &
  2.5509 &
  \textbf{2.2749} &
  1.5221 &
  1.2340 &
  1.1781 &
  1.1824 &
  \textbf{1.0979} \\
 &
  GRU &
  2.8924 &
  2.5257 &
  2.3690 &
  2.4846 &
  \textbf{2.1150} &
  1.5269 &
  1.2002 &
  1.0993 &
  1.1547 &
  \textbf{1.0054} \\
 &
  LSTM &
  2.8894 &
  2.6052 &
  2.5443 &
  2.5016 &
  \textbf{2.0167} &
  1.5253 &
  1.2347 &
  1.1806 &
  1.2072 &
  \textbf{0.9663} \\
 &
  IndRNN &
  2.8975 &
  2.6022 &
  2.6404 &
  2.5119 &
  \textbf{1.9912} &
  1.5273 &
  1.2371 &
  1.2269 &
  1.2125 &
  \textbf{0.9755} \\
 &
  phLSTM &
  2.9839 &
  2.5958 &
  2.7224 &
  2.5885 &
  \textbf{2.4847} &
  1.5711 &
  1.2325 &
  1.2649 &
  \textbf{1.2008} &
  1.2013 \\
 &
  pLSTM &
  2.9739 &
  2.5358 &
  2.6357 &
  2.4309 &
  \textbf{2.2223} &
  1.5688 &
  1.2014 &
  1.2233 &
  1.1293 &
  \textbf{1.0687} \\
 &
  xLSTM &
  2.9158 &
  2.3727 &
  2.2642 &
  2.3406 &
  \textbf{2.0552} &
  1.5359 &
  1.1282 &
  \textbf{1.0494} &
  1.0873 &
  1.0805 \\ \midrule
\multirow{8}{*}{48} &
  RNN &
  2.9834 &
  2.4931 &
  \textbf{2.3067} &
  2.4124 &
  2.3958 &
  1.6607 &
  1.2480 &
  1.2256 &
  1.2786 &
  \textbf{1.2153} \\
 &
  MGU &
  3.0290 &
  2.7380 &
  2.8790 &
  2.5033 &
  \textbf{2.2778} &
  1.6831 &
  1.3702 &
  1.4091 &
  1.2243 &
  \textbf{1.1513} \\
 &
  GRU &
  3.0021 &
  2.8581 &
  2.9732 &
  2.5464 &
  \textbf{2.2779} &
  1.6691 &
  1.4314 &
  1.4549 &
  1.2457 &
  \textbf{1.1471} \\
 &
  LSTM &
  3.0050 &
  2.7262 &
  \textbf{2.5428} &
  2.6820 &
  2.5619 &
  1.6714 &
  1.3674 &
  \textbf{1.2842} &
  1.3158 &
  1.2968 \\
 &
  IndRNN &
  3.1719 &
  2.8630 &
  2.6721 &
  2.7338 &
  \textbf{2.2225} &
  1.7640 &
  1.4352 &
  1.3094 &
  1.3400 &
  \textbf{1.1167} \\
 &
  phLSTM &
  3.1549 &
  2.9590 &
  2.9581 &
  2.5917 &
  \textbf{2.2940} &
  1.7568 &
  1.4809 &
  1.4476 &
  1.2708 &
  \textbf{1.1621} \\
 &
  pLSTM &
  3.0984 &
  2.5455 &
  2.5953 &
  2.5836 &
  \textbf{2.4081} &
  1.7228 &
  1.2757 &
  1.5812 &
  1.2643 &
  \textbf{1.2193} \\
 &
  xLSTM &
  2.8393 &
  2.5617 &
  2.4205 &
  2.3731 &
  \textbf{2.1893} &
  1.5796 &
  1.2814 &
  1.6389 &
  1.2133 &
  \textbf{1.1572} \\ \midrule
\multicolumn{2}{c|}{Average} &
  2.9101 &
  2.5155 &
  2.4430 &
  2.3931 &
  \textbf{2.1790} &
  1.5394 &
  1.1996 &
  1.1865 &
  1.1416 &
  \textbf{1.0583} \\ \midrule
\multicolumn{2}{c|}{Improvement} &
  \textbf{25.12\%} &
  \textbf{13.38\%} &
  \textbf{10.81\%} &
  \textbf{8.95\% }&
  - &
  \textbf{31.25\%} &
  \textbf{11.78\%} &
  \textbf{10.80\%} &
  \textbf{7.30\%} &
  - \\ \midrule
\multicolumn{2}{c|}{Count} &
  0 &
  0 &
  4 &
  0 &
  \textbf{20} &
  0 &
  0 &
  4 &
  1 &
  \textbf{19} \\ \bottomrule
\end{tabular}%
}
\end{table}

\end{document}